\documentclass[10pt,journal,compsoc]{IEEEtran}

\usepackage{graphicx}
\usepackage{epstopdf}
\usepackage{amsmath}
\usepackage{subfigure}
\usepackage{algorithmic}
\usepackage{algorithm}
\usepackage{bm}
\usepackage{multirow}
\usepackage[colorlinks,linkcolor=red]{hyperref}
\usepackage{colortbl}
\usepackage{url}
\definecolor{mygray}{gray}{.9}

\begin{document}

\title{Searching for Representative Modes on Hypergraphs for Robust Geometric Model Fitting}

\author{Hanzi Wang,~\IEEEmembership{Senior Member,~IEEE,}
        Guobao Xiao,
        Yan Yan,~\IEEEmembership{Member,~IEEE,}
        and~David Suter
\IEEEcompsocitemizethanks{\IEEEcompsocthanksitem H. Wang and Y. Yan are with the Fujian Key Laboratory of Sensing and Computing for Smart City, School of Information Science and Engineering, Xiamen University, China.
~G. Xiao is with the School of Aerospace Engineering, Xiamen University, China. D. Suter is with the School of Computer Science, The University of Adelaide, Australia.
\IEEEcompsocthanksitem H. Wang and G. Xiao contributed equally.}
\thanks{Manuscript received XXX; revised XXX.}}

\markboth{IEEE TRANSACTIONS ON PATTERN ANALYSIS AND MACHINE INTELLIGENCE,~Vol.~XX, No.~XX, XXX}%
{Shell \MakeLowercase{\textit{et al.}}: Bare Demo of IEEEtran.cls for Computer Society Journals}
\IEEEtitleabstractindextext{%
\begin{abstract}
In this paper, we propose a simple and effective geometric model fitting method to fit and segment multi-structure data even in the presence of severe outliers. We cast the task of geometric model fitting as a representative mode-seeking problem on hypergraphs. Specifically, a hypergraph is firstly constructed, where the vertices represent model hypotheses and the hyperedges denote data points. The hypergraph involves higher-order similarities (instead of pairwise similarities used on a simple graph), and it can characterize complex relationships between model hypotheses and data points. In addition, we develop a hypergraph reduction technique to remove ``insignificant" vertices while retaining as many ``significant" vertices as possible in the hypergraph. Based on the simplified hypergraph, we then propose a novel mode-seeking algorithm to search for representative modes within reasonable time. Finally, the proposed mode-seeking algorithm detects modes according to two key elements, i.e., the weighting scores of vertices and the similarity analysis between vertices. Overall, the proposed fitting method is able to efficiently and effectively estimate the number and the parameters of model instances in the data simultaneously. Experimental results demonstrate that the proposed method achieves significant superiority over several state-of-the-art model fitting methods on both synthetic data and real images.
\end{abstract}

}

\maketitle
\IEEEdisplaynontitleabstractindextext
\IEEEpeerreviewmaketitle
\IEEEraisesectionheading{\section{Introduction}\label{sec:introduction}}

\IEEEPARstart{G}{eometric} model fitting is a challenging research problem for a variety of applications in computer vision, such as optical flow calculation, motion segmentation and homography/fundamental matrix estimation. Given that data usually contain outliers, the task of geometric model fitting is to robustly estimate {the number} and the parameters of model instances in data. A number of robust geometric model fitting methods (e.g., { \cite{chin2009robust,fischler1981random,pearl:ijcv12,Magri_2014_CVPR,toldo2008robust,wang2012simultaneously,tennakoon2016robust,fredriksson2015practical}}) have been proposed. One of the most popular robust fitting methods is RANSAC~\cite{fischler1981random} due to its efficiency and simplicity. However, RANSAC is sensitive to a threshold specified by a user and it is originally designed to fit single-structure data. During the past few decades, many robust {model} fitting methods have been proposed to deal with multi-structure data, such as KF~\cite{chin2009robust}, PEARL~\cite{pearl:ijcv12}, AKSWH~\cite{wang2012simultaneously}, T-linkage~\cite{Magri_2014_CVPR}, SCAMS~\cite{li2014scams} and PM~\cite{tennakoon2016robust}. However, current fitting methods are still far from being practical to deal with real-world problems, due to the limitations of speed or accuracy. {In this paper, we aim to accurately detect model instances in data within reasonable time}.

\begin{figure}
\centering
\begin{minipage}{.24\textwidth}
\centerline{\includegraphics[width=1.0\textwidth]{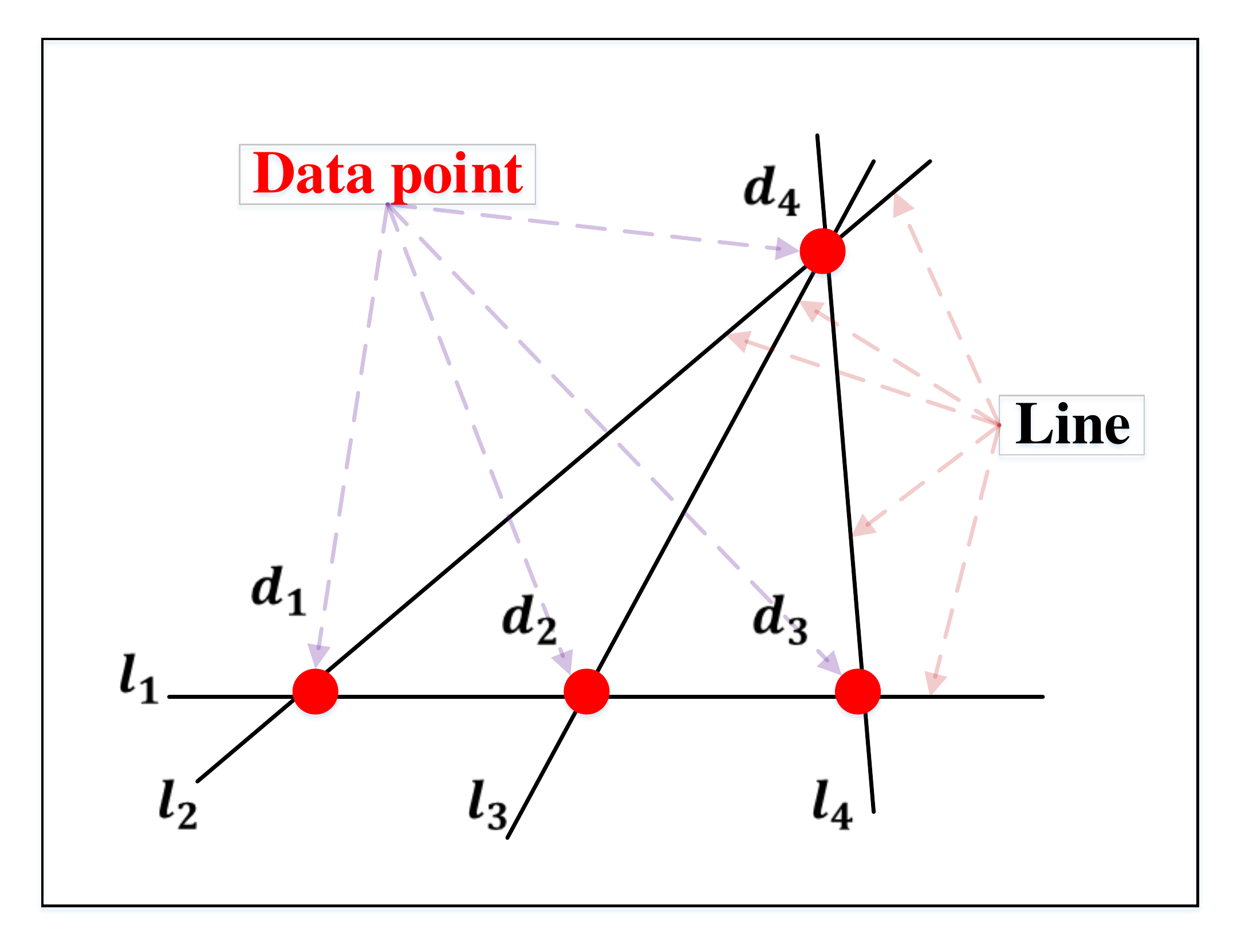}}
  \centerline{(a)}
\end{minipage}
\begin{minipage}{.24\textwidth}
\centerline{\includegraphics[width=1.0\textwidth]{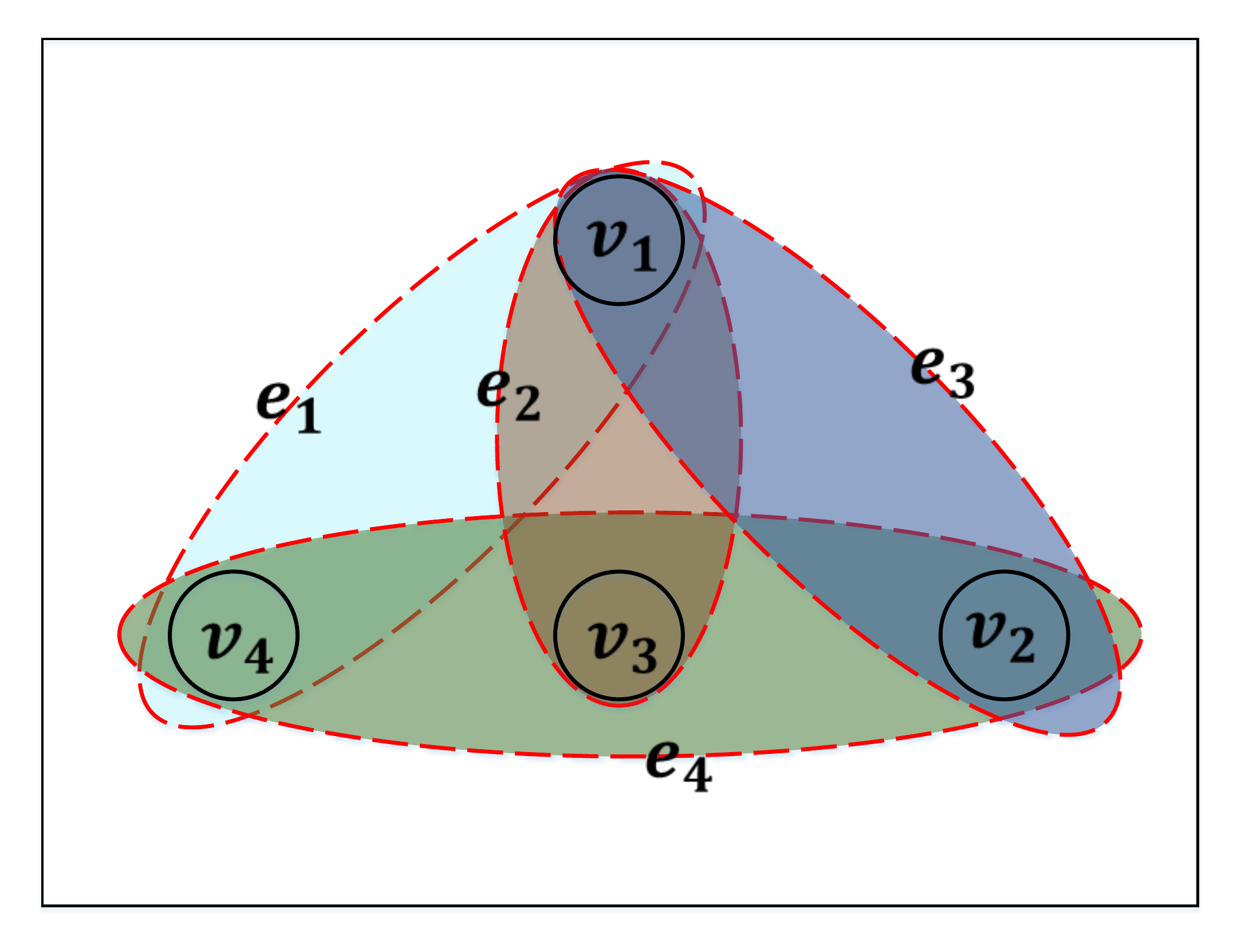}}
  \centerline{(b)}
\end{minipage}
\caption{An example of hypergraph construction for line fitting. (a) The input data including four data points and four model hypotheses (i.e., lines). (b) A hypergraph with four vertices $\{v_i\}_{i=1}^4$ and four hyperedges $\{e_i\}_{i=1}^4$. In the hypergraph, each vertex $v_i$ and each hyperedge $e_i$ denote a model hypothesis $l_i$ and a data point $d_i$ in (a), respectively.}
\label{fig:hypergraphexample}
\end{figure}
{Note that a hypergraph involves high-order similarities, and some works have been proposed to deal with the model fitting problem based on hypergraphs, e.g., \cite{jain2013efficient,liu2012efficient,ochs2012higher,pulak2014clustering}.} Although using a hypergraph is beneficial for a model fitting method in terms of the fitting accuracy, it also causes {the problem of} high computational complexity due to complex relationships in the hypergraph. {To reduce the computational complexity, these hypergraph based fitting} {methods usually fix the degree of each hyperedge in the hypergraph to be a small constant value. However, such a way cannot characterize the complex relationships between model hypotheses and data points (for the model fitting problem) very well.}

In this paper, we propose a simple and effective Mode-Seeking on Hypergraphs Fitting method (MSHF) to fit and segment multi-structure data. {The proposed method (MSHF) starts from hypergraph construction, where vertices and hyperedges respectively correspond to model hypotheses and data points (as shown in Fig.~\ref{fig:hypergraphexample}). We also develop a novel hypergraph reduction technique to remove insignificant vertices, which improves the effectiveness of the constructed hypergraph. After that, we propose a novel mode-seeking algorithm to {search for} representative modes on the hypergraph. Finally, MSHF simultaneously estimates the number} and the parameters of all model instances in data (according to the detected modes).

The proposed MSHF method has three main advantages over previous model fitting methods. {First, the constructed hypergraph is able to effectively characterize the complex relationships between model hypotheses and data points, and the {size} of each hyperedge in the hypergraph is data-driven. Moreover, the hypergraph can be directly used for geometric model fitting. That is, MSHF avoids constructing a pairwise affinity matrix as used in~\cite{pu2012hypergraph} and~\cite{bulo2009game}. Note that the projection from a hypergraph to an induced graph usually causes information-loss (except for the methods using the affinity tensor)}. Second, MSHF deals with geometric model fitting in the parameter space, {which can effectively handle severe unbalanced data}  (i.e., the numbers of inliers belonging to different model instances in data are significantly different). Third, MSHF performs mode seeking by analyzing the weighting scores and the similarity between vertices on a hypergraph, which shows great scalability to solve {the model fitting problem}. We demonstrate that MSHF is a highly robust method for geometric model fitting by conducting extensive experimental evaluations and comparisons in Sec.~\ref{sec:experiments}.

This paper is an extension of our previous work in~\cite{wang2015mode}. We have made {several significant improvements}{, including a novel hypergraph reduction technique to improve the performance of the original proposed method on fitting accuracy (Sec.~\ref{sec:hypergraphreduction}), more theoretical analyses (Sec.~\ref{sec:relatework}, Sec.~\ref{sec:analysis} and Sec.~\ref{sec:limitations}) and more experimental justification (Sec.~\ref{sec:experiments}). We have also used the neighboring constraint to reduce the computational cost of MSHF (Sec.~\ref{sec:algorihmanalysis}).}


The rest of the paper is organized as follows: We {provide an overview of the related work in Sec.~\ref{sec:relatework}}. {We describe the components of the proposed fitting method in Sec.~\ref{sec:HypergraphModelling} and summarize the complete fitting method in Sec.~\ref{sec:completealgorithm}. We present the experimental results on both synthetic and real data in Sec.~\ref{sec:experiments}}. We analyze the limitations of the proposed method in Sec.~\ref{sec:limitations}, and draw conclusions in Sec.~\ref{sec:conclusion}.
\section{Related Work}
\label{sec:relatework}
In this section, we briefly review the related work on robust geometric model fitting including { the hypergraph-based fitting methods, the mode-seeking based fitting methods, and several other state-of-the-art fitting methods}.
\subsection{Hypergraph Based Fitting Methods}
Recently, some hypergraph based methods, e.g., \cite{jain2013efficient,liu2012efficient,ochs2012higher,pulak2014clustering}, have been proposed for robust model fitting {due to its effectiveness}. For example, Liu and Yan \cite{liu2012efficient} proposed the random consensus graph (RCG) to fit multiple structures in data. Purkait et al.~\cite{pulak2014clustering} proposed to use large hyperedges for face clustering and motion segmentation.

Compared with the hypergraph constructed in the previous methods~(e.g., \cite{jain2013efficient,liu2012efficient,ochs2012higher,pulak2014clustering}), where a hyperedge is constrained to connect with a fixed number of vertices, the hyperedge of hypergraphs constructed in this paper can connect with a varying number of vertices {(that is we construct non-uniform hypergraphs as those in \cite{Debarghya2017})}. { In addition, the vertices of the hypergraph constructed in the previous methods (e.g., \cite{jain2013efficient,liu2012efficient,ochs2012higher,pulak2014clustering}) represent data points, while the vertices of the hypergraph constructed in this paper denote model hypotheses. Therefore, we can directly deal with the model fitting problem in the parameter space.}

\subsection{Mode-seeking Based Fitting Methods}
Mode-seeking is a simple and effective data analysis technique, and it can be extended to deal with model fitting problems (e.g., \cite{comaniciu2002mean, subbarao2009nonlinear,hough1962method,xu1990new}). These mode-seeking based fitting methods select model instances by {seeking the peaks of the underlying distributions in} the parameter space. Each point in the parameter space corresponds to a model hypothesis, and the detected modes represent the estimated model instances. For example, { Mean Shift \cite{comaniciu2002mean} and its variant~\cite{subbarao2009nonlinear} attempt to find peaks in the parameter space to estimate the model instances in the data.} Hough~\cite{hough1962method} proposed a robust fitting method, called the Hough Transform (HT), which discretizes the parameter space into bins {and then} votes for these bins according to the information derived from a set of sampled data points. The bins with higher votes correspond to the estimated model instances {in the data}. Xu et al.~\cite{xu1990new} proposed an {extended version} of HT, i.e., Randomized Hough Transform (RHT). RHT uses model hypotheses to vote for the bins in the parameter space to reduce the computational cost of HT.

{The above-mentioned mode-seeking based fitting methods can estimate the number of model instances in data}, but their performance largely depends on the {proportion} of good model hypotheses in {the} generated model hypotheses derived from a set of sampled data points. As a result, these fitting methods {often wrongly} estimate the number of model instances when the {proportion} of good model hypotheses is low. In contrast, the proposed mode-seeking based fitting method {alleviates this} drawback. Specifically,  the proposed method can effectively seek modes by analyzing {both} the weighting scores of vertices, and the similarity between the vertices of hypergraphs. The vertices corresponding to good model hypotheses {usually show unique characteristics} even when the {proportion} of good model hypotheses is low. {Thus, we can select these vertices as modes for model fitting.}

In this paper, we integrate hypergraph construction with mode seeking for solving the model fitting problems. The constructed hypergraph {can effectively capture} the correlation information of model hypotheses and data points, and the proposed mode-seeking method {can efficiently search for} representative modes, which correspond to model instances in data, on the hypergraph. {The proposed method tightly couples both hypergraph construction and mode-seeking, by which it yields better performance for model fitting.}

\subsection{Other Related Fitting Methods}
In addition to the above-mentioned robust fitting methods, there are {several} other related fitting methods, such as KF~\cite{chin2009robust}, J-linkage~\cite{toldo2008robust}, T-linkage~\cite{Magri_2014_CVPR}, {SCAMS~\cite{li2014scams}, PM~\cite{tennakoon2016robust}, PEARL~\cite{pearl:ijcv12}}, AKSWH~\cite{wang2012simultaneously}, HS~\cite{wang2014shifting}, RELRT~\cite{cohen2015likelihood} and GMD~\cite{litman2015inverting}. {KF, J-linkage, T-linkage and SCAMS directly deal with data points for model fitting but they are sensitive to unbalanced data distributions that are quite common in practical applications}. In addition, these methods have difficulties in dealing with the data points near the intersection of model instances. {The computational costs of J-linkage and T-linkage are} high due to the use of the agglomerative clustering procedure. The other robust fitting methods also have {some} problems. For example, {PM requires {the input of} the number of model instances in data; PEARL is sensitive to the initial generated hypotheses; AKSWH may remove some good model hypotheses corresponding to the correct model instances involving a {small} number of data points, during the procedure of selecting significant hypotheses;} HS encounters the computational complexity problem due to the expansion and dropping strategy used; both RELRT and GMD only work for single-structure data.

The proposed method in this paper is based on the mode-seeking technique, which is related to the clustering technique. However, {the proposed method directly searches for the cluster centers in the parameter space}, which can avoid dealing with the data points near the intersection of model instances. In addition, the proposed method can achieve more accurate fitting results for multiple-structure data within reasonable time.

\section{The Methodology}
\label{sec:HypergraphModelling}
In this paper, the geometric model fitting problem is formulated as a mode-seeking problem on a hypergraph. {We describe the details of the proposed MSHF method in this section. Specifically, {we first construct hypergraphs for model fitting} in Sec.~\ref{sec:problemsetting}. Then, we develop a novel hypergraph reduction technique to remove the ``insignificant" vertices in the hypergraph in Sec.~\ref{sec:hypergraphreduction}. After that, we propose a novel mode-seeking algorithm to search for representative modes on the hypergraph in Sec.~\ref{sec:algorihmanalysis}.}
\subsection{Hypergraph Construction}
\label{sec:problemsetting}
A hypergraph $G=(\mathcal{V},\mathcal{E}, \mathcal{W})$ consists of vertices $\mathcal{V}$, hyperedges  $\mathcal{E}$, and weights $\mathcal{W}$. Each vertex $v$ is {weighed} by a weighting score $w(v)$. When $v\in e$, a hyperedge $e$ is incident with a vertex $v$. Then an incident matrix $\mathbf{H}$, whose entry at $(v, e)$ satisfies $h(v,e)=1$ if $v\in e$ and 0 otherwise, is used to represent the relationships between vertices and hyperedges in the hypergraph $G$. For a vertex $v\in \mathcal{V}$, its degree is defined by $\delta(v)=\sum_{e\in \mathcal{E}} h(v,e)$.

{In our case, a vertex in a hypergraph} represents a model hypothesis and a hyperedge denotes a data point. {The detailed procedure of hypergraph construction is described} as follows: Given a set of data points $\mathbf{X}=\{\mathbf{x}_i\}_{i=1}^n$, we first sample a set of minimal subsets from $\mathbf{X}$. A minimal subset contains the minimum number of data points, which is necessary to estimate a model hypothesis (e.g., $2$ for line fitting and $4$ for homography fitting). Then we generate a set of model hypotheses using the minimal subsets and estimate their inlier noise scales. In this paper, we use IKOSE~\cite{wang2012simultaneously} as the inlier noise scale estimator due to its efficiency. After that, we connect each vertex (i.e., a model hypothesis) to the corresponding hyperedges (i.e., the inliers of the model hypothesis). {We can see that, the constructed hypergraph effectively characterizes the relationship between model hypotheses and data points.} In this manner, we can directly perform mode-seeking on the hypergraph for model fitting. 

 {The constructed hypergraph usually includes a large number of vertices, and we assign a weighting score $w(v)$ to each vertex $v$ to measure its quality. Inspired by~\cite{wang2012simultaneously}, we employ the density estimate technique through the following equation to compute $w(v)$ (see Section 3.2 in~\cite{wang2012simultaneously})}
\begin{align}
\label{equ:score1}
w(v)=\frac{1}{n}\sum_{e\in\mathcal{E}} \frac{\Psi(r_e(v)/b(v))}{\hat{s}(v)b(v)},
\end{align}
where $\Psi(\cdot)$ is a kernel function (such as the Epanechnikov kernel); $r_e(v)$ is a residual measured with the {Sampson} distance~\cite{torr1997development} from the model hypothesis {to a data point; $n$ and $\hat{s}(v)$ are the number of hyperedges} and the inlier noise scale of the model hypothesis, respectively{, and $b(v)$ is the window radius (bandwidth)}, which is estimated using~\cite{wand1994kernel}
\begin{align}
\label{equ:bandwith}
b(v)=\left[\frac{243\int_{-1}^1{\Psi(\lambda)}^2d\lambda}{35n\int_{-1}^1{\lambda}^2\Psi(\lambda)d\lambda}\right]^{0.2}\hat{s}(v).
\end{align}

Since the good model hypotheses corresponding to the model instances {in the data} have significantly more inliers with {smaller absolute} residuals than the bad model hypotheses, the weighting scores of the vertices corresponding to the good model hypotheses should be higher than those of {the vertices corresponding to the bad model hypotheses}. However, {weighing} a vertex based on residuals may not be robust to outliers, especially for extreme outliers. To weaken the influence of outliers, we only consider the residuals of the corresponding {inliers} belonging to the model hypotheses {(note that \cite{wang2012simultaneously} considers the residuals of all data points, which is less effective and robust)}. Thus, based on a hypergraph $G$, Eq.~(\ref{equ:score1}) can be reformulated as
\begin{align}
\label{equ:score}
w(v)=\frac{1}{\delta(v)}\sum_{e\in\mathcal{E}} \frac{h(v,e)\Psi(r_e(v)/b(v))}{\hat{s}(v)b(v)},
\end{align}
where $\delta(v)$ is the degree of a vertex $v$ and $h(v,e)$ is {an entry} of the incident matrix $\mathbf{H}$ corresponding to the hypergraph $G$. Recall that $h(v,e)$ will be assigned $0$ if the corresponding data point is not an inlier belonging to the corresponding model hypothesis. Thus, {compared with Eq.~(\ref{equ:score1}),} the weighting score computed by Eq.~(\ref{equ:score}) {is not} greatly influenced by outliers.
\subsection{{Hypergraph Reduction}}
\label{sec:hypergraphreduction}
In~\cite{wang2015mode}, we originally use a weight-aware sampling technique (WAS) to sample vertices according to their weighting scores to avoid ineffective mode-seeking results.  The main task of WAS is to remove a few insignificant vertices corresponding to bad model hypotheses with low weighting scores. However, although WAS can effectively remove some bad model hypotheses, some good model hypotheses may be also discarded in some cases, which will affect the accuracy of the mode-seeking algorithm (see Sec.~\ref{sec:algorihmanalysis}). {Furthermore, WAS has a {low} probability to sample some bad model hypotheses. This will make the whole algorithm unstable}. Therefore, we propose a new hypergraph reduction technique in this paper, which is inspired by the information theoretic approach proposed in \cite{ferraz2007density}, to remove insignificant vertices corresponding to bad model hypotheses with low weighting scores while preserving most of the significant vertices corresponding to good model hypotheses.

Given a hypergraph $G$ with vertices $\mathcal{V}=\{v_1,v_2,\ldots,v_m\}$ and the associated weighting scores $\mathcal{W}=\{w_1,w_2,\ldots,w_m\}$, where $m$ is the number of vertices, i.e.,  the number of the generated model hypotheses, let $q_i=mean(\mathcal{W})-w_i$ denote the gap between the average weighting score of vertices and the weight of $v_i$. Thus, as in~\cite{ferraz2007density}, we can compute the {prior probability} $p_i$ of the vertex $v_i$ by normalizing $q_i$ as follows:
\begin{align}
\label{equ:probability}
 p_i &=\left\{ \begin{array}    {r@{\quad \quad} l}
\frac{q_i}{\sum_{j=1}^m q_j}, & if~q_i>0,\\
{\xi}~~~~,&otherwise,
\end{array}\right.
\end{align}
{where $\xi$ denotes a small positive value to avoid zero division.}
Then we can obtain the entropy of the {prior probability}, which is used as an adaptive threshold to {select} significant vertices:
 \begin{align}
\label{equ:entropy}
{E}= -\sum_{i=1}^mp_i\log p_i.
\end{align}
Finally, we retain the vertices with {higher quantities} of information than the entropy ${E}$, and the retained vertices $\mathcal{V}$ are defined as:
 \begin{align}
\label{equ:condition}
\mathcal{V}=\{v_i|-\log p_i>E\}.
\end{align}

Note that \cite{wang2012simultaneously} also applies the information theoretic approach to select significant hypotheses. However, in \cite{wang2012simultaneously}, the gap between the weight of {a model hypothesis} and the maximum weight is used to select significant hypotheses. However, {such a} strategy may remove not only many bad model hypotheses but also some good model hypotheses {having less number of inliers}. In contrast, the proposed method removes less significant model hypotheses than \cite{wang2012simultaneously} and retains more good ones. Thus, it is more effective than the method used in \cite{wang2012simultaneously} for removing insignificant model hypotheses while {preserving} significant ones.
\subsection{{The Mode-seeking Algorithm}}
\label{sec:algorihmanalysis}
{Given a hypergraph $G^*$, we aim to seek modes by searching for the authority peaks that correspond to model instances in data. The ``authority peaks" in a hypergraph can be defined as follows.}

\textbf{Definition 1}~~\emph{Authority peaks are the vertices that have the local maximum values of weighting scores in the hypergraph.}

{The vertices that have the local maximum values of weighting scores correspond to the modes in a hypergraph. Here, ``local" refers to the neighbors of a vertex in a hypergraph. } This definition is consistent with the conventional concept of modes, which are defined as the significant peaks of the density distribution in the parameter space~\cite{comaniciu2002mean,xu1990new,cho2012mode}.

Inspired by \cite{rodriguez2014clustering}, where each cluster center is characterized by two attributes (i.e., a higher local density than {its} neighbors and a relatively large distance from any point that has higher densities to the cluster center itself), we search for {the} authority peaks, {which are the vertices that are surrounded by their neighbors with the lower local weighting scores, and are significantly dissimilar to any other vertices that have higher local weighting scores}.

{To describe the relationships between two vertices in a hypergraph, we propose an effective similarity measure}  based on the Tanimoto distance \cite{tanimoto1957internal} (referred to as T-distance), {which is able to effectively} measure the degree of overlap between two hyperedge sets connected by the two vertices. {Given two vertices $v_p$ and $v_q$, their T-distance is computed as: }
{
\begin{equation}
\label{equ:tdistance}
\mathcal{T}(\bm{\mathcal{C}}_{v_p},\bm{\mathcal{C}}_{v_q})
=1-\frac{\langle\bm{\mathcal{C}}_{v_p},\bm{\mathcal{C}}_{v_q}\rangle}{\|\bm{\mathcal{C}}_{v_p}\|^2+\|\bm{\mathcal{C}}_{v_q}\|^2-\langle\bm{\mathcal{C}}_{v_p},\bm{\mathcal{C}}_{v_q}\rangle},
\end{equation}
where} $\langle\cdot,\cdot\rangle$ and $\|\cdot\|$ indicate the standard inner product and the corresponding induced norm, respectively. {$\bm{\mathcal{C}}_{v_p}$ and $\bm{\mathcal{C}}_{v_q}$ denote the preference function of $v_p$ and $v_q$ to hyperedges $\mathcal{E}$, respectively.}

We define the preference function {(see Section 2.1 in \cite{Magri_2014_CVPR}) of a vertex $v_p$ to a hyperedge $e\in\mathcal{E}$ as}
\begin{align}
\label{equ:preferencefunction1}
  \mathcal{C}_{v_p}^e &=\left\{ \begin{array}    {r@{\quad \quad} l}
\exp\{-\frac{r_e(v_p)}{\hat{s}(v_p)}\}, & if~r_e(v_p)\leq E\hat{s}(v_p),\\
0~~~~~~~~~~~,&otherwise,
\end{array}\right.
\end{align}
where $E$ is a threshold (the value of $E$ is usually set to 2.5 to include $98\%$ inliers of a Gaussian distribution). {The preference function is used to compute a rank of a vertex that indicates the degree of preference of the vertex to a hyperedge, where the most preferred hyperedge is ranked in the top (a high value), and the least preferred hyperedge is ranked in the last (a low value).}
Note that the preference function of each vertex can be effectively expressed by Eq.~(\ref{equ:preferencefunction1}), which takes advantages of the information of residuals of data points with regard to model hypotheses. {That is, a vertex prefers to {hyperedges that correspond} to a data point with a small absolute residual.}

Considering a hypergraph, we rewrite Eq.~(\ref{equ:preferencefunction1}) {for the preference function of each vertex $v_p$ to hyperedges $\mathcal{E}$ as
\begin{align}
\label{equ:preferencefunction}
\bm{\mathcal{C}}_{v_p}=h(v_p,e)\exp\{-\frac{r_e(v_p)}{\hat{s}(v_p)}\}, {\forall e\in\mathcal{E}}.
\end{align}}

Although \cite{Magri_2014_CVPR} also employs the T-distance as a similarity measure, the T-distance defined in this paper has significant differences: 1) {We define the preference function of  a vertex (i.e., a model hypothesis) towards a hyperedge set (i.e., the inliers), while the authors in~\cite{Magri_2014_CVPR} define the preference function of a data point towards model hypotheses. Thus, the similarity between model hypotheses can be more effectively discovered by the corresponding {preference functions} in our case.} 2) The T-distance used in the proposed method is calculated without using iterative processes. In contrast, the T-distance in \cite{Magri_2014_CVPR} is iteratively calculated until an agglomerative clustering algorithm segments all data points. Therefore, the T-distance used in this paper is much more efficient than that in \cite{Magri_2014_CVPR}.
\begin{figure*}[t]
\centering
\begin{minipage}{.16\textwidth}
\centerline{\includegraphics[width=1.0\textwidth]{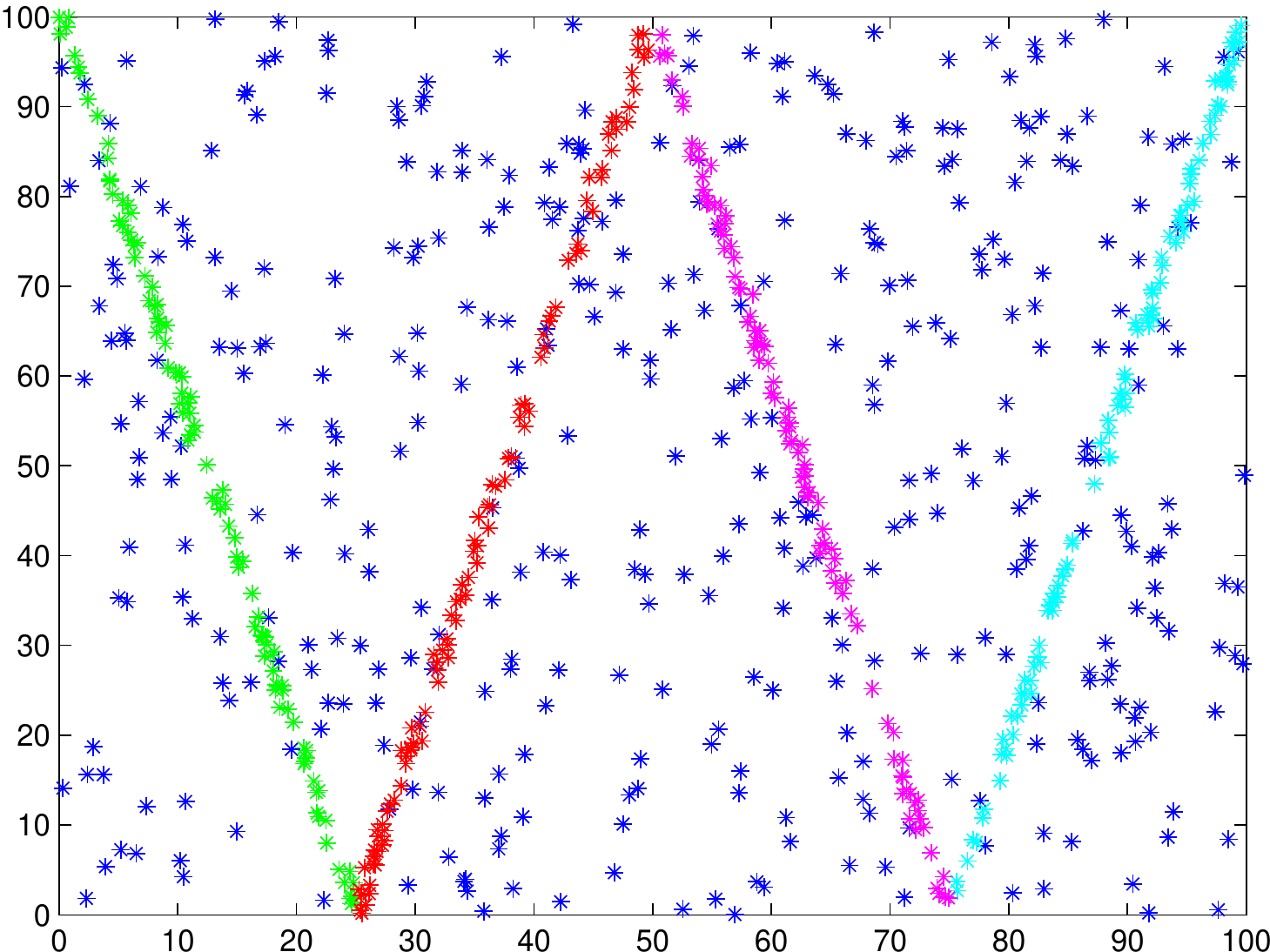}}
 \centerline{(a)}
\end{minipage}
\begin{minipage}{.16\textwidth}
\centerline{\includegraphics[width=1.0\textwidth]{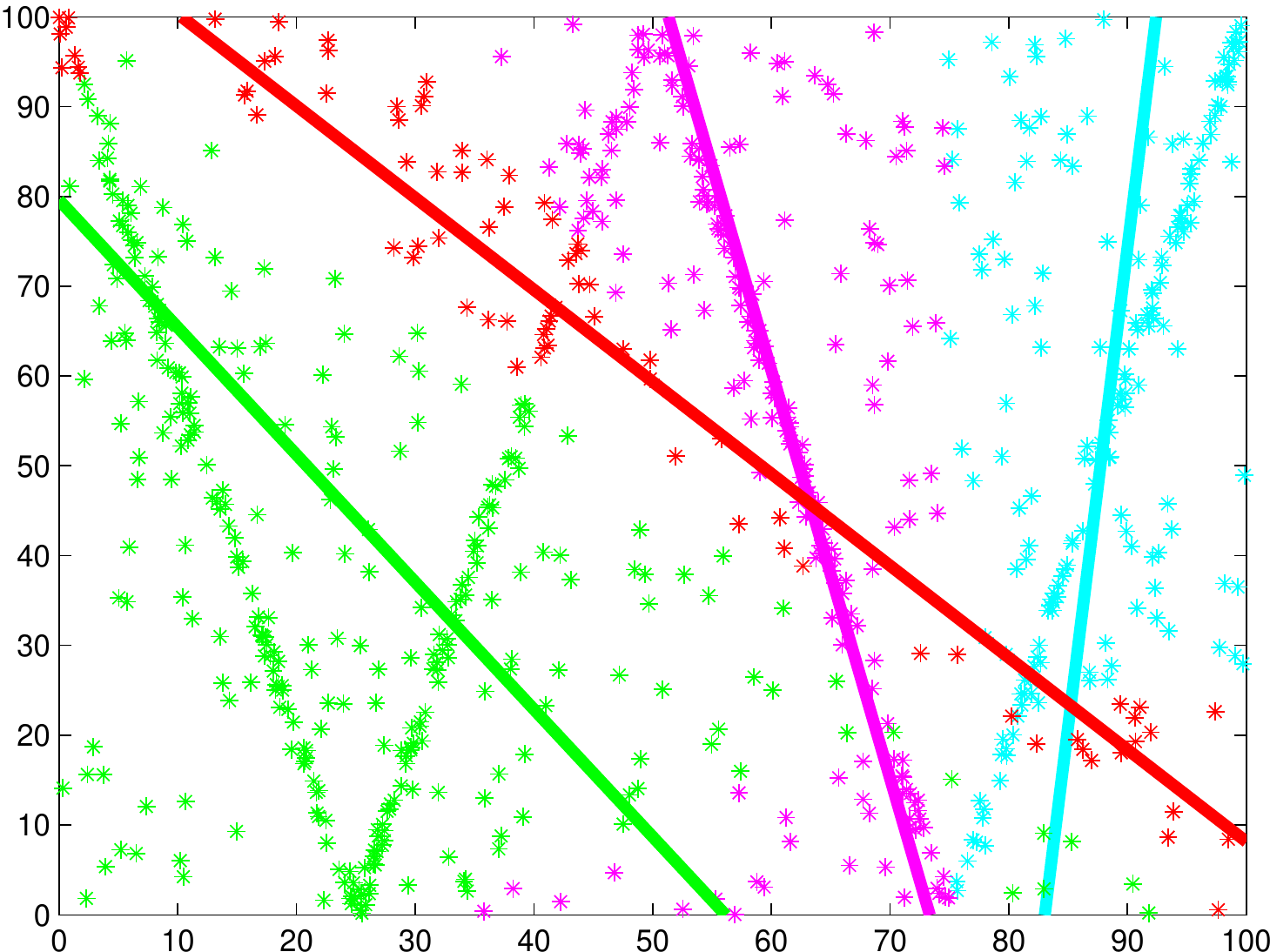}}
 \centerline{(b)}
\end{minipage}
\begin{minipage}{.16\textwidth}
\centerline{\includegraphics[width=1.0\textwidth]{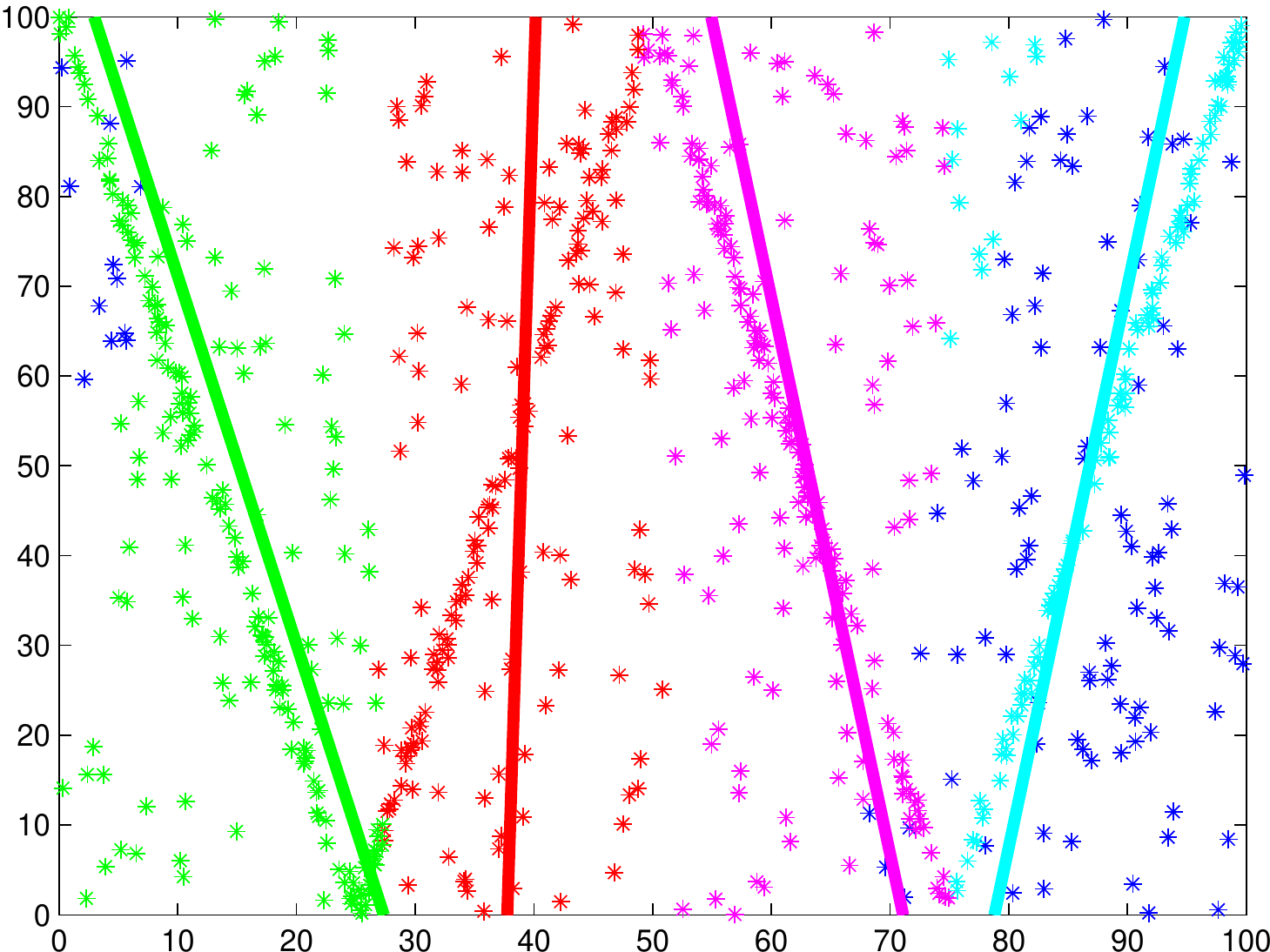}}
 \centerline{(c)}
\end{minipage}
\begin{minipage}{.16\textwidth}
 \centerline{\includegraphics[width=1.0\textwidth]{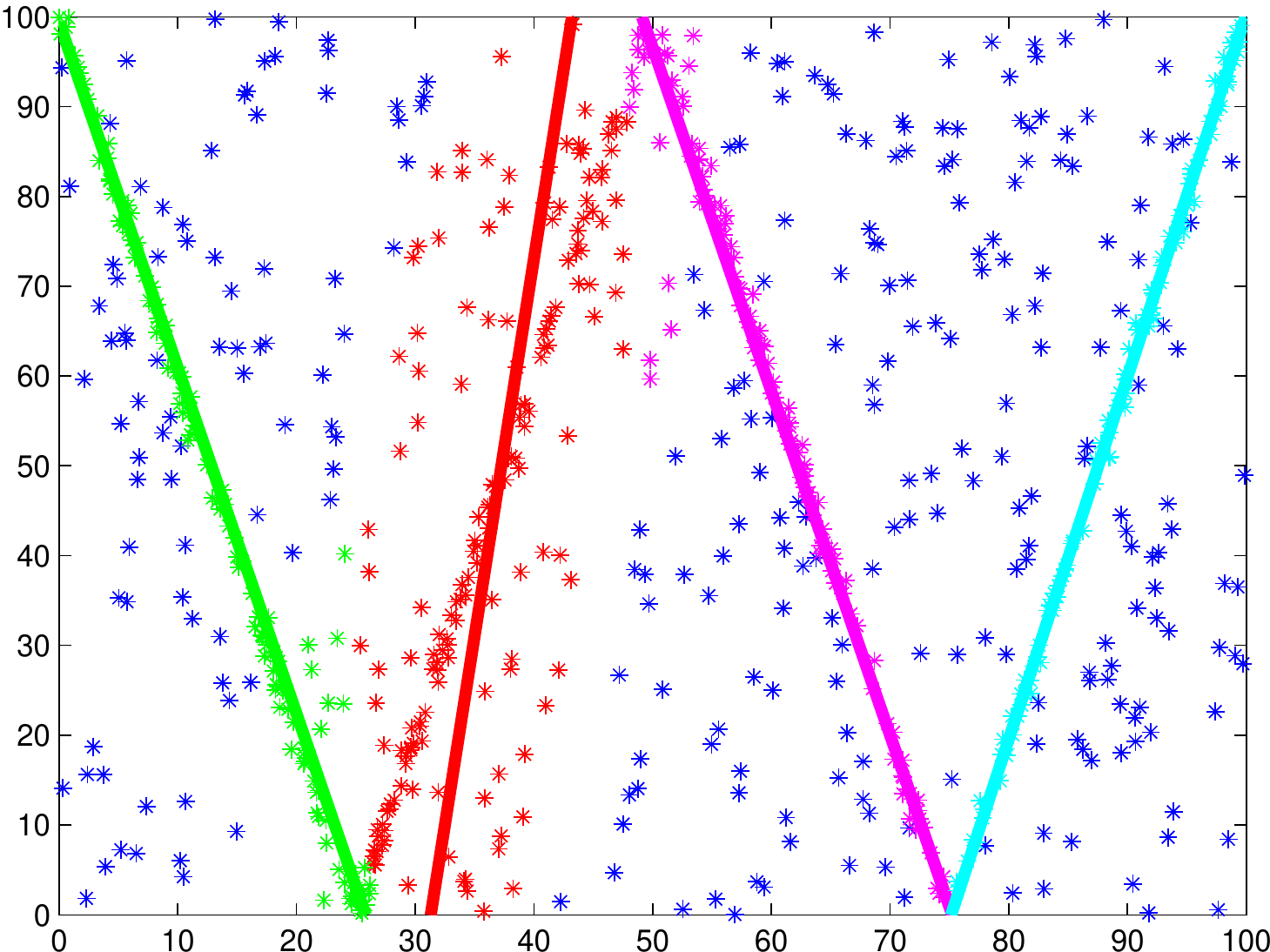}}
 \centerline{(d)}
\end{minipage}
\begin{minipage}{.16\textwidth}
 \centerline{\includegraphics[width=1.0\textwidth]{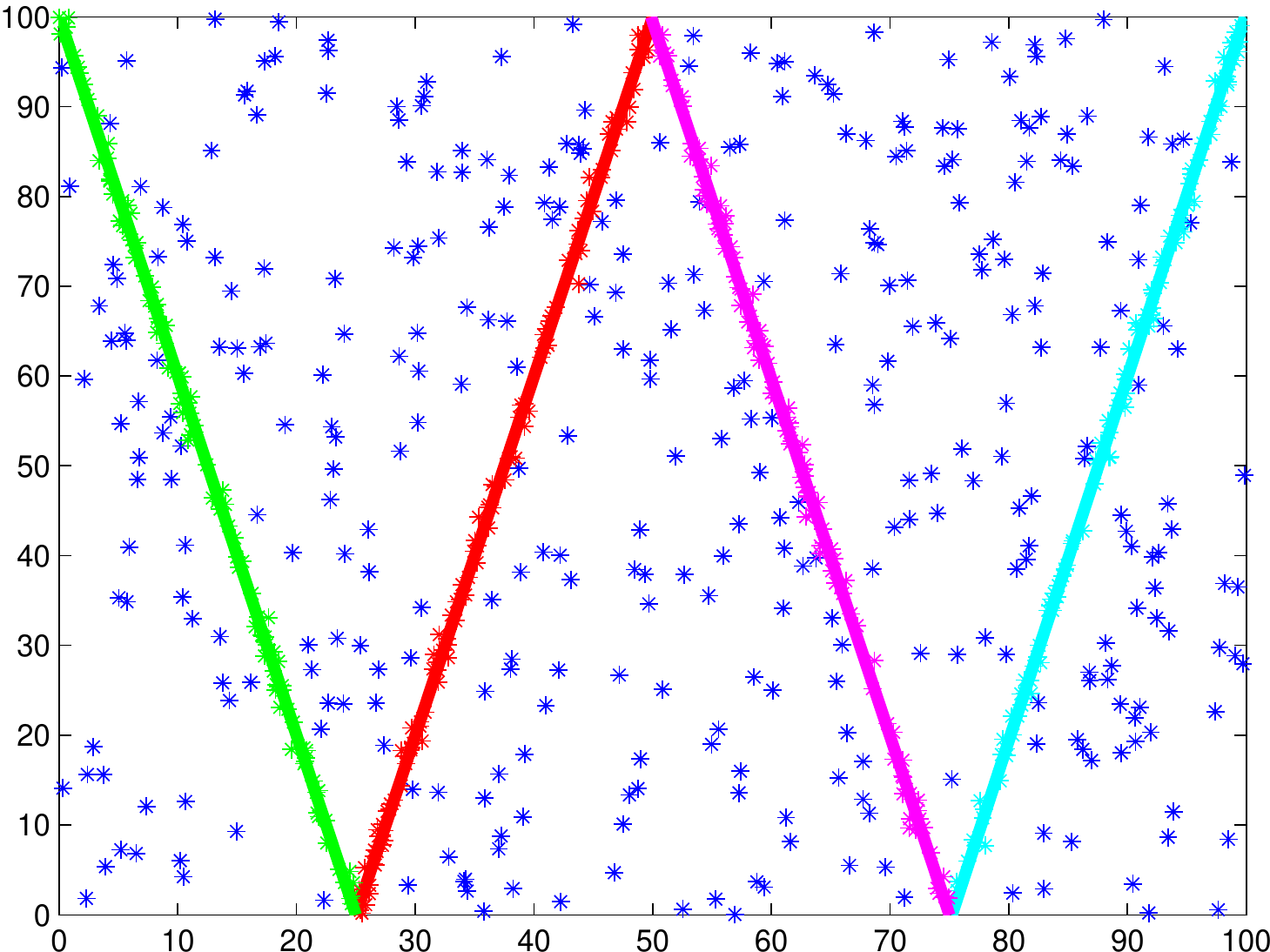}}
 \centerline{(e)}
\end{minipage}
\begin{minipage}{.16\textwidth}
 \centerline{\includegraphics[width=1.0\textwidth]{Ncut_degree_ours-eps-converted-to.pdf}}
 \centerline{(f)}
\end{minipage}
\caption{{Some results obtained by NCut based on different hypergraphs for line fitting. (a) The input data. The data points with blue color are outliers, and the other data points with the same color belong to the inliers of the same model instance. (b) to (e) The results obtained by NCut based on the uniform hypergraphs with three, six, nine and twelve degrees, respectively. (f) The results obtained by NCut based on the proposed non-uniform hypergraph.}}
\label{fig:ncut}
\end{figure*}

Based on the similarity measure and weighting scores, {we then compute the Minimum T-Distance (MTD) $\eta^{v_i}_{min}$ of a vertex $v_i$ in $G^*$ as follows:
\begin{align}
\label{equ:minimumtdistance}
\eta^{v_i}_{min}=\min_{v_j\in \Omega(v_i)}\{\mathcal{T}(\bm{\mathcal{C}}_{v_i},\bm{\mathcal{C}}_{v_j})\},
\end{align}
where
\begin{align}
\Omega(v_i)=\{v_j\big|w(v_j)>w(v_i),v_j\in\mathcal{N}(v_i)\},
\end{align}
\begin{align}
\label{equ:neighbor}
\mathcal{N}(v_i)=\{v_j\big| \frac{\sum_{e\in \mathcal{E}} h(v_i,e)h(v_j,e)}{\sum_{e\in \mathcal{E}} (h(v_i,e)+h(v_j,e))}>\epsilon, {v_j}\in \mathcal{V}\}.
\end{align}
That is, $\Omega(v_i)$ contains all the vertices that have higher weighting scores than $w(v_i)$ in the neighbors of $v_i$. $\mathcal{N}(v_i)$ contains the neighbors of $v_i$ in $G^*$.
$ \frac{\sum_{e\in \mathcal{E}} h(v_i,e)h(v_j,e)}{\sum_{e\in \mathcal{E}} (h(v_i,e)+h(v_j,e))}$ denotes the ratio of the common hyperedges connected by two vertices ($v_i$ and $v_j$). We fix $\epsilon=0.8$, which means that two vertices share at least $80\%$ of common hyperedges in a hypergraph.
For the vertex $v_{max}$ with the highest weighting score, we set $\eta^{v_{max}}_{min}=\max\{\mathcal{T}(\bm{\mathcal{C}}_{v_{max}},\bm{\mathcal{C}}_{v_i})\}_{v_i\in\mathcal{N}(v_{max})}$.}

Note that a vertex with the local maximum value of weighting score, {usually} has a larger MTD value than the other vertices in $G^*$. Therefore, we propose to seek modes by searching for the authority peaks, i.e., the vertices with significantly large MTD values.
\section{The Complete Method and Analysis}
\label{sec:completealgorithm}
 {Based on the components described in the previous section, we present the complete fitting method in Sec.~\ref{sec:completealgorithm2}. We also analyze how MSHF is able to perform well for the model fitting problem in Sec.~\ref{sec:analysis}.}
\subsection{The Complete Method}
\label{sec:completealgorithm2}
 {We summarize the proposed Mode-Seeking on Hypergraphs Fitting (MSHF) method in Algorithm \ref{alg:MSH}}. The proposed MSHF seeks modes by directly {searching for} authority peaks (i.e., representative modes) without requiring iterative processes. As a result, {the number} and the parameters of model instances can be {simultaneously} derived from the detected modes.

 The computational complexity of MSHF is mainly governed by Step \ref{alg:state:t-distance} of the algorithm for computing the T-distance between pairs of vertices (here, we do not consider the time for generating model hypotheses since we focus on model selection for model fitting). The other steps in MSHF take much less time than Step \ref{alg:state:t-distance}. { For Step \ref{alg:state:t-distance}, the complexity of computing the neighbors of each vertex and the T-distance between all pairs of vertices are  $O(M\log M)$ and $O(MM')$, respectively. Here, $M$ is the number of vertices in $G^*$ ($M$ is empirically about $15\%\sim 30\%$ of the total number of vertices in $G$), and $M'~(>>\log M)$ is the average number of the neighbors of vertices in $G$. Therefore, the total complexity of MSHF approximately amounts to $O(MM')$.}
\begin{algorithm} 
\renewcommand{\algorithmicrequire}{\textbf{Input:}}
\renewcommand\algorithmicensure {\textbf{Output:} }
\caption{The mode-seeking on hypergraphs fitting (MSHF) method for geometric model fitting} 
\label{alg:MSH} 
\begin{algorithmic}[1] 
\REQUIRE 
Data points $X$, the $K$ value for IKOSE
\STATE Construct a hypergraph $G$ and compute the weighting score for each vertex (described in Sec.~\ref{sec:problemsetting}).
\STATE  Use the information theoretic approach for hypergraph reduction and generate a new hypergraph $G^*$ (described in Sec.~\ref{sec:hypergraphreduction}).
\STATE Compute the minimum T-distance $\eta^v_{min}$ for {each vertex ${v}$ of $G^*$ } by Eq.~(\ref{equ:minimumtdistance}).
\label{alg:state:t-distance}
\STATE Sort the vertices in $G^*$ according to their MTD values satisfying $\eta^{v_1}_{min}\geq \eta^{v_2}_{min} \geq \cdots$.
\STATE Find the vertex $v_i$ whose MTD value ($\eta^{v_i}_{min}$) has the largest drop from $\eta^{v_i}_{min}$ to $\eta^{v_i+1}_{min}$ and reject the vertices whose values of $\eta^v_{min}$ are smaller than $\eta^{v_i}_{min}$.
\STATE Derive the inliers/outliers dichotomy from the hypergraph $G^*$ and the remaining vertices (modes).
\ENSURE The modes (model instances) and the hyperedges (inliers) connected by the modes.
\end{algorithmic}
\end{algorithm}
\subsection{How Does The Proposed Method Find Representative Modes on Hypergraphs}
\label{sec:analysis}
\begin{figure*}[t]
\centering
\begin{minipage}{.25\textwidth}
\centerline{\includegraphics[width=1.0\textwidth]{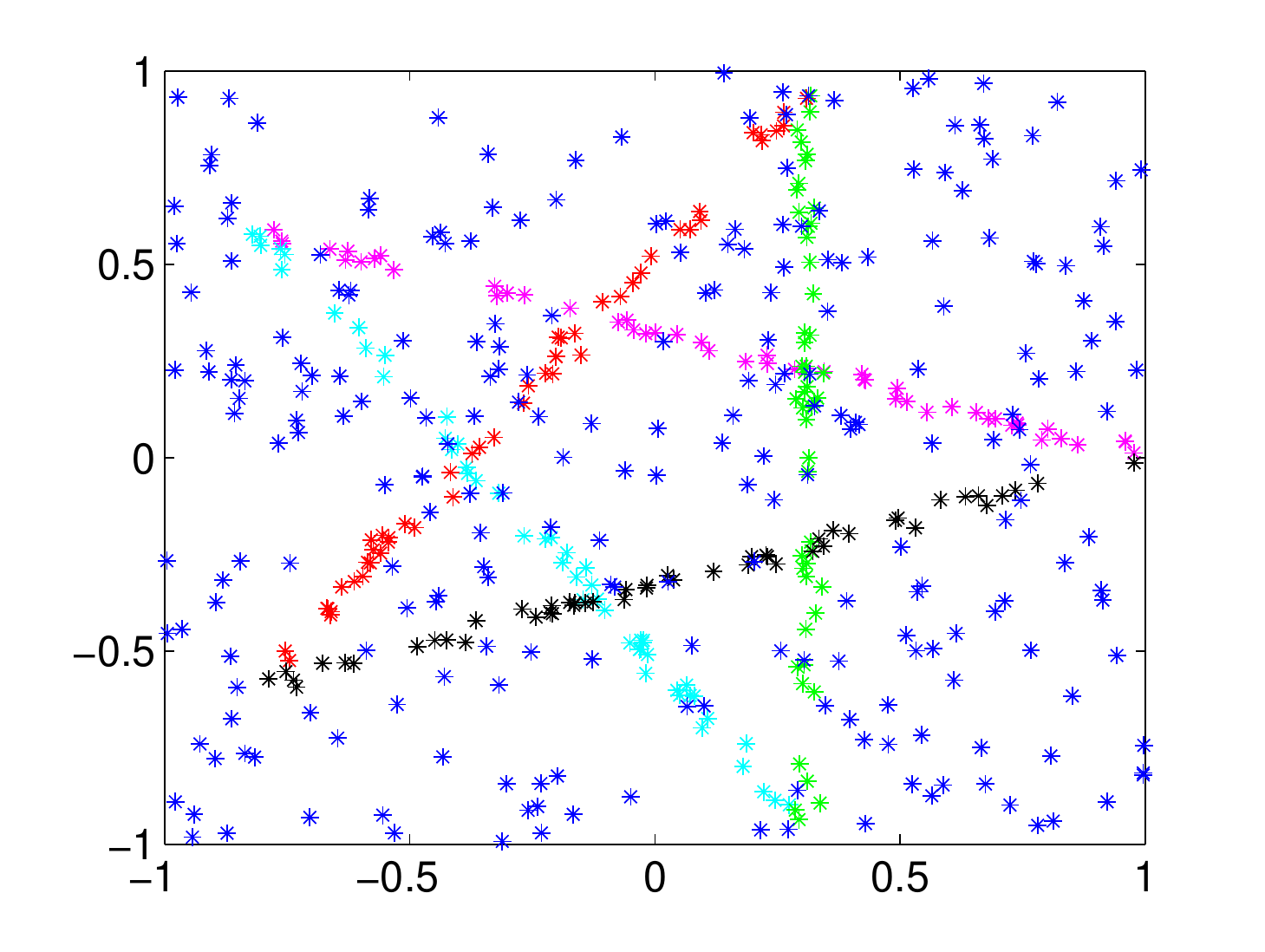}}
 \centerline{(a)}
\end{minipage}
\begin{minipage}{.31\textwidth}
\centerline{\includegraphics[width=1.00\textwidth]{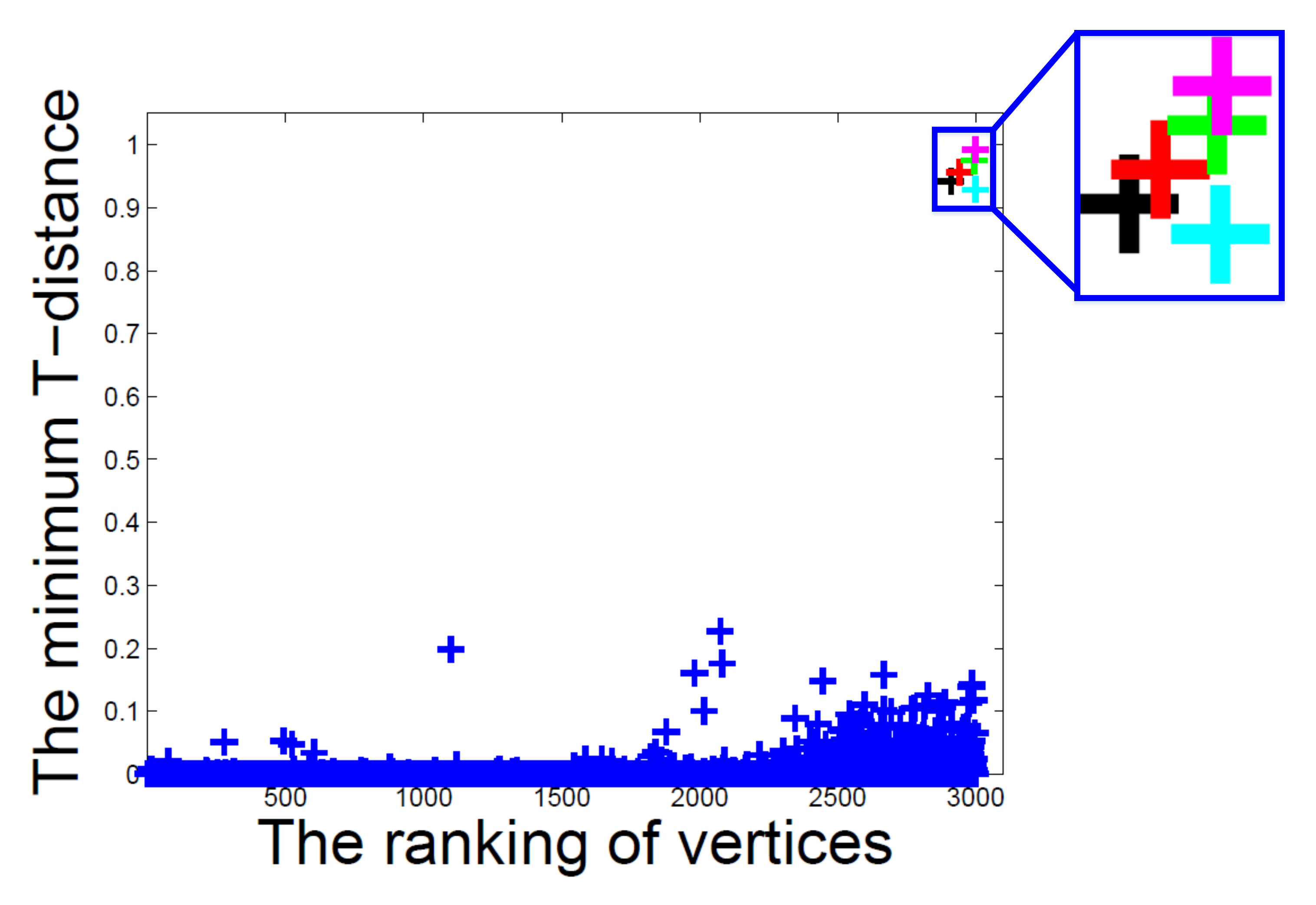}}
 \centerline{(b)}
\end{minipage}
\begin{minipage}{.25\textwidth}
\centerline{\includegraphics[width=1.0\textwidth]{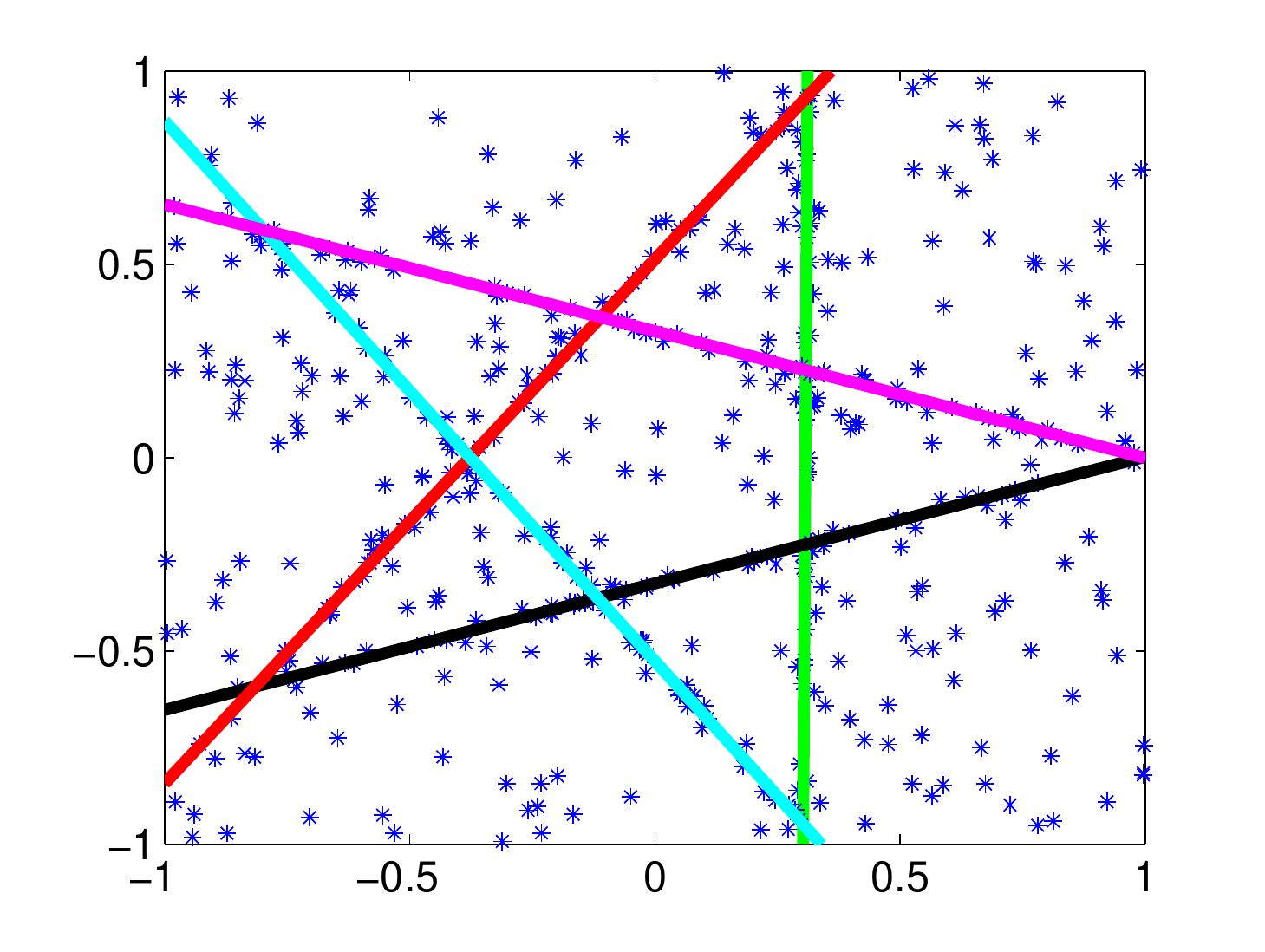}}
 \centerline{(c)}
\end{minipage}
\caption{An example shows that MSHF fits the five lines on {the ``star5" data}. (a) The input data. The data points with blue color are outliers, and the other data points with the same {given} color belong to the inliers of the same model instance. (b) The obtained decision graph. The vertices are ranked according to {their} weighting scores in non-decreasing order. The five vertices with the first five highest values of the minimum T-distance (shown in different colors {in the zoomed portion of the figure} except for blue) are the sought modes. (c) The five lines corresponding to the five sought modes.}
\label{fig:star5}
\end{figure*}
{MSHF includes three main parts: hypergraph construction, hypergraph reduction and mode-seeking. MSHF tightly combines these three parts, by which it can effectively search for representative modes {on hypergraphs for model fitting}.

For the hypergraph construction, we construct a non-uniform hypergraph to represent the relationships between model hypotheses and data points. As mentioned in~\cite{pulak2014clustering}, we argue that using larger hyperedges is more effective for model fitting. In Fig.~\ref{fig:ncut}, we show some results obtained by NCut~\cite{zhou2007learning} based on several uniform hypergraphs with different degrees and one non-uniform hypergraph constructed by the proposed MSHF method for line fitting. From the results, we can see that NCut achieves better accuracy based on the hypergraphs with larger degrees than that based on the hypergraphs with smaller degrees. However, {how large to set the size of the hyperedges is still an unsolved problem and unaddressed in those most works.} {Recall that the proposed hypergraph construction can adaptively estimate the degree of each hyperedge. It is worth pointing out that, NCut tends to find a balanced cut, and it cannot effectively deal with unbalanced data.}
}

MSHF searches for representative modes on hypergraphs which shares the similar idea of detecting cluster centers in~\cite{rodriguez2014clustering}. Specifically, \cite{rodriguez2014clustering} computes the density of each data point and the minimum T-distance between the data point and any other data point with higher density, to detect cluster centers. Similarly, MSHF computes the weighting score $w(v)$ of each vertex in a hypergraph and the minimum T-distance (MTD) $\eta^v_{min}$ between the vertex and  {its neighbors with higher weighting scores}. We show an example of mode-seeking on hypergraphs for line fitting on {the ``star5" data} in Fig.~\ref{fig:star5}. We show the plot of $\eta^v_{min}$ {with respect to the weighting scores of vertices in non-decreasing order} in Fig.~\ref{fig:star5}(b), and this representation is called the decision graph. {MSHF} can find the representative modes according to the decision graph, and {then} estimate the model instances {in the data} (as shown in Fig.~\ref{fig:star5}(c)).

In \cite{rodriguez2014clustering}, cluster centers can be intuitively determined by the corresponding decision graph. However, it is nontrivial to detect cluster centers {in our case,} since some isolated data points also show large values of the minimum distance.  For the model fitting problem, the proposed mode-seeking algorithm works well for line fitting. This is because the distribution of model hypotheses generated for line fitting is dense in the parameter space, and there do not exist any isolated vertices (corresponding to bad model hypotheses with low weighting scores) showing large MTD values. However, the distribution of model hypotheses generated for higher-order model fitting applications, such as homography based segmentation or two-view based motion segmentation, is often sparse, {where} a few isolated vertices corresponding to bad model hypotheses may also {have} anomalously large MTD values as good model hypotheses (with high weighting scores). This problem will cause the proposed {mode-seeking} algorithm to work ineffectively.

\begin{figure}[t]
\centering
\begin{minipage}{.241\textwidth}
\centerline{\includegraphics[width=1.05\textwidth]{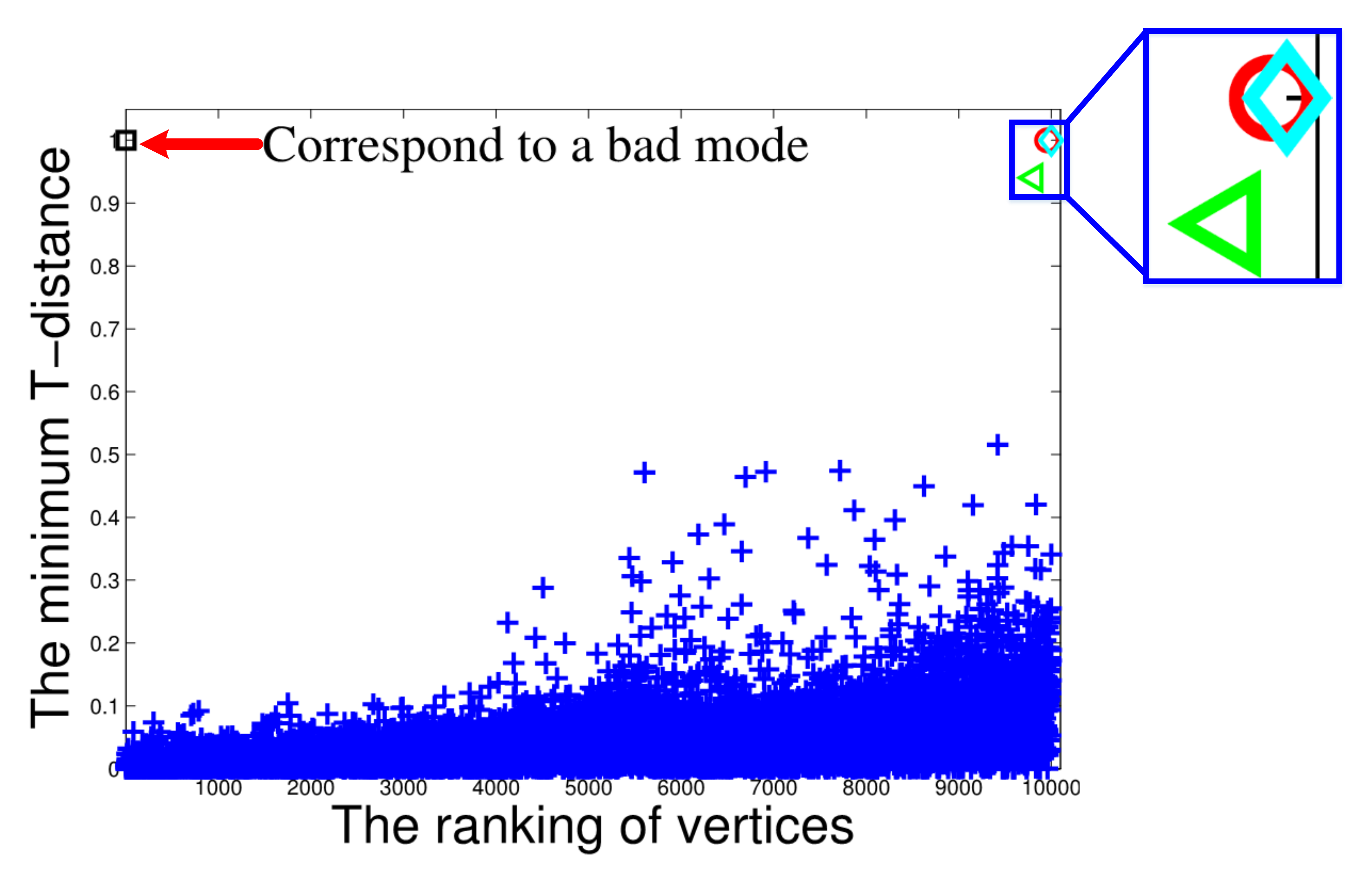}}
  \centerline{(a)}
  \end{minipage}
  \begin{minipage}{.241\textwidth}
\centerline{\includegraphics[width=1.05\textwidth]{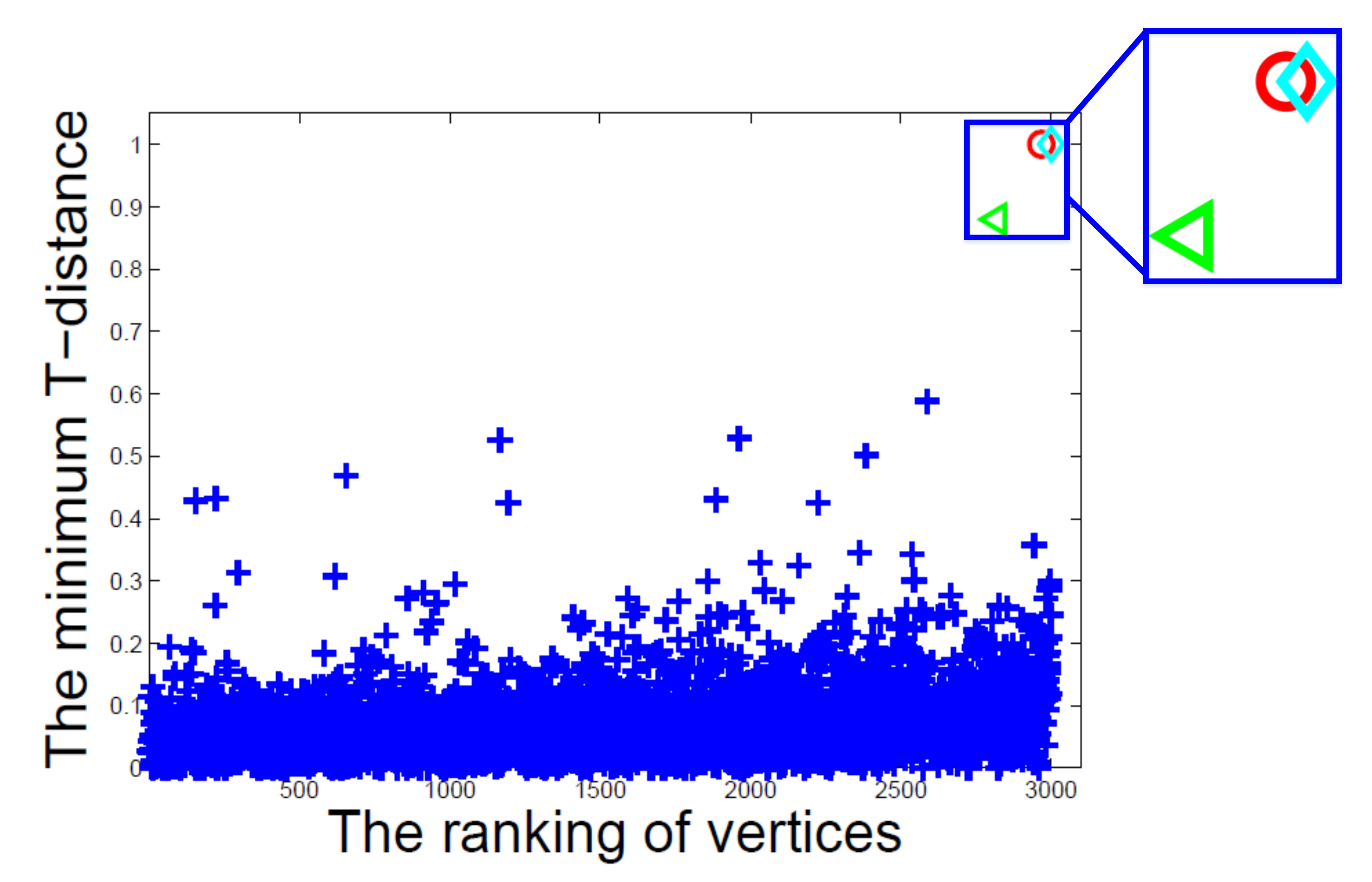}}
  \centerline{(b)}
  \end{minipage}
\begin{minipage}{.21\textwidth}
  \centerline{\includegraphics[width=1.00\textwidth]{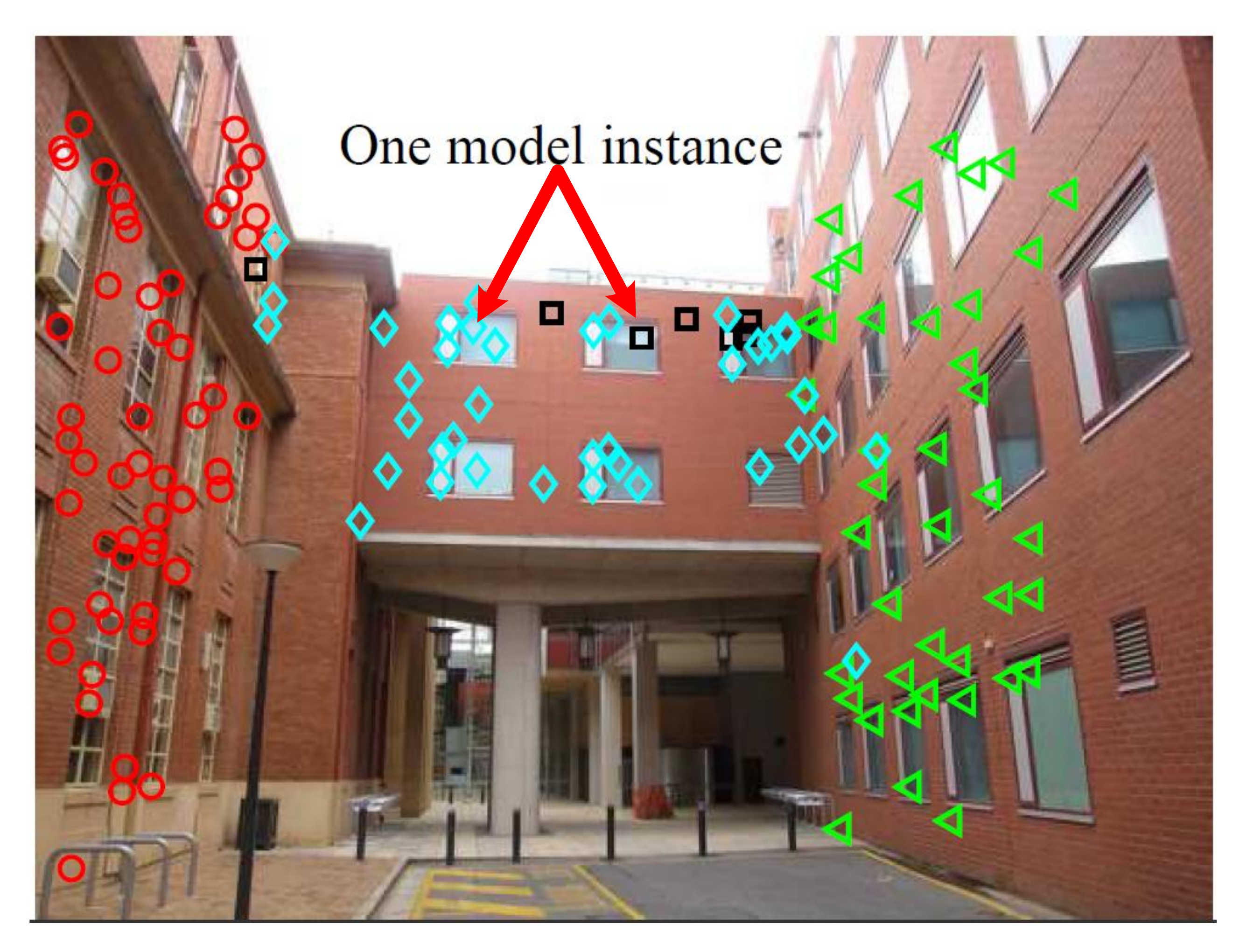}}
  \centerline{(c)}
\end{minipage}
\begin{minipage}{.03\textwidth}
~
\end{minipage}
\begin{minipage}{.21\textwidth}
  \centerline{\includegraphics[width=1.0\textwidth]{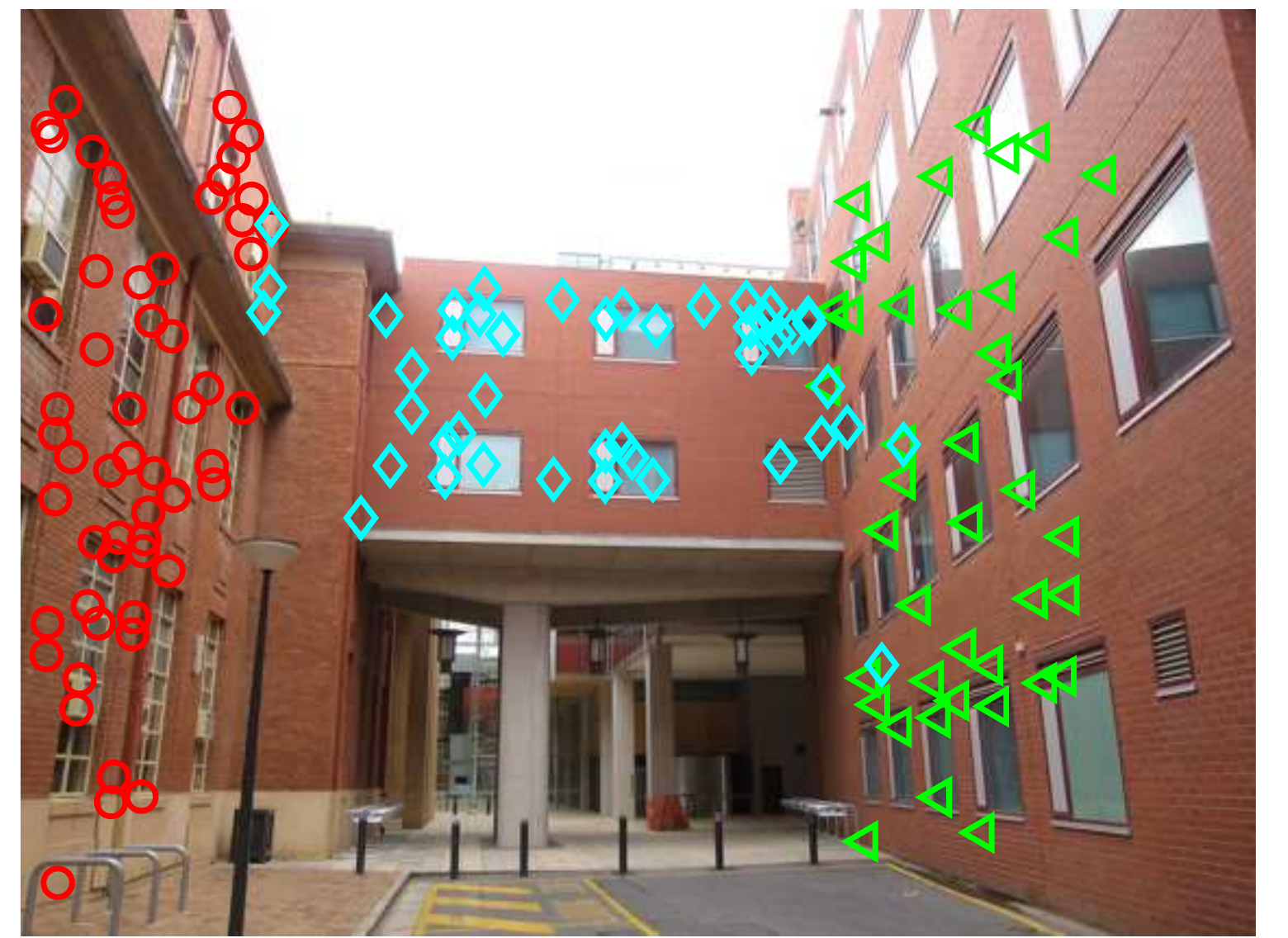}}
  \centerline{(d)}
\end{minipage}
\caption{Homography based segmentation on ``Neem''~\cite{wong2011dynamic}. (a) and (b) The decision graphs obtained by the proposed mode-seeking algorithm based on $G$ and $G^*$, respectively. (c) and (d) The segmentation results obtained by the proposed MSHF method based on $G$ and $G^*$, respectively.}
\label{fig:homotop10}
\end{figure}
To solve the above problem, we propose to remove some vertices corresponding to bad model hypotheses in Step 2 of Algorithm \ref{alg:MSH}. Therefore, hypergraph reduction (described in Sec.~\ref{sec:hypergraphreduction}) is a critical step to improve the effectiveness of the proposed MSHF algorithm. To show the importance of hypergraph reduction on the performance of the proposed mode-seeking algorithm, we evaluate the algorithm for fitting multiple homographies based on the two constructed hypergraphs, i.e., $G$ (the hypergraph without hypergraph reduction) and $G^*$ (the hypergraph with hypergraph reduction), as shown in Fig.~\ref{fig:homotop10}. We show the { obtained decision graphs in Fig.~\ref{fig:homotop10}(a) and \ref{fig:homotop10}(b), which respectively correspond to $G$ and $G^*$}. We can see that the proposed mode-seeking algorithm based on $G$ has difficulty {in distinguishing} the three significant model hypotheses according to {the} MTD values. This is because a vertex corresponding to {a bad} model hypothesis with a low weighting score also {has} a large MTD value {(as pointed by the arrow in Fig.~\ref{fig:homotop10}(a))}. Therefore, for the feature points of the middle wall (shown in cyan {in Fig.~\ref{fig:homotop10}(c)}), there are {two model instances estimated} by the proposed MSHF method based on $G$. In contrast, {the vertex corresponding to a bad model hypothesis is successfully removed by the step of hypergraph reduction, and} the proposed mode-seeking algorithm based on $G^*$ can correctly find all the three significant model hypotheses by seeking the largest drop in the MTD values. As shown in Fig.~\ref{fig:homotop10}(c) and \ref{fig:homotop10}(d), the segmentation results further show the {importance} of hypergraph reduction for the proposed MSHF method -- leading to more accurate results.

\begin{figure*}[t]
\centering
\begin{minipage}[t]{.145\textwidth}
  \centerline{\includegraphics[width=1.10\textwidth]{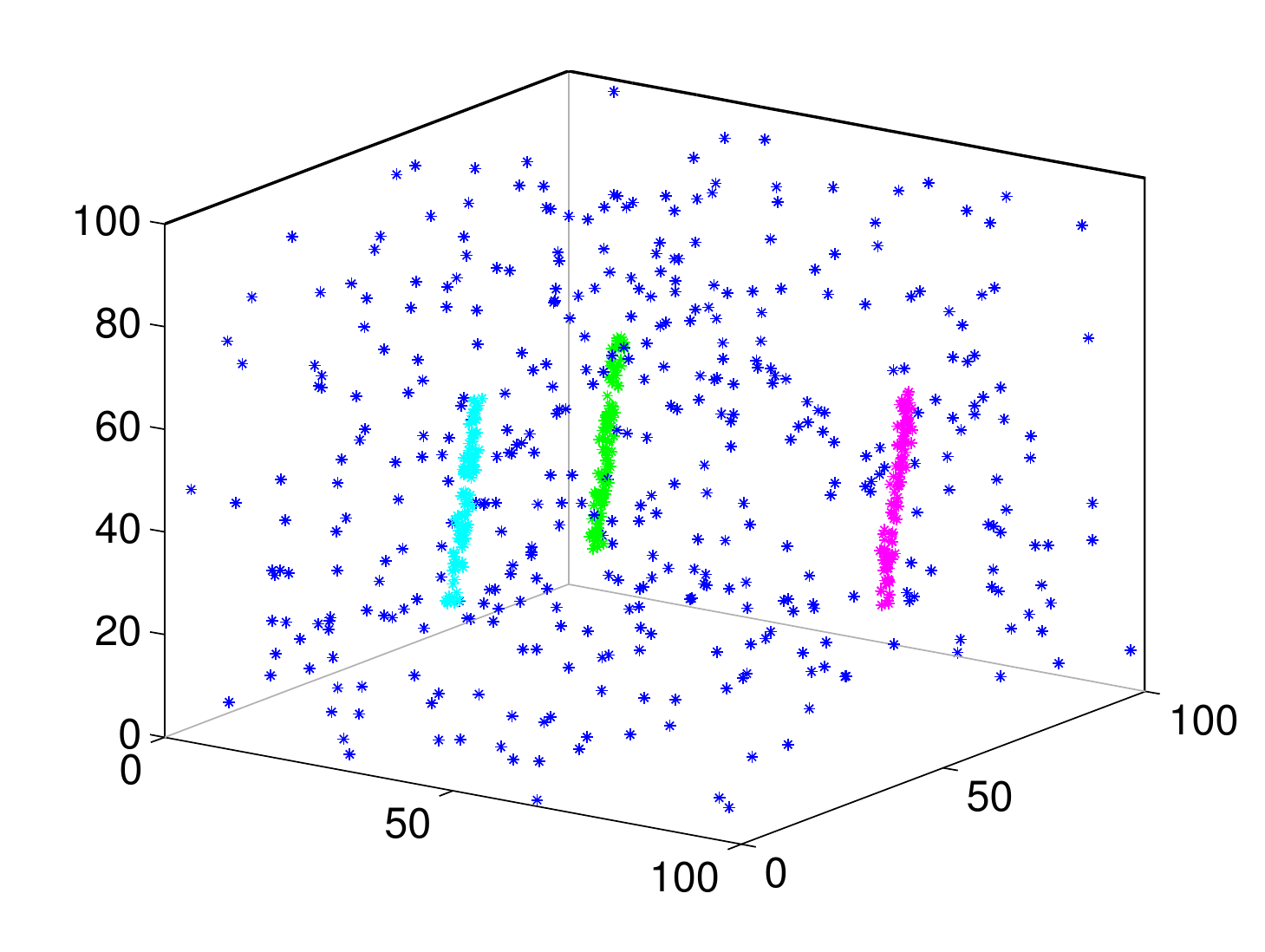}}
  \centerline{\includegraphics[width=1.100\textwidth]{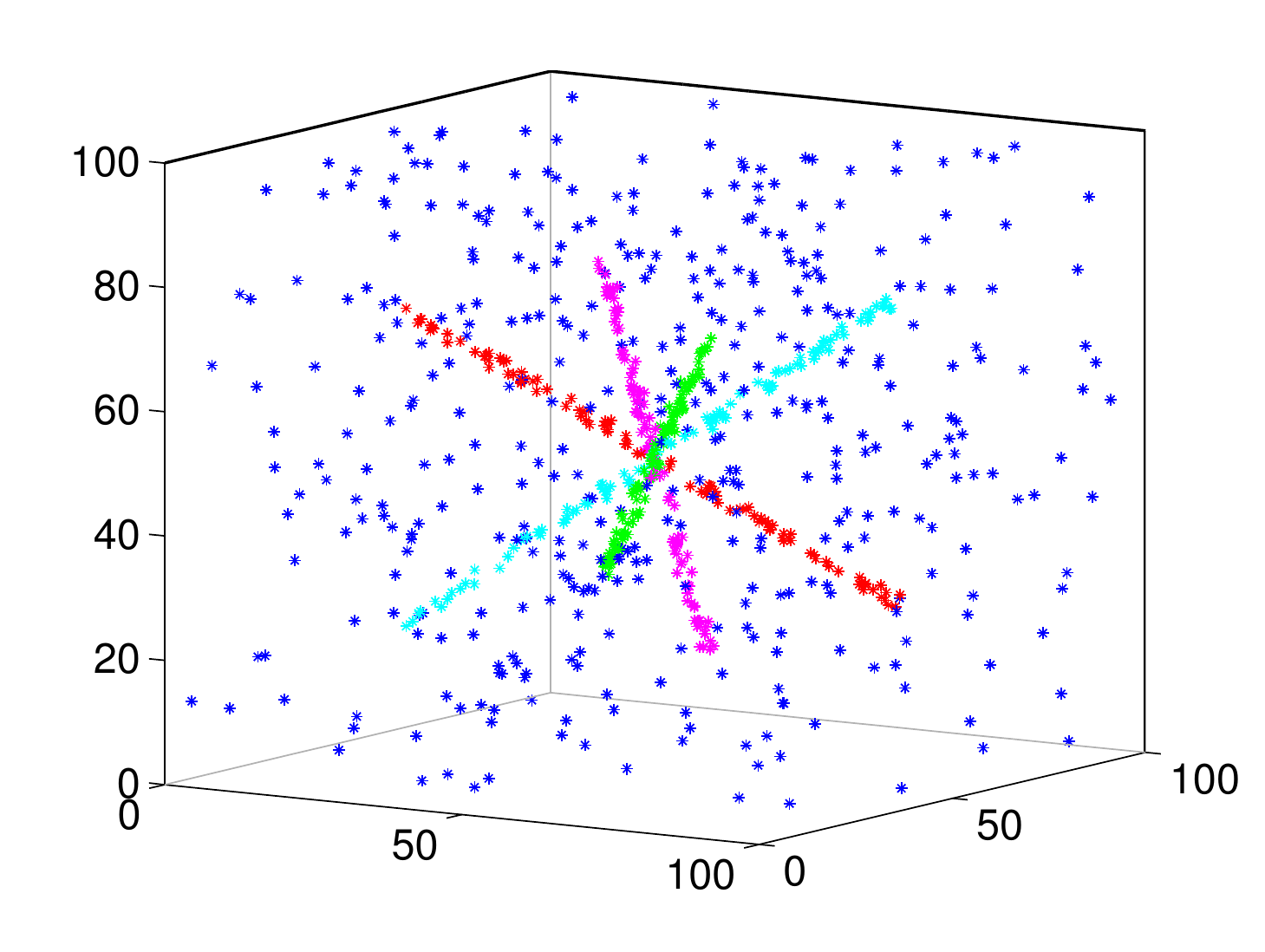}}
  \centerline{\includegraphics[width=1.100\textwidth]{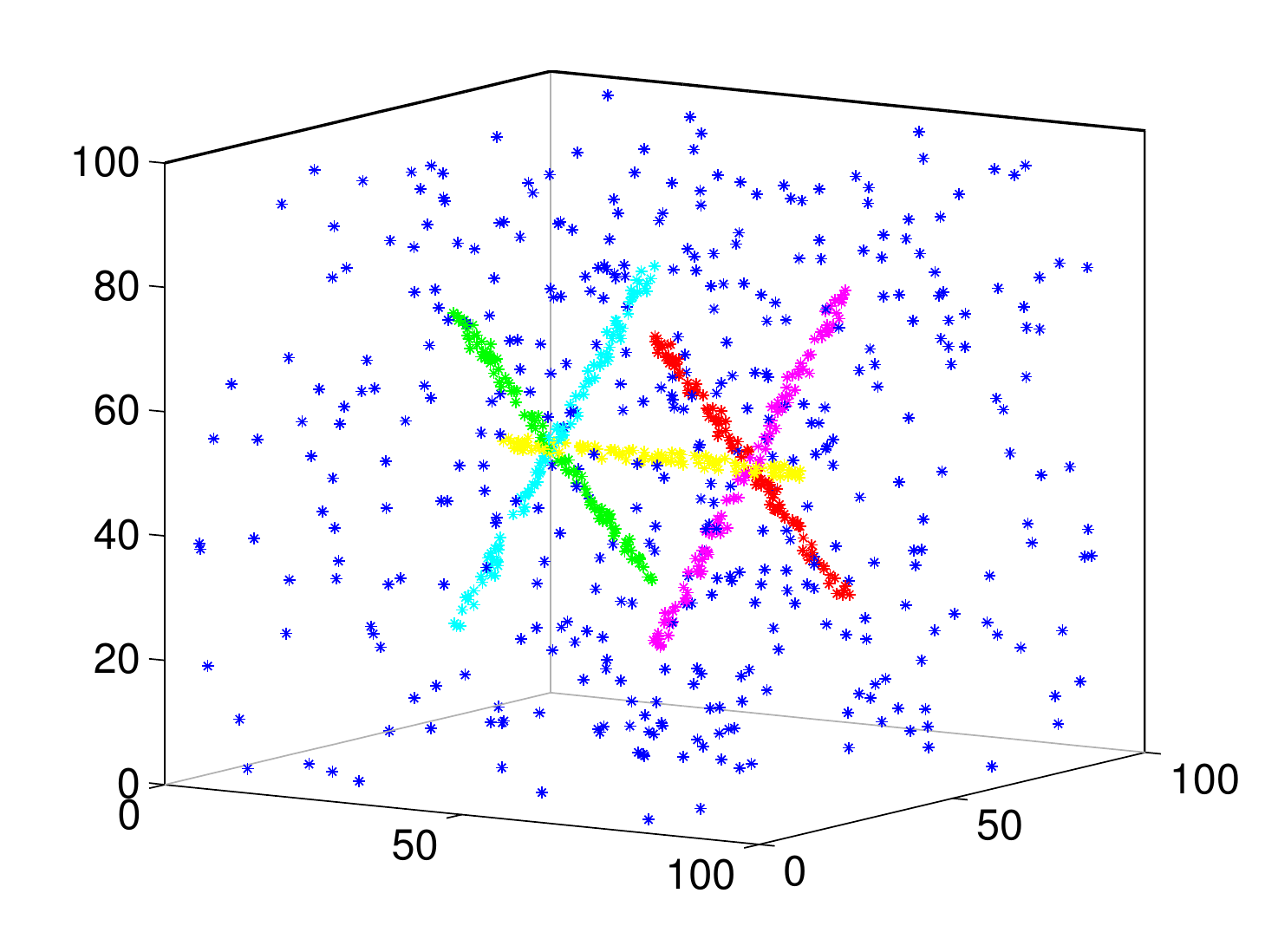}}
  \centerline{\includegraphics[width=1.100\textwidth]{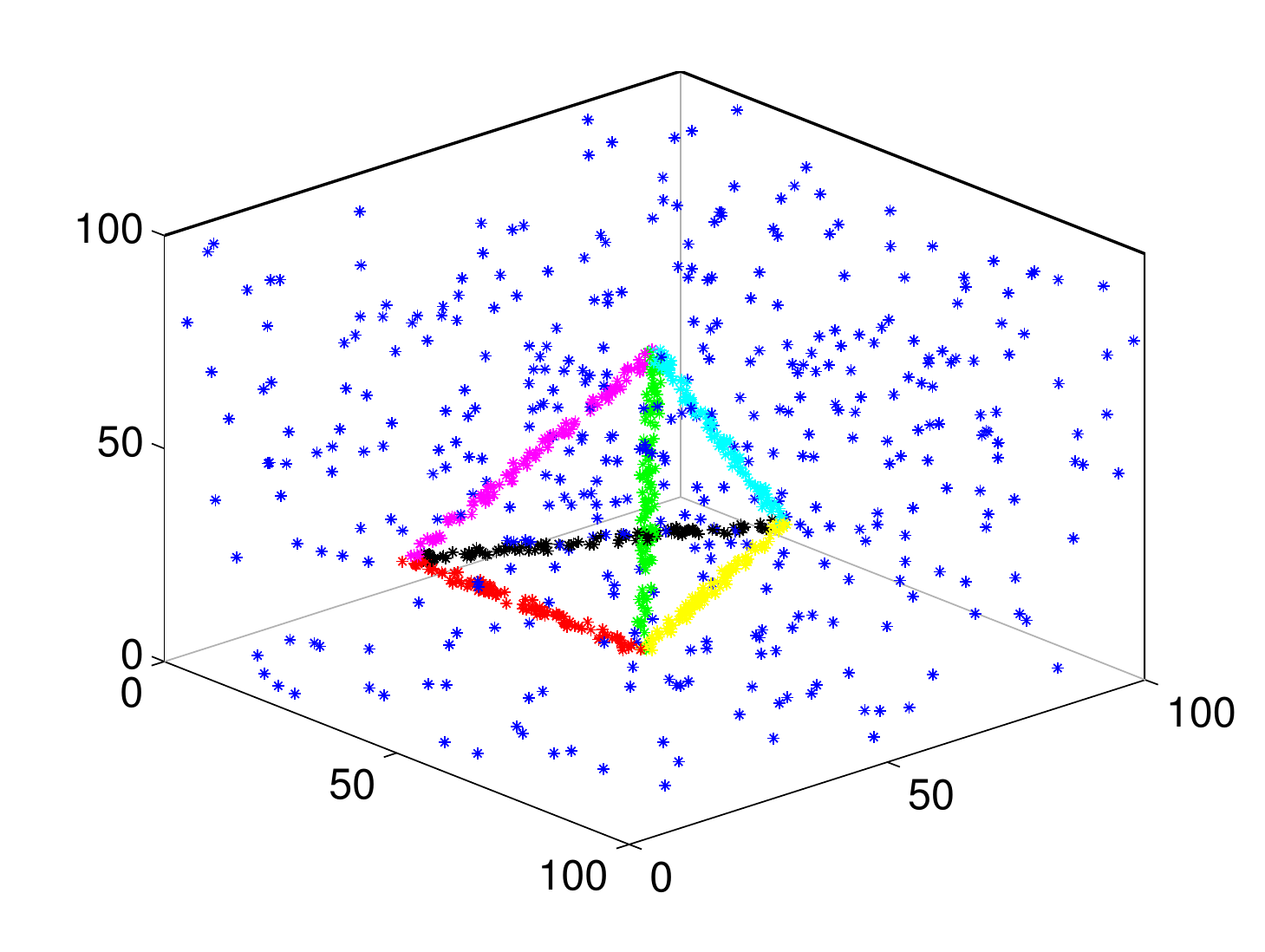}}
  \centerline{\footnotesize(a) Data }
\end{minipage}
\begin{minipage}[t]{.145\textwidth}

 \centerline{\includegraphics[width=1.10\textwidth]{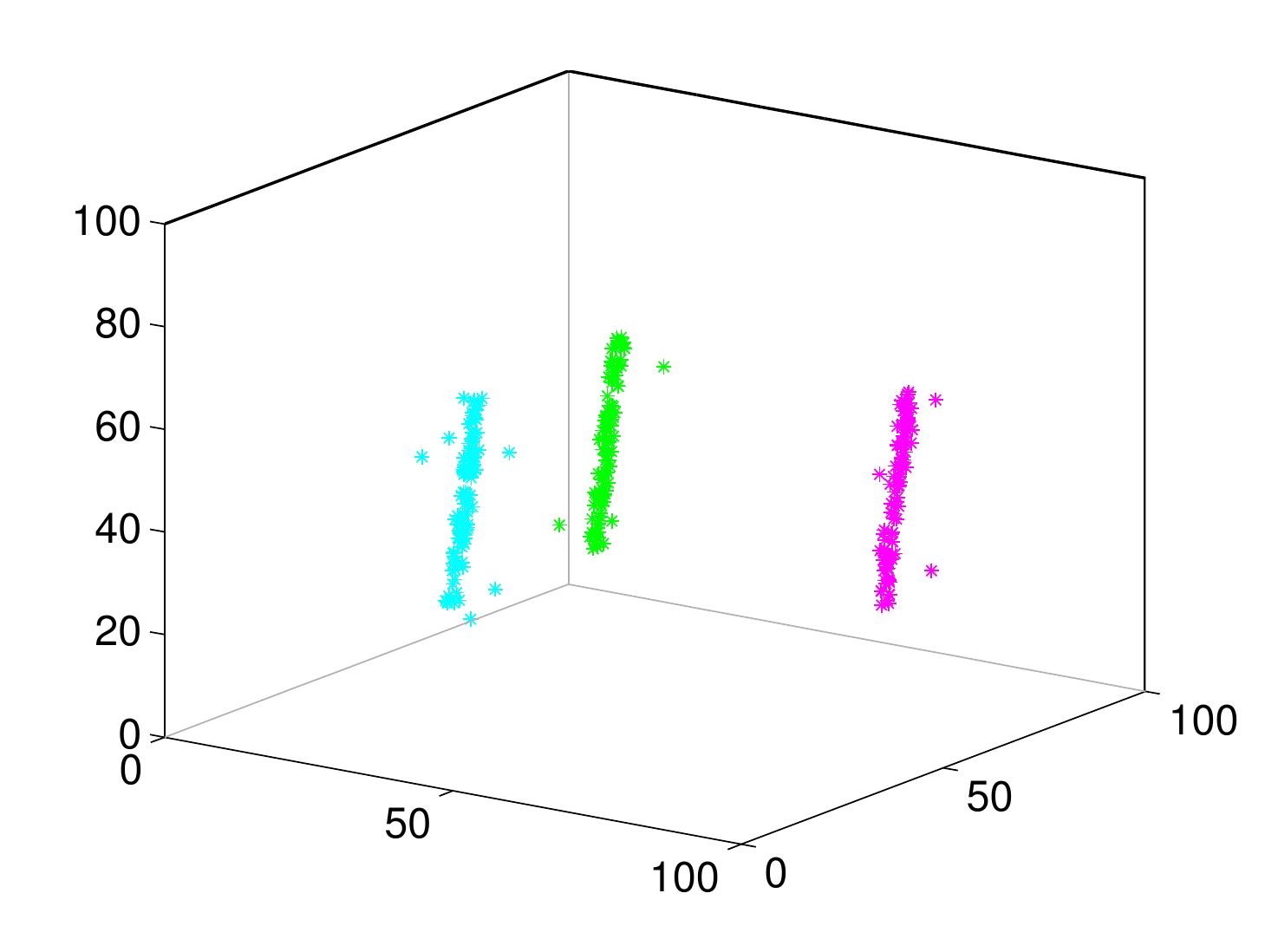}}
  \centerline{\includegraphics[width=1.100\textwidth]{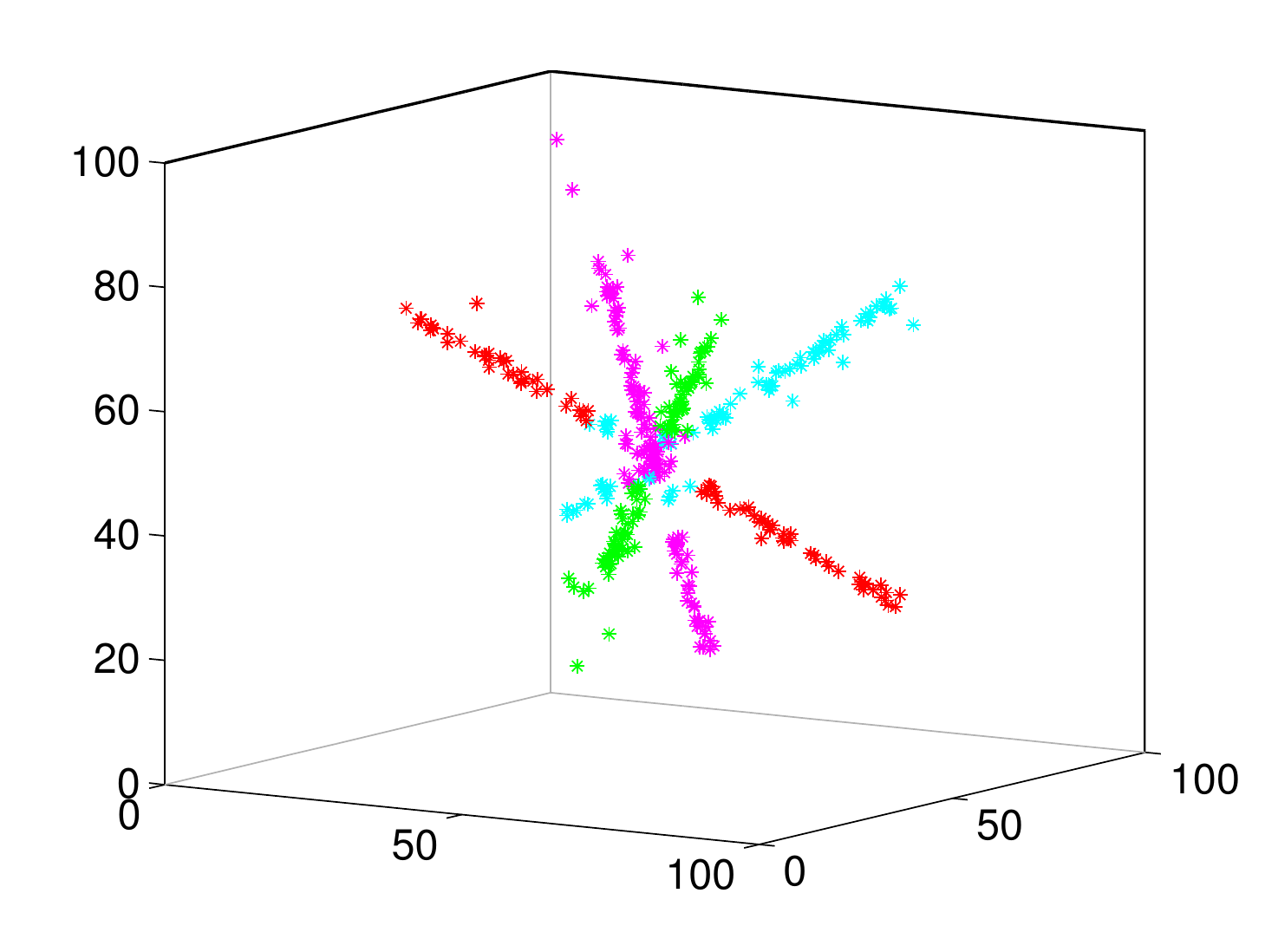}}
  \centerline{\includegraphics[width=1.10\textwidth]{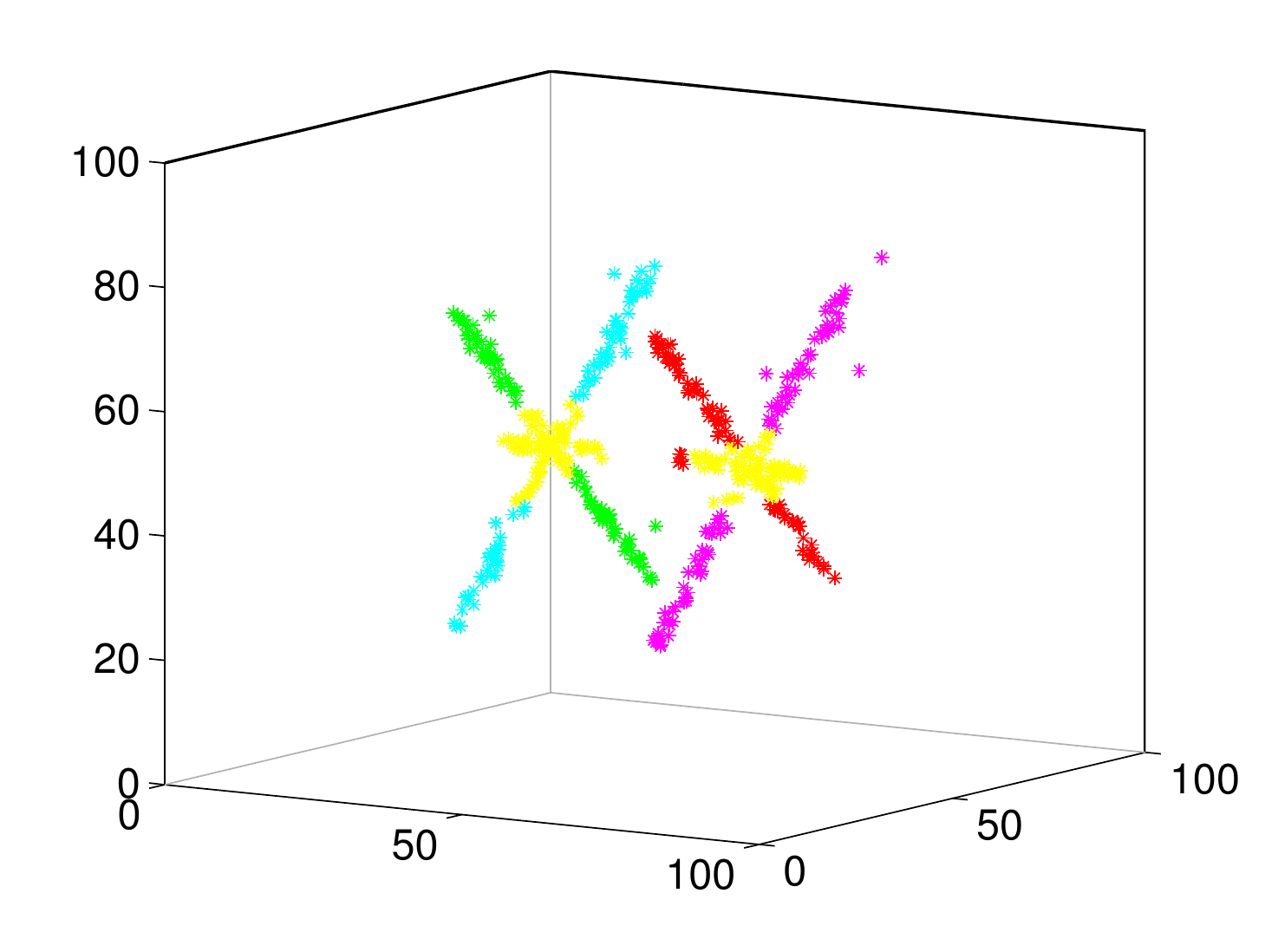}}
  \centerline{\includegraphics[width=1.10\textwidth]{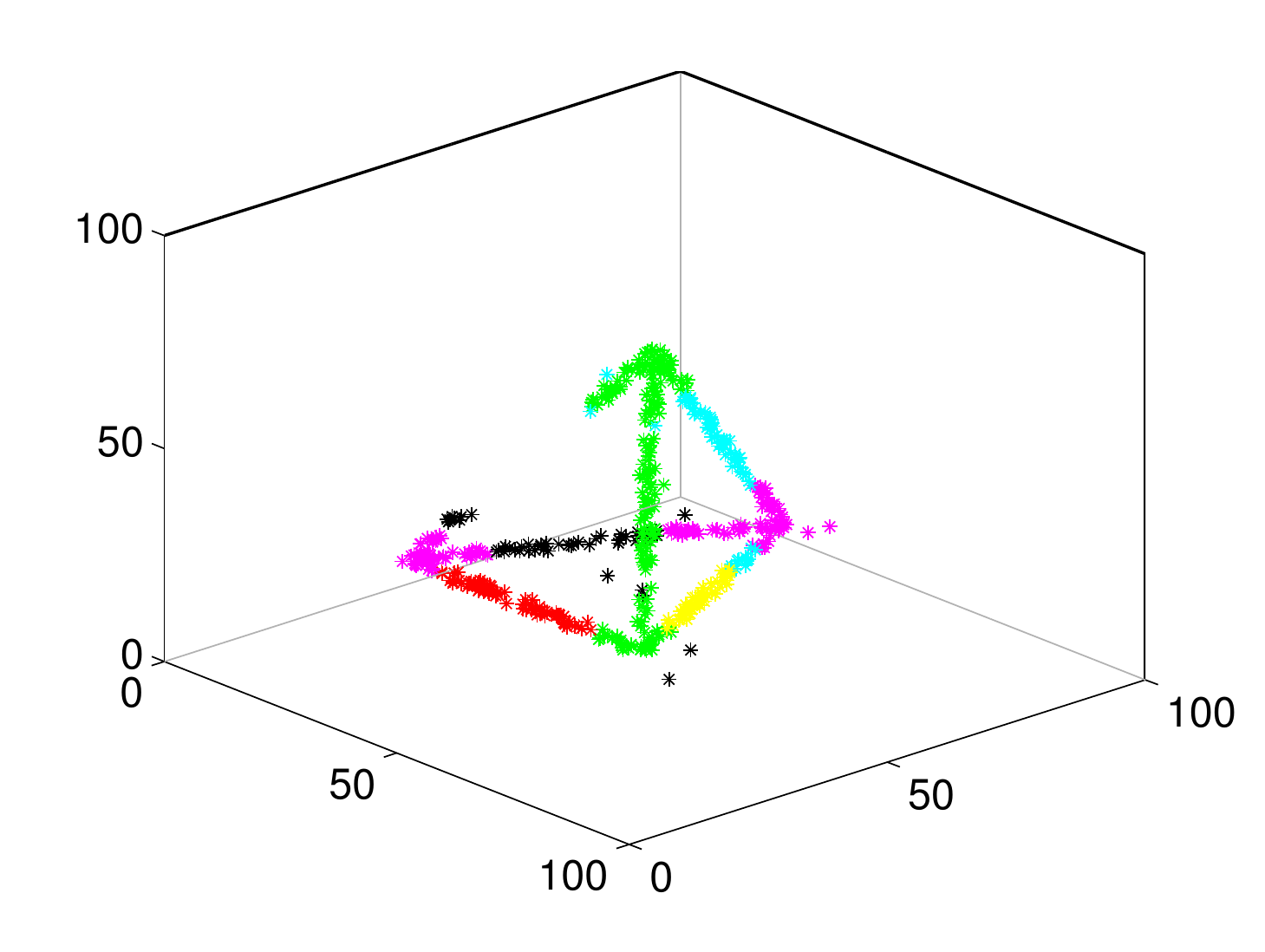}}
  \centerline{\footnotesize(b) KF }
\end{minipage}
\begin{minipage}[t]{.145\textwidth}
\centerline{\includegraphics[width=1.1\textwidth]{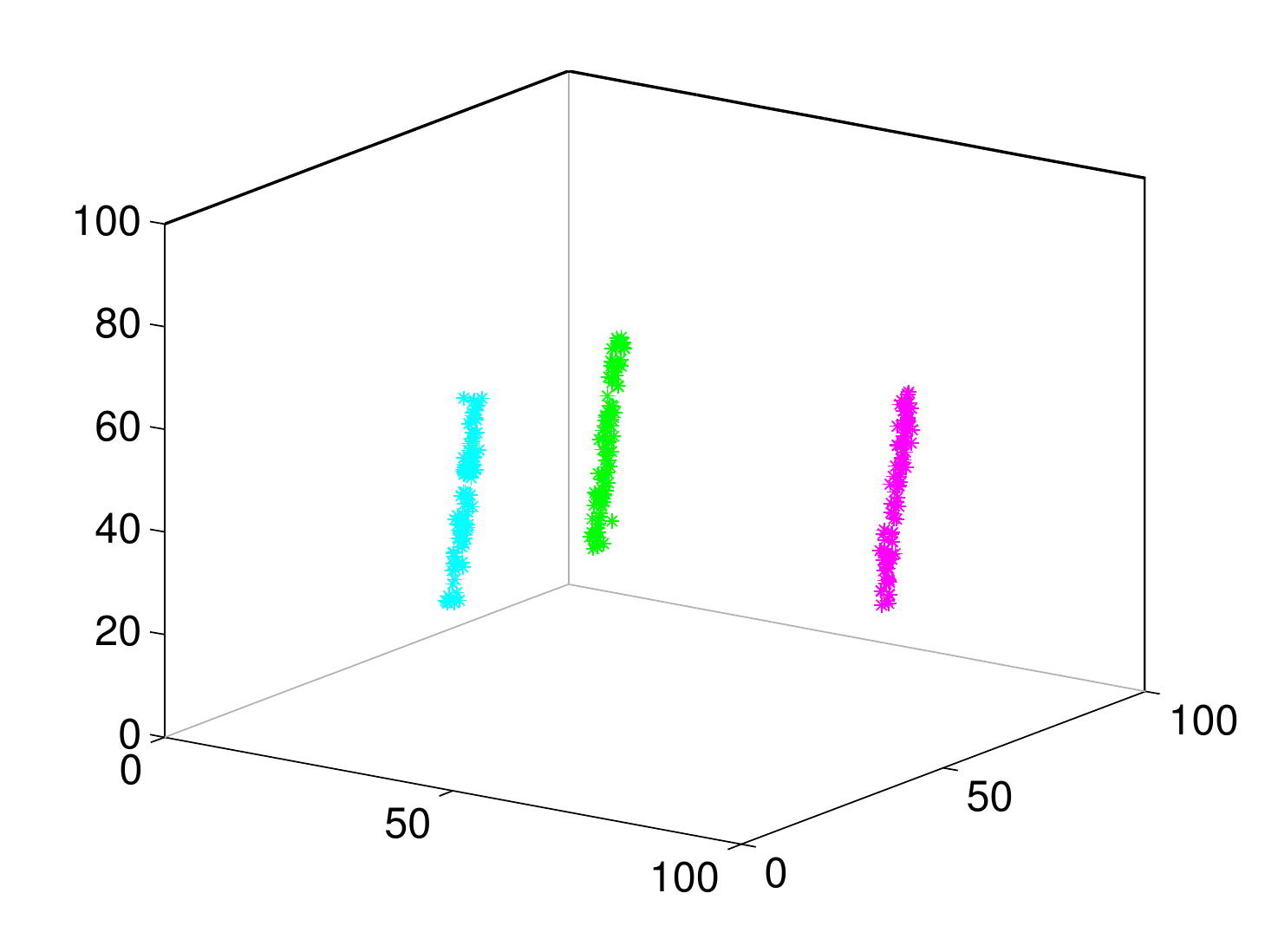}}
  \centerline{\includegraphics[width=1.10\textwidth]{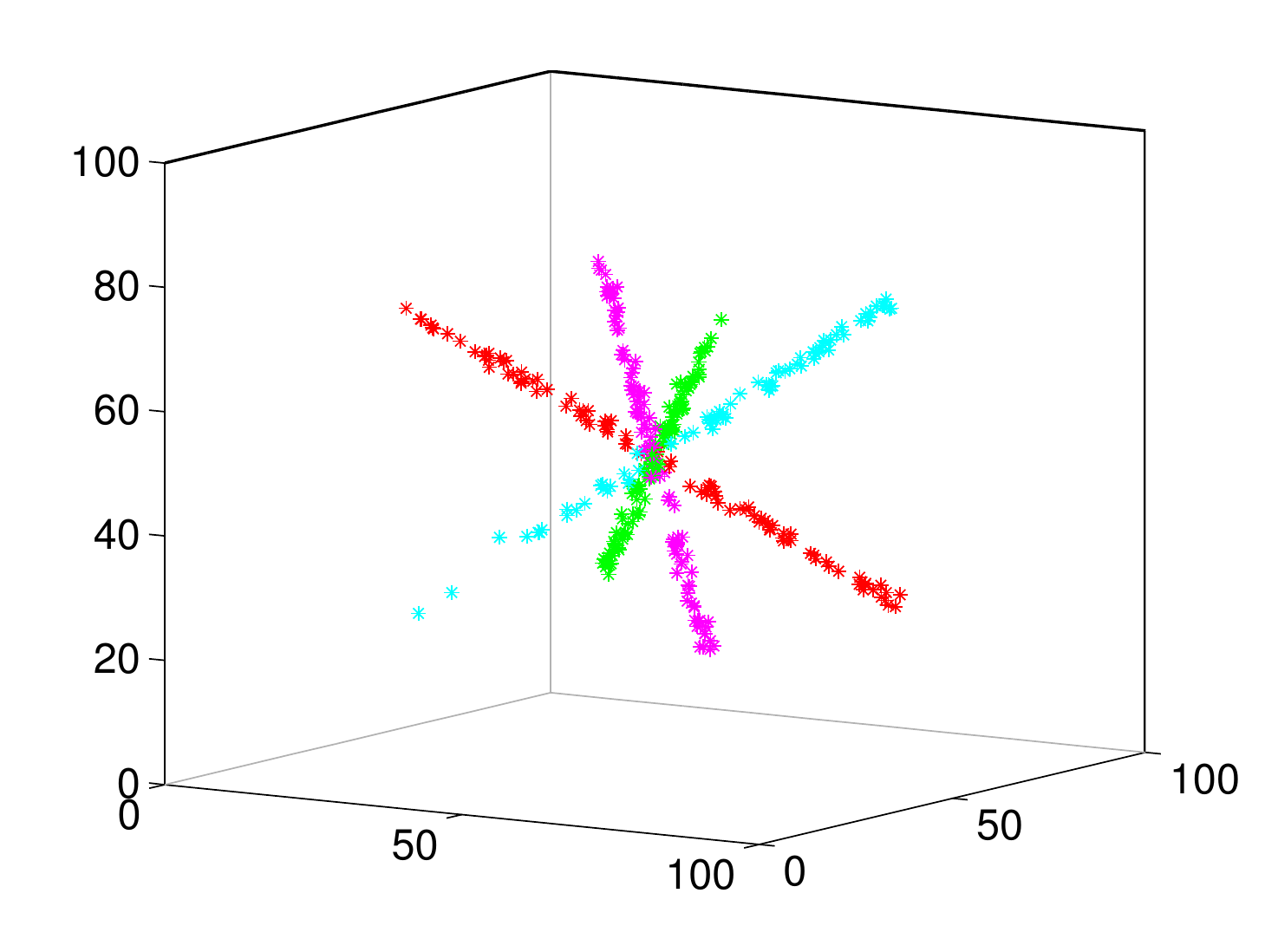}}
  \centerline{\includegraphics[width=1.10\textwidth]{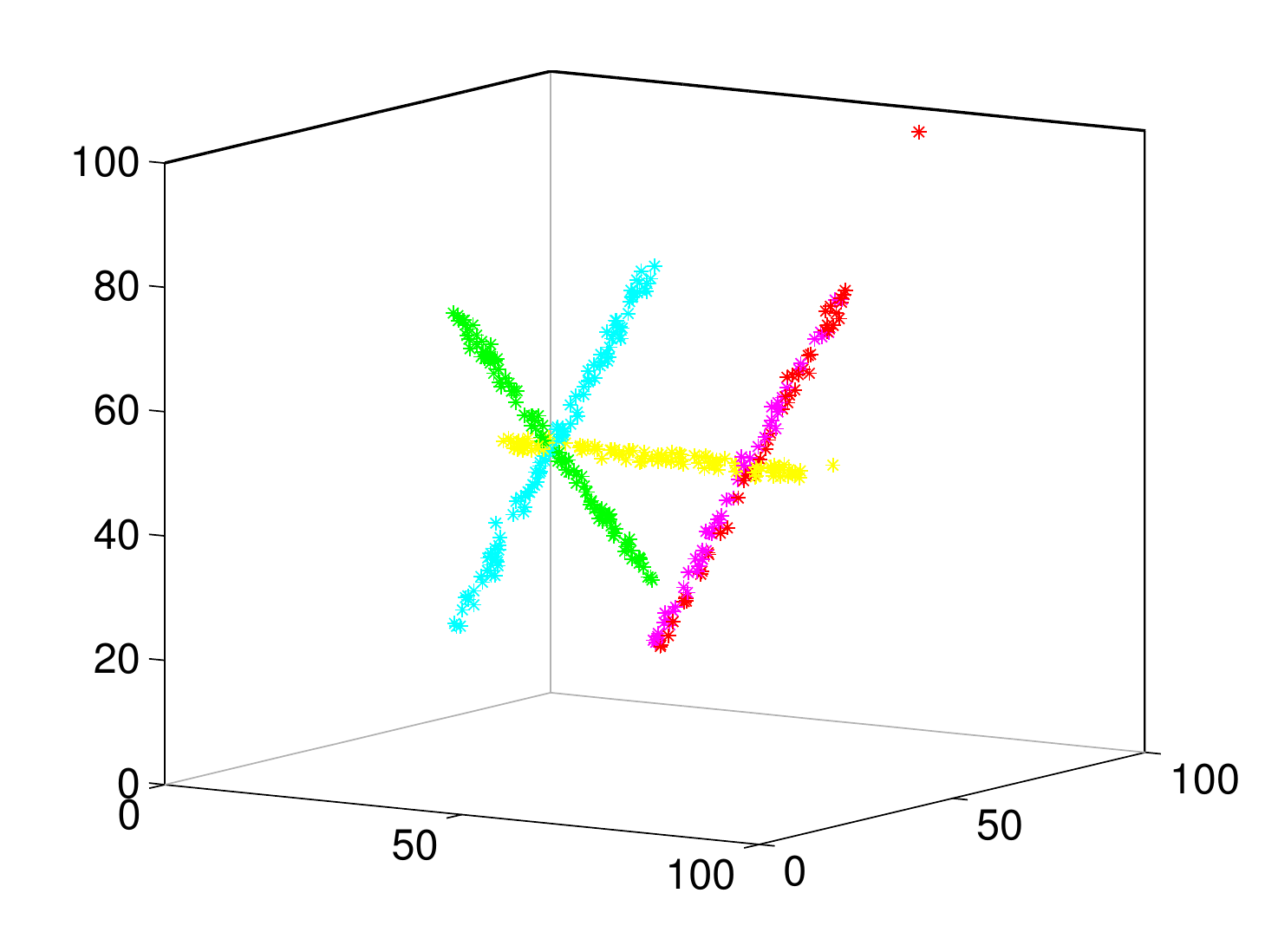}}
  \centerline{\includegraphics[width=1.10\textwidth]{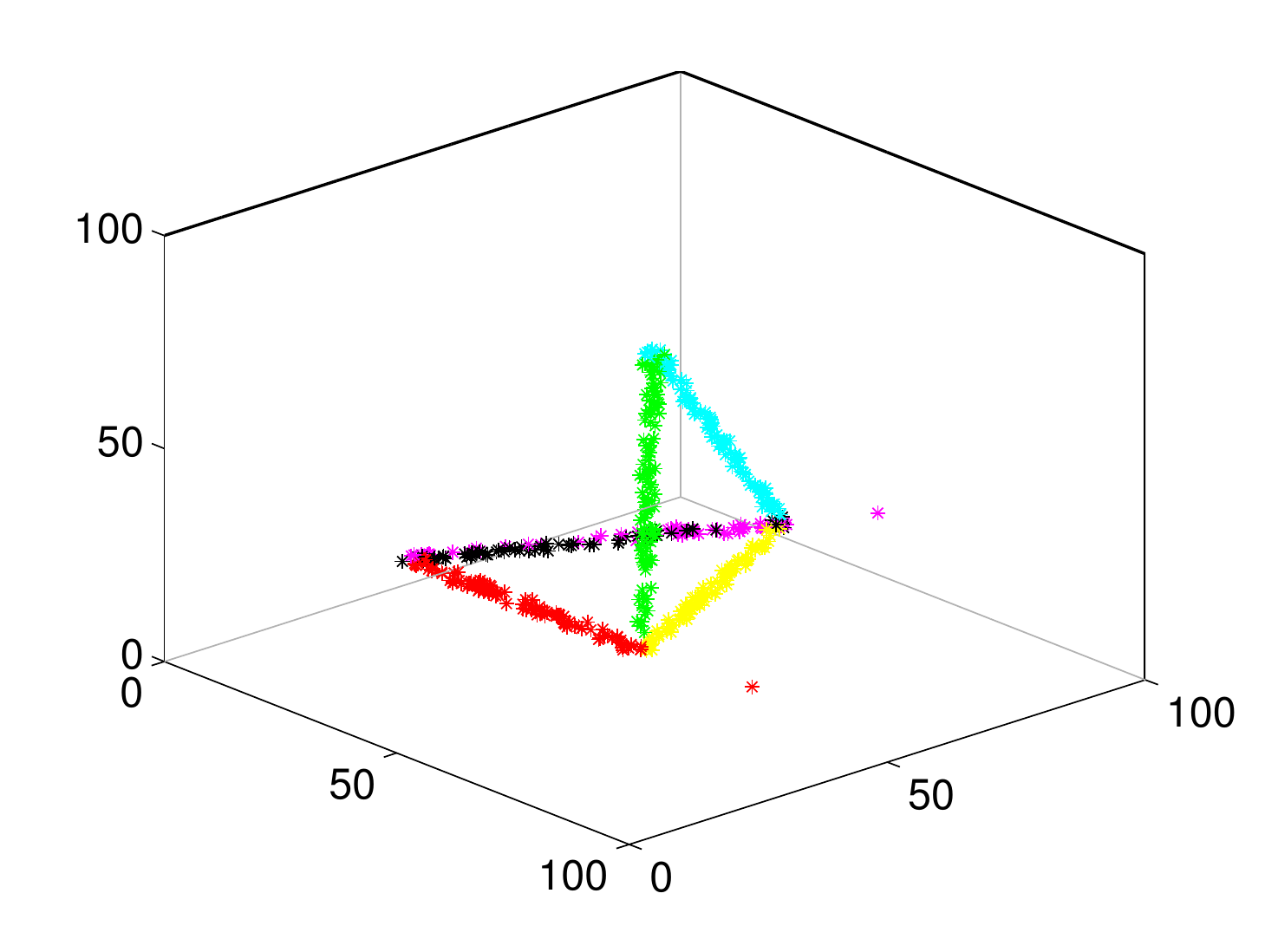}}
 \centerline{\footnotesize(c) RCG }
\end{minipage}
\begin{minipage}[t]{.145\textwidth}
\centerline{\includegraphics[width=1.10\textwidth]{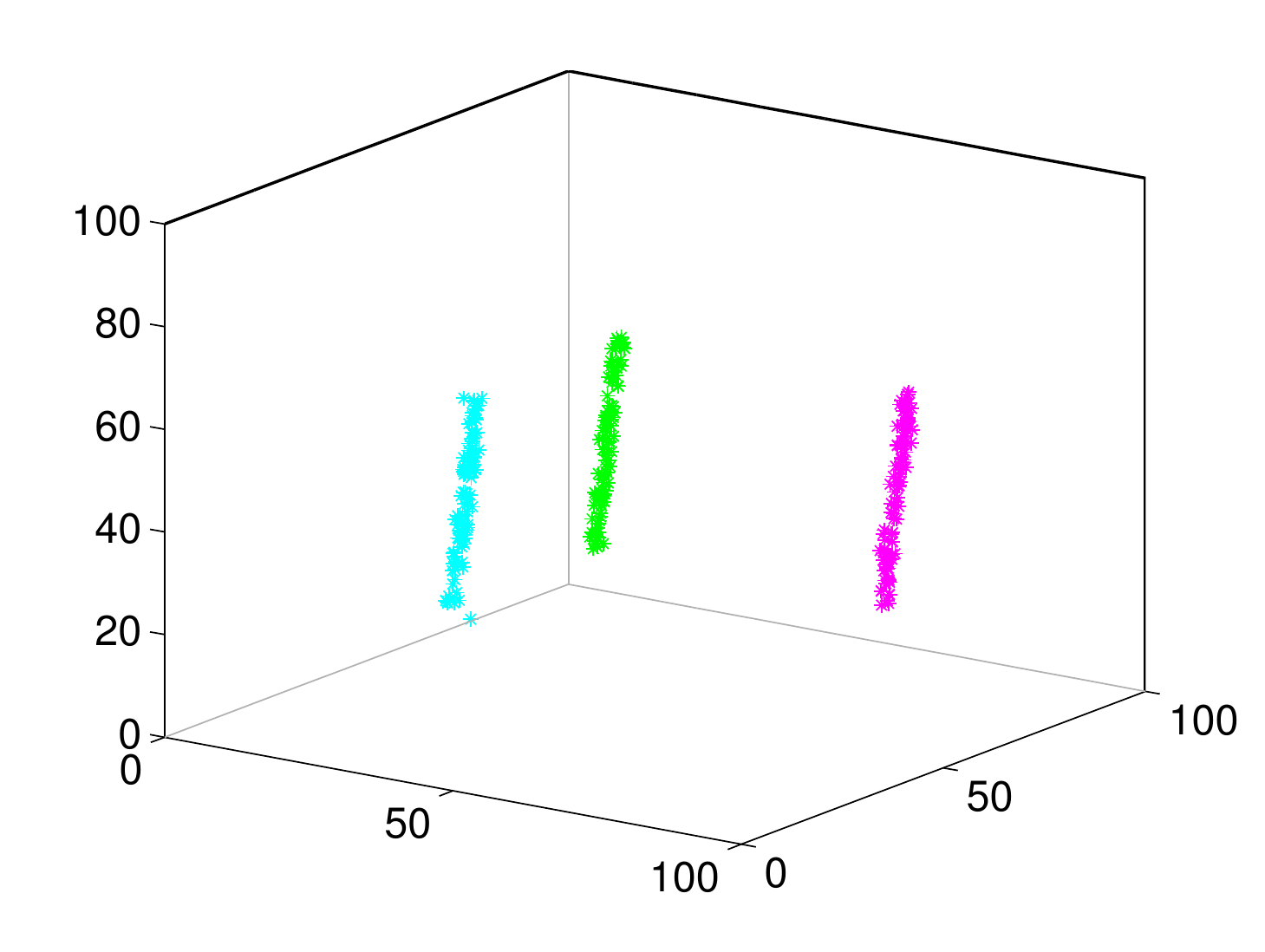}}
  \centerline{\includegraphics[width=1.10\textwidth]{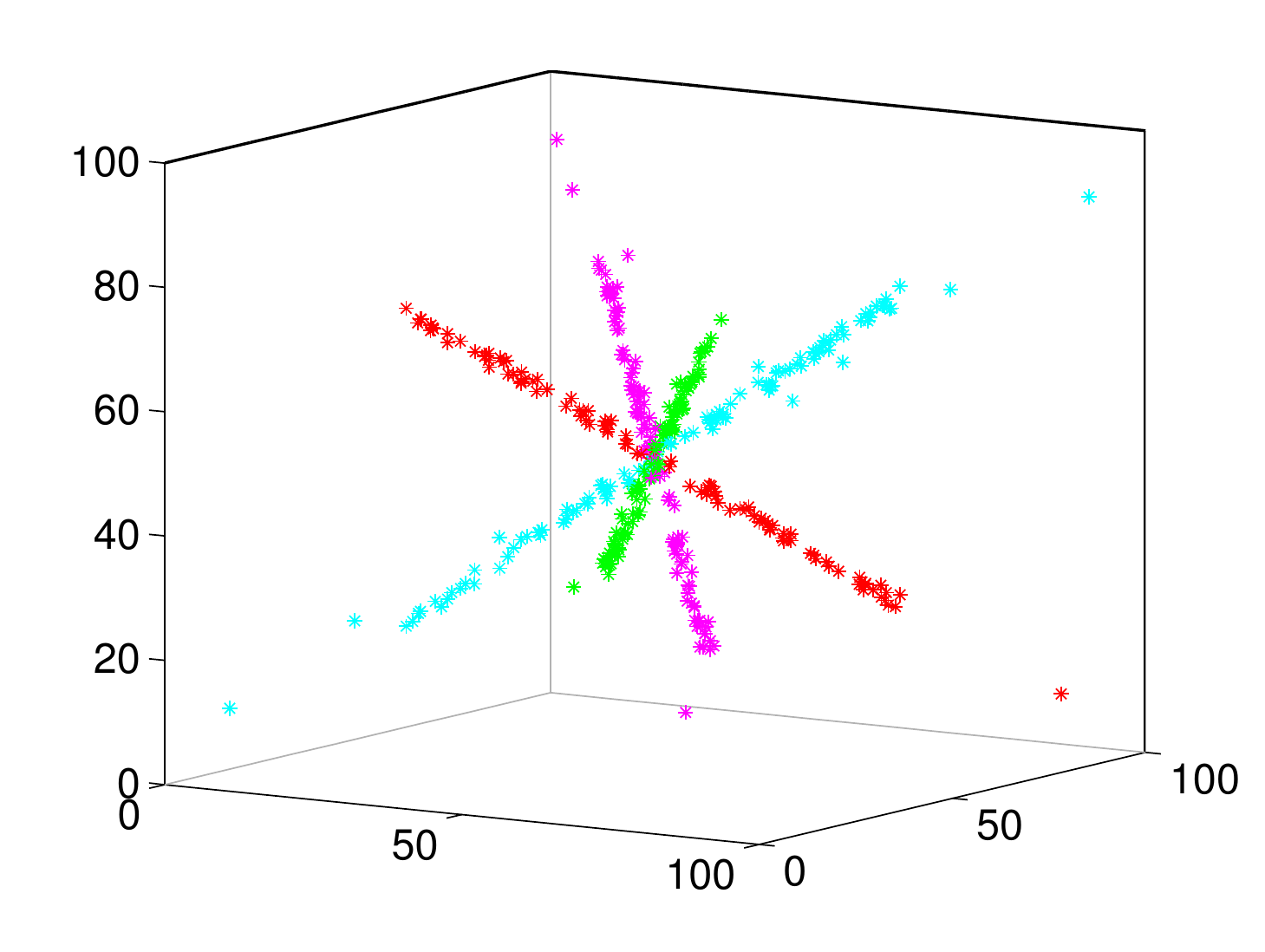}}
  \centerline{\includegraphics[width=1.10\textwidth]{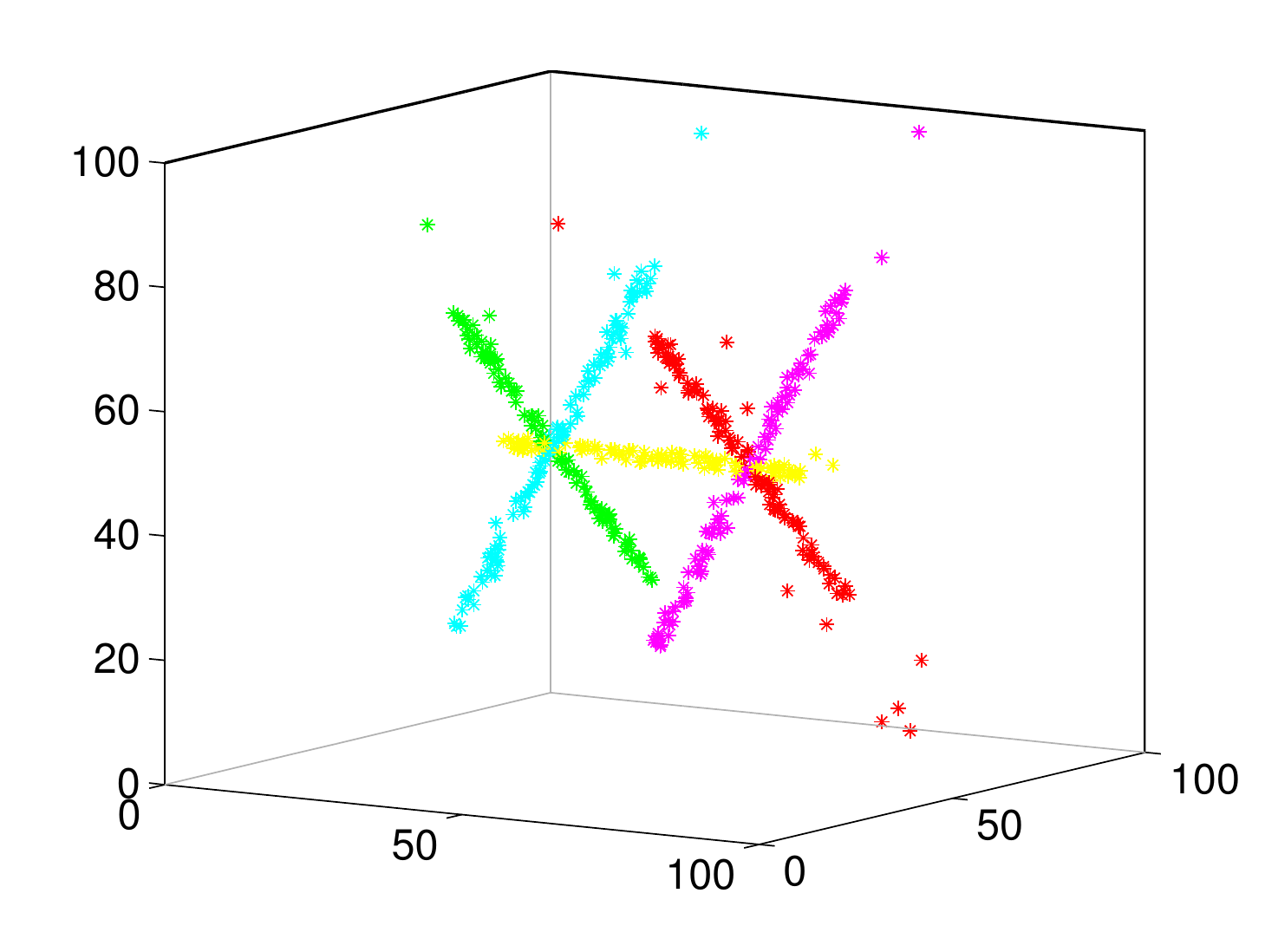}}
  \centerline{\includegraphics[width=1.10\textwidth]{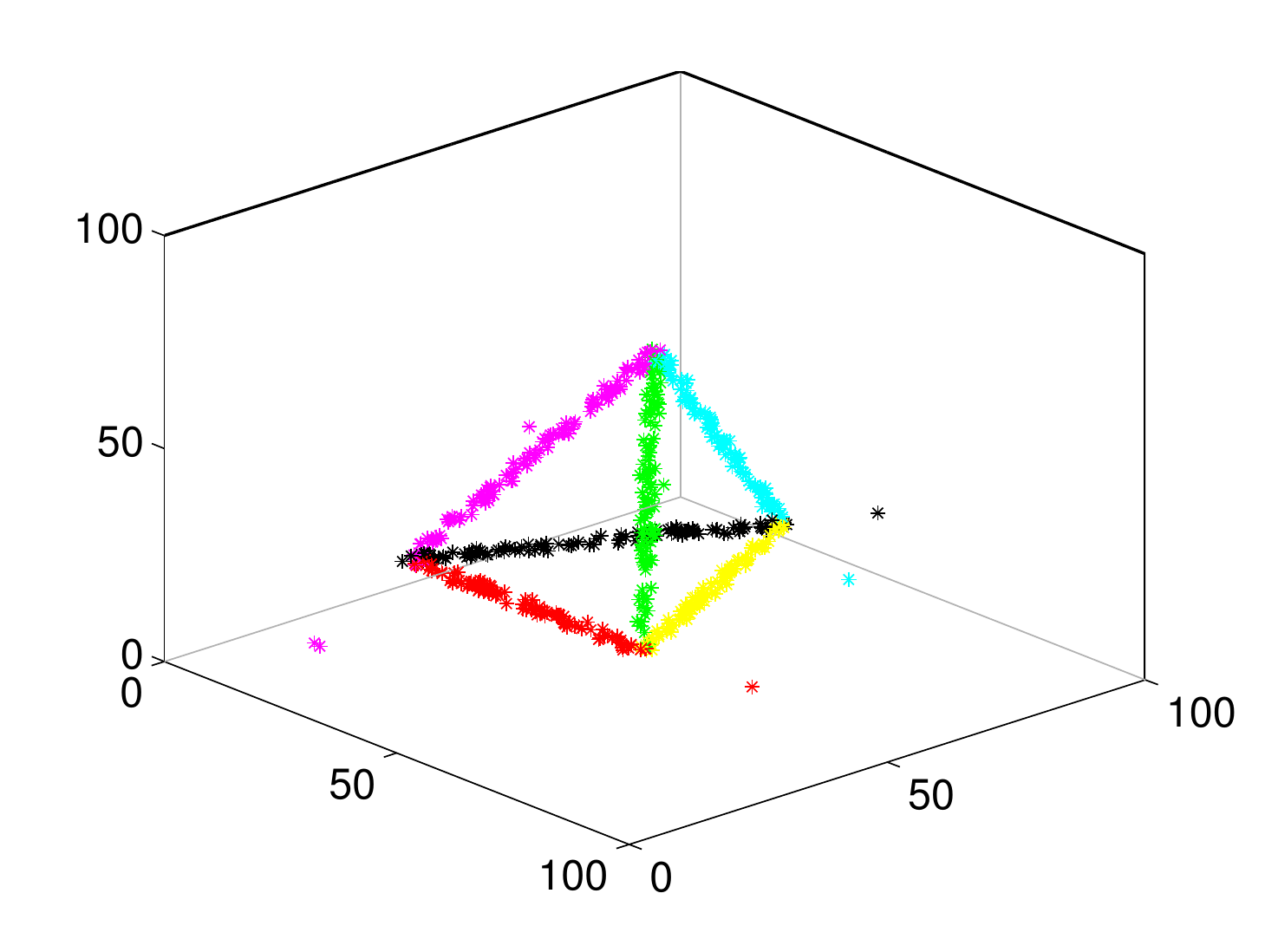}}
  \centerline{\footnotesize(d) AKSWH }
\end{minipage}
\begin{minipage}[t]{.145\textwidth}
\centerline{\includegraphics[width=1.10\textwidth]{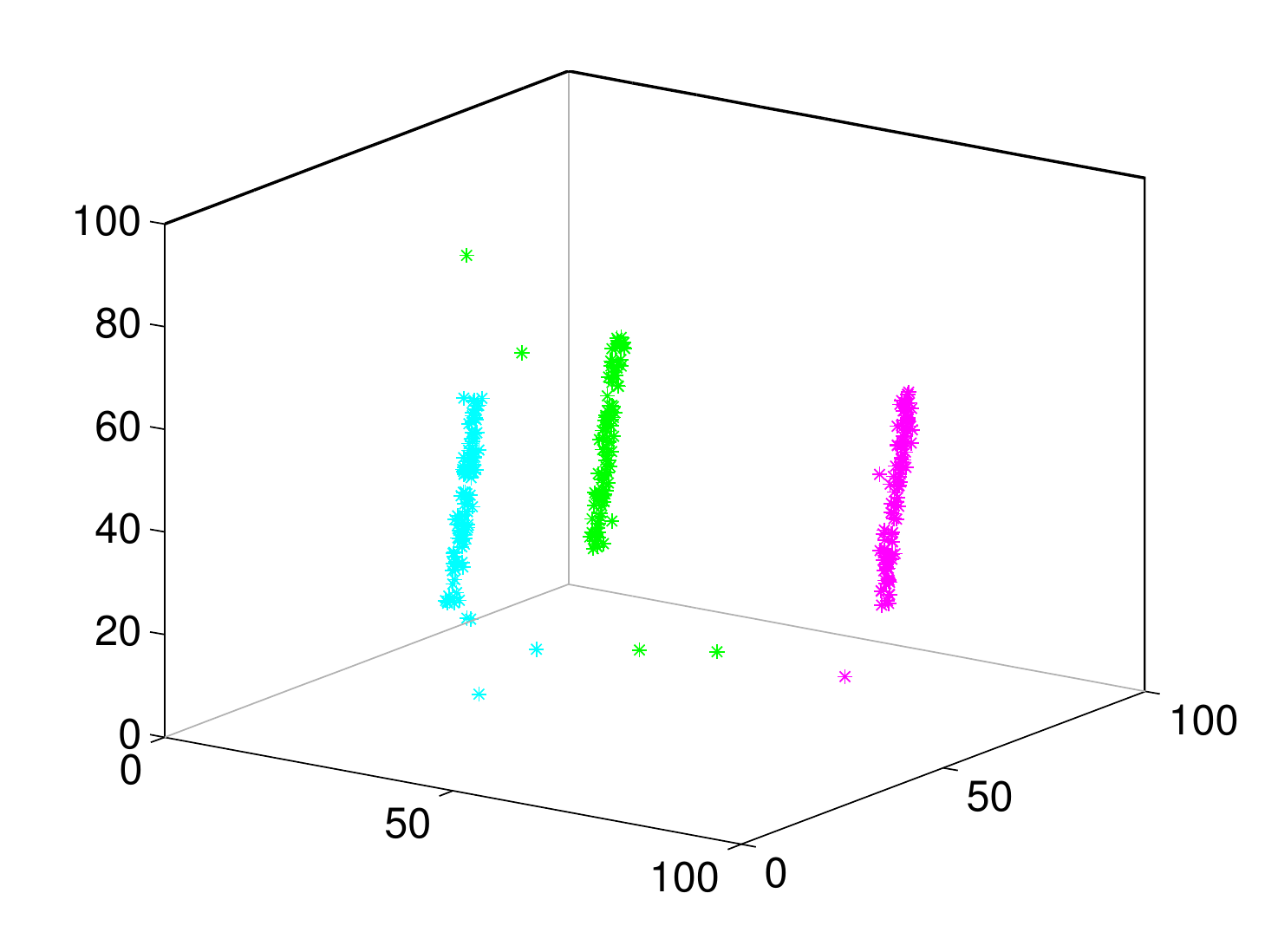}}
  \centerline{\includegraphics[width=1.10\textwidth]{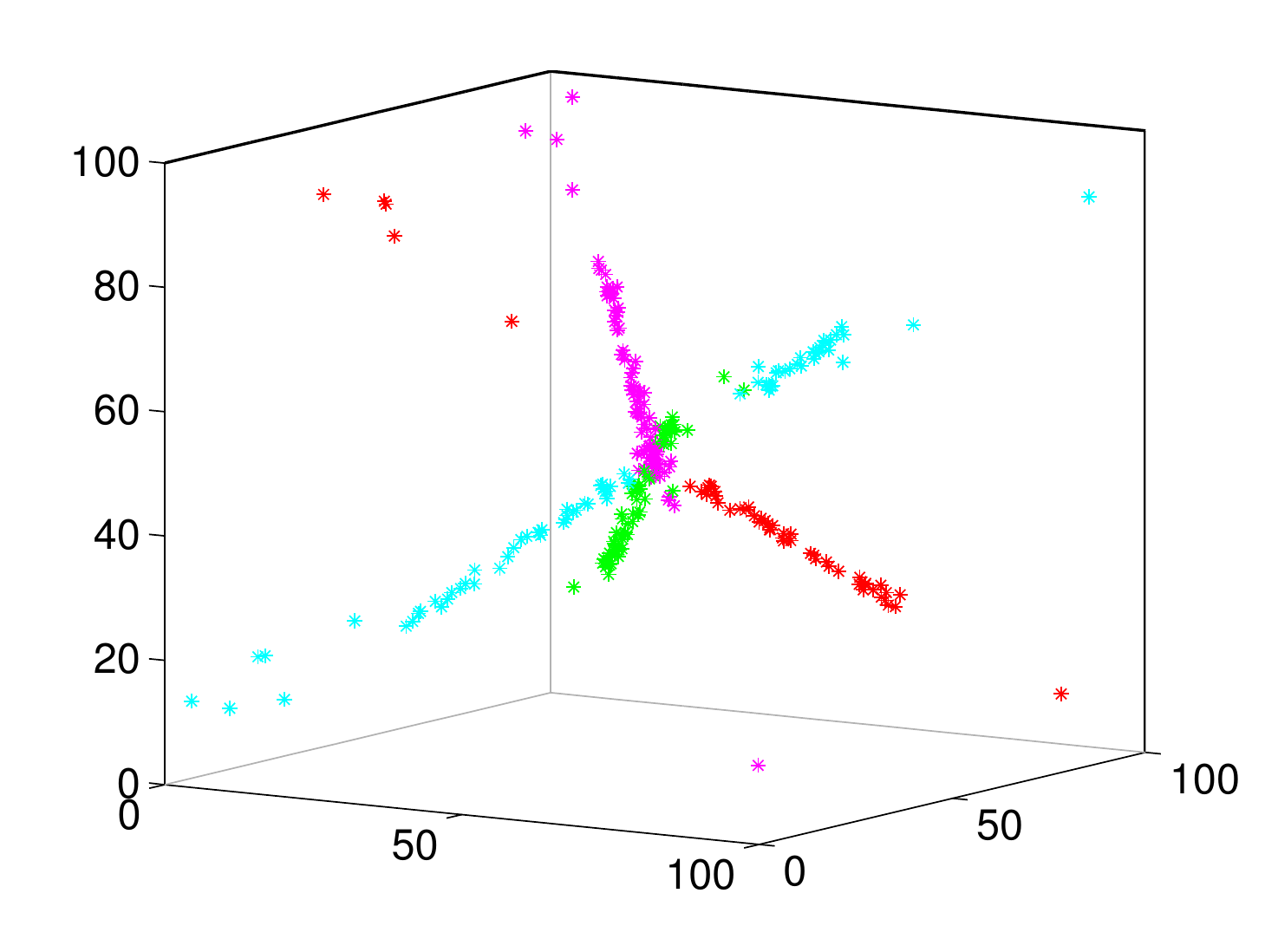}}
  \centerline{\includegraphics[width=1.10\textwidth]{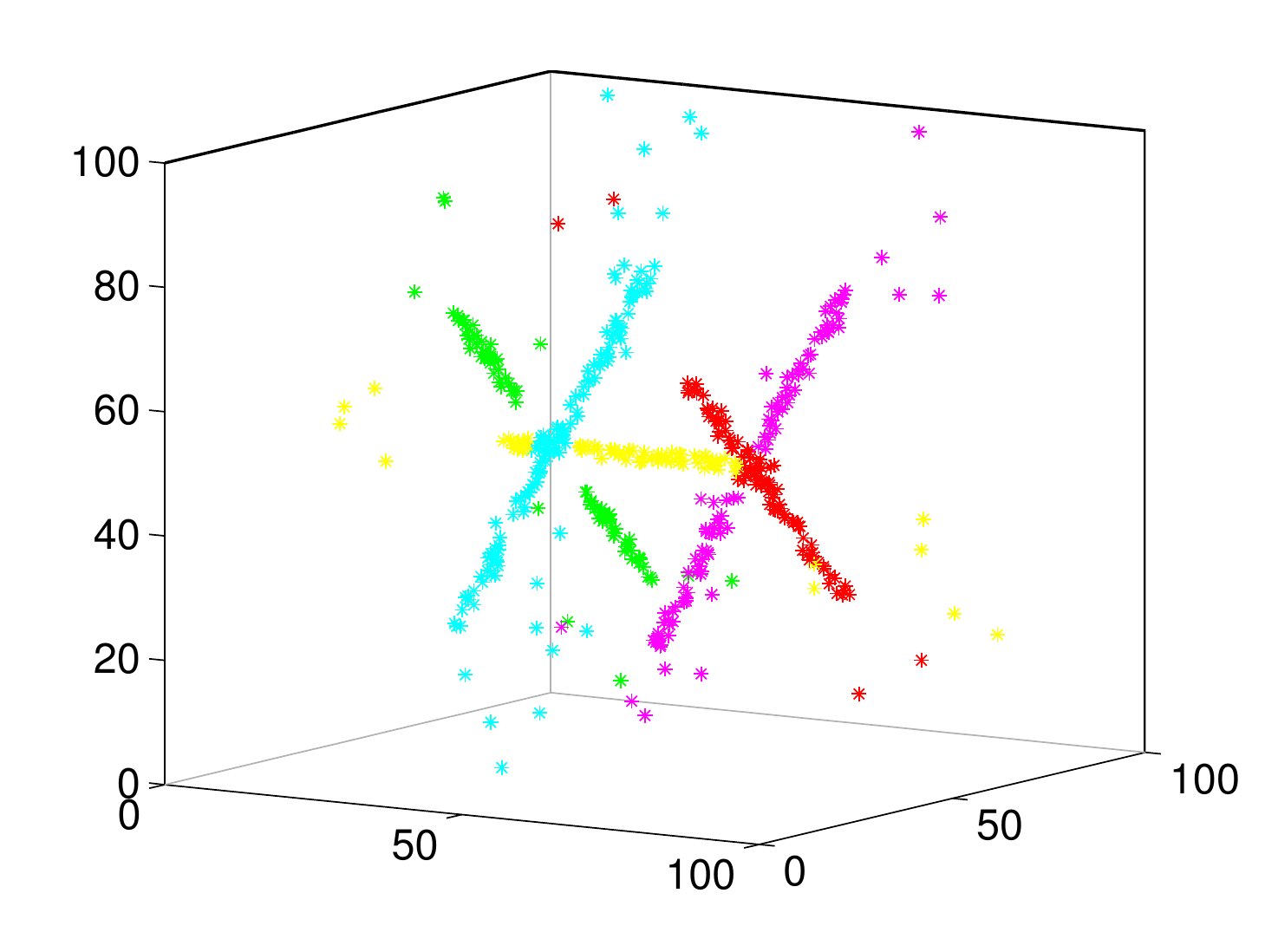}}
  \centerline{\includegraphics[width=1.10\textwidth]{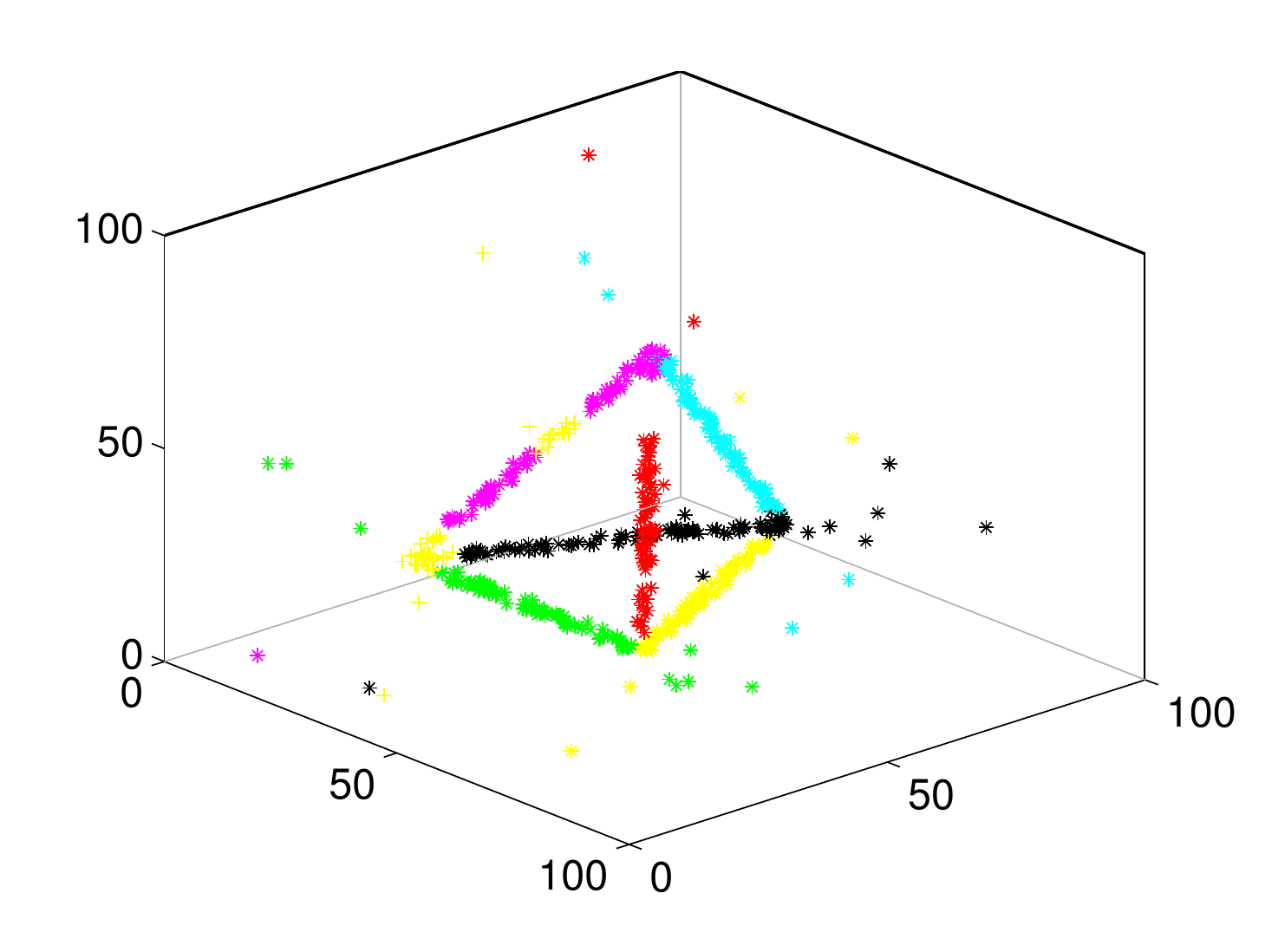}}
  \centerline{\footnotesize(e) T-linkage }
\end{minipage}
\begin{minipage}[t]{.145\textwidth}
  \centerline{\includegraphics[width=1.10\textwidth]{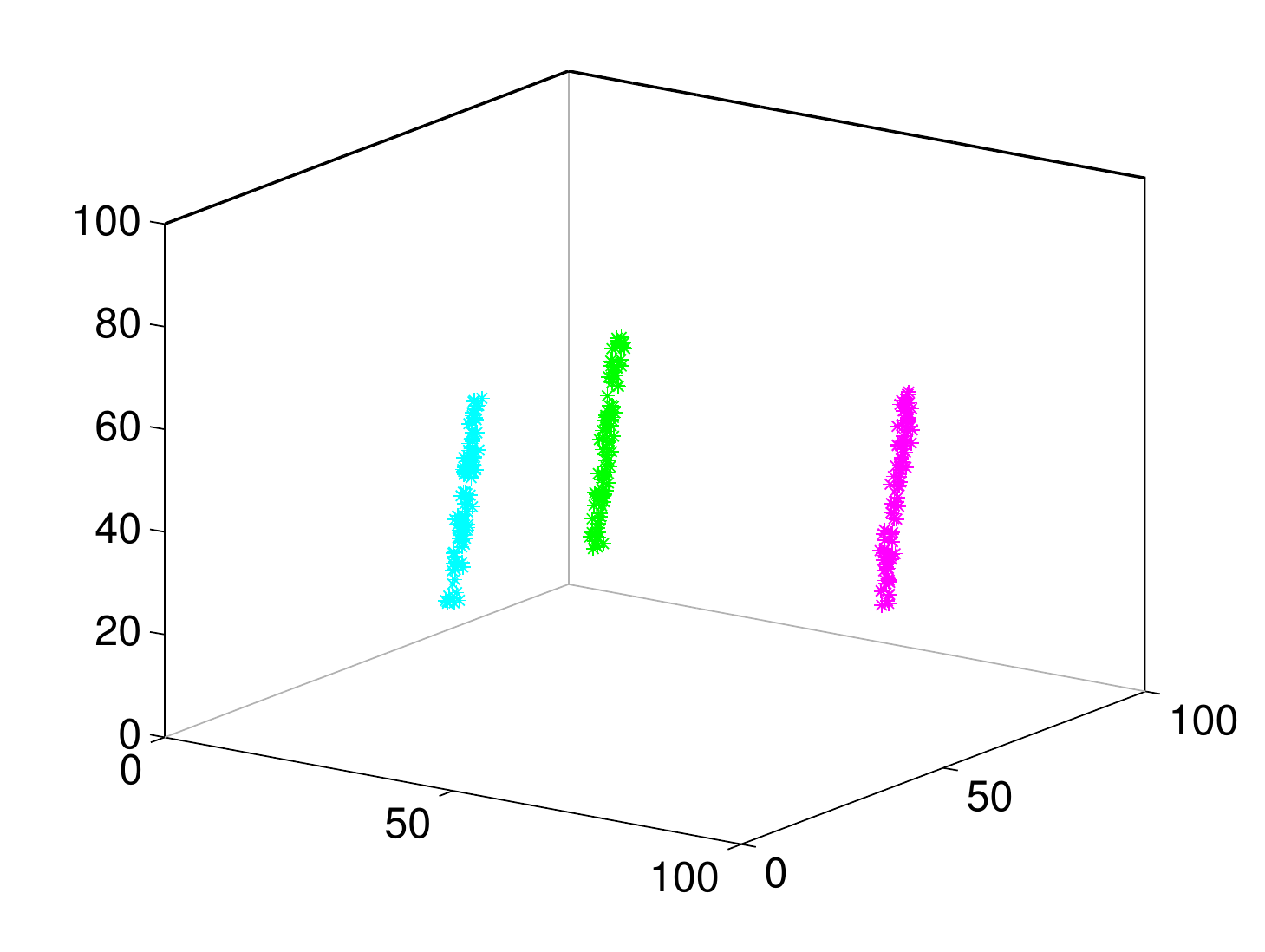}}
  \centerline{\includegraphics[width=1.10\textwidth]{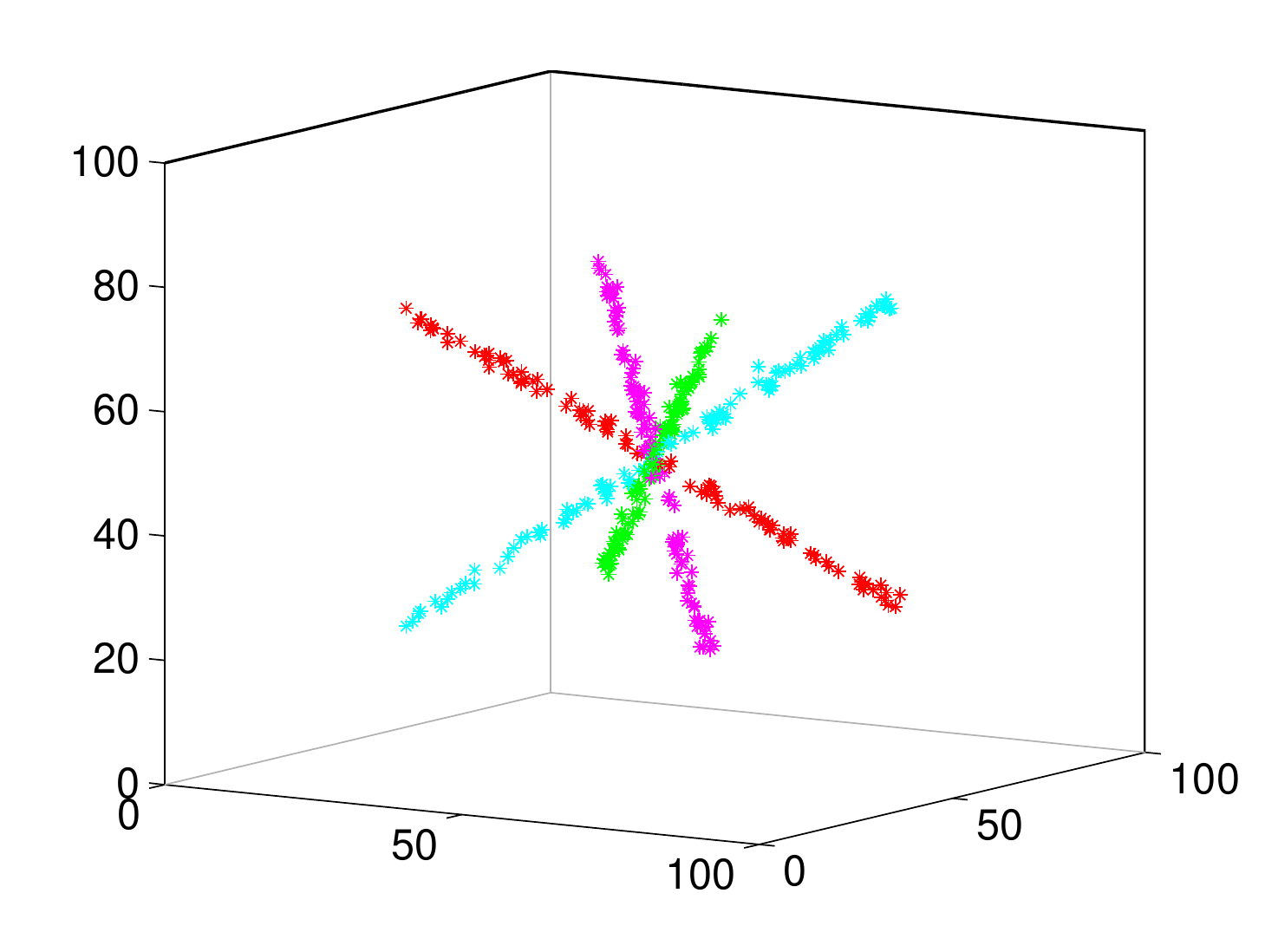}}
  \centerline{\includegraphics[width=1.10\textwidth]{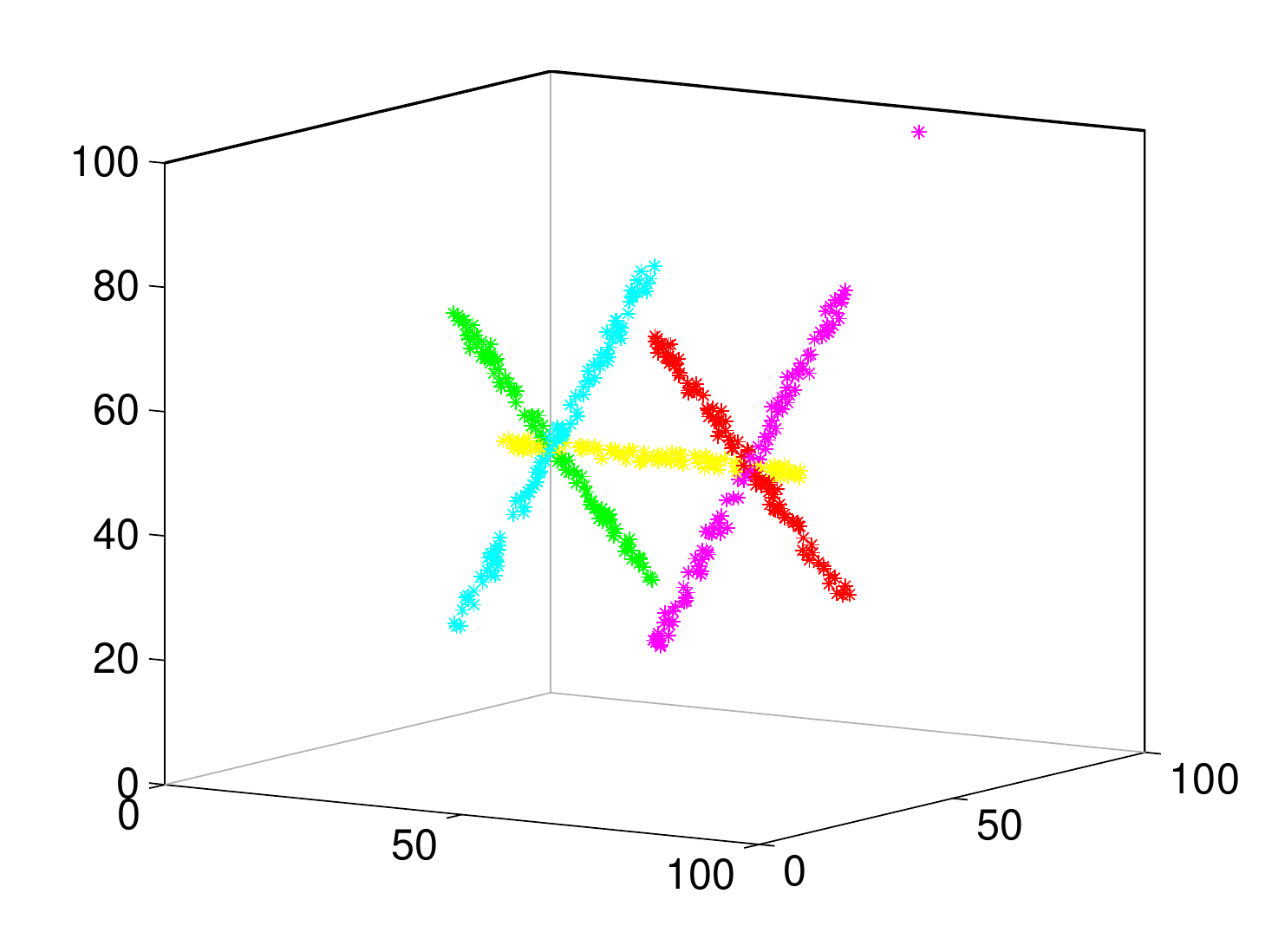}}
  \centerline{\includegraphics[width=1.10\textwidth]{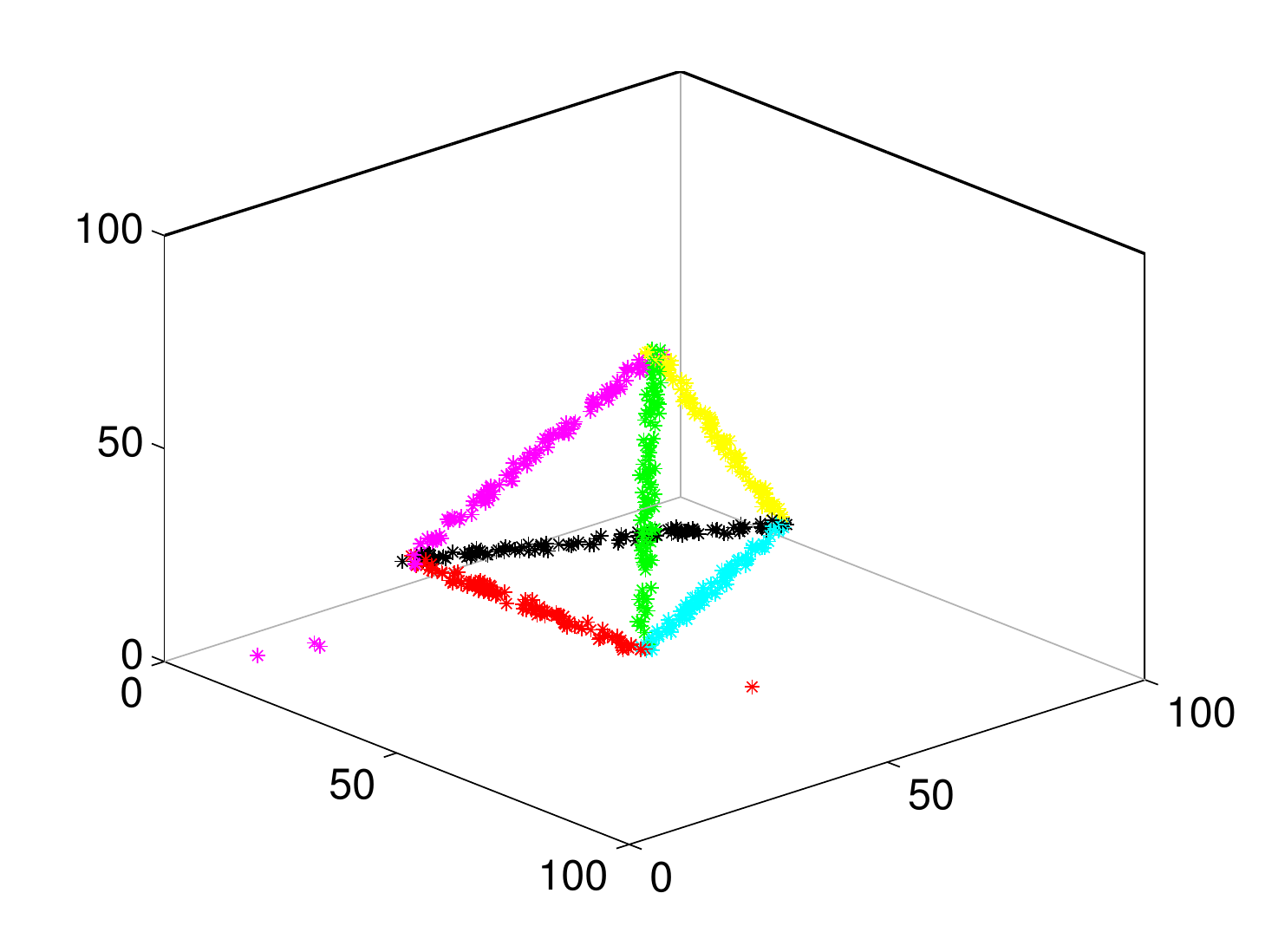}}
  \centerline{\footnotesize(f) MSHF2 }
\end{minipage}
\caption{Examples for line fitting in the $3$D space. $1^{st}$ to $4^{th}$ rows respectively fit three, four, five and six lines. The corresponding outlier percentages are respectively $86\%$, $88\%$, $89\%$ and $90\%$. {The inlier noise scale is set to $1.0$ and each line includes $100$ inliers. Each data includes $400$ outliers.} We do not show the results of MSH/MSHF1, which are similar to those of MSHF2, due to the space limit.}
\label{fig:fivelines}
\end{figure*}

\section{Experiments}
\label{sec:experiments}
In this section, we compare the proposed MSHF with several state-of-the-art model fitting methods, including KF~\cite{chin2009robust}, RCG~\cite{liu2012efficient}, AKSWH~\cite{wang2012simultaneously}, and T-linkage~\cite{Magri_2014_CVPR}, on both synthetic data and real images. We choose these representative methods because KF is a data clustering based method, RCG is a hypergraph based method, and AKSWH is a parameter space based method. These fitting methods are {closely} related to the proposed method (recall that MSHF seeks modes on hypergraphs and it fits multi-structure data in the parameter space). We also choose T-linkage as a competing method due to its good performance. Moreover, we compare with our original method (MSH) in~\cite{wang2015mode} to show the improvements of the proposed MSHF. {For MSHF, we test two versions: MSHF1, which does not use the neighboring constraint in Eq.~(\ref{equ:minimumtdistance}) and MSHF2, which uses the neighboring constraint in Eq.~(\ref{equ:minimumtdistance}).}

To be fair, we first generate a set of model hypotheses by using the proximity sampling~\cite{kanazawa2004detection,toldo2008robust} for all the competing algorithms in each experiment.  Then all the competing methods perform model fitting based on the same set of model hypotheses. We generate {the same number} of model hypotheses as~\cite{wang2012simultaneously}, i.e., there are $5,000$ model hypotheses generated for line fitting (Sec.~\ref{sec:sylinefitting} and Sec.~\ref{sec:linefitting}) and circle fitting (Sec.~\ref{sec:sycirclefitting} and Sec.~\ref{sec:circlefitting}), $10,000$ model hypotheses generated for homography based segmentation (Sec.~\ref{sec:homographbasedsegmentation}), and $20,000$ model hypotheses generated for two-view based motion segmentation (Sec.~\ref{sec:motionsegmentation}).

We have optimized the parameters of all the competing fitting methods\footnote{For KF and T-linkage, we use the code published on the web: \url{http://cs.adelaide.edu.au/~tjchin/doku.php} and \url{http://www.diegm.uniud.it/fusiello/demo/jlk/}, respectively. For RCG and AKSWH, we use the code provided by the authors.} on each dataset for the best performance. {For our methods (i.e., MSH and MSHF1/MSHF2), we only slightly adjust the value of $K$ for IKOSE. In most cases we fix $K=10\%*n$, where $n$ is the number of input data points. In some challenging cases (i.e., where the data do not include at least $10\%$ inliers), we adjust the value of $K$ to obtain good inlier noise scales.} All experiments are run on MS Windows $7$ with Intel Core i$7$-$3630$ CPU $2.4GHz$ and $16GB$ RAM. The fitting error is computed as follows~\cite{Magri_2014_CVPR,mittal2012generalized}:
\begin{align}
\label{equ:fittingerror}
error=\frac{\#~mislabeled~data~points}{\#~data~points}\times 100\%.
\end{align}

\subsection{Synthetic Data}
\label{sec:syntheticData}
\subsubsection{Line Fitting}
\label{sec:sylinefitting}
\begin{table}[ht]
  \caption{Quantitative comparison results of line fitting on four synthetic {data}. The best results are boldfaced.}
\centering
\scalebox{0.96}{{
\begin{tabular}{|c|c|c|c|c|c|>{\columncolor{mygray}}c|>{\columncolor{mygray}}c|>{\columncolor{mygray}}c|}
\hline
Data        &                  & M1&M2&M3&M4&M5&M6 &{M7}\\
\hline
\hline
              &Std.    &{0.01}  & {0.01}& {0.01} & {0.01}& {0.01}& {0.01}& {0.01}  \\
   3         &Avg.   & {1.76} & {0.33} & {0.34}& {1.87}& {0.16}& {\bf0.14}&{\bf0.14}\\
lines       &Min.  & {1.71}& {0.29} & {0.29}& {1.71}& {\bf0.14}& {\bf0.14}& {\bf0.14}\\
             &Time   &{13.74}&{\bf0.41} &{1.17}&{155.92}& {0.99}& {1.77}& {1.04}\\
                       \hline
             &Std.    & {3.16}& {2.11}& {1.02}& {4.73}& {0.59} & {\bf0.26} & {\bf0.26}  \\
   4       &Avg.    & {18.25}& {4.13}& {3.00}& {31.40}& {1.29} & {\bf1.23} & {\bf1.23} \\
lines      &Min.  & {13.25}& {1.63}& {2.88}& {23.75}& {0.88}& {\bf0.75}& {\bf0.75}\\
             &Time   &{17.09}&{\bf0.53}&{1.18}&{210.39} &{1.68} &{2.92}&{2.57}  \\
                       \hline
               &Std.  & {3.07}& {7.42}& {5.34}& {4.53}& {0.21}  & {\bf0.17}& {\bf0.17}  \\
   5          &Avg.   & {15.27} & {18.00} & {3.78} & {17.29} & {1.76}& {\bf1.72} & {\bf1.72}\\
   lines     &Min.  & {11.42}& {2.44}& {2.67}& {11.89}& {1.44}& {\bf1.22}& {\bf1.22} \\
               &Time    &{20.36}&{\bf0.68}&{1.23}&{274.08}&{1.88}&{2.81}&{2.61}\\
                       \hline
               &Std.    & {3.32}&{5.63}& {2.87}& {3.15} & {\bf0.47} & {\bf0.47}& {\bf0.47} \\
   6         &Avg.    & {33.71} & {15.69}& {4.57}& {16.26}& {3.34}& {\bf3.32}& {\bf3.32}  \\
 lines      &Min.  & {27.10}& {5.00}& {2.70}& {11.70}& {\bf2.30}& {\bf2.30}& {\bf2.30}\\
              &Time   &{25.53}&{\bf0.69}&{1.37}&{326.48}&{2.07} &{3.06}&{2.77}\\
                       \hline
\end{tabular}}}
\\
\medskip
{(M1-KF; M2-RCG; M3-AKSWH; M4-T-linkage; M5-MSH; M6-MSHF1; M7-MSHF2. {M1-M7 in the following tables denote the same meaning.})}
 \label{table:3Dlinetable}
\end{table}
We evaluate the performance of the {seven} fitting methods on line fitting using four challenging synthetic data in the $3$D space (as shown in Fig.~\ref{fig:fivelines}). We repeat the experiment $50$ times and report the standard variances, the average and the best results of the fitting errors {(in percentage)} and the average CPU time (in seconds) obtained by the competing methods in Table~\ref{table:3Dlinetable} (we exclude the time used for sampling and generating potential hypotheses for all the fitting methods). We also show the corresponding fitting results obtained by all the competing methods from Fig.~\ref{fig:fivelines}(b) to Fig.~\ref{fig:fivelines}(f).

From Fig.~\ref{fig:fivelines} and Table~\ref{table:3Dlinetable}, we can see that: (1) For the ``three lines" data, the three lines are completely separable in the $3$D space, and the {seven} fitting methods succeed in fitting all the three lines. However, {MSH and MSHF1/MSHF2} achieve the best performance on the fitting accuracy among the {seven} fitting methods due to their high robustness to outliers. (2) For the ``four lines" data, the four lines intersect at one point. The {seven} fitting methods succeed in estimating the number of the lines {in the data}, but the data clustering based methods (i.e., KF and T-linkage) cannot effectively segment the data points near the intersection. In contrast, RCG, AKSWH, {MSH and MSHF1/MSHF2} correctly fit {all} the four lines with lower fitting errors, {while MSHF1/MSHF2} achieve the lowest fitting error. (3) For the ``five lines" data, there exist two intersections. The data points near the intersections are not correctly segmented by both KF and T-linkage, which causes these two methods to obtain high fitting errors. RCG correctly fits four lines but it wrongly fits one. This is because the dense subgraph representing a potential structure {in the data} is not effectively detected by RCG. In contrast, the parameter space based methods (i.e., AKSWH, {MSH and MSHF1/MSHF2}) {do not directly deal with data points}. AKSWH, {MSH and MSHF1/MSHF2} correctly fit all the five lines with low fitting errors. (4) For the ``six lines" data, RCG only correctly fits five of the six lines. T-linkage wrongly estimates the number of lines {in the data}. KF achieves the worst performance among the five fitting methods. In contrast, AKSWH, {MSH and MSHF1/MSHF2} correctly fit the six lines. These results on the challenging dataset further shows the superiority of the parameter space based methods over the other competing fitting methods.

For the performance of computational time, RCG achieves the fastest speed among the {seven} fitting methods, but it cannot achieve good fitting accuracy. AKSWH, {MSH and MSHF1/MSHF2} achieve similar computational speed. {Here, the speed of these four fitting methods depends on the number of the selected significant model hypotheses/vertices. MSHF1/MSHF2 retain the maximum number of significant vertices to avoid missing model instances with less number of inliers. Thus, MSHF1/MSHF2 are slower than AKSWH and MSH. MSHF2 is faster than MSHF1 due to the use of the neighboring constraint.} MSHF1 is faster than KF and T-linkage for all four {data} (about $5.85$$\sim$$8.34$ times faster than KF and about $72.05$$\sim$$106.69$ times faster than T-linkage).

\begin{figure}[t]
\centering
\begin{minipage}{.23\textwidth}
\centerline{\includegraphics[width=1.0\textwidth]{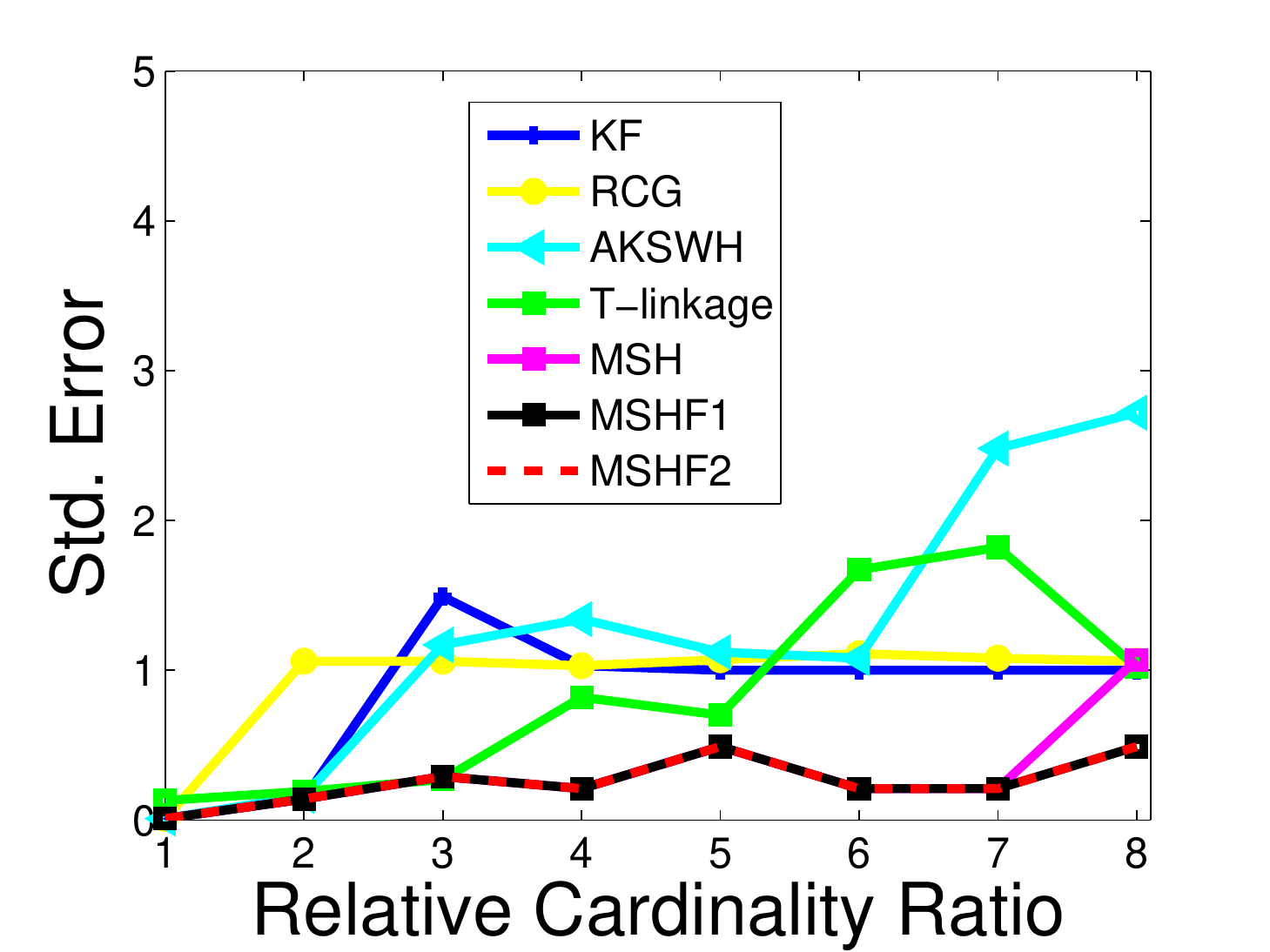}}
  \centerline{(a)}
\end{minipage}
\begin{minipage}{.23\textwidth}
\centerline{\includegraphics[width=1.0\textwidth]{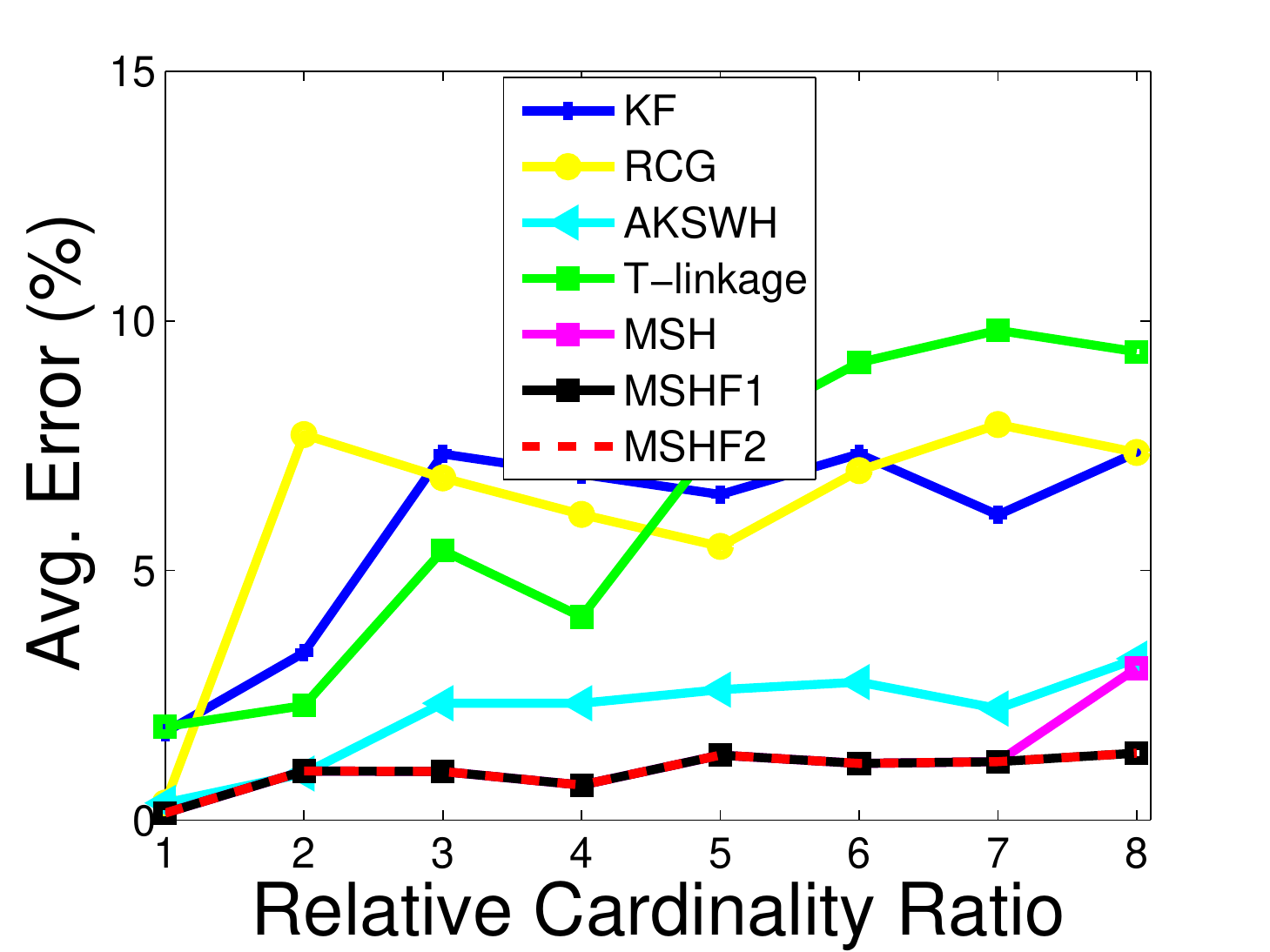}}
  \centerline{(b)}
\end{minipage}
\caption{{The fitting errors obtained by the seven competing methods for data with different cardinality ratios of inliers: (a) and (b) show the performance comparison of the standard variances and the average fitting errors for data with different inlier cardinality ratios, respectively.}}
\label{fig:diffratio}
\end{figure}
\begin{figure*}[t]
\centering
\begin{minipage}[t]{.145\textwidth}
  \centerline{\includegraphics[width=1.10\textwidth]{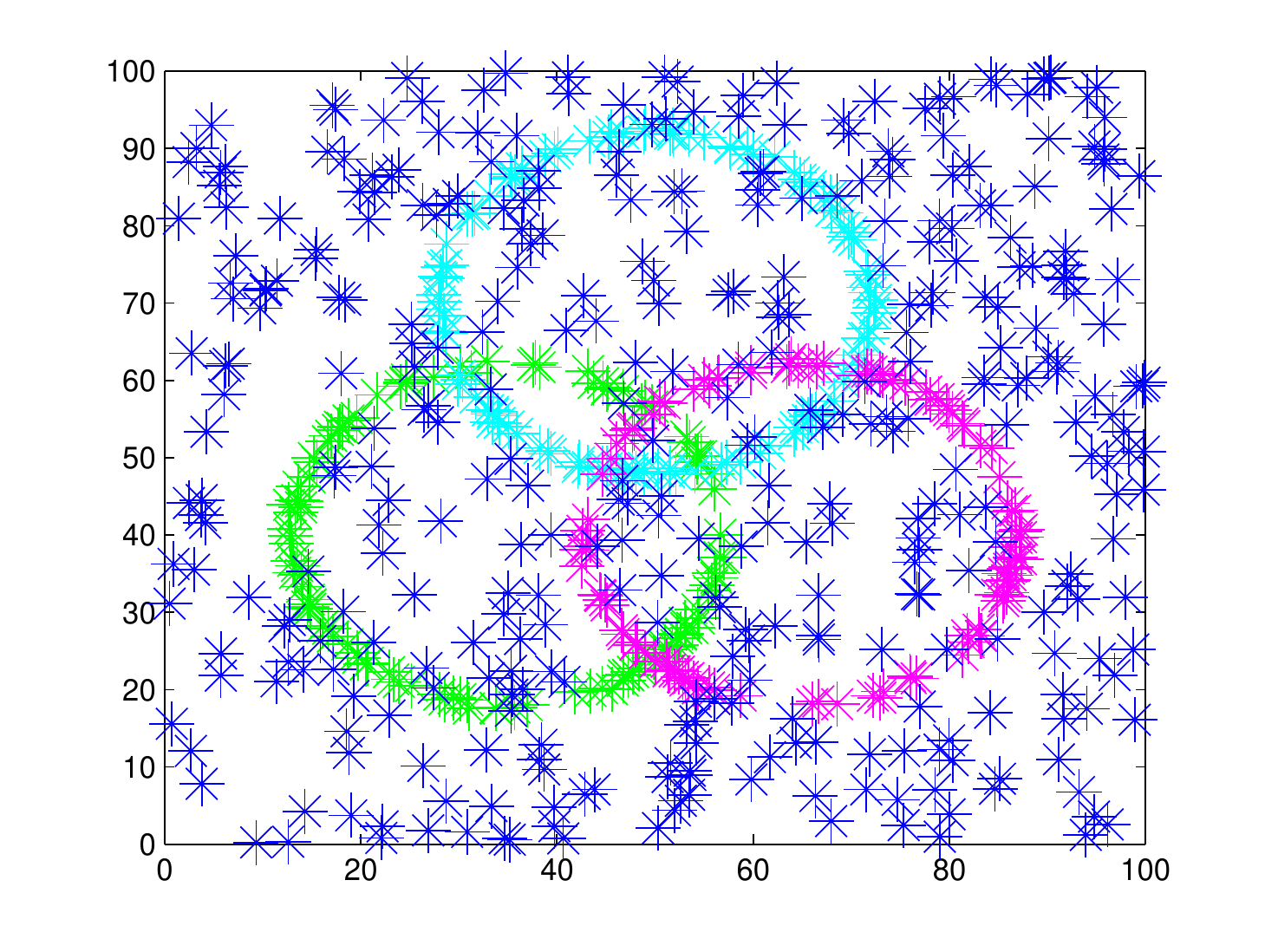}}
  \centerline{\includegraphics[width=1.100\textwidth]{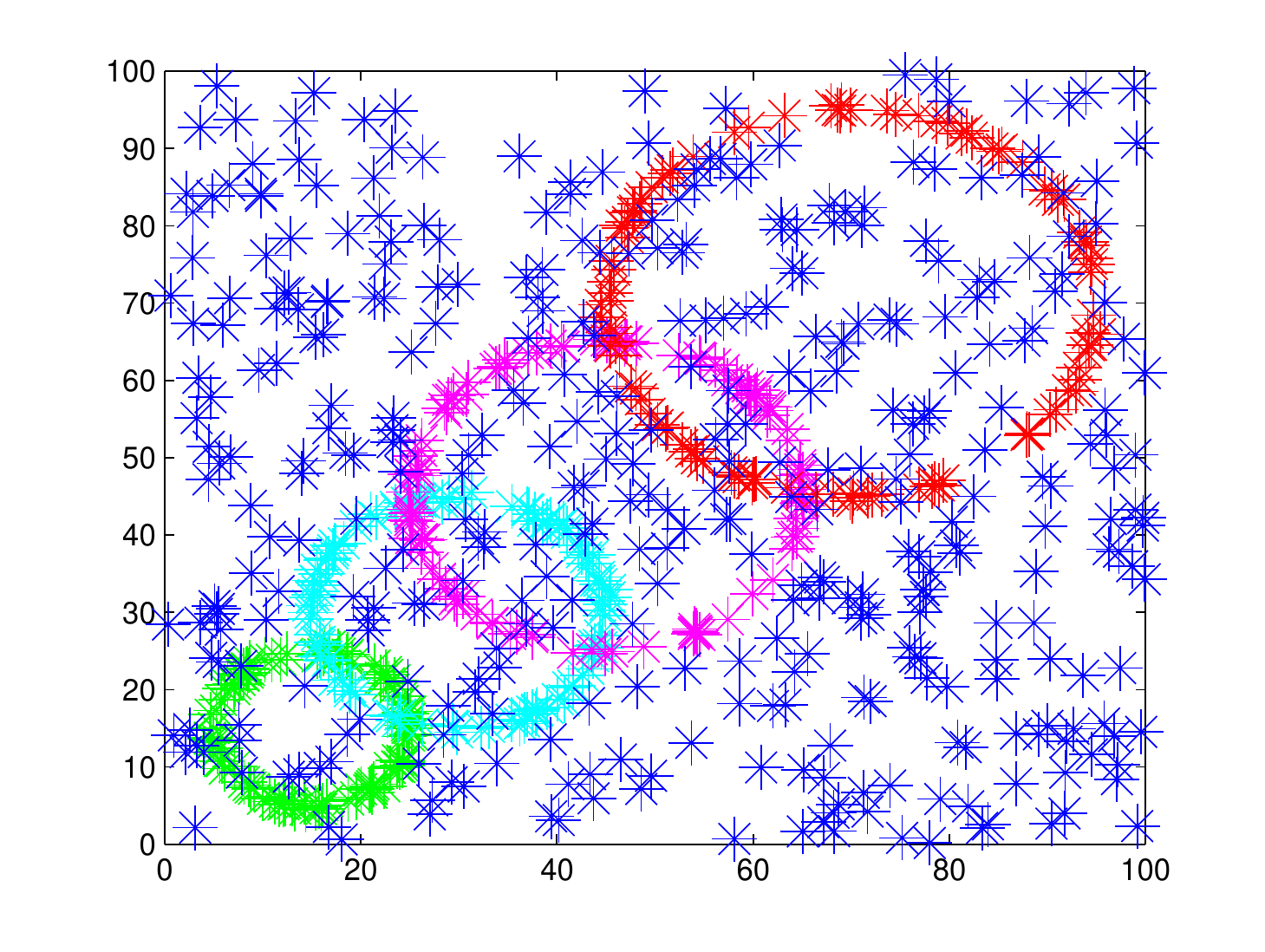}}
  \centerline{\includegraphics[width=1.100\textwidth]{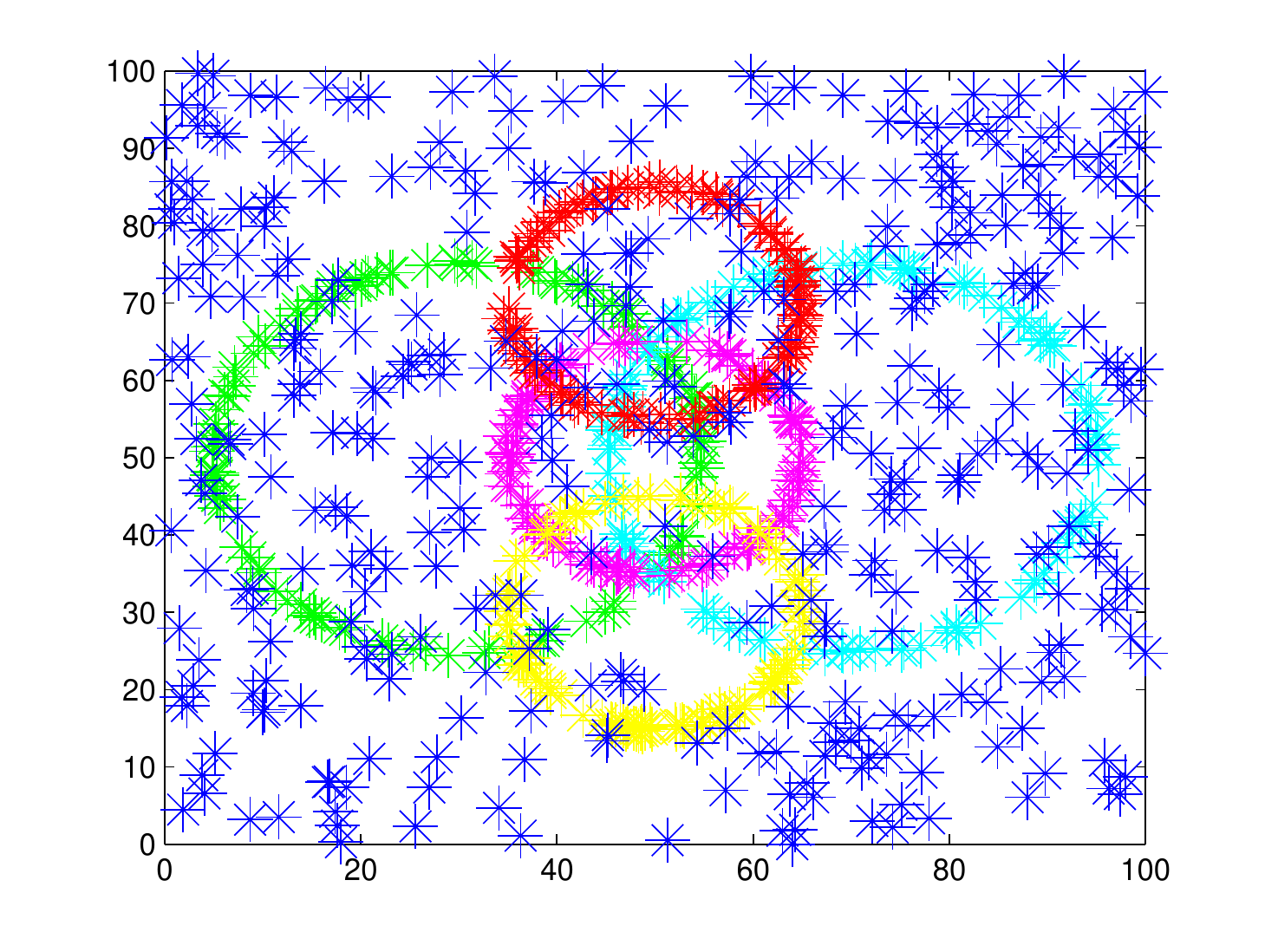}}
  \centerline{\includegraphics[width=1.100\textwidth]{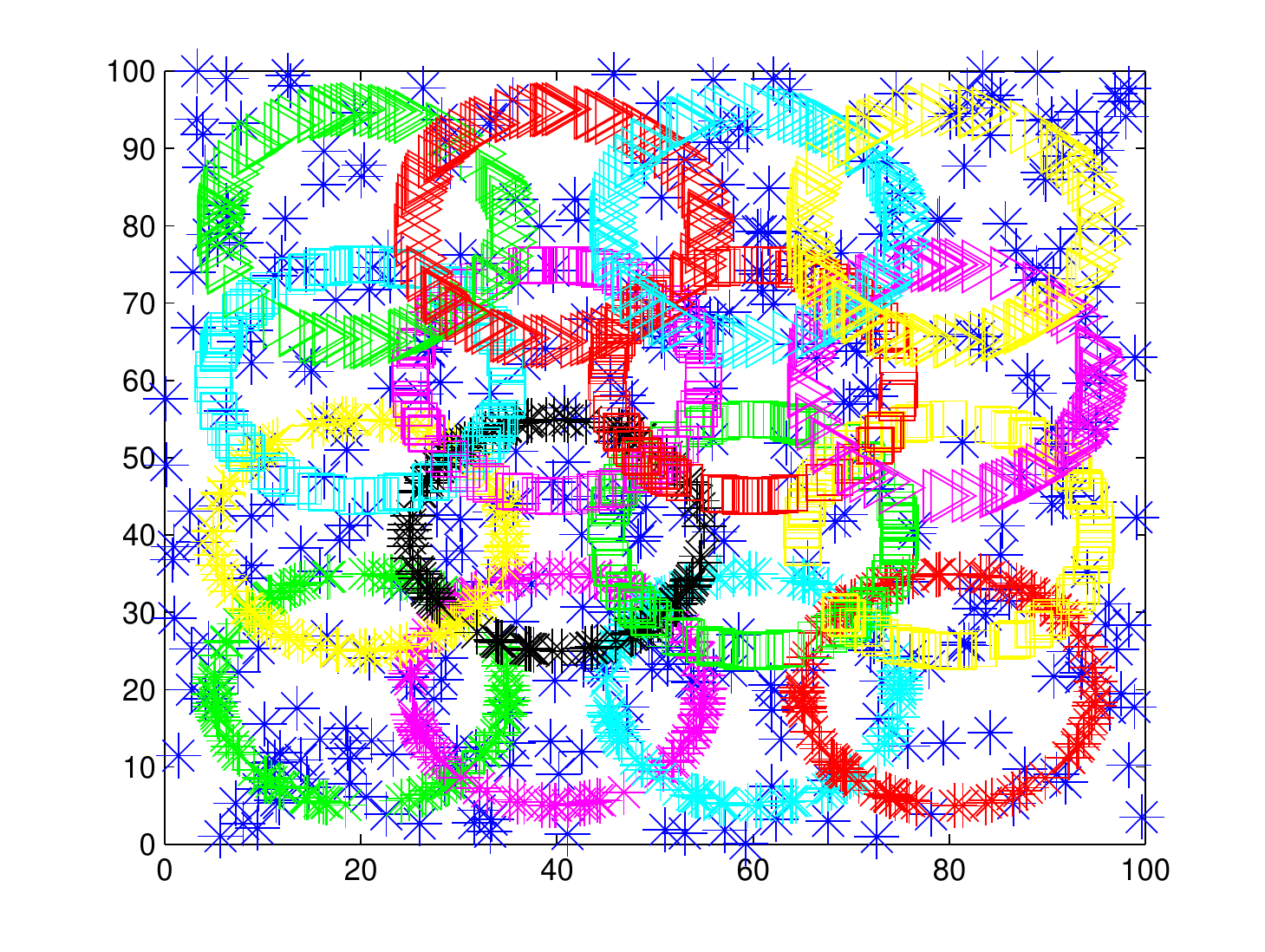}}
  \centerline{\footnotesize(a) Data }
\end{minipage}
\begin{minipage}[t]{.145\textwidth}

 \centerline{\includegraphics[width=1.10\textwidth]{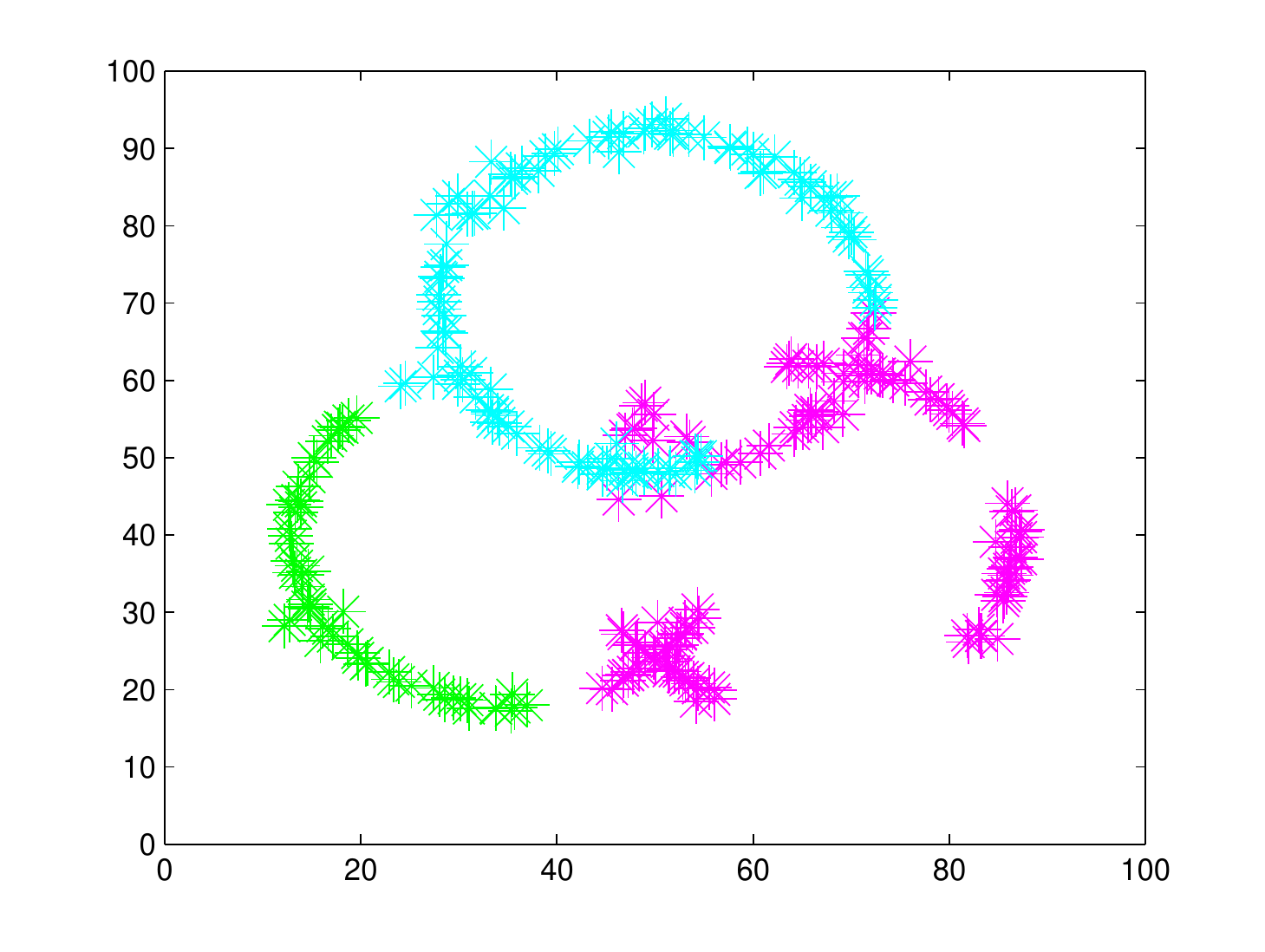}}
  \centerline{\includegraphics[width=1.100\textwidth]{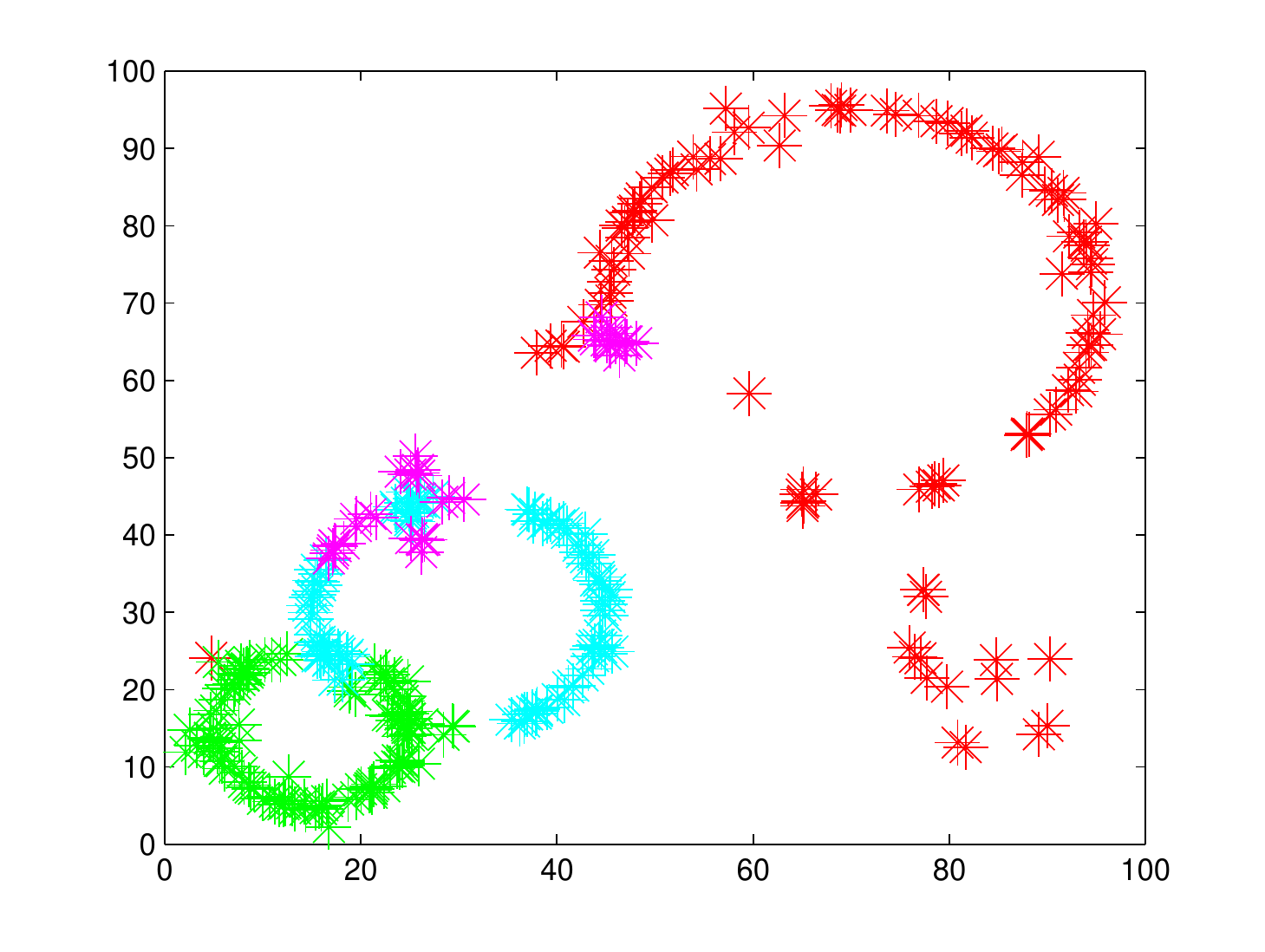}}
  \centerline{\includegraphics[width=1.10\textwidth]{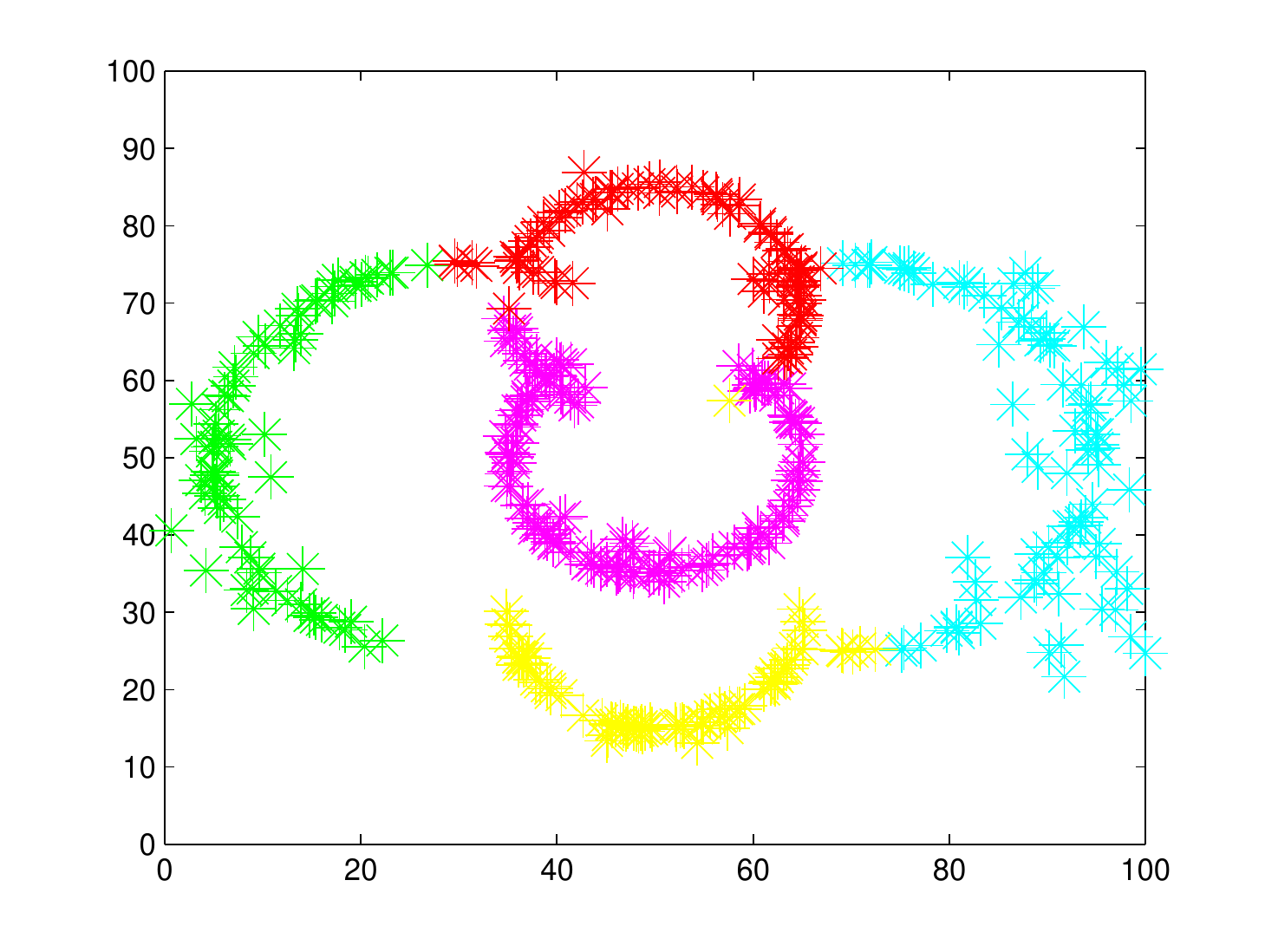}}
  \centerline{\includegraphics[width=1.10\textwidth]{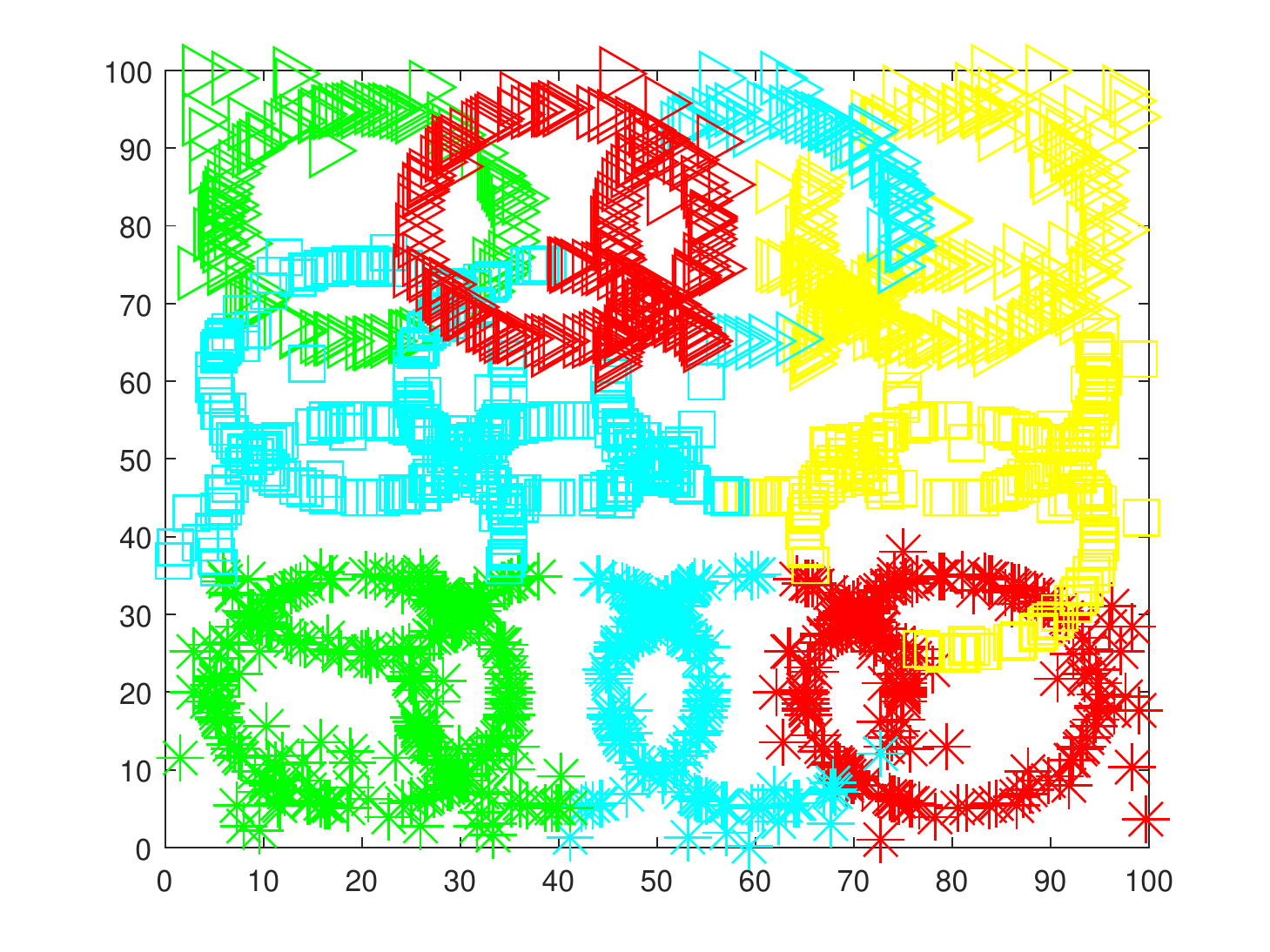}}
  \centerline{\footnotesize(b) KF }
\end{minipage}
\begin{minipage}[t]{.145\textwidth}
\centerline{\includegraphics[width=1.1\textwidth]{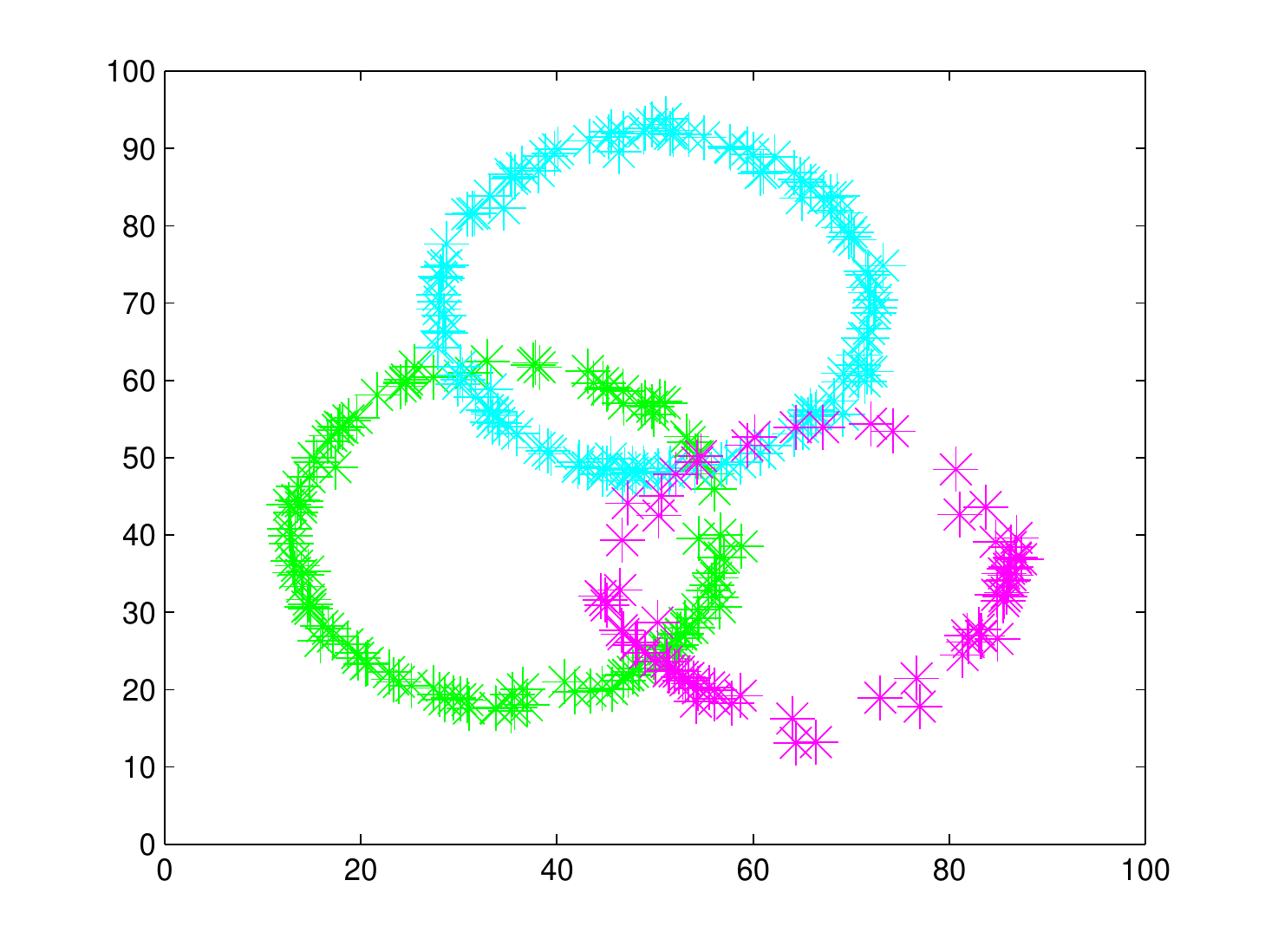}}
  \centerline{\includegraphics[width=1.10\textwidth]{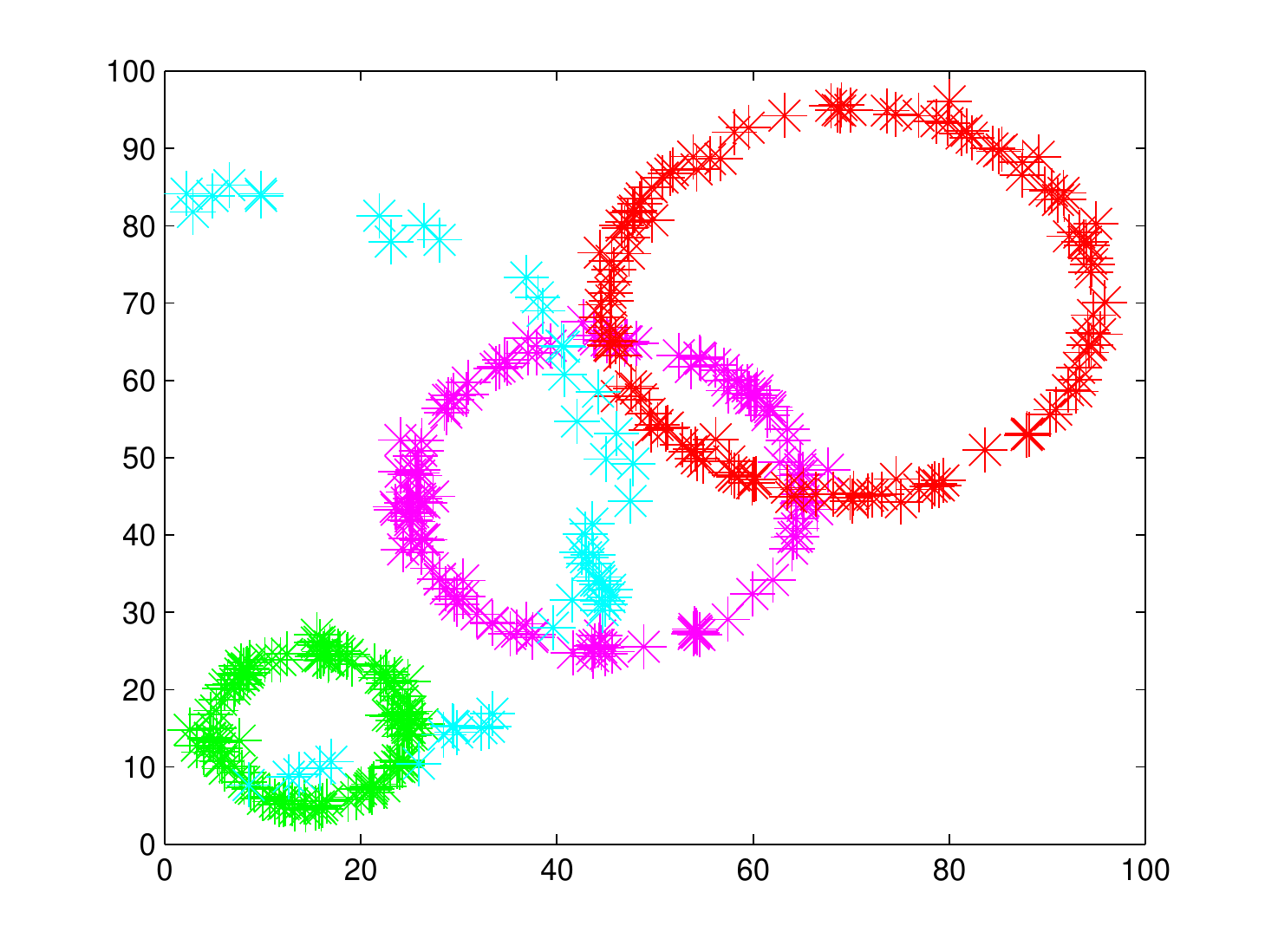}}
  \centerline{\includegraphics[width=1.10\textwidth]{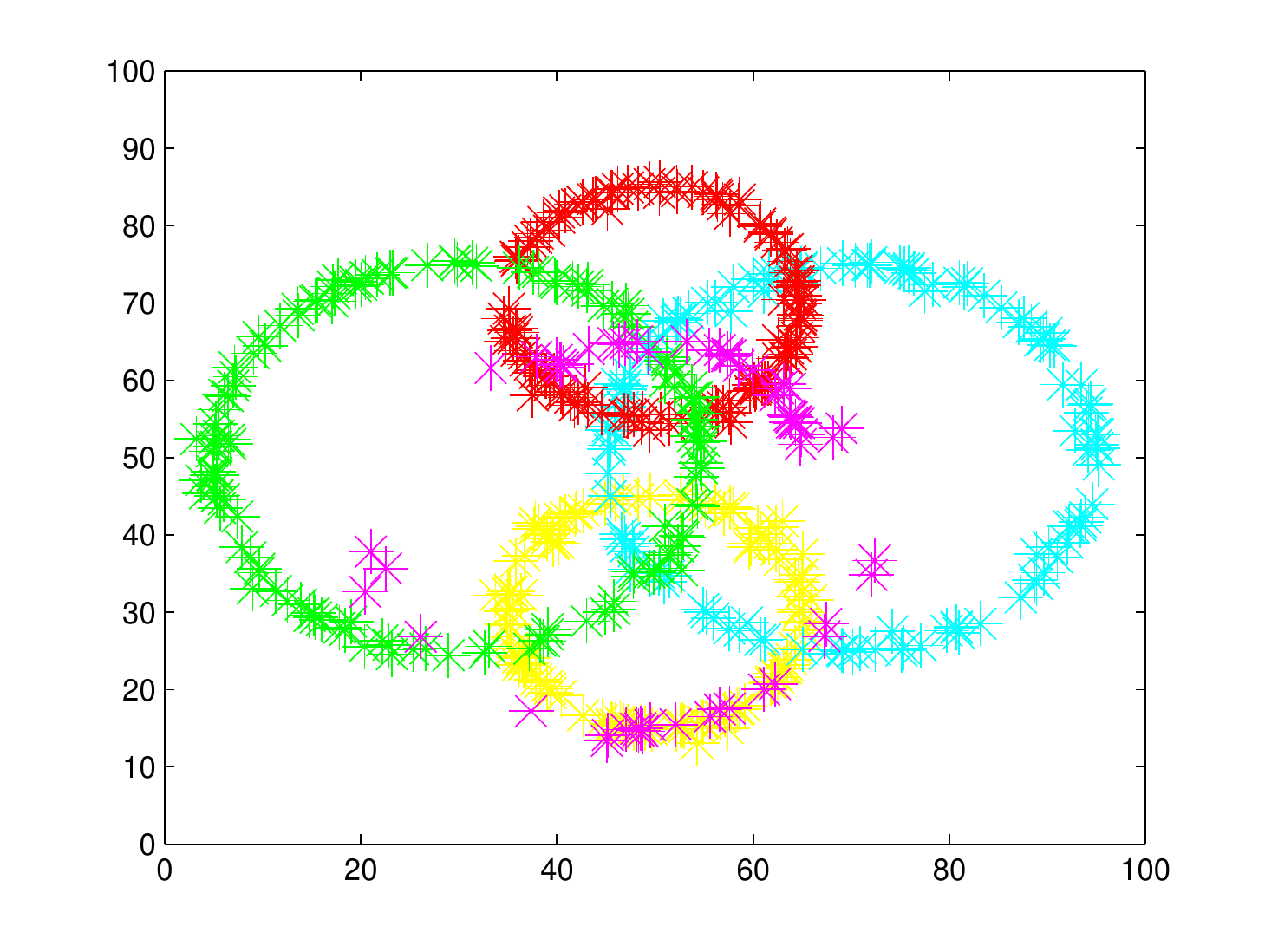}}
  \centerline{\includegraphics[width=1.10\textwidth]{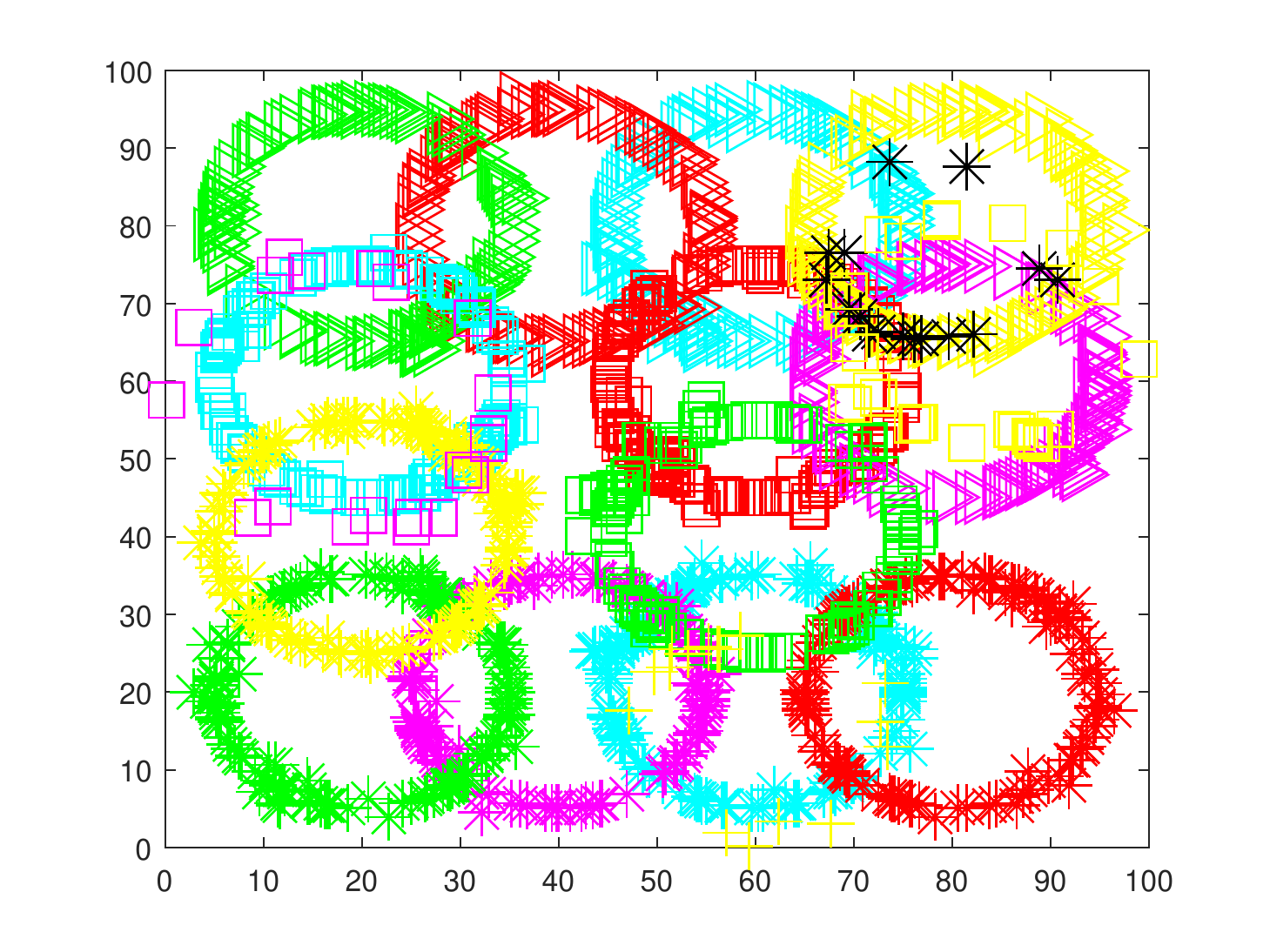}}
 \centerline{\footnotesize(c) RCG }
\end{minipage}
\begin{minipage}[t]{.145\textwidth}
\centerline{\includegraphics[width=1.10\textwidth]{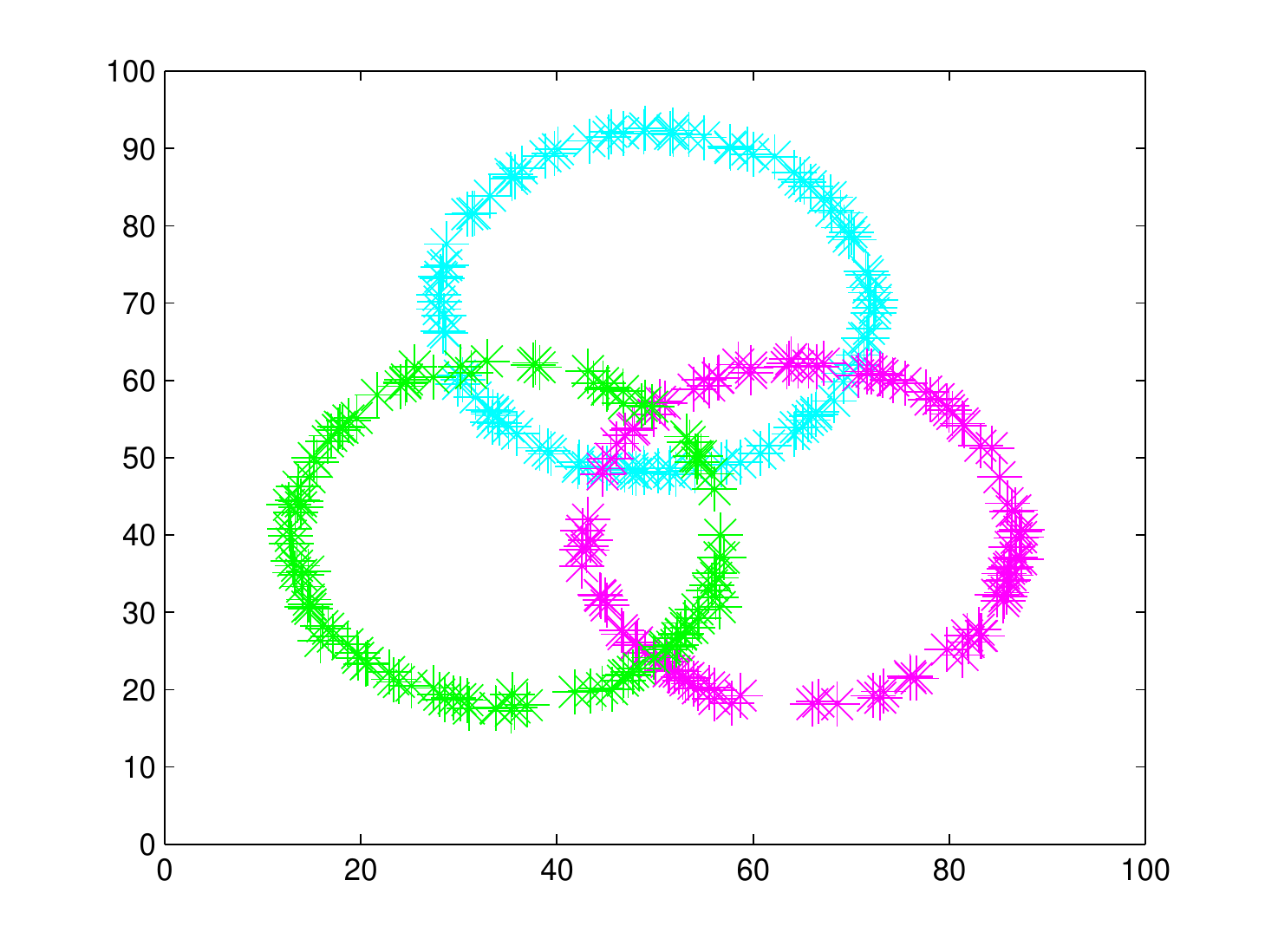}}
  \centerline{\includegraphics[width=1.10\textwidth]{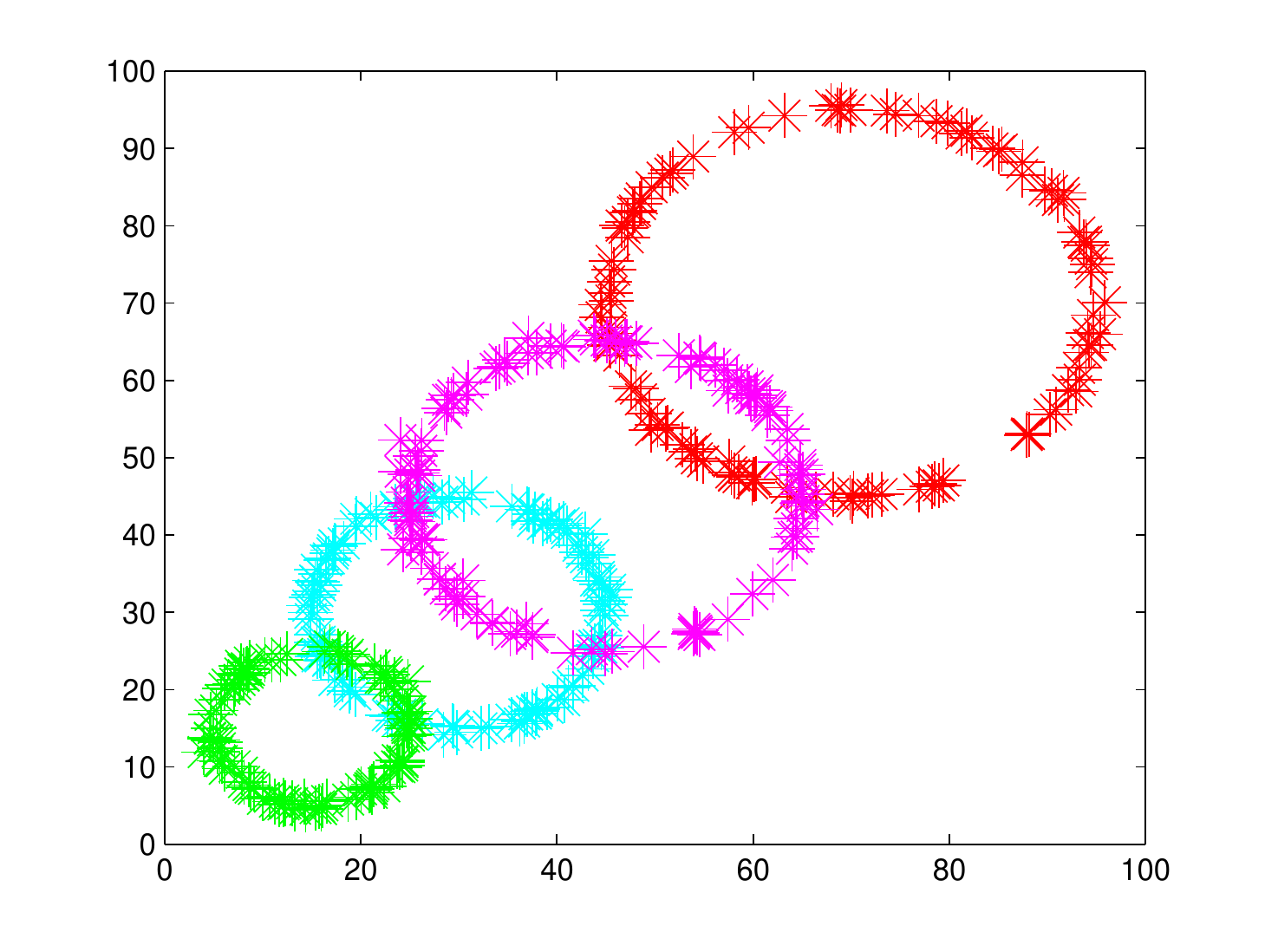}}
  \centerline{\includegraphics[width=1.10\textwidth]{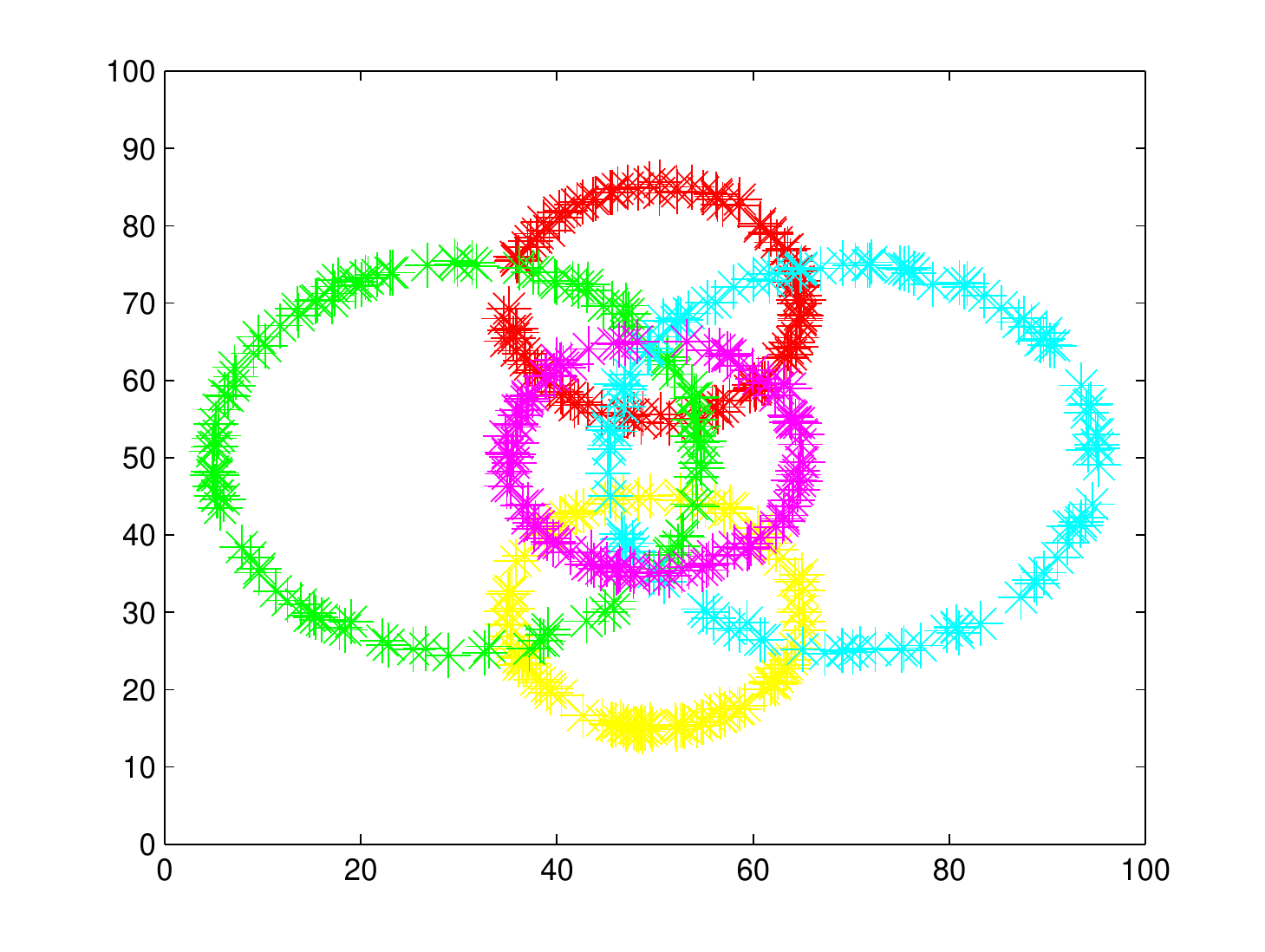}}
  \centerline{\includegraphics[width=1.10\textwidth]{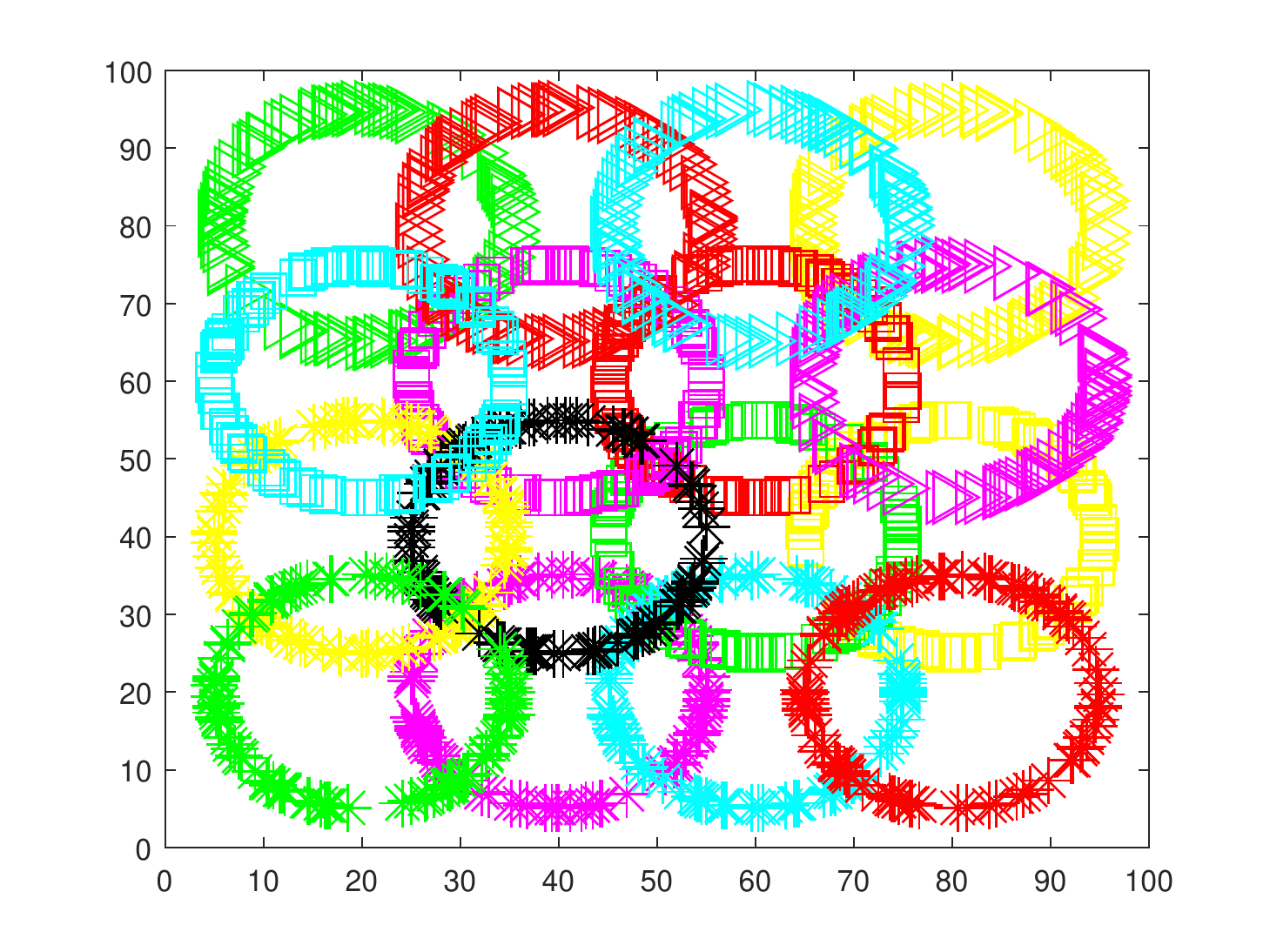}}
  \centerline{\footnotesize(d) AKSWH }
\end{minipage}
\begin{minipage}[t]{.145\textwidth}
\centerline{\includegraphics[width=1.10\textwidth]{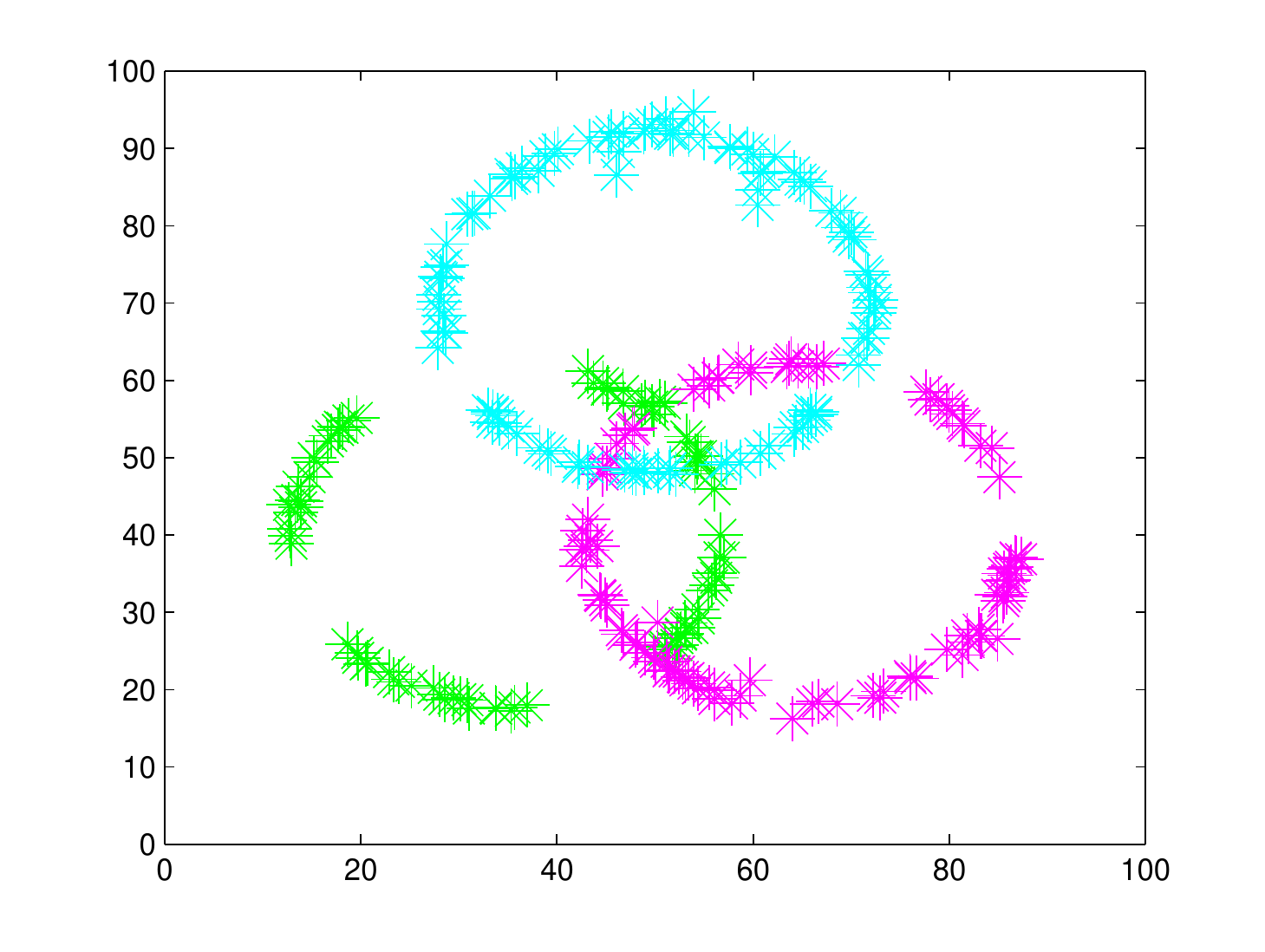}}
  \centerline{\includegraphics[width=1.10\textwidth]{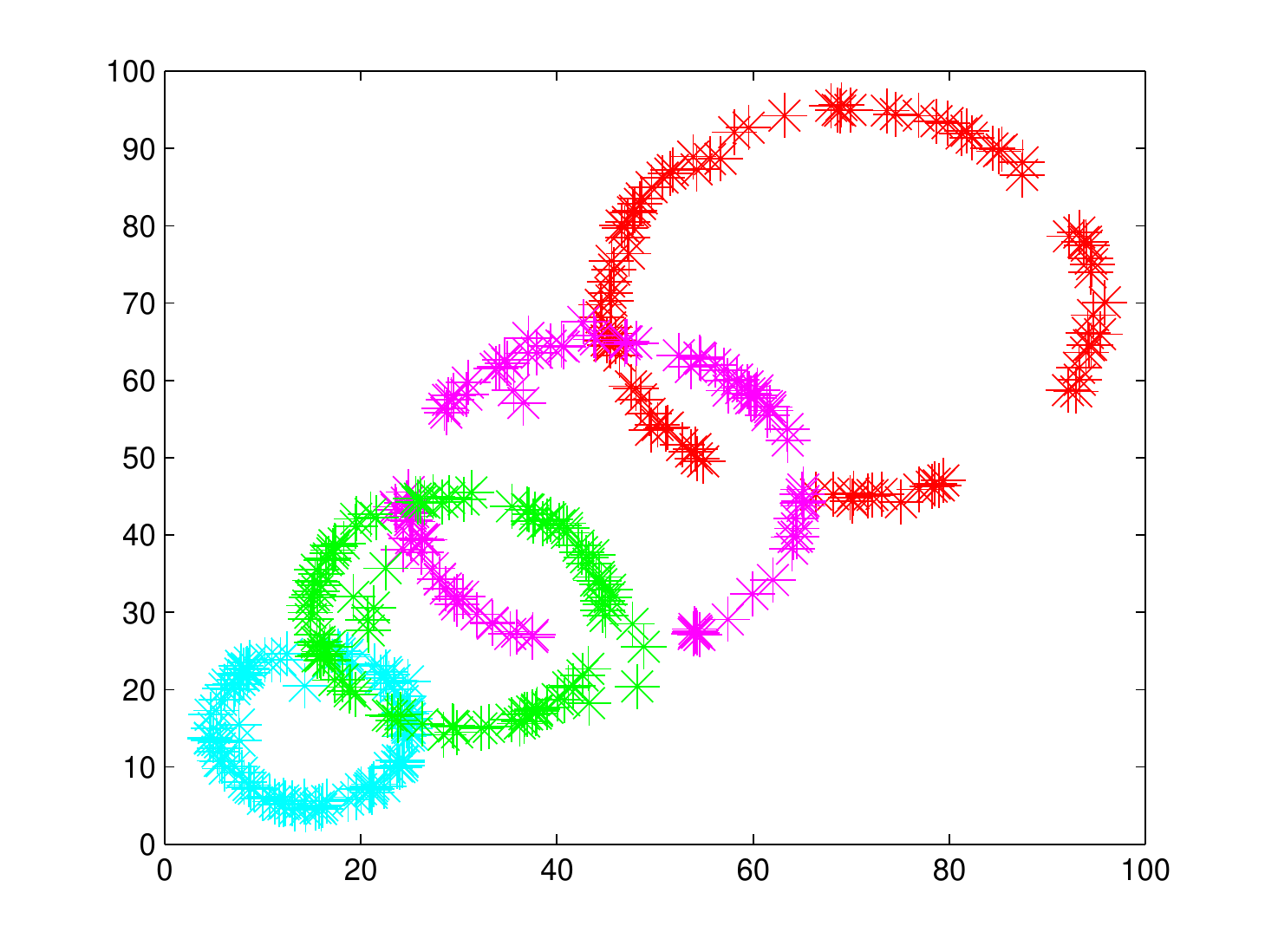}}
  \centerline{\includegraphics[width=1.10\textwidth]{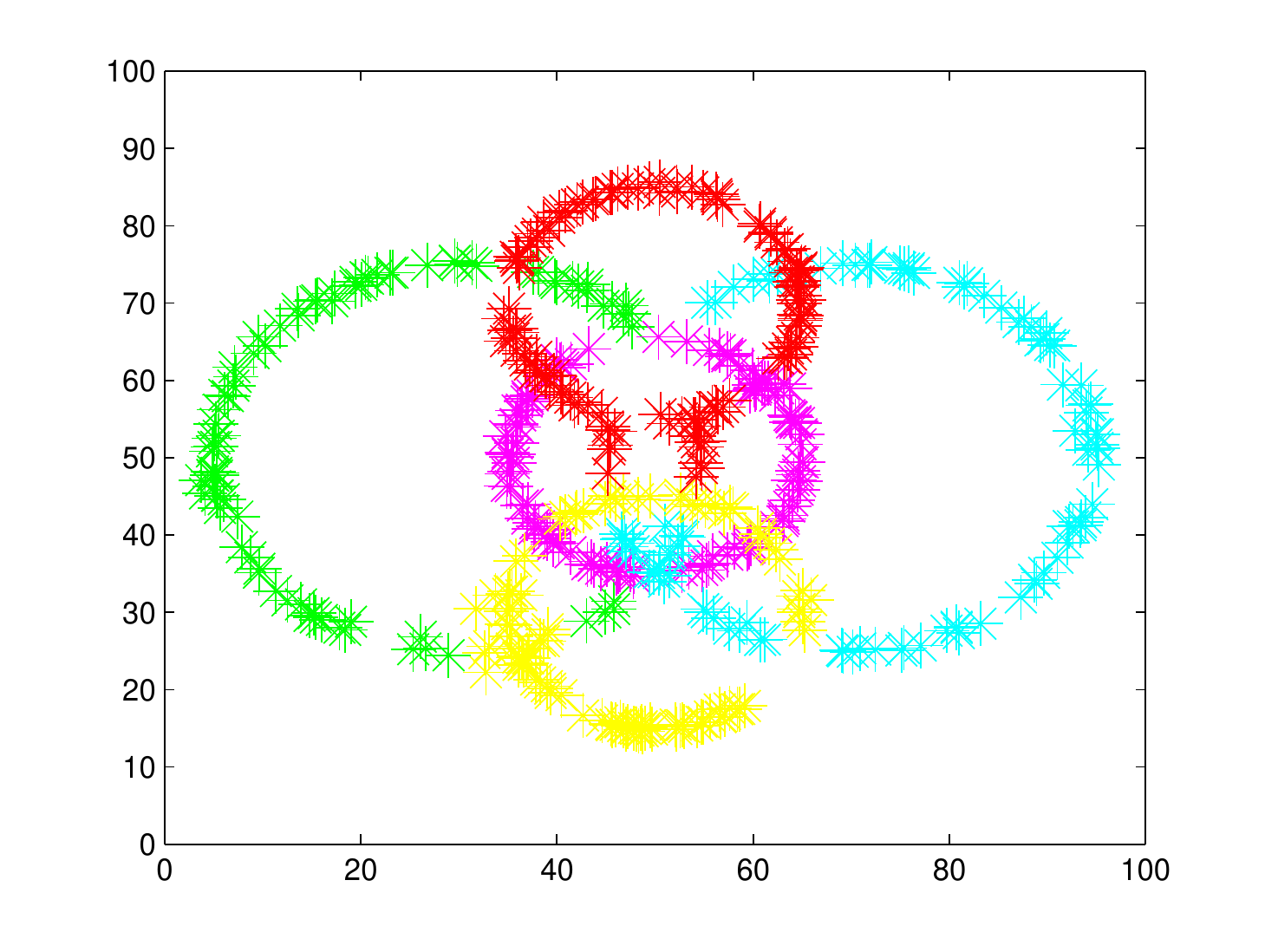}}
  \centerline{\includegraphics[width=1.10\textwidth]{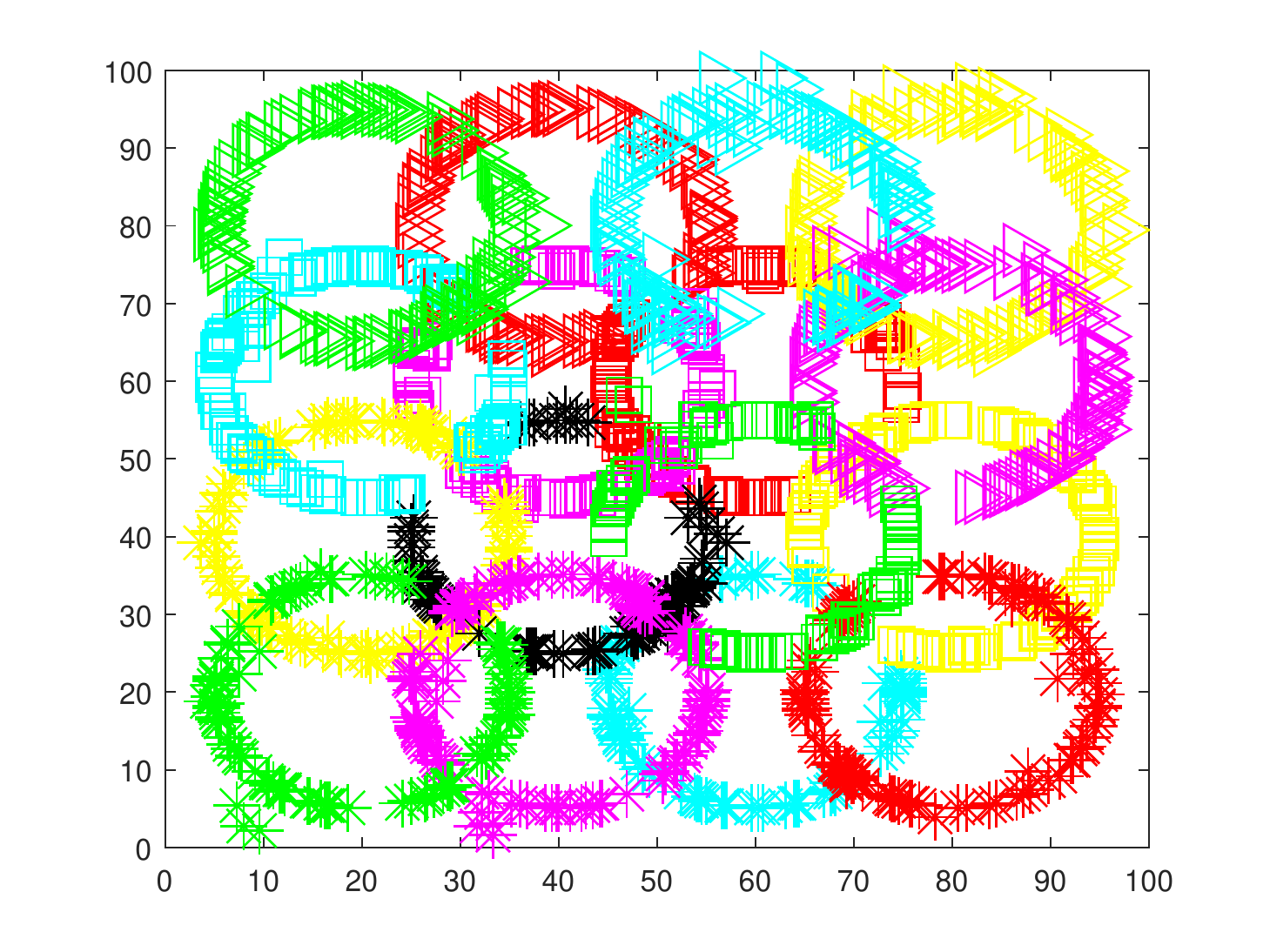}}
  \centerline{\footnotesize(e) T-linkage }
\end{minipage}
\begin{minipage}[t]{.145\textwidth}
  \centerline{\includegraphics[width=1.10\textwidth]{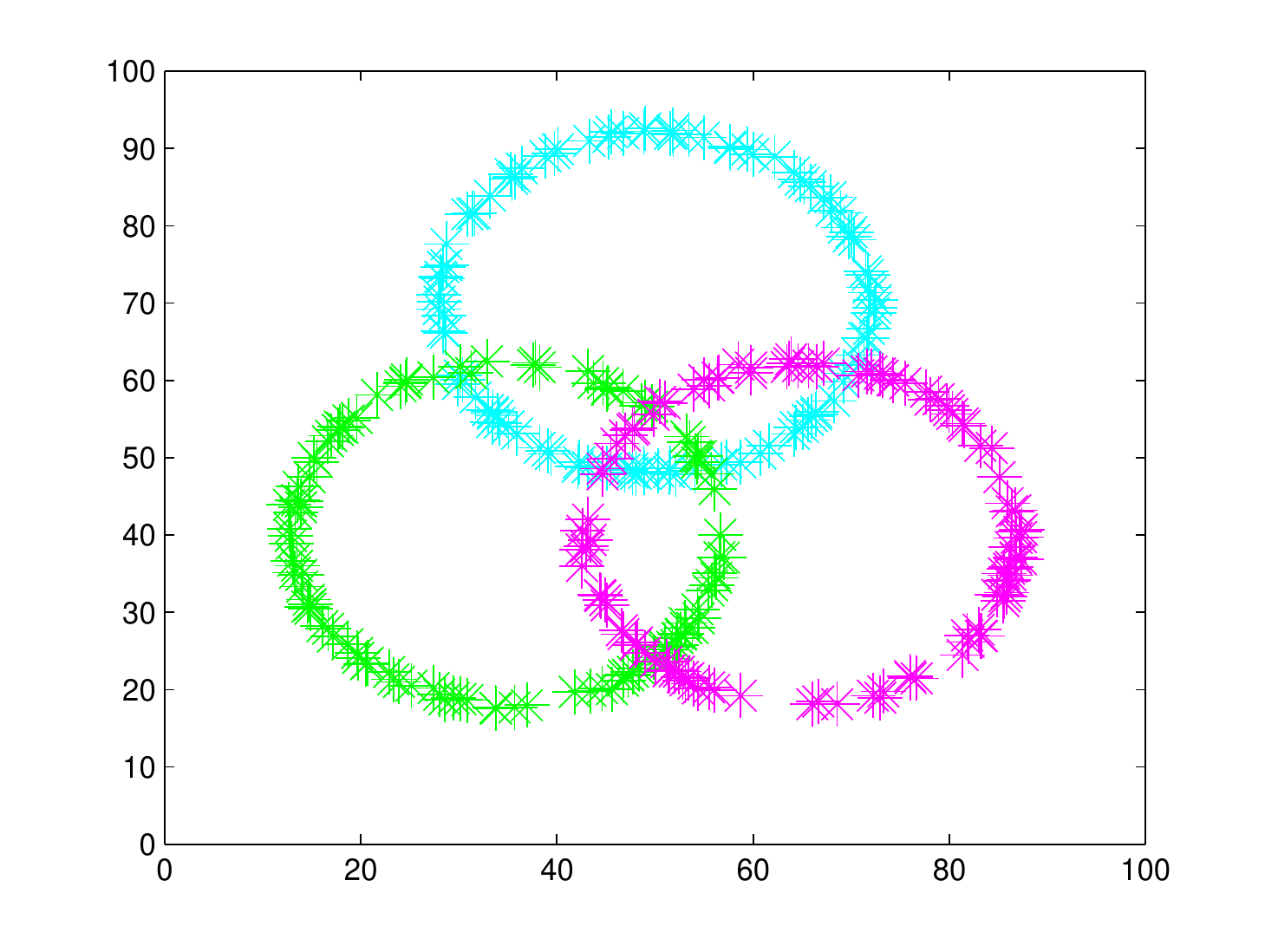}}
  \centerline{\includegraphics[width=1.10\textwidth]{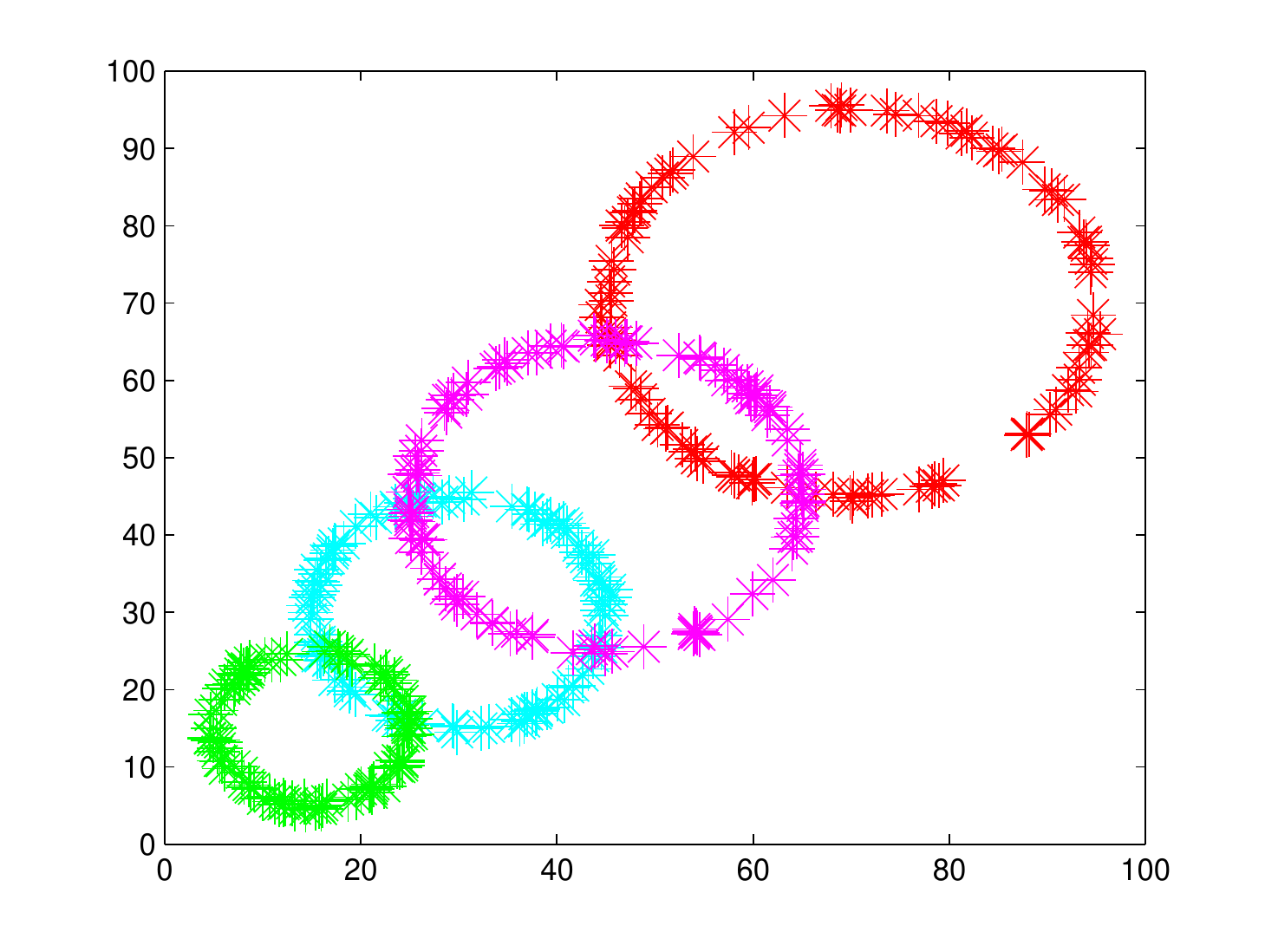}}
  \centerline{\includegraphics[width=1.10\textwidth]{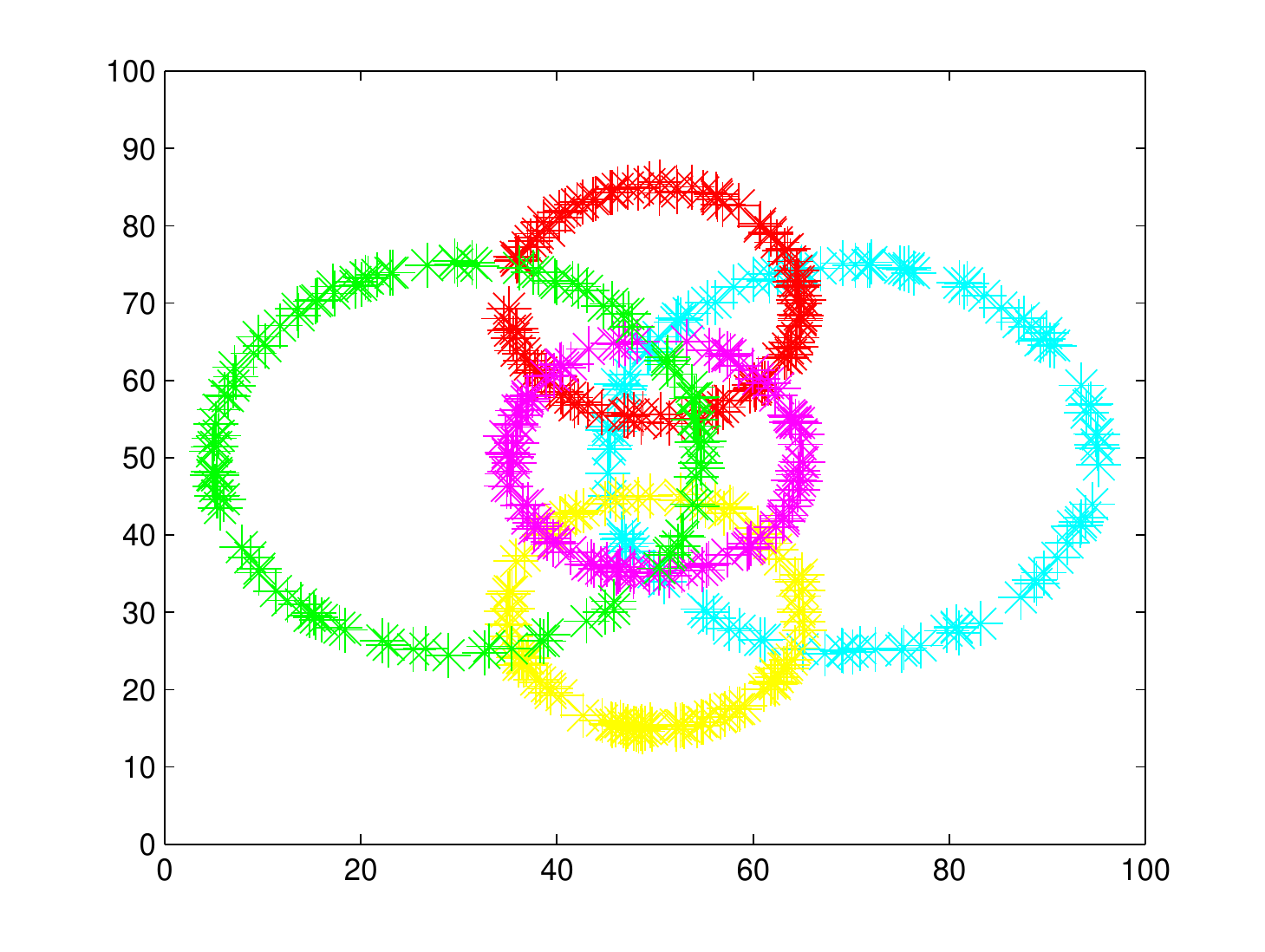}}
  \centerline{\includegraphics[width=1.10\textwidth]{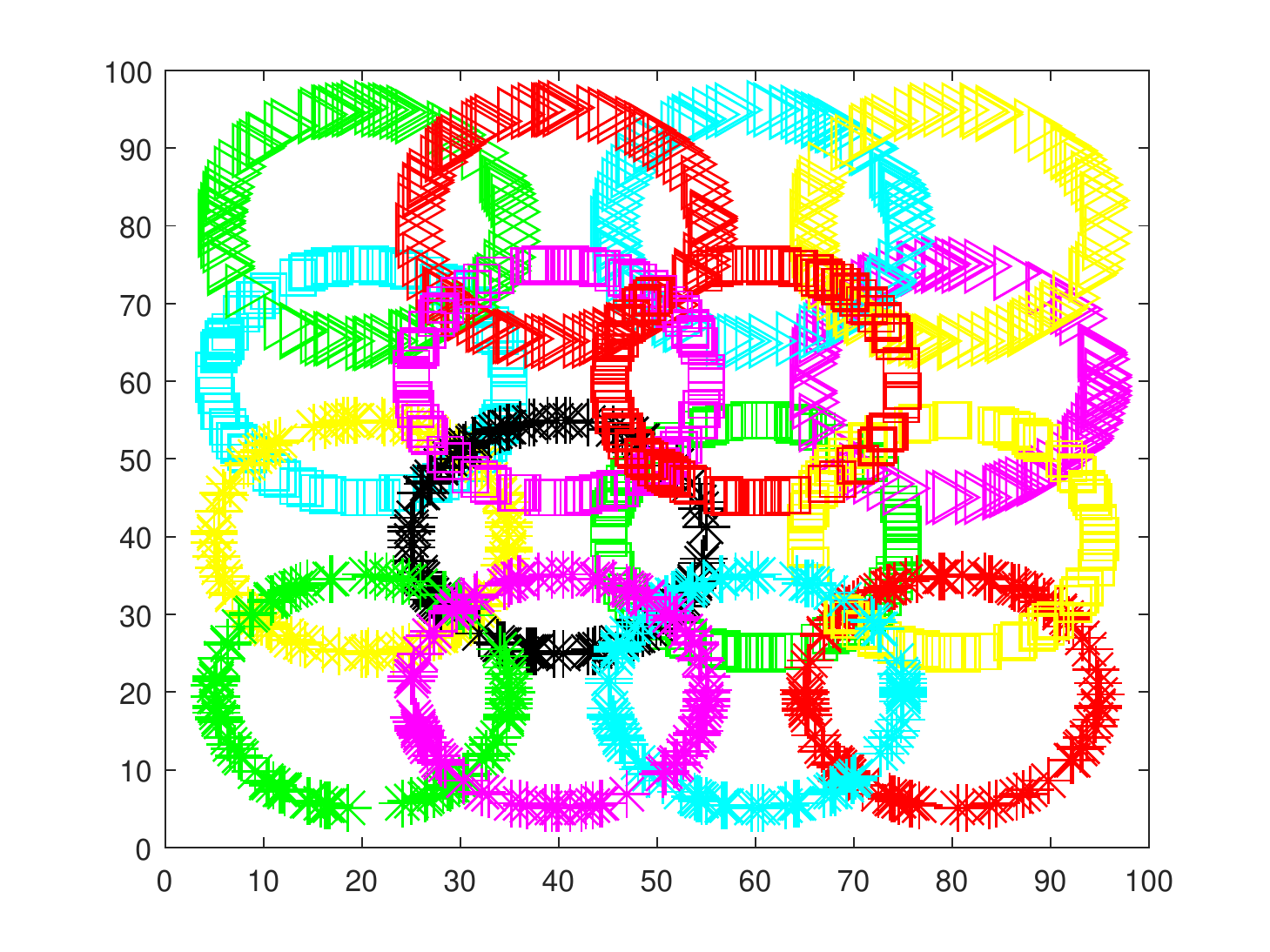}}
  \centerline{\footnotesize(f) MSHF2 }
\end{minipage}
\caption{Examples for circle fitting in the $2$D space. $1^{st}$ to $4^{th}$ rows respectively fit three, four, five and sixteen circles. {The inlier noise scale is set to $0.5$ and each circle has $100$ inliers. Each data includes $400$ outliers.} We do not show the results of MSH/MSHF1, which are similar to those of MSHF2, due to the space limit.}
\label{fig:circlefitting}
\end{figure*}
{We also evaluate the performance obtained by the seven fitting methods for the data with different cardinality ratios between the inliers of each line, to show the ability of the model fitting methods to deal with unbalanced data. We use the ``three lines" data from Fig.~\ref{fig:fivelines} for evaluation since all competing methods can successfully estimate the three lines when the cardinality ratio is low. We set the {inlier numbers of the three lines} to be the same at the beginning, and we gradually increase the inlier numbers of two lines while reducing the inlier number of the third line to make the cardinality ratios between the inliers of lines increase from $1.0$ to $8.0$. We repeat each experiment $20$ times and show the standard variances and the average fitting errors in Fig.~\ref{fig:diffratio}.

From Fig.~\ref{fig:diffratio}, we can see that MSHF1/MSHF2 achieve the same results and both achieve low standard variances and average fitting errors for the ``three lines" data with different inlier cardinality ratios. In contrast, KF, RCG and T-linkage achieve low average fitting errors when the cardinality ratio of inliers is smaller than $2.0$, but they begin to break down when the inlier cardinality ratios are larger than $2.0$, $3.0$ and $3.0$, respectively. AKSWH and MSH obtain large fitting errors when the inlier cardinality ratios are larger than $2.0$ and $7.0$, respectively. As a result, the parameter space based methods (i.e., AKSWH, MSH and MSHF1/MSHF2) show better performance than the other competing fitting methods for the unbalanced data.
}

\subsubsection{Circle Fitting}
\label{sec:sycirclefitting}
\begin{table}[ht]
  \caption{Quantitative comparison results of circle fitting on four synthetic {data}. }
\centering
\scalebox{0.96}{\tabcolsep0.07in
\begin{tabular}{|c|c|c|c|c|c|>{\columncolor{mygray}}c|>{\columncolor{mygray}}c|>{\columncolor{mygray}}c|}
\hline
Data        &                  & M1&M2&M3&M4&M5&M6 &{M7}\\
\hline
\hline
               &Std.    &{3.79}  & {3.93}& {2.78} & {2.04}& {6.33}& {\bf0.63} & {\bf0.63}\\
   3          &Avg.   & {27.85} & {24.01} & {4.63}& {15.50}& {3.07}& {\bf1.64}& {\bf1.64}\\
   circles  &Min.  & {22.00}& {17.28} & {2.00}& {13.00}& {\bf0.57}& {\bf0.57}& {\bf0.57}\\
               &Time   &{51.66}&{\bf1.50} &{2.91}&{164.27}& {2.49}& {3.70}& {1.76}\\
                       \hline
              &Std.    & {4.38}& {5.65}& {5.62}& {4.53}& {0.68} & {\bf0.60} & {\bf0.60}  \\
   4        &Avg.    & {32.79}& {23.98}& {9.12}& {16.62}& {2.46} & {\bf2.14}& {\bf2.14}  \\
   circles&Min.  & {26.25}& {18.12}& {3.75}& {10.37}& {1.50}& {\bf1.37}& {\bf1.37}\\
             &Time   &{64.95}&{2.21}&{2.00}&{235.34} &{2.91} &{3.03} &{\bf1.48} \\
                       \hline
               &Std.  & {3.74}& {4.87}& {1.87}& {5.54}& {6.34}  & {\bf1.14}& {\bf1.14}  \\
   5         &Avg.   & {36.17} & {24.78} & {6.50} & {18.86} & {4.72}& {\bf2.95}& {\bf2.95} \\
   circles &Min.  & {31.11}& {18.22}& {3.33}& {13.88}& {1.55}& {\bf1.44}& {\bf1.44} \\
             &Time    &{70.94}&{2.80}&{3.40}&{287.94}&{4.54}&{4.72}&{\bf1.83}\\
                       \hline
             &Std.    & {3.00}&{6.30}& {7.58}& {1.62} & {3.83} & {\bf1.56}  & {\bf1.56}\\
   16     &Avg.    & {57.92} & {32.72}& {14.46}& {24.75}& {4.46}& {\bf3.59}  & {\bf3.59}\\
circles  &Min.  & {54.35}& {23.70}& {\bf1.55}& {23.15}& {2.25}& {1.90}& {1.90}\\
             &Time   &{254.43}&{7.38}&{3.97}&{1589.13}&{7.75} &{9.46}&{\bf1.47}\\
                       \hline
\end{tabular}}
 \label{table:circlefittingtable}
\end{table}
We further evaluate the performance of the {seven} fitting methods on circle fitting using four challenging synthetic data in the $2$D space (see Fig.~\ref{fig:circlefitting}). We repeat the experiment $50$ times and report the standard variances, the average and the best results of the fitting errors {(in percentage)} and the average CPU time (in seconds) obtained by the {seven} competing methods, in Table~\ref{table:circlefittingtable} (we exclude the time used for sampling and generating potential hypotheses for all the fitting methods). We also show the corresponding fitting results obtained by all the competing methods from Fig.~\ref{fig:circlefitting}(b) to Fig.~\ref{fig:circlefitting}(f).

From Fig.~\ref{fig:circlefitting} and Table~\ref{table:circlefittingtable}, we can see that: (1) For the ``three circles" data, the three circles with the same diameter {intersect each other}. All of the {seven} competing methods can correctly estimate the number of circles in the data. However, {MSH and MSHF1/MSHF2} achieve the {top-three} lowest average and minimal fitting errors among all competing methods. {And MSHF1/MSHF2} are the most stable methods ({achieving} the lowest standard deviation of fitting errors). This is because that {MSHF1/MSHF2} can keep the representative modes corresponding to model instances in most cases, while MSH may remove some representative modes in some cases, which will increase the average fitting errors. AKSWH can also achieve a low average fitting error with the third lowest standard deviation. In contrast, KF, RCG and T-linkage {obtain} high average fitting errors. KF removes some inliers during the procedure of outlier removal. RCG cannot effectively find the representative sub-graphs corresponding to the model instances. T-linkage cannot segment the data points of intersection with high accuracy. (2) For the ``four circles" data, the four circles with different {diameters intersect} each other. {MSH and MSHF1/MSHF2} achieve the lowest average and minimal fitting errors again. {Among the other four competing methods,} AKSWH and T-linkage successfully estimate all four circles, while KF and RCG miss one of the circles. (3) For the ``five circles" data, the five circles with different {diameters} intersect together. All of AKSWH, T-linkage, {MSH and MSHF1/MSHF2} successfully estimate {the} five circles, but {MSHF1/MSHF2} achieve the best {results of} the standard deviation, {the} average and minimal fitting errors. In contrast, KF and RCG {obtain} high fitting errors. (4) For the ``sixteen circles" data, the sixteen circles with the same diameter intersect together. This data include a large number of model instances. {MSHF1/MSHF2} can effectively estimate the number of model instances. MSH wrongly estimates the number during the repeating experiments in $1$ out of $50$ times. {However,} MSH still achieves a low average fitting error. In contrast, KF, RCG, AKSWH and T-linkage cannot achieve low average fitting errors, especially for KF, which fails to estimate the number of model instances in most cases.

{For the performance of computational time, MSHF2 achieves the fastest speed among the seven fitting methods for three data (i.e., the ``four circles", ``five circles" and ``sixteen circles" data). AKSWH and MSH/MSHF1 achieve the similar computational speed for the ``three circles", ``four circles" and ``five circles" data. KF and T-linkage are relatively slow. Especially for T-linkage which takes $1,589.13$ seconds for the ``sixteen circles" data, it is more than $1,000$ times slower than the proposed MSHF2. Among MSH and MSHF1/MSHF2, MSHF2 is significantly faster than MSH and MSHF1 (MSHF2 is about $1.41$$\sim$$5.27$ and $2.04$$\sim$$6.44$ times faster than MSH and MSHF1, respectively). This is because MSHF2 uses the neighboring constraint to reduce the computational cost. {RCG achieves fast speed for the ``three circles", ``four circles" and ``five circles" data, but obtains slow speed for the ``sixteen circles" data due to the large number of data points.}}
\begin{figure*}[t]
\centering
\begin{minipage}[t]{.15\textwidth}
  \centering
  \centerline{\includegraphics[width=1.0\textwidth]{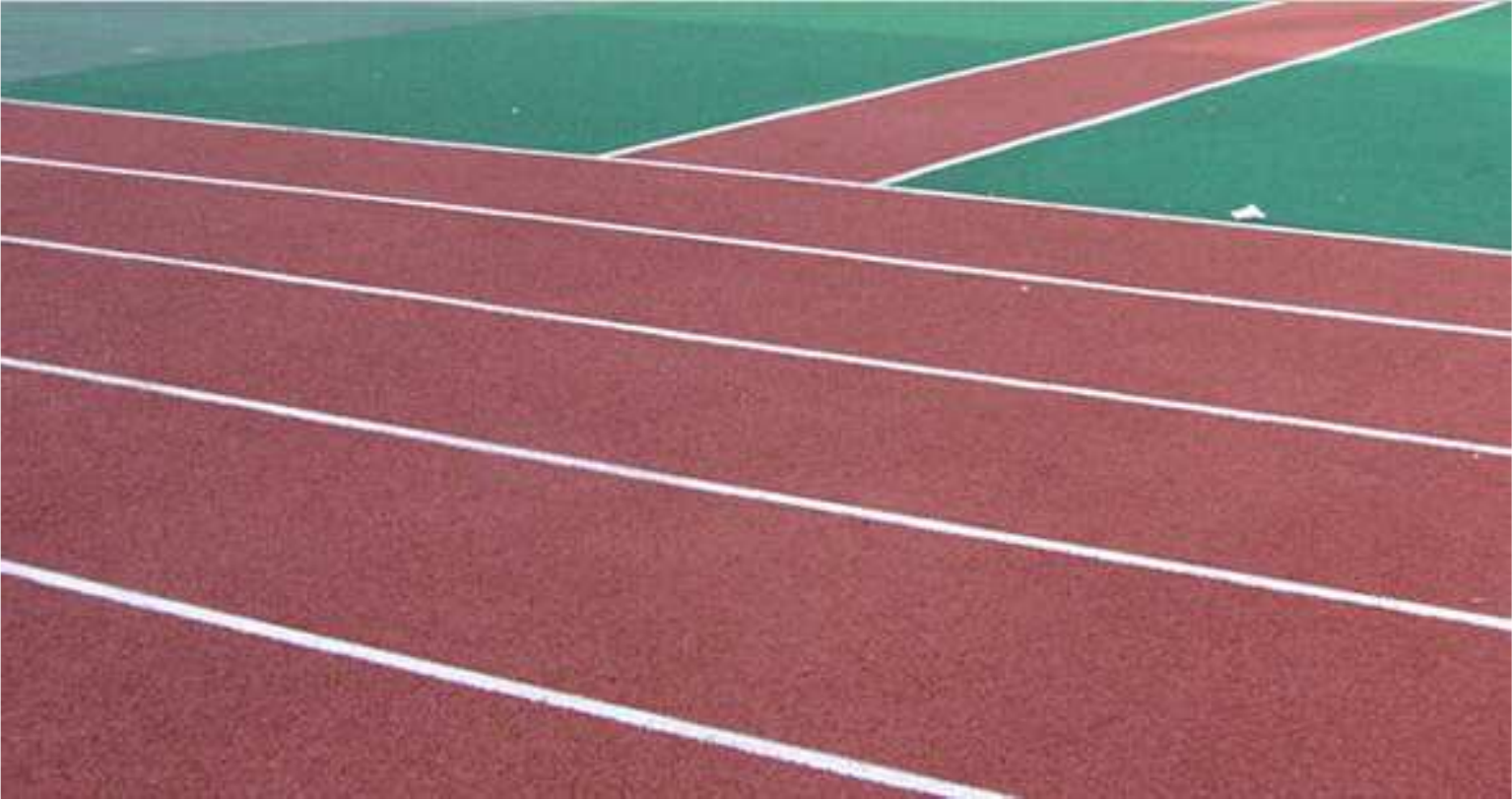}}
  \centerline{}
  \centerline{\includegraphics[width=1.0\textwidth]{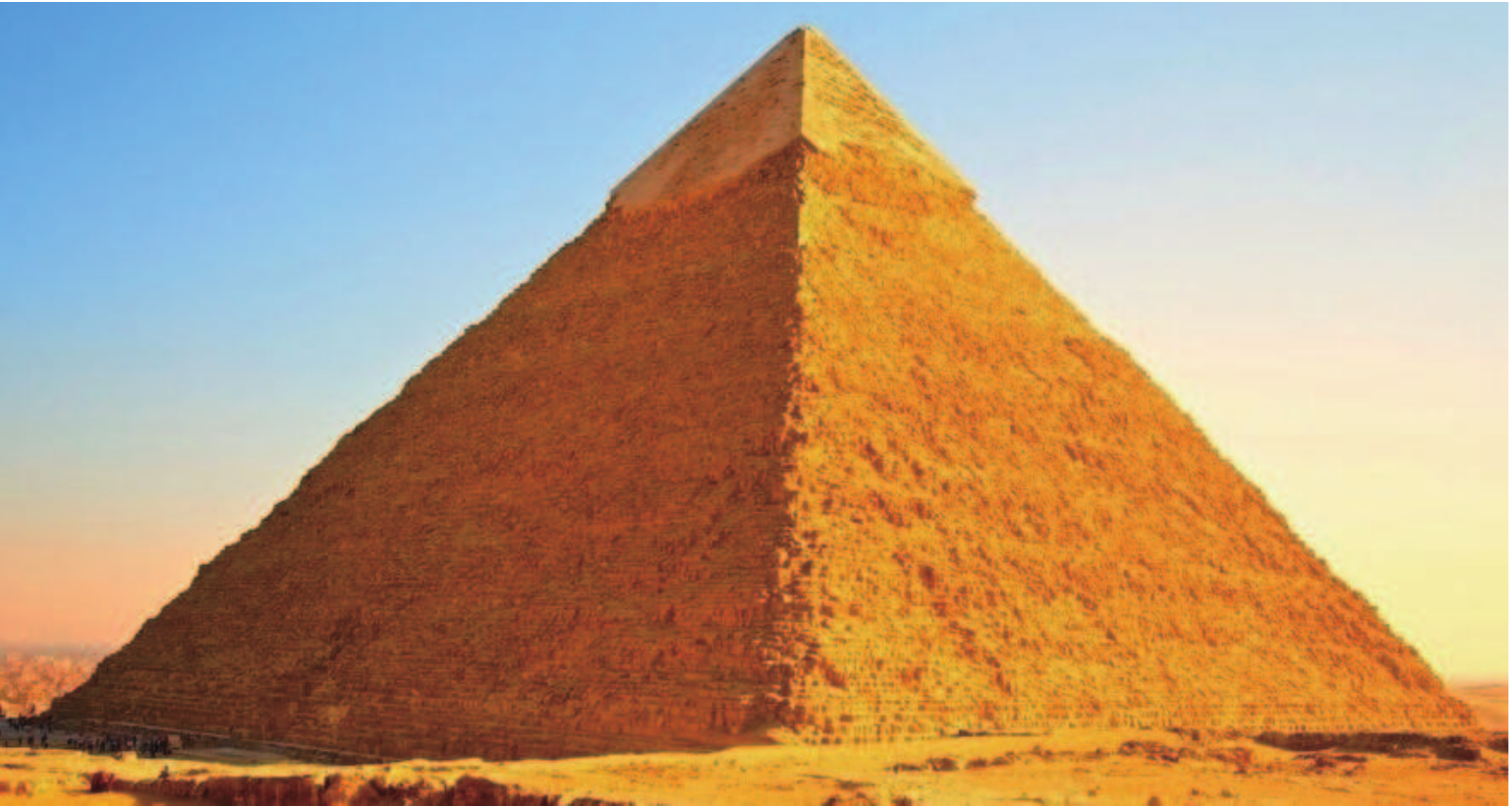}}
 \centerline{\footnotesize(a) Data }
\end{minipage}
\begin{minipage}[t]{.15\textwidth}
  \centering
  \centerline{\includegraphics[width=1.0\textwidth]{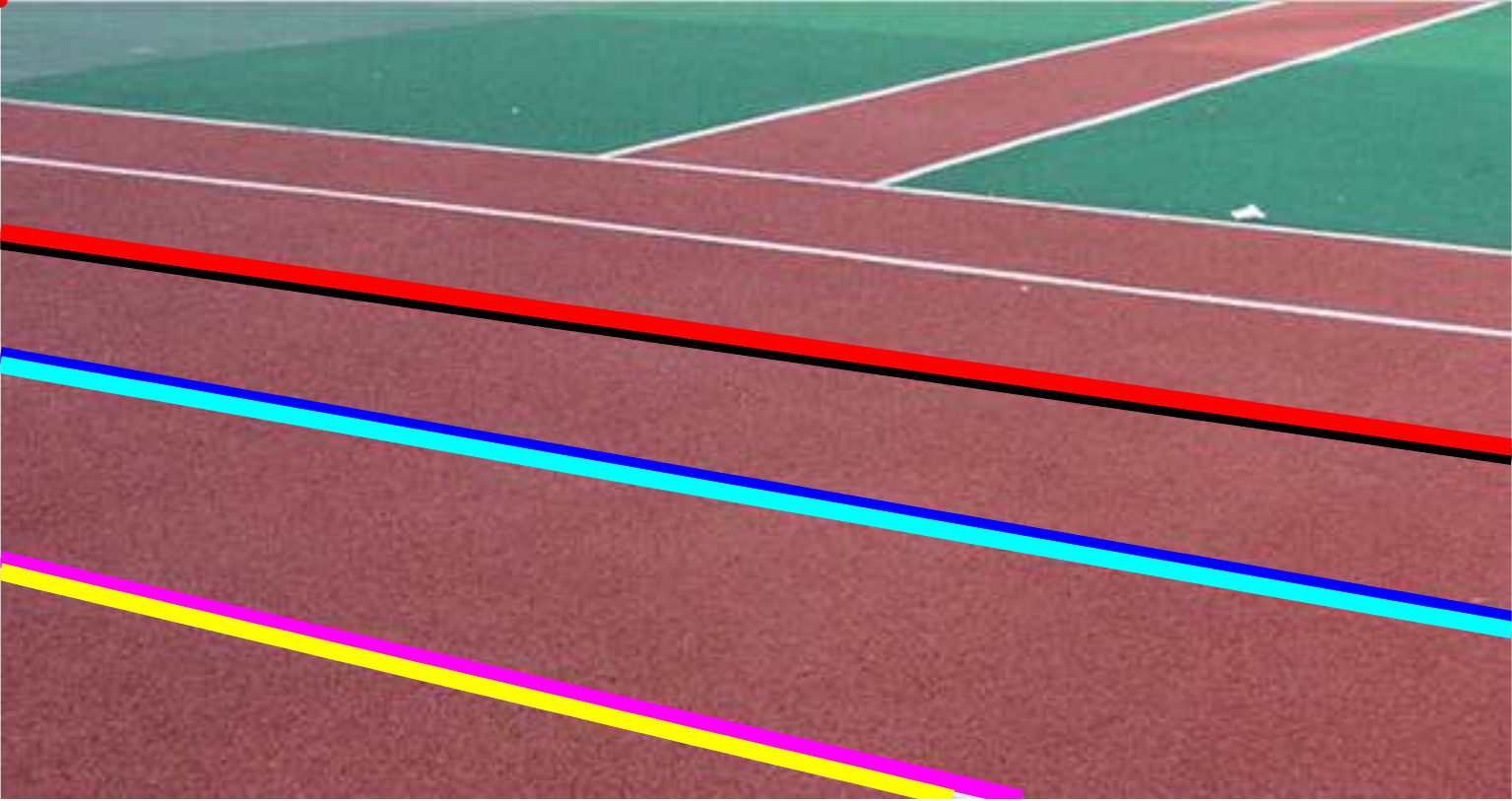}}
  \centerline{}
  \centerline{\includegraphics[width=1.0\textwidth]{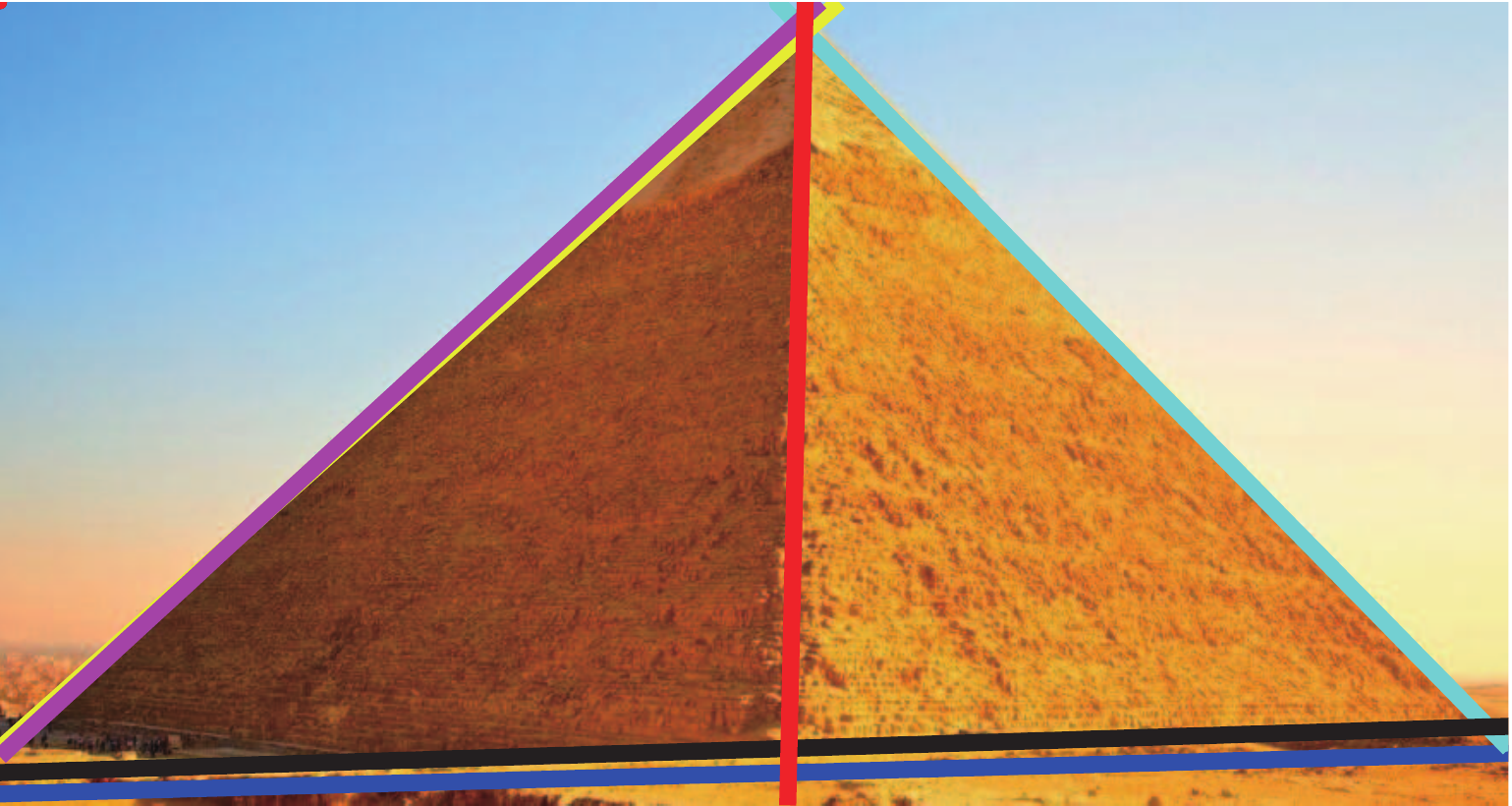}}
\centerline{\footnotesize(b) KF }
\end{minipage}
\begin{minipage}[t]{.15\textwidth}
  \centering
  \centerline{\includegraphics[width=1.0\textwidth]{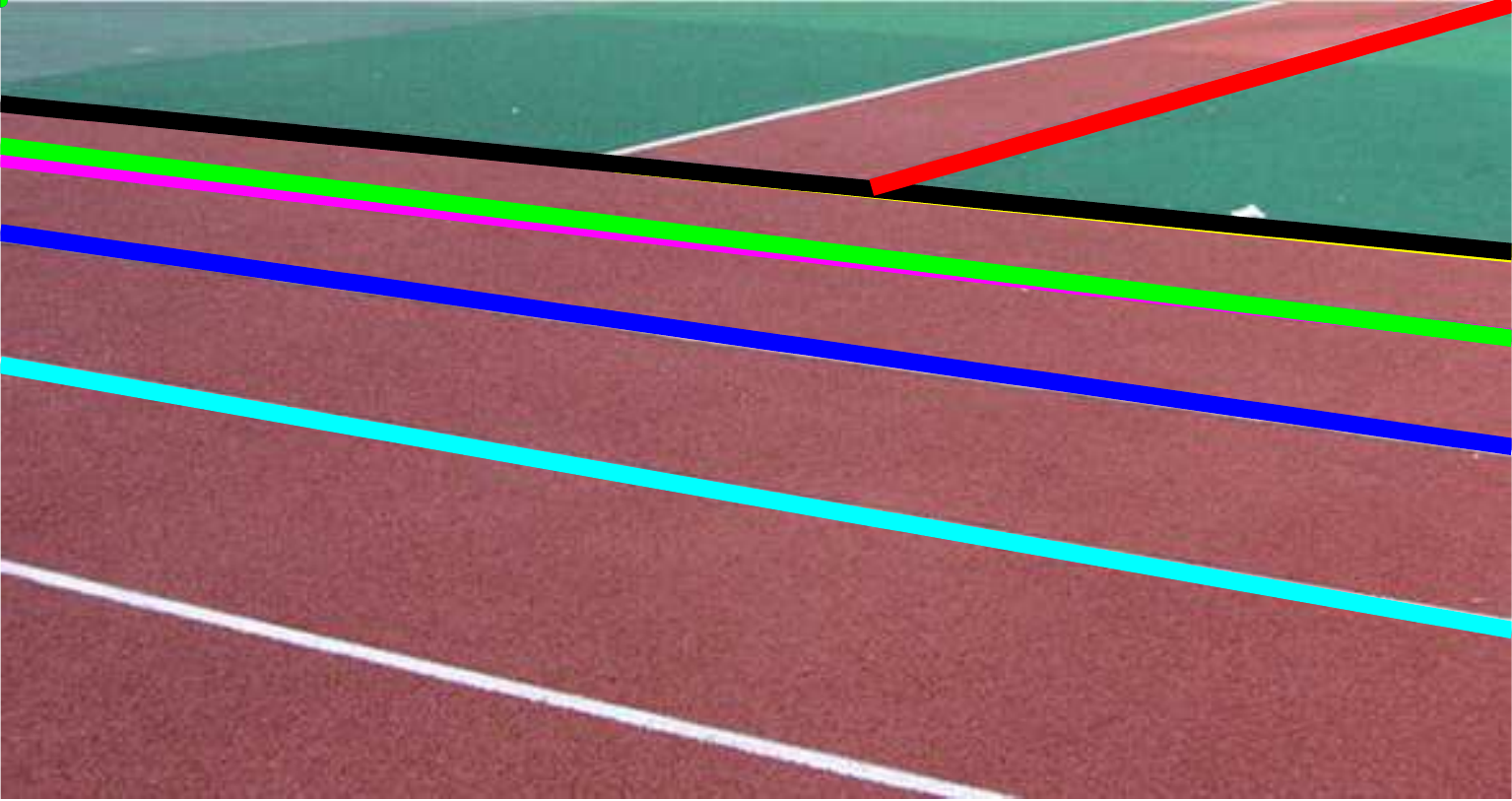}}
  \centerline{}
  \centerline{\includegraphics[width=1.0\textwidth]{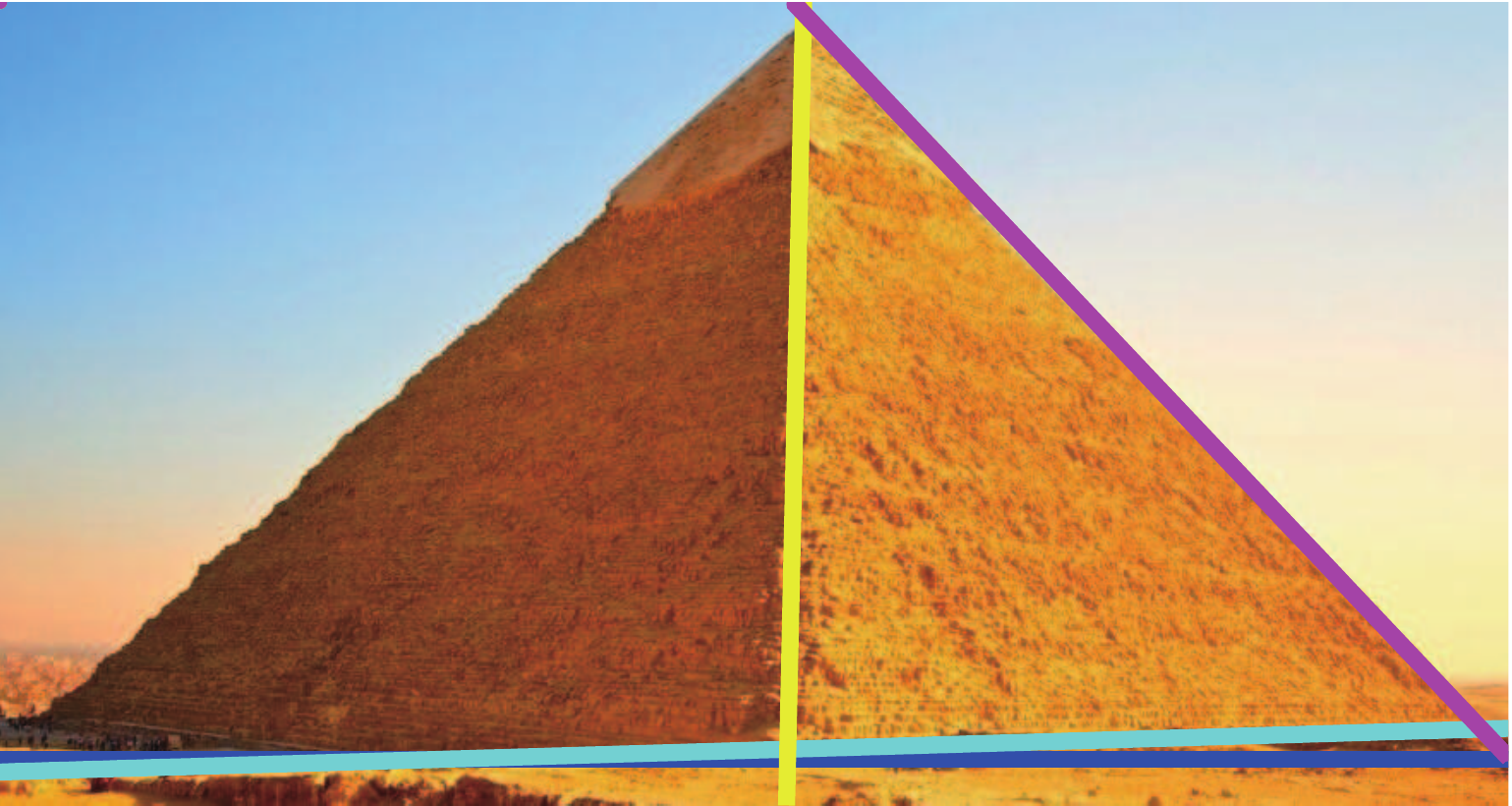}}
\centerline{\footnotesize (c) RCG}
\end{minipage}
\begin{minipage}[t]{.15\textwidth}
  \centering
  \centerline{\includegraphics[width=1.0\textwidth]{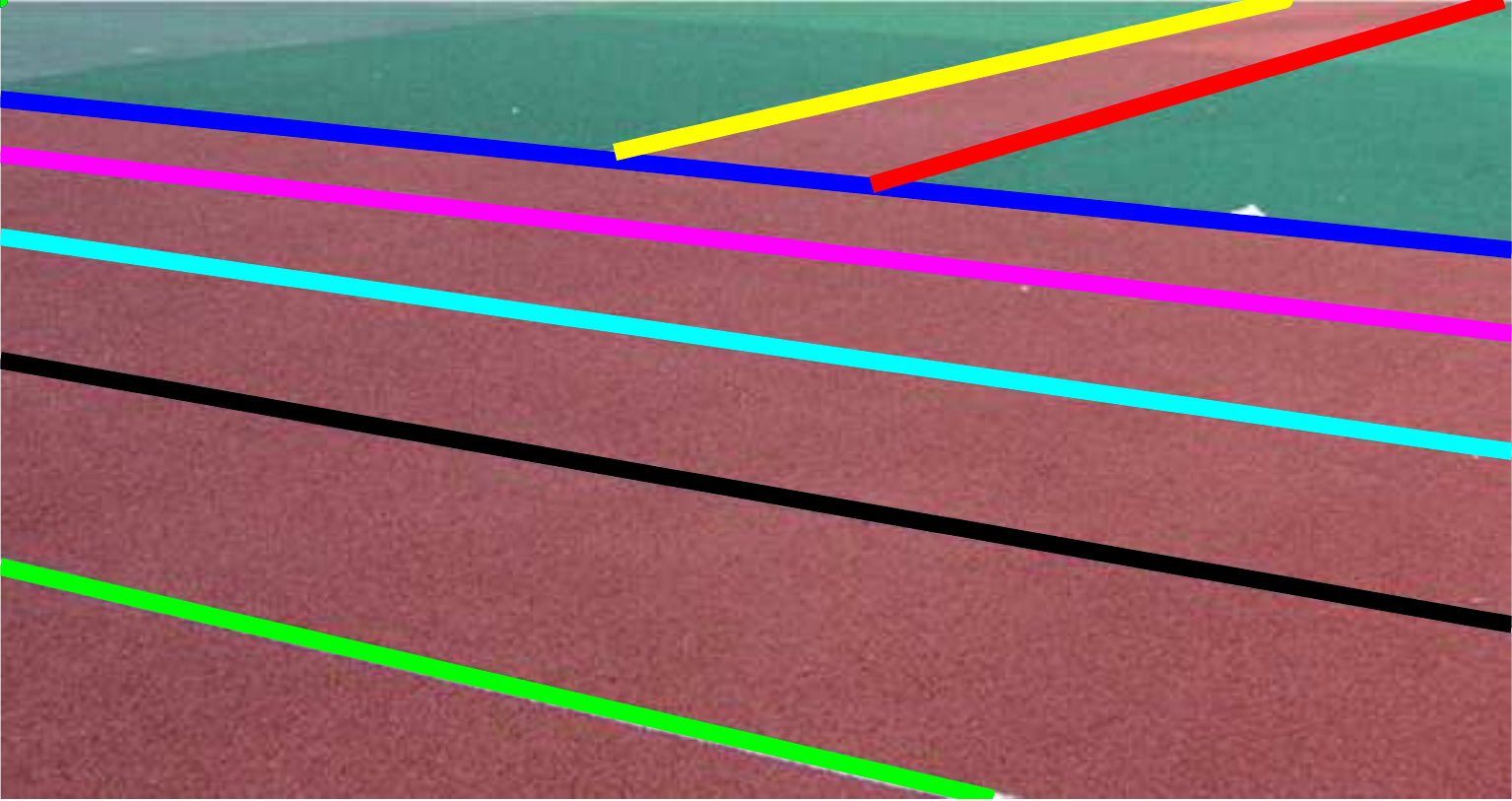}}
  \centerline{}
  \centerline{\includegraphics[width=1.0\textwidth]{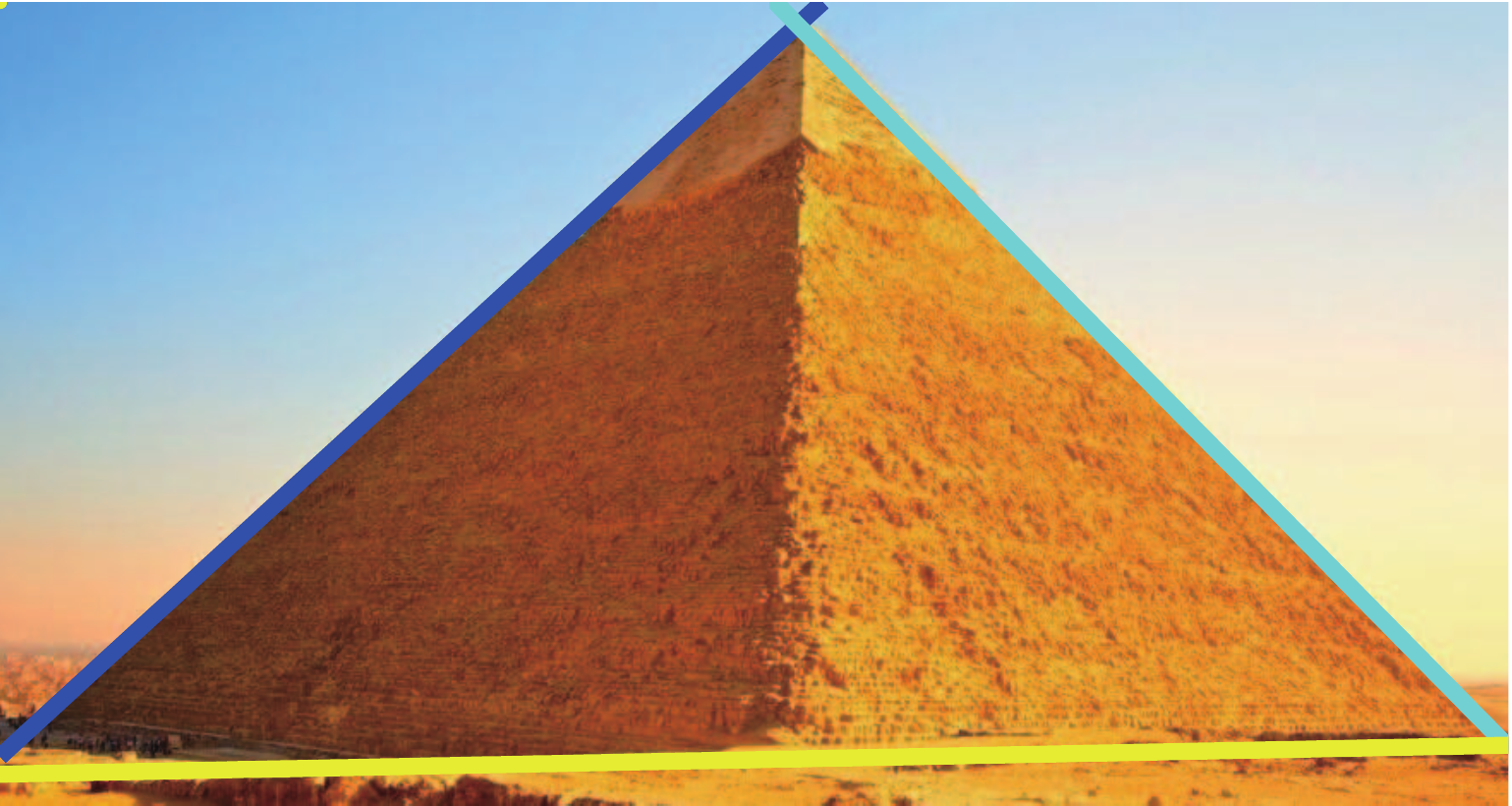}}
  \centerline{\footnotesize(d) AKSWH }
\end{minipage}
\begin{minipage}[t]{.15\textwidth}
  \centering
  \centerline{\includegraphics[width=1.0\textwidth]{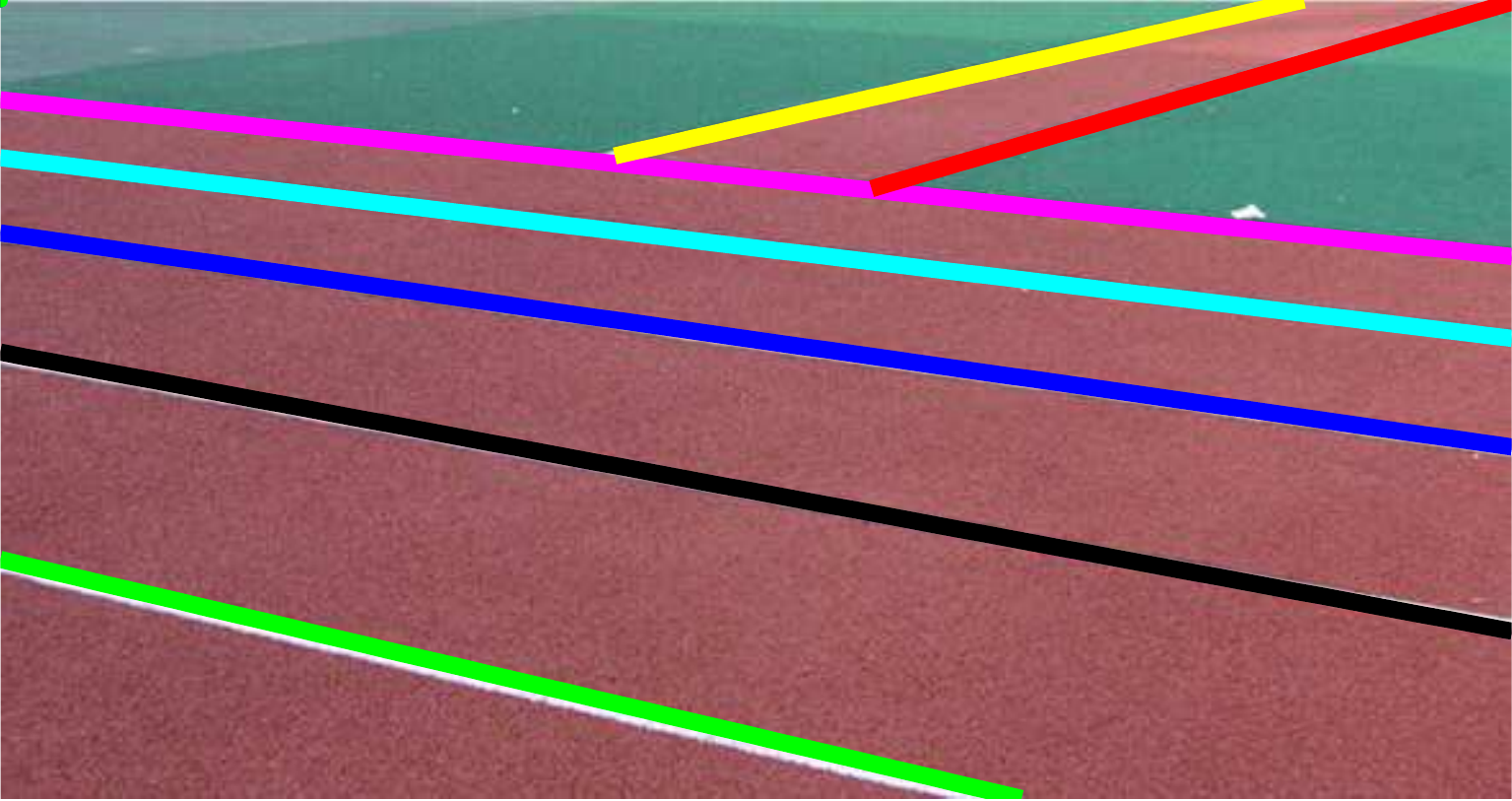}}
  \centerline{}
  \centerline{\includegraphics[width=1.0\textwidth]{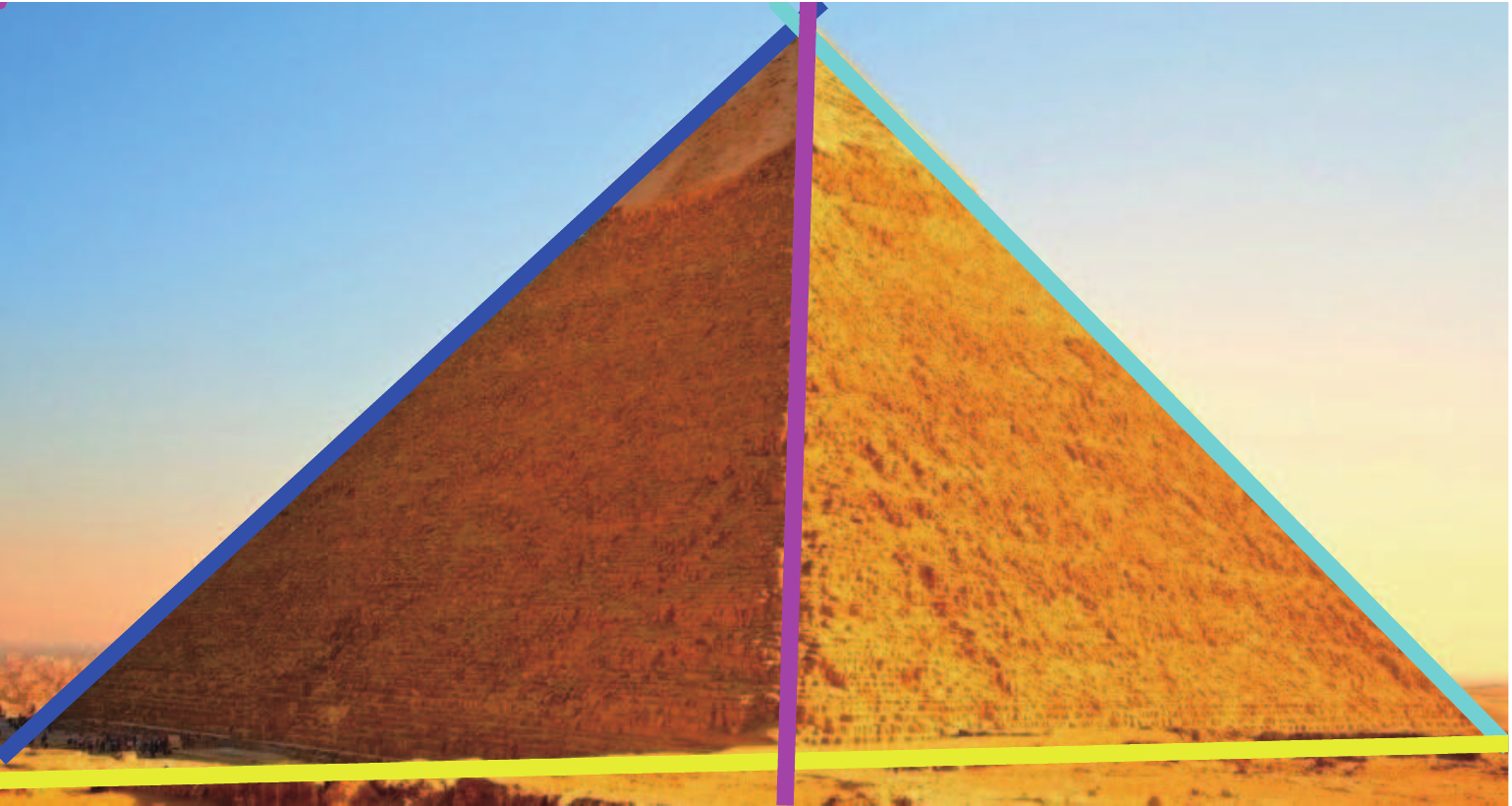}}
  \centerline{\footnotesize(e) T-linkage }
\end{minipage}
\begin{minipage}[t]{.15\textwidth}
  \centering
  \centerline{\includegraphics[width=1.0\textwidth]{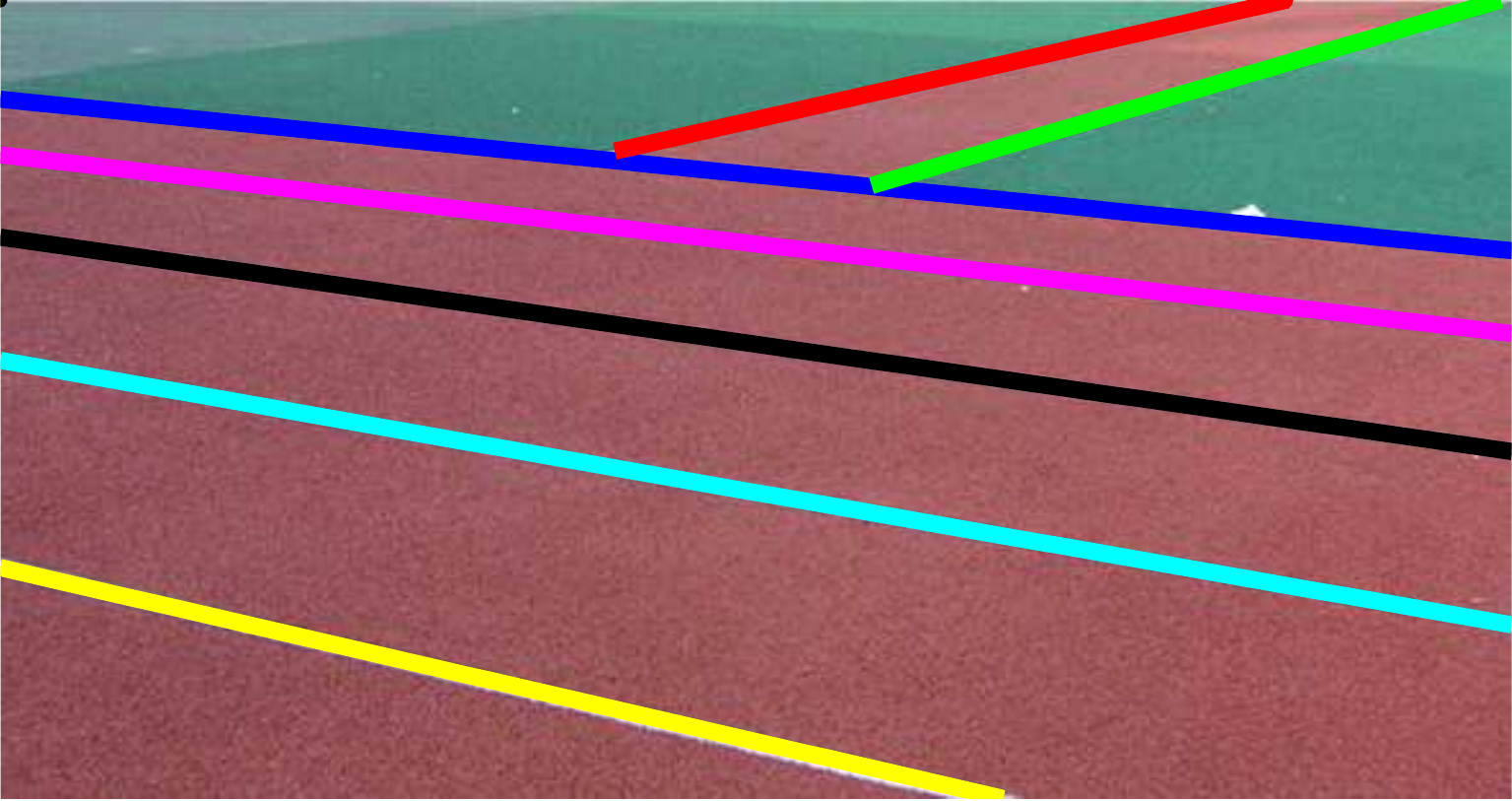}}
  \centerline{}
  \centerline{\includegraphics[width=1.0\textwidth]{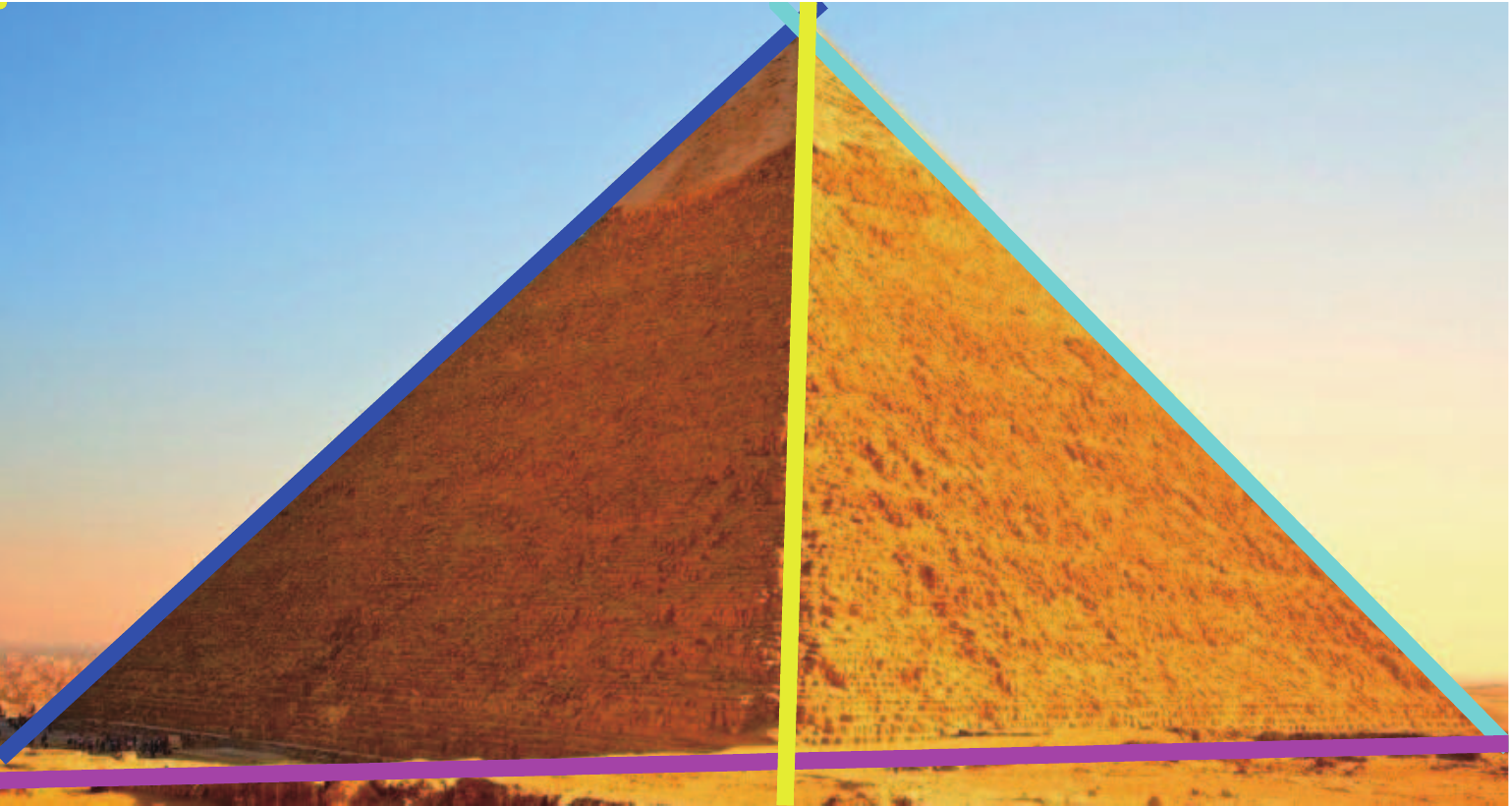}}
  \centerline{\footnotesize(f) MSHF2 }
\end{minipage}
\caption{Examples for line fitting. First (``Tracks") and second (``Pyramid") rows respectively fit seven and four lines. {We do not show the results of MSH/MSHF1, which are similar to those of MSHF2, due to the space limit.}}
\label{fig:linefiting}
\end{figure*}
\begin{figure*}[t]
\centering
\begin{minipage}[t]{.15\textwidth}
  \centering
  \centerline{\includegraphics[width=1.0\textwidth]{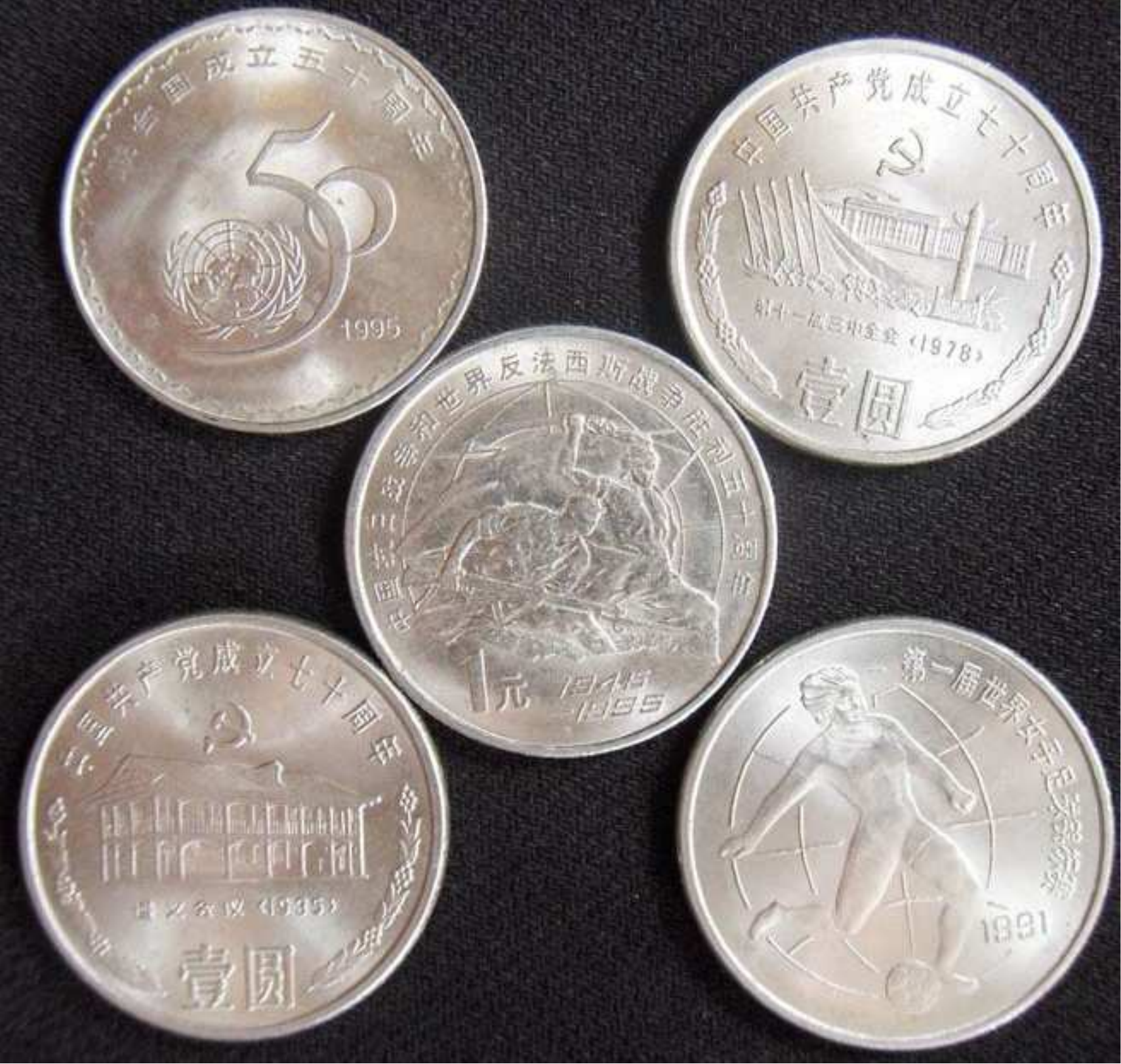}}
  \centerline{}
  \centerline{\includegraphics[width=1.0\textwidth]{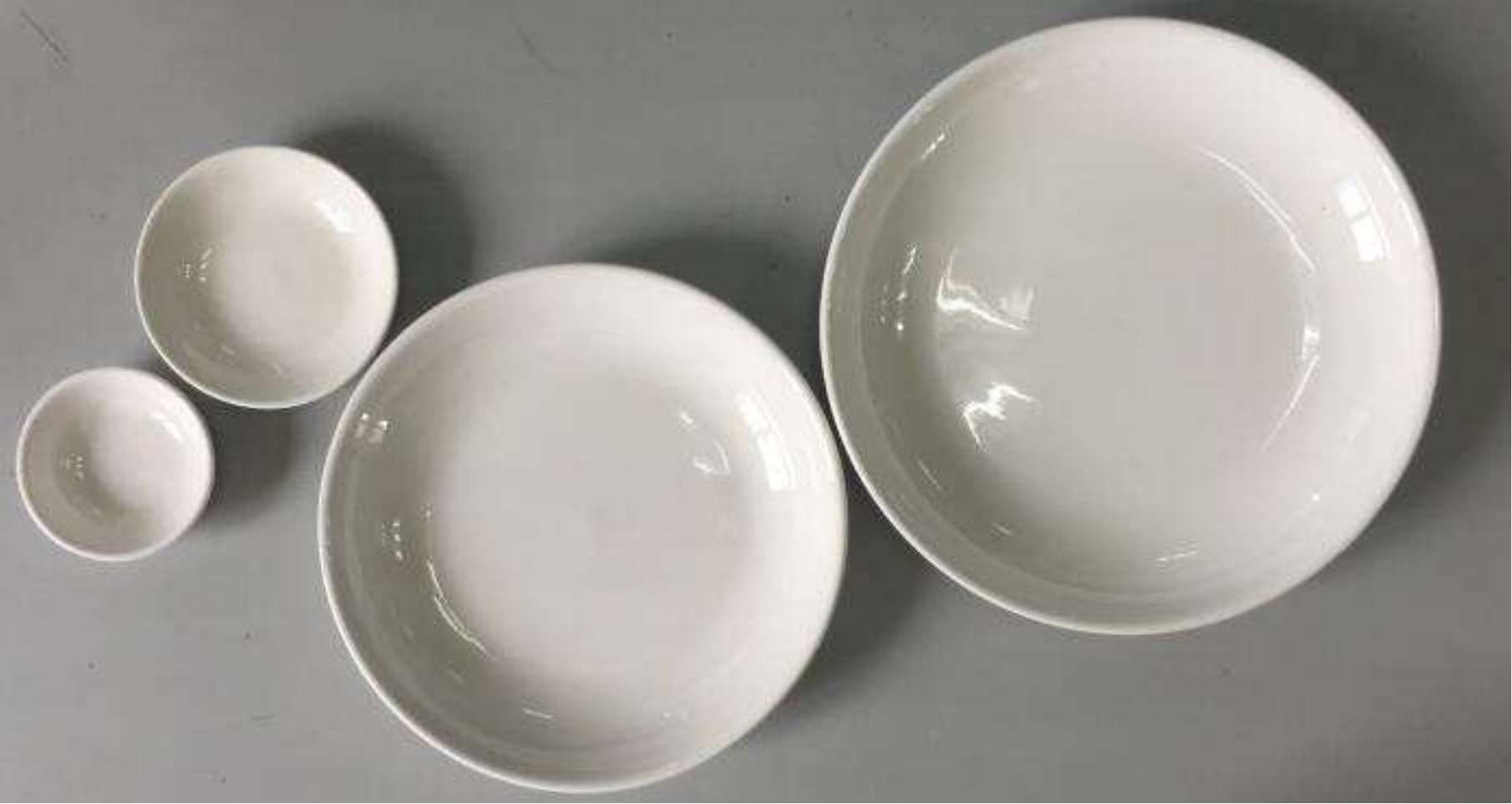}}
 \centerline{ \footnotesize(a) Data}
\end{minipage}
\begin{minipage}[t]{.15\textwidth}
  \centering
  \centerline{\includegraphics[width=1.0\textwidth]{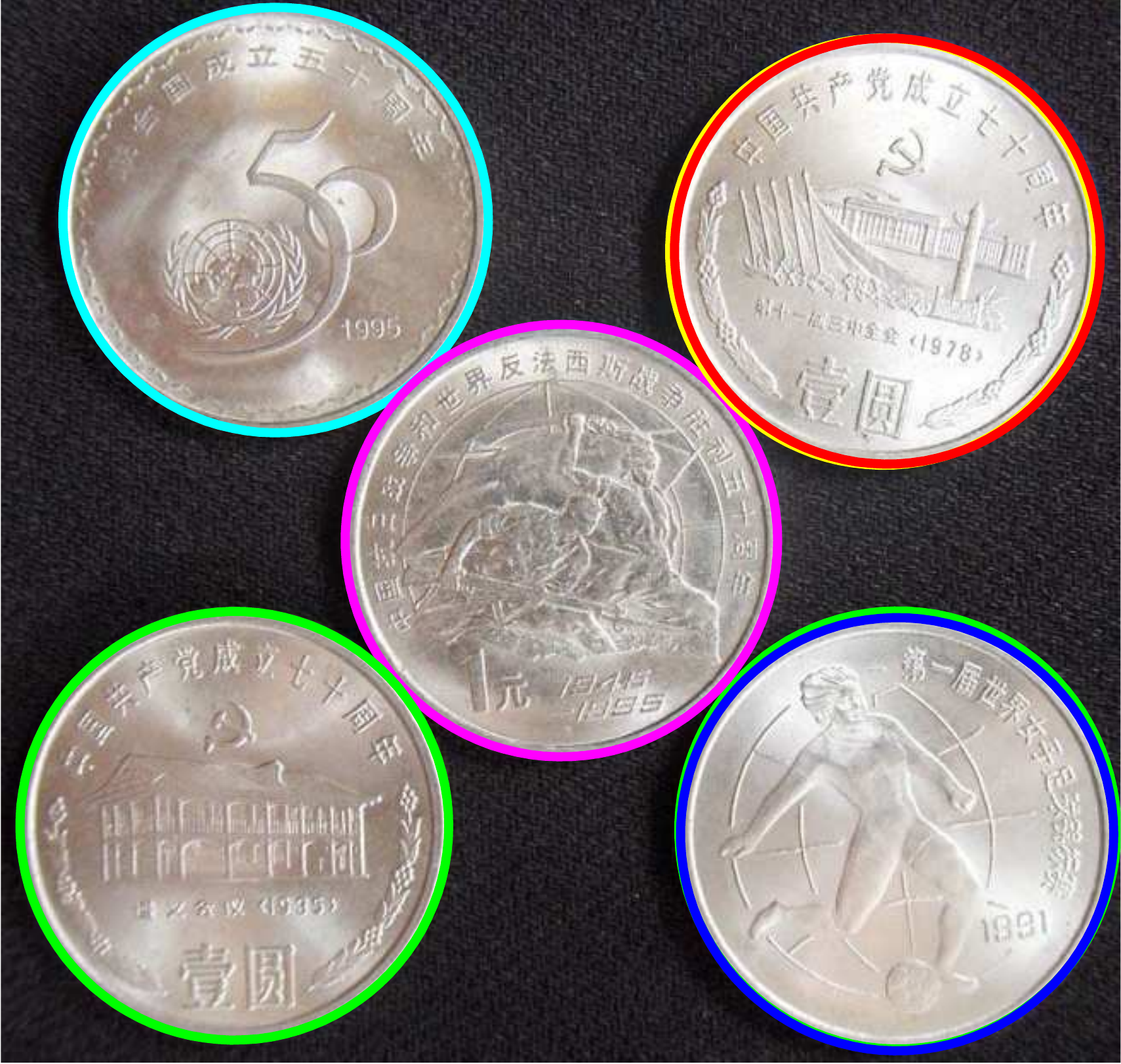}}
  \centerline{}
  \centerline{\includegraphics[width=1.0\textwidth]{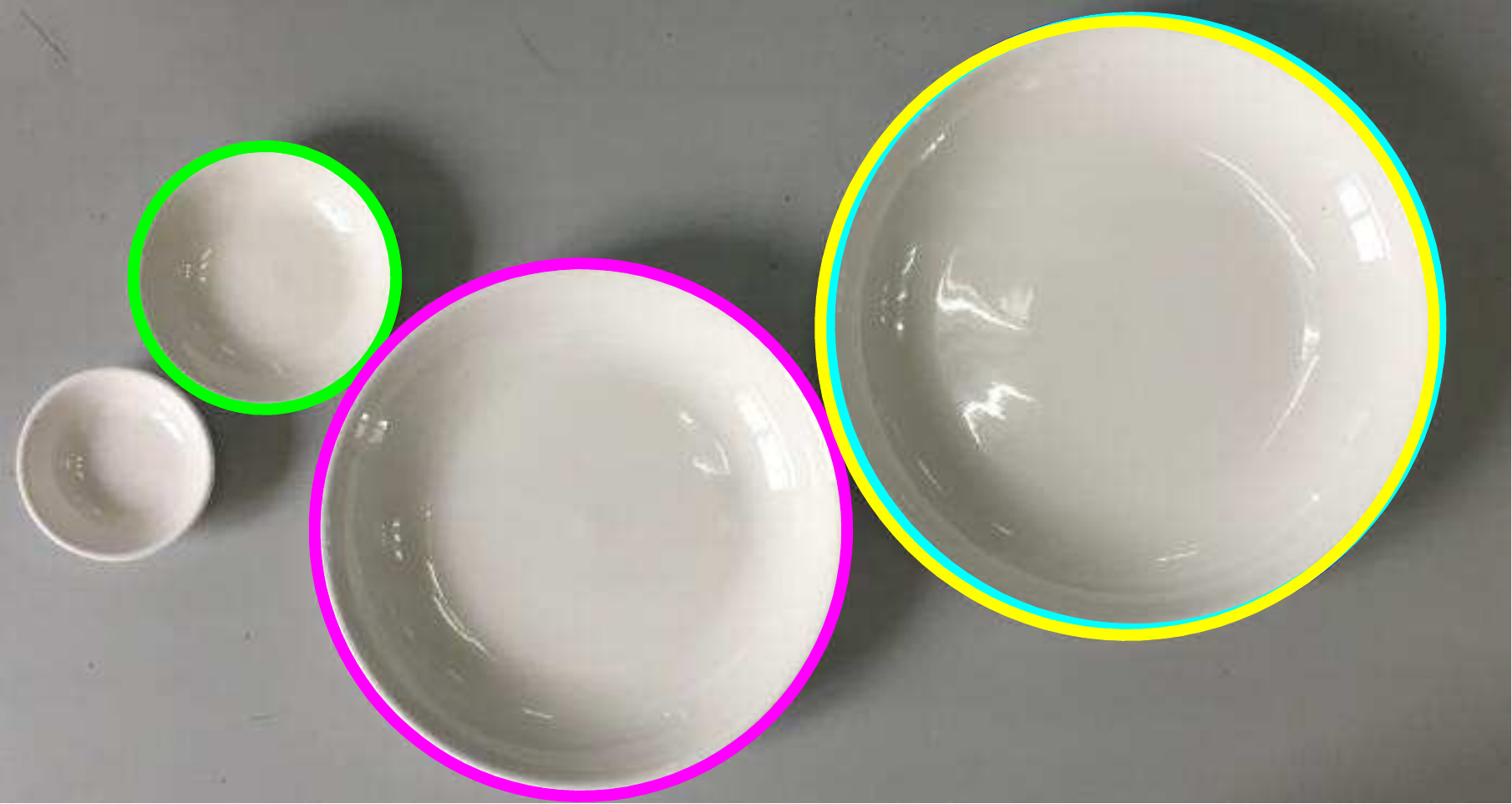}}
 \centerline{\footnotesize (b) KF }
\end{minipage}
\begin{minipage}[t]{.15\textwidth}
  \centering
  \centerline{\includegraphics[width=1.0\textwidth]{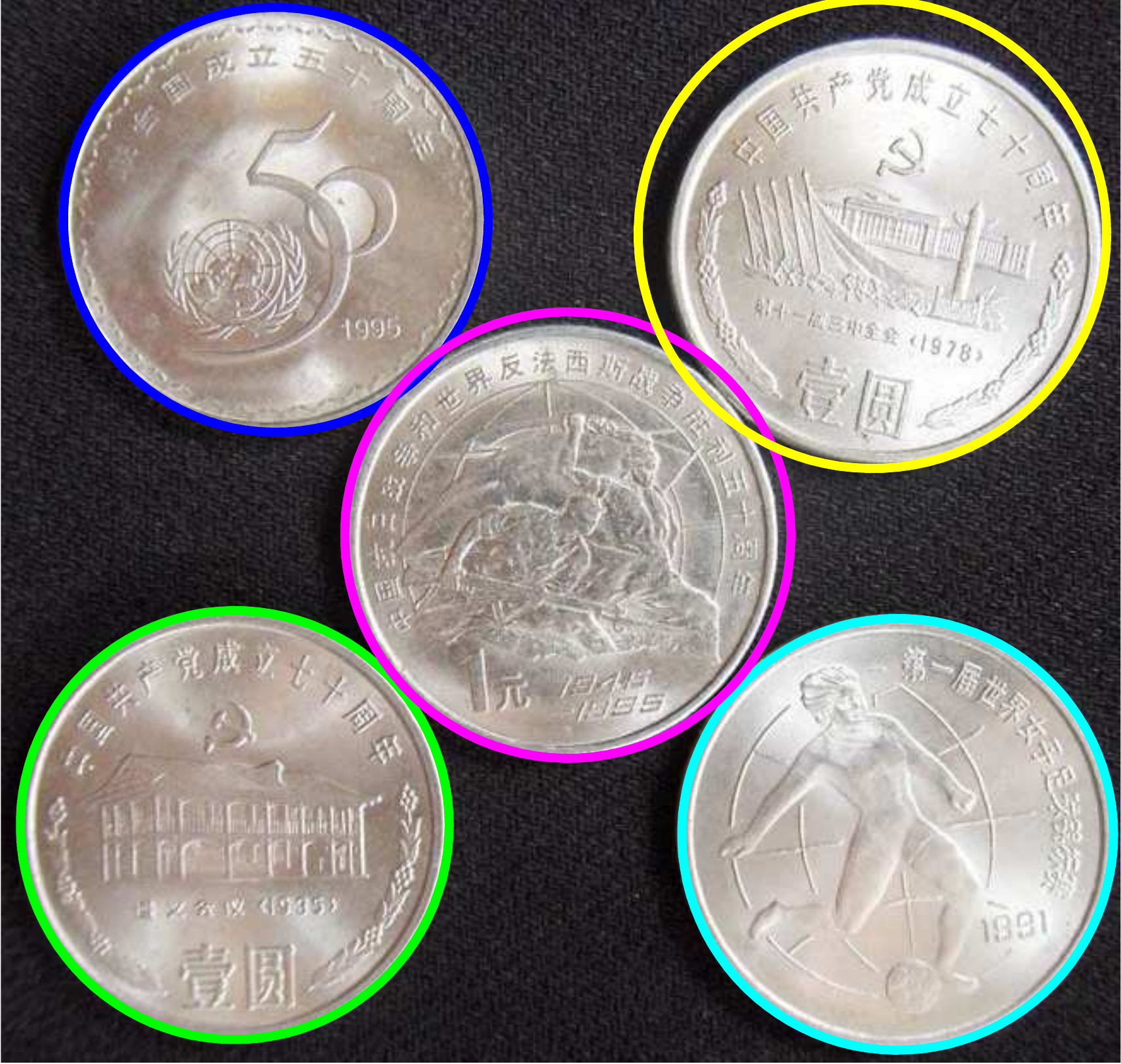}}
  \centerline{}
  \centerline{\includegraphics[width=1.0\textwidth]{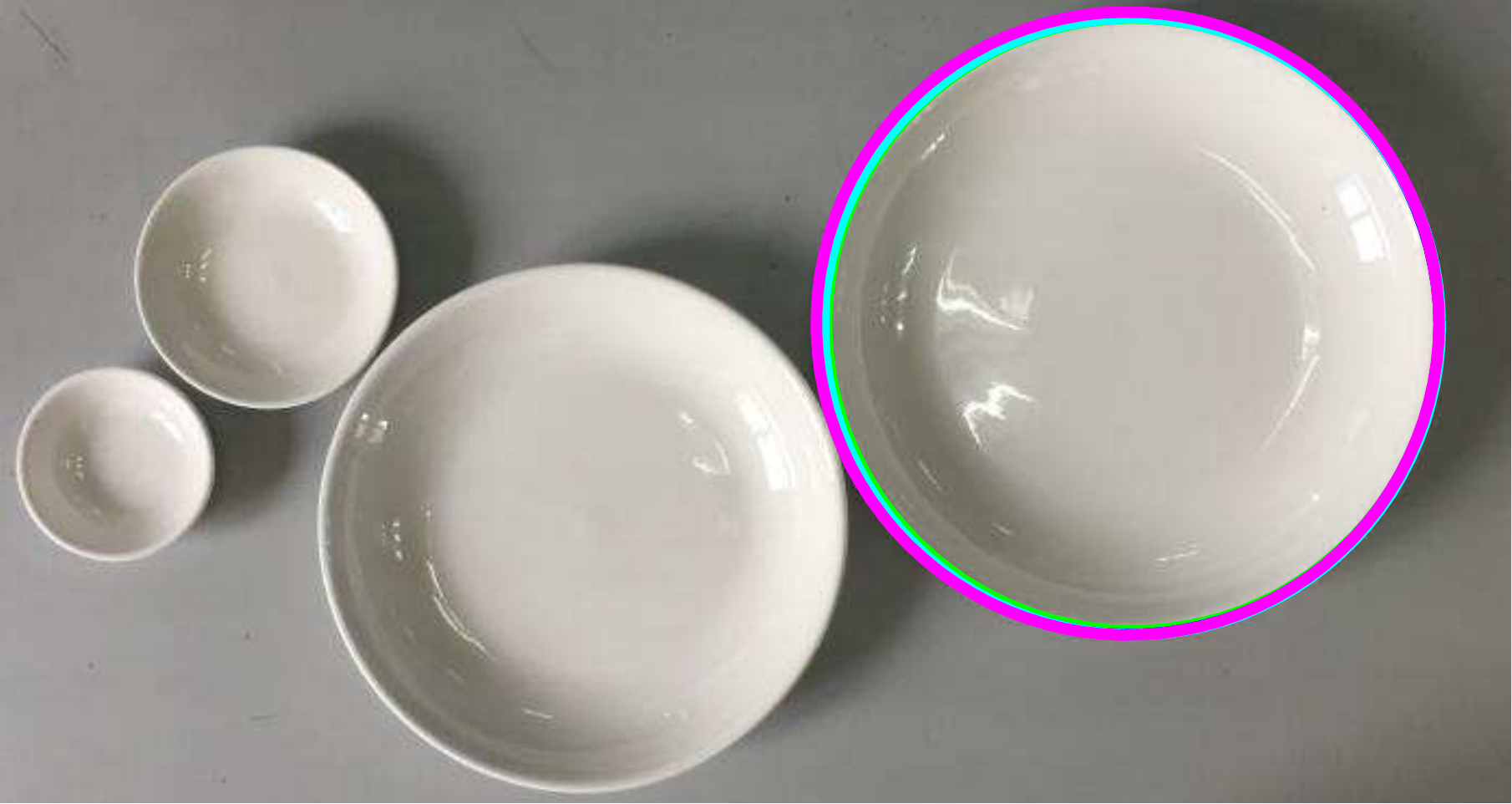}}
 \centerline{ \footnotesize(c) RCG}
\end{minipage}
\begin{minipage}[t]{.15\textwidth}
  \centering
  \centerline{\includegraphics[width=1.0\textwidth]{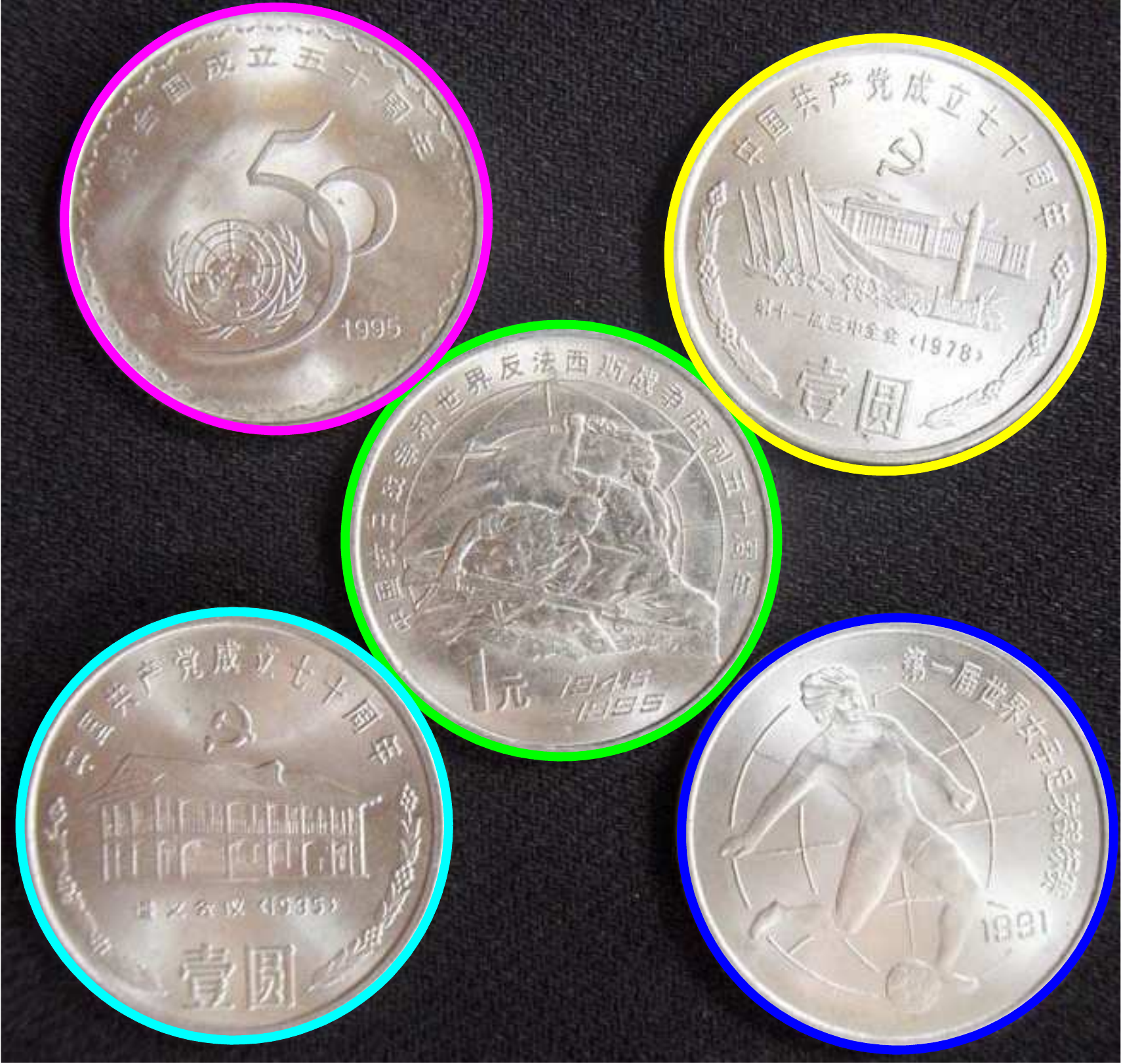}}
  \centerline{}
  \centerline{\includegraphics[width=1.0\textwidth]{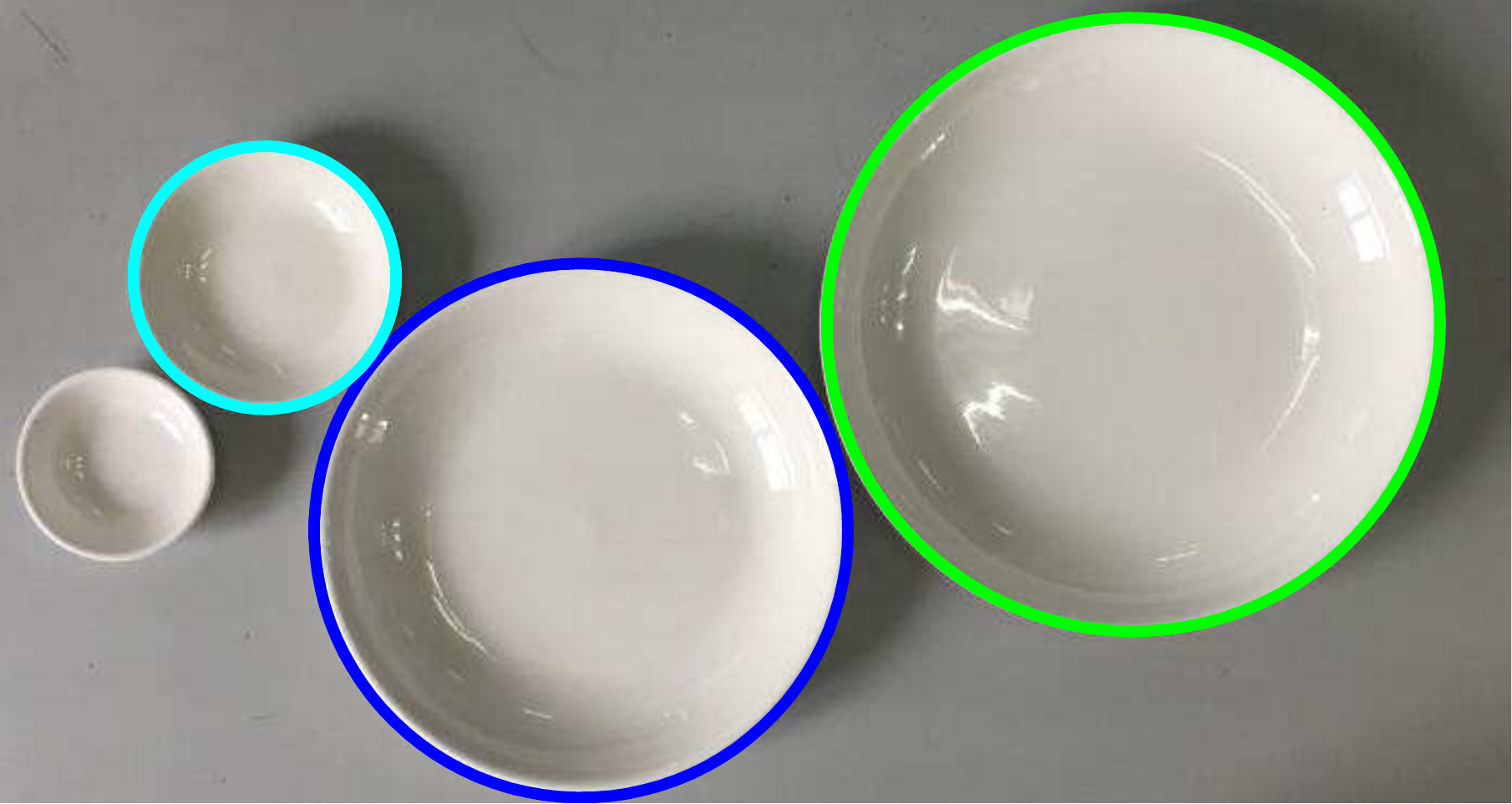}}
  \centerline{\footnotesize(d) AKSWH }
\end{minipage}
\begin{minipage}[t]{.15\textwidth}
  \centering
  \centerline{\includegraphics[width=1.0\textwidth]{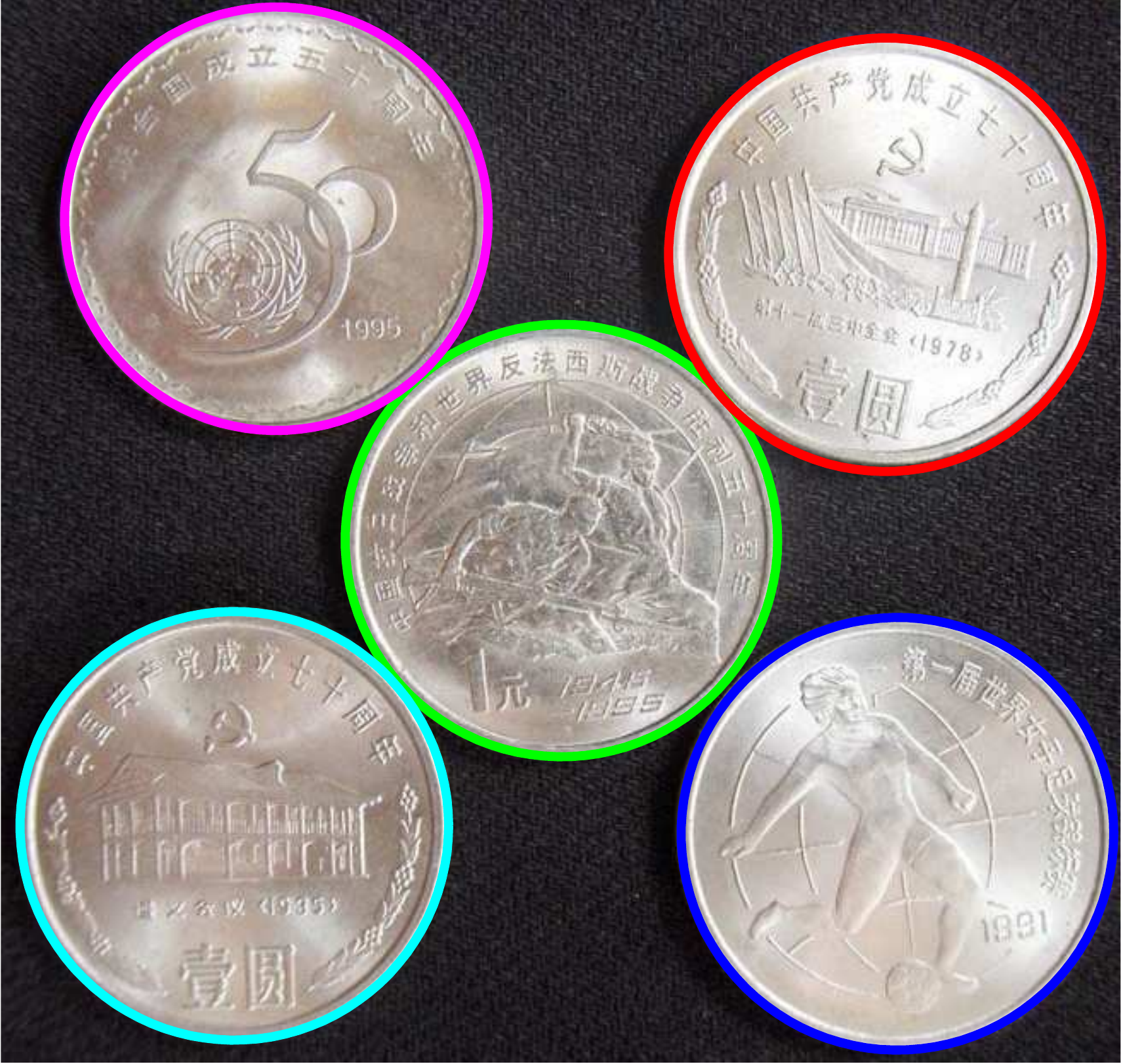}}
  \centerline{}
  \centerline{\includegraphics[width=1.0\textwidth]{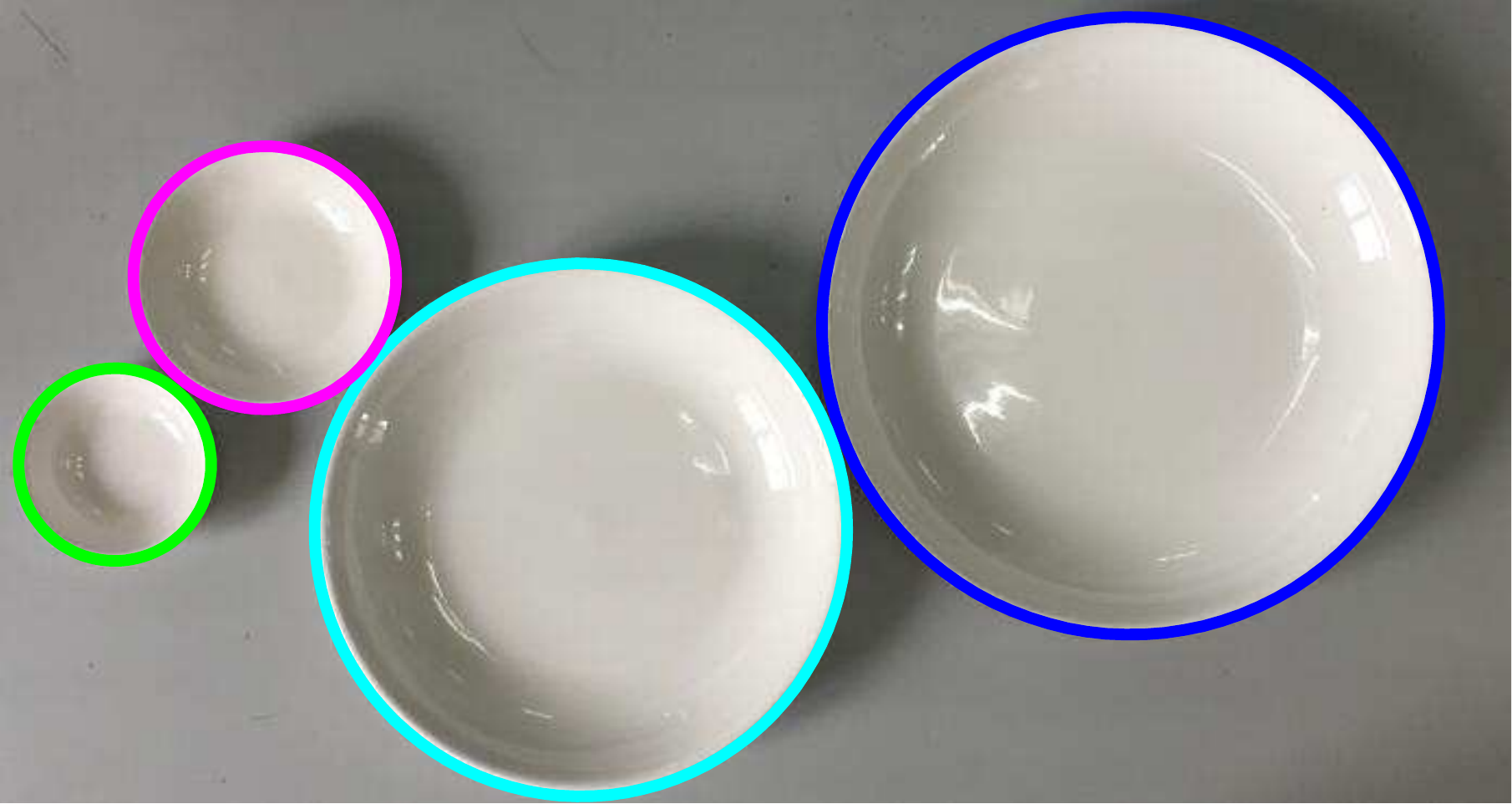}}
  \centerline{\footnotesize(e) T-linkage }
\end{minipage}
\begin{minipage}[t]{.15\textwidth}
  \centering
  \centerline{\includegraphics[width=1.0\textwidth]{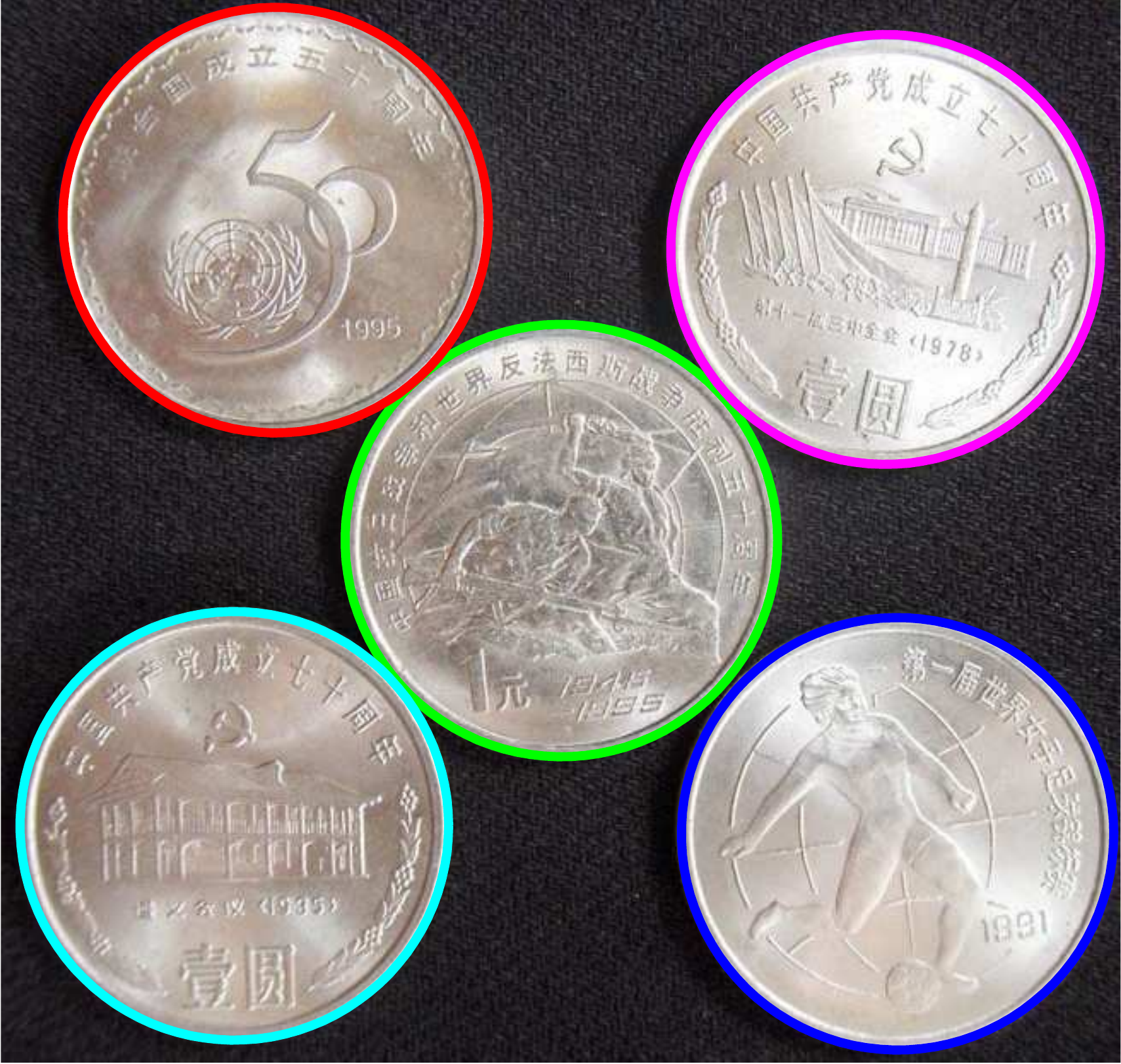}}
  \centerline{}
  \centerline{\includegraphics[width=1.0\textwidth]{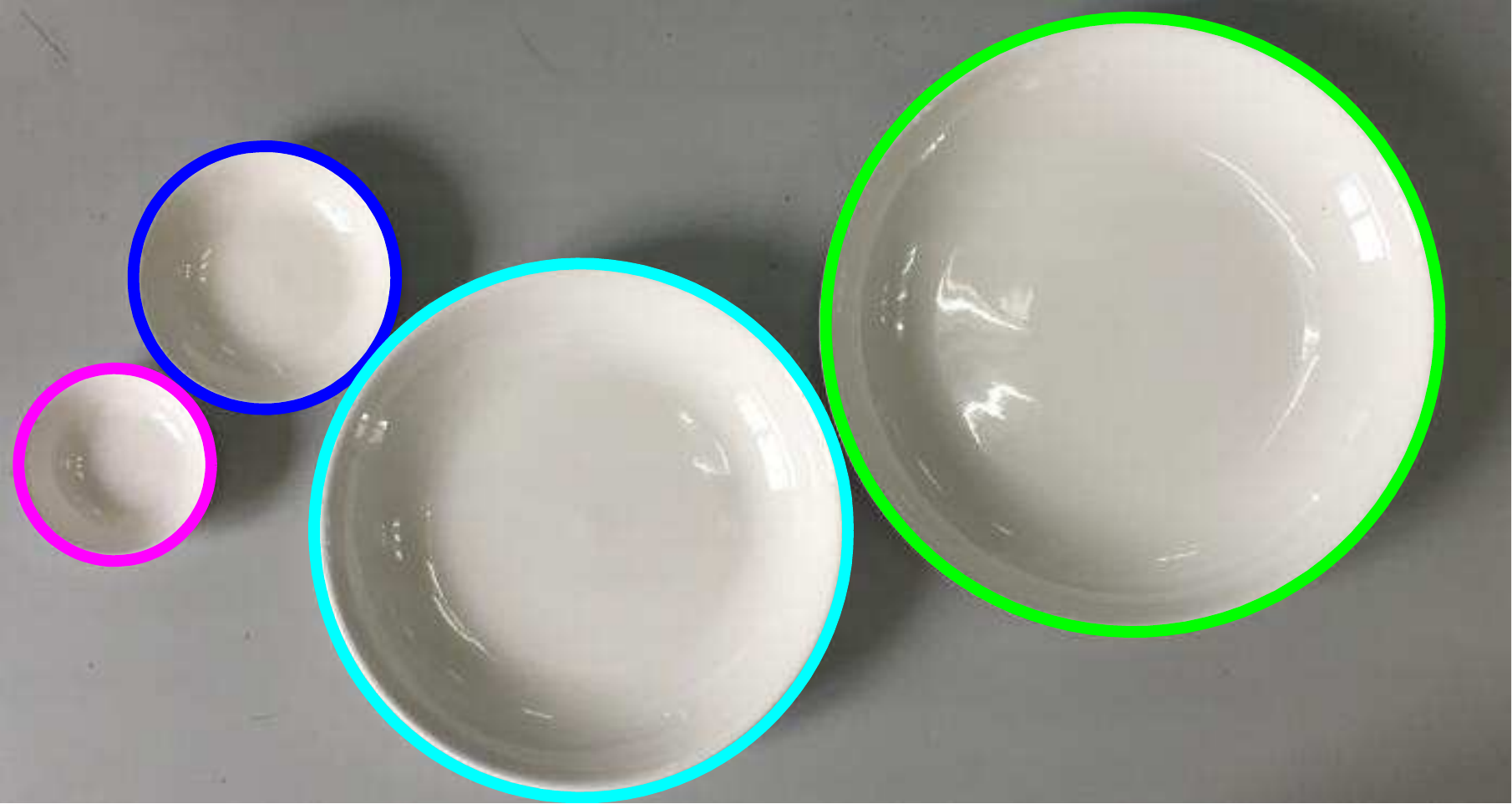}}
  \centerline{\footnotesize (f) MSHF2 }
\end{minipage}
\caption{Examples for circle fitting. First (``Coins") and second (``Bowls") rows respectively fit five and four circles. {We do not show the results of MSH/MSHF1, which are similar to those of MSHF2, due to the space limit.}}
\label{fig:circlefiting}
\end{figure*}
\subsection{Real Images}
\subsubsection{Line Fitting}
\label{sec:linefitting}
\begin{table}[t]
  \caption{{The CPU time used by the seven fitting methods (in seconds). }}
\centering
\scalebox{0.98}{\tabcolsep0.08in
\begin{tabular}{|c|c|c|c|c|>{\columncolor{mygray}}c|>{\columncolor{mygray}}c|>{\columncolor{mygray}}c|}
\hline
Data        & M1&M2&M3&M4&M5&M6 &{M7}\\
\hline
\hline
   Tracks          & {133.71} & {9.76} & {8.81}& {23,256.00}& {7.21}& {8.26}& {\bf7.12}\\
                       \hline
   Pyramid           & {79.28}& {8.12}& {\bf7.23}& {13,600.00}& {8.05} & {8.65}& {7.62}  \\
                       \hline
   Coins           & {65.02} & {6.29} & {5.84} & {8,746.50} & {5.02}& {5.16}& {\bf4.48} \\
                       \hline
  Bowls        & {9.01} & {4.33}& {3.68}& {862.34}& {3.81}& {3.98}  & {\bf3.64}\\
                       \hline
\end{tabular}}
\\
\medskip
 \label{table:realimagetable}
\end{table}
We evaluate the performance of all the competing fitting methods using real images for line fitting (see Fig.~\ref{fig:linefiting}). For the ``Tracks" image, which includes seven lines, there are $6,704$ edge points detected by the Canny operator \cite{canny1986computational}. As shown in Fig.~\ref{fig:linefiting} {and Table~\ref{table:realimagetable},} AKSWH, T-linkage, {MSH and MSHF1/MSHF2} correctly fit all the seven lines. {However, MSH, MSHF1/2 are faster than AKSWH and T-linkage. T-linkage is very slow due to the large number of input data points.} RCG correctly estimates the number of lines, but some {estimated} lines are overlapped and two lines are missed{. This is} because the potential structures {in the data} are not correctly estimated {by RCG during the detection of} dense subgraphs. KF only correctly fits three out of the seven lines because many inliers belonging to the other four lines are wrongly removed.

For the ``Pyramid" image {shown in Fig.~\ref{fig:linefiting}}, which includes four lines with a large number of outliers, there are $5,576$ edge points detected by the Canny operator. Only T-linkage, {MSH and MSHF1/MSHF2} succeed in fitting all the four lines. In contrast, although KF also fits the four lines, it wrongly estimates the number of lines (which is five lines instead of four lines). Both RCG and AKSWH only correctly fit three out of the four lines, although RCG successfully estimates the number of lines {in the data}. AKSWH can detect four lines after the clustering step, but two lines are wrongly fused during the fusion step.  {For the CPU time (see Table~\ref{table:realimagetable}), RCG, AKSWH, MSH and MSHF1/MSHF2 achieve similar time, but KF and T-linkage are much slower than the other five methods.}
\begin{figure*}[t]
\centering
\begin{minipage}[t]{.19\textwidth}
  \centering
  \centerline{\includegraphics[width=1.16\textwidth]{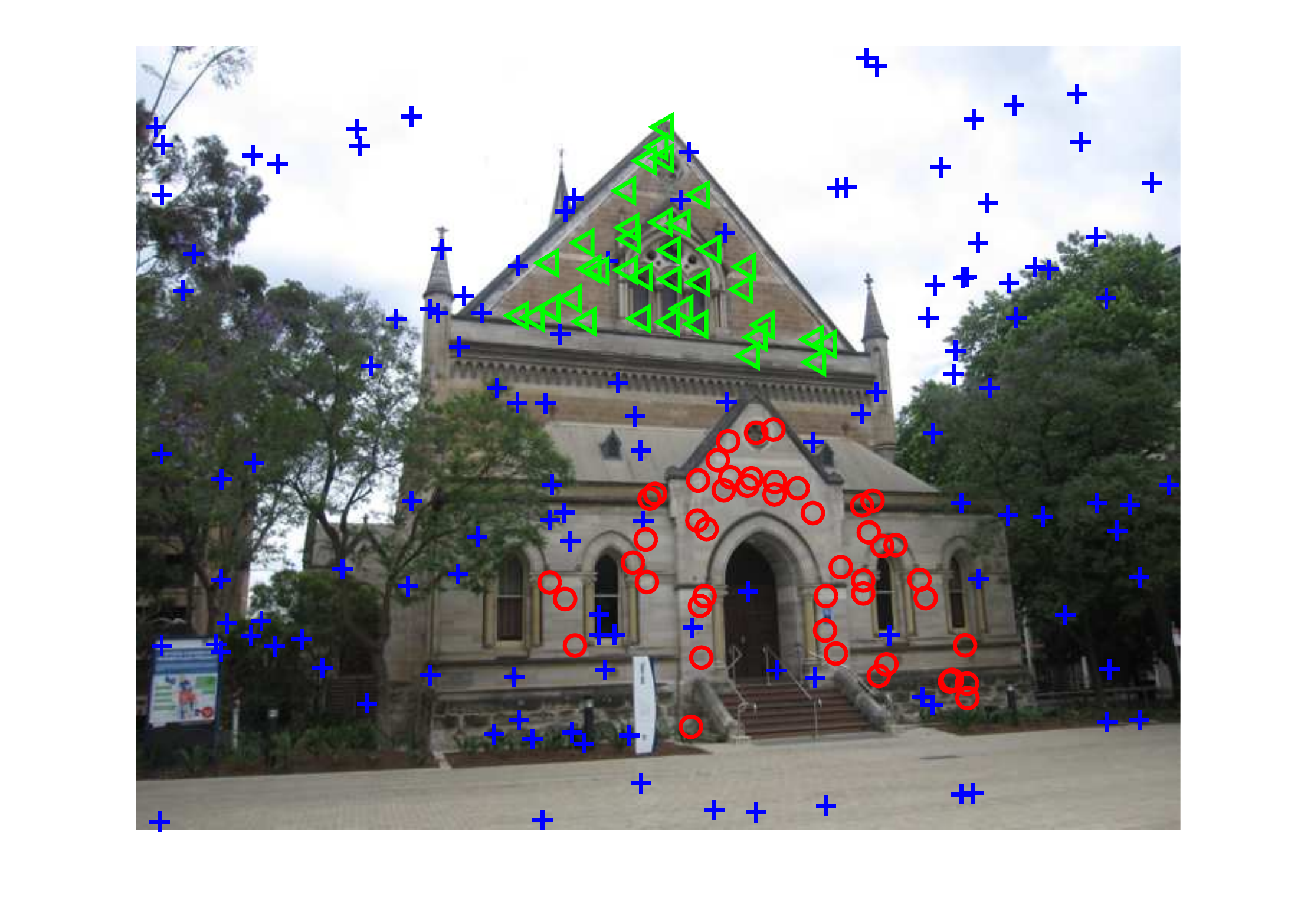}}
\centerline{ \footnotesize(a) Elderhalla}
  \centerline{\includegraphics[width=1.16\textwidth]{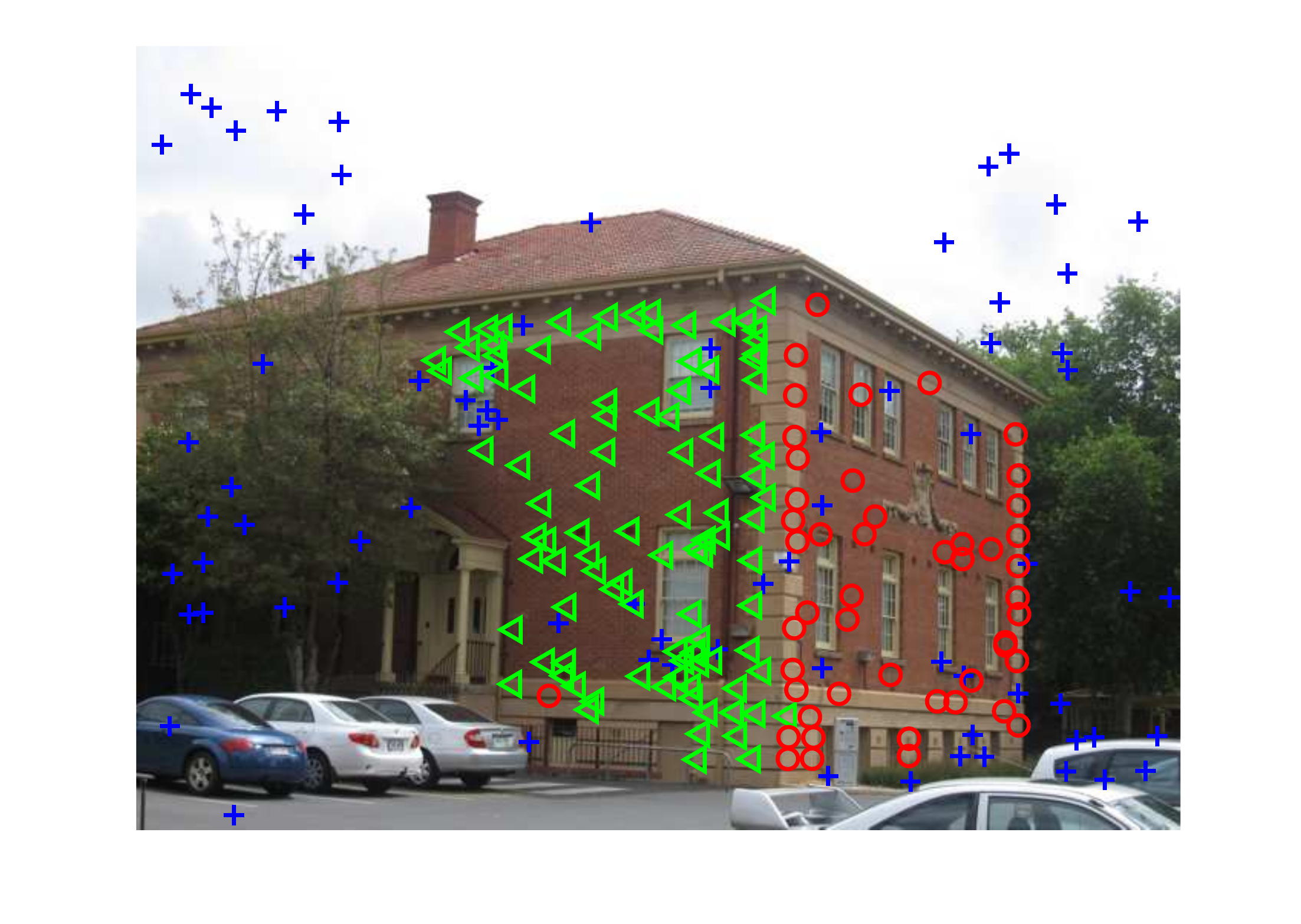}}
  \centerline{\footnotesize (f) Ladysymon}
\end{minipage}
\begin{minipage}[t]{.19\textwidth}
  \centering
  \centerline{\includegraphics[width=1.16\textwidth]{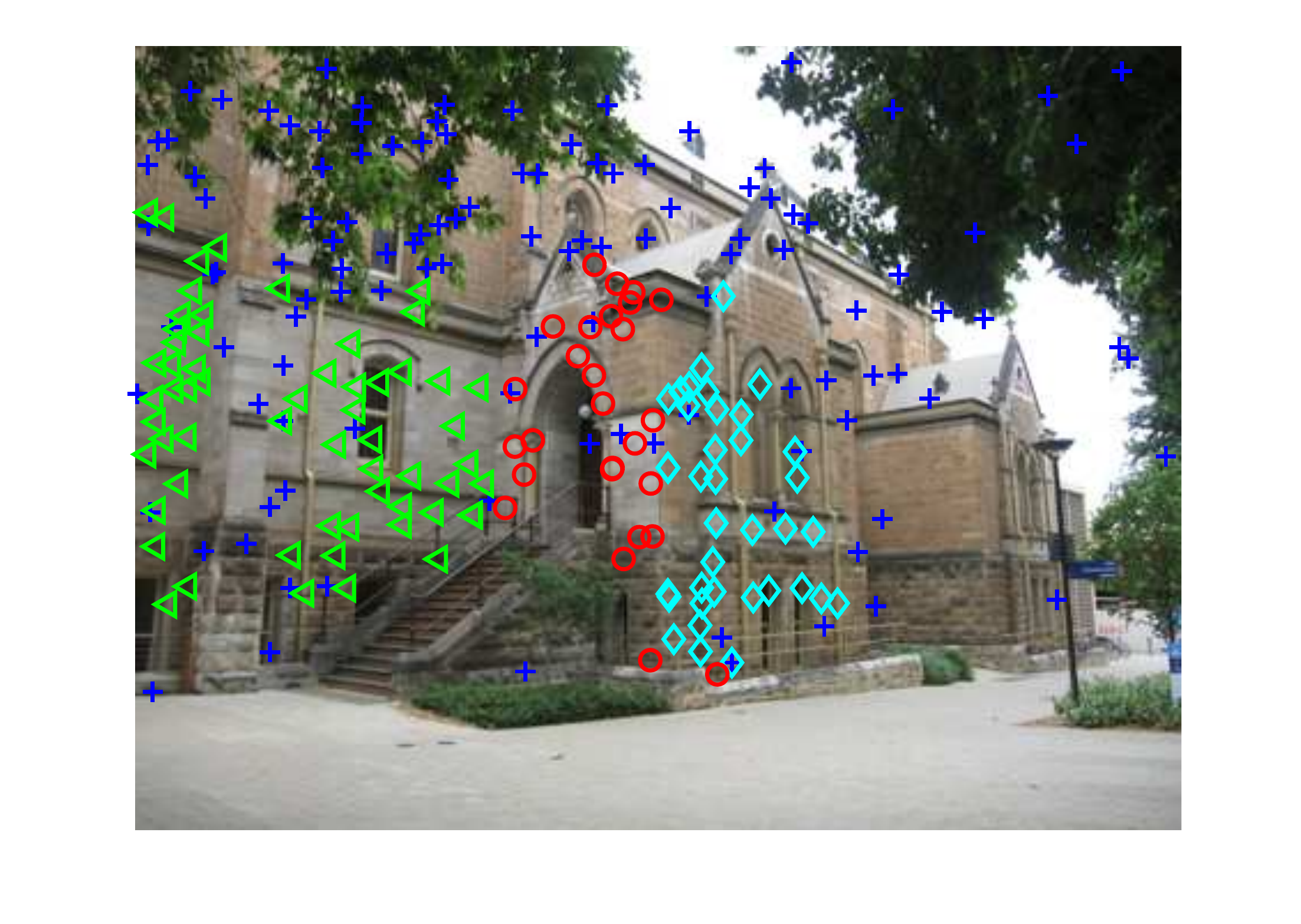}}
\centerline{\footnotesize (b) Elderhallb }
  \centerline{\includegraphics[width=1.16\textwidth]{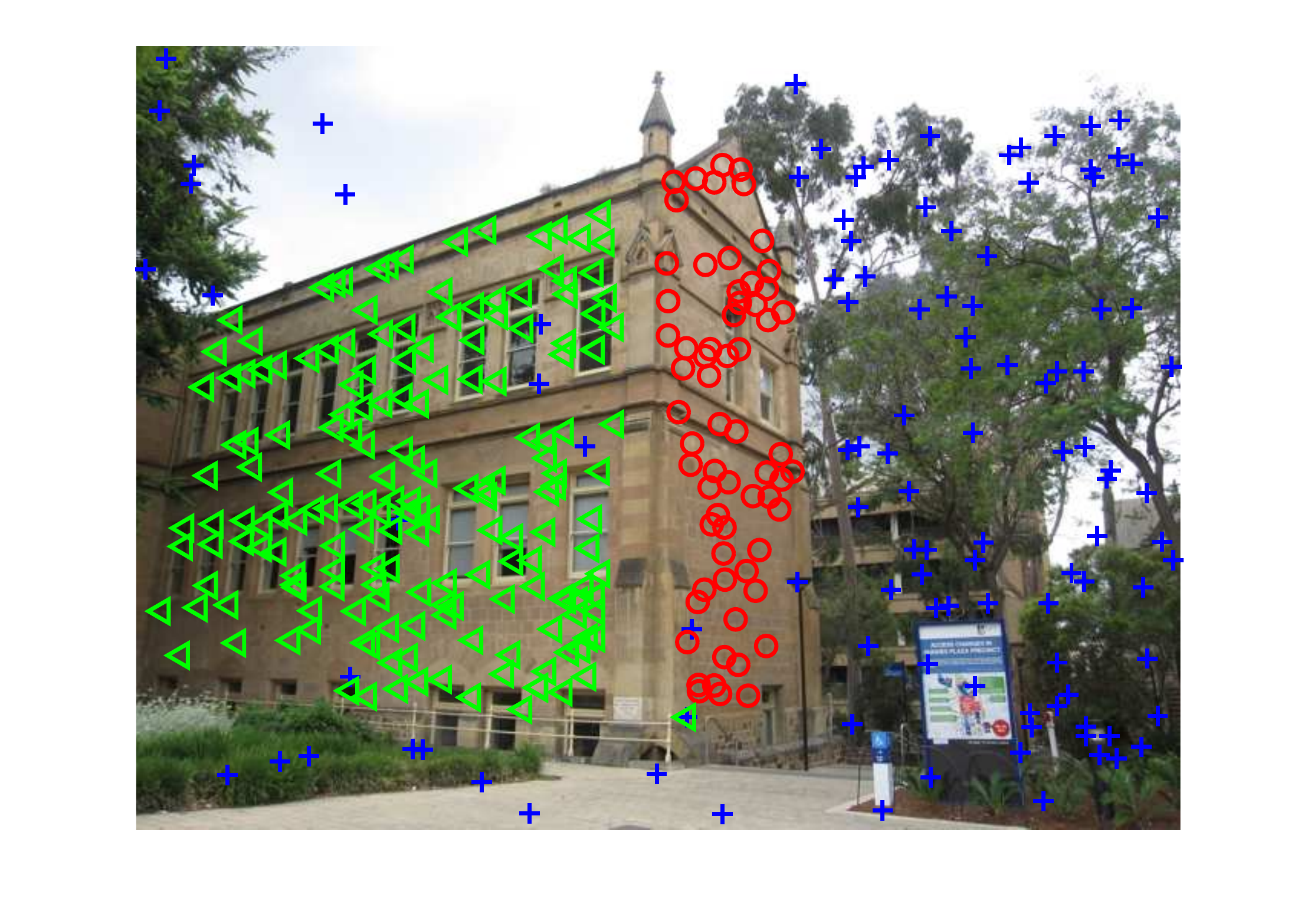}}
\centerline{\footnotesize (g) Oldclassicswing}
\end{minipage}
\begin{minipage}[t]{.19\textwidth}
  \centering
  \centerline{\includegraphics[width=1.16\textwidth]{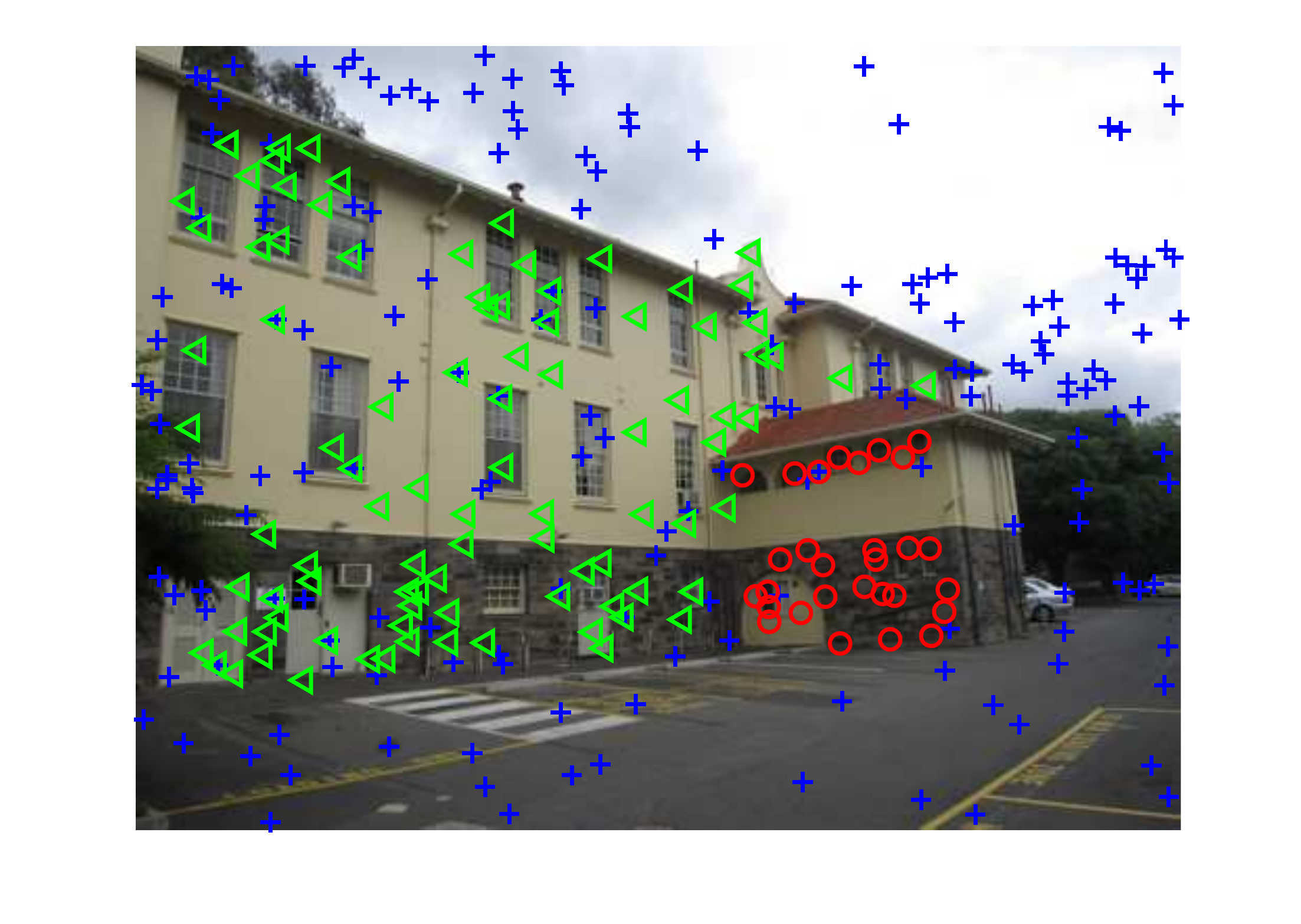}}
\centerline{\footnotesize (c) Hartley }
  \centerline{\includegraphics[width=1.16\textwidth]{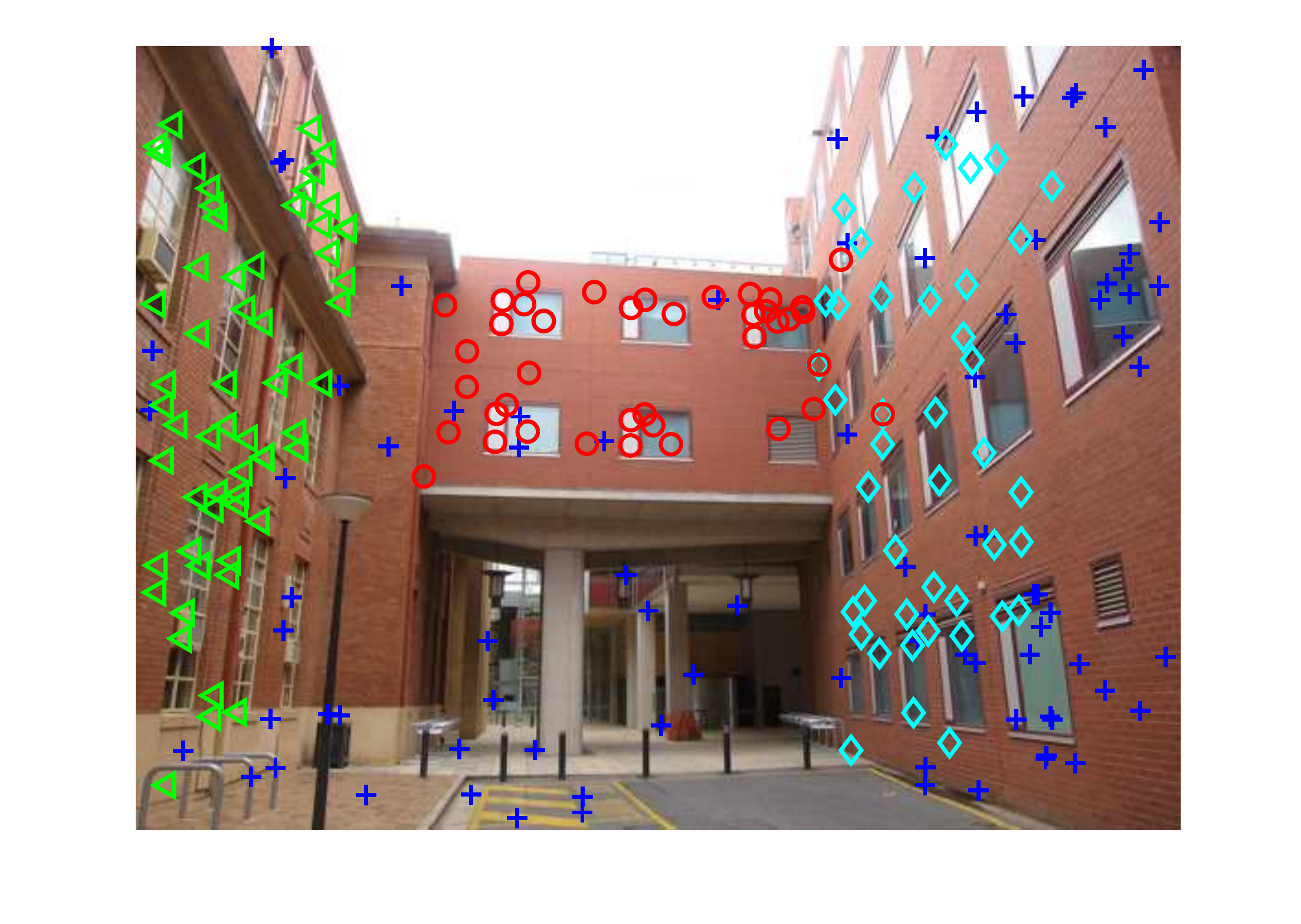}}
\centerline{\footnotesize (h) Neem}
\end{minipage}
\begin{minipage}[t]{.19\textwidth}
  \centering
  \centerline{\includegraphics[width=1.16\textwidth]{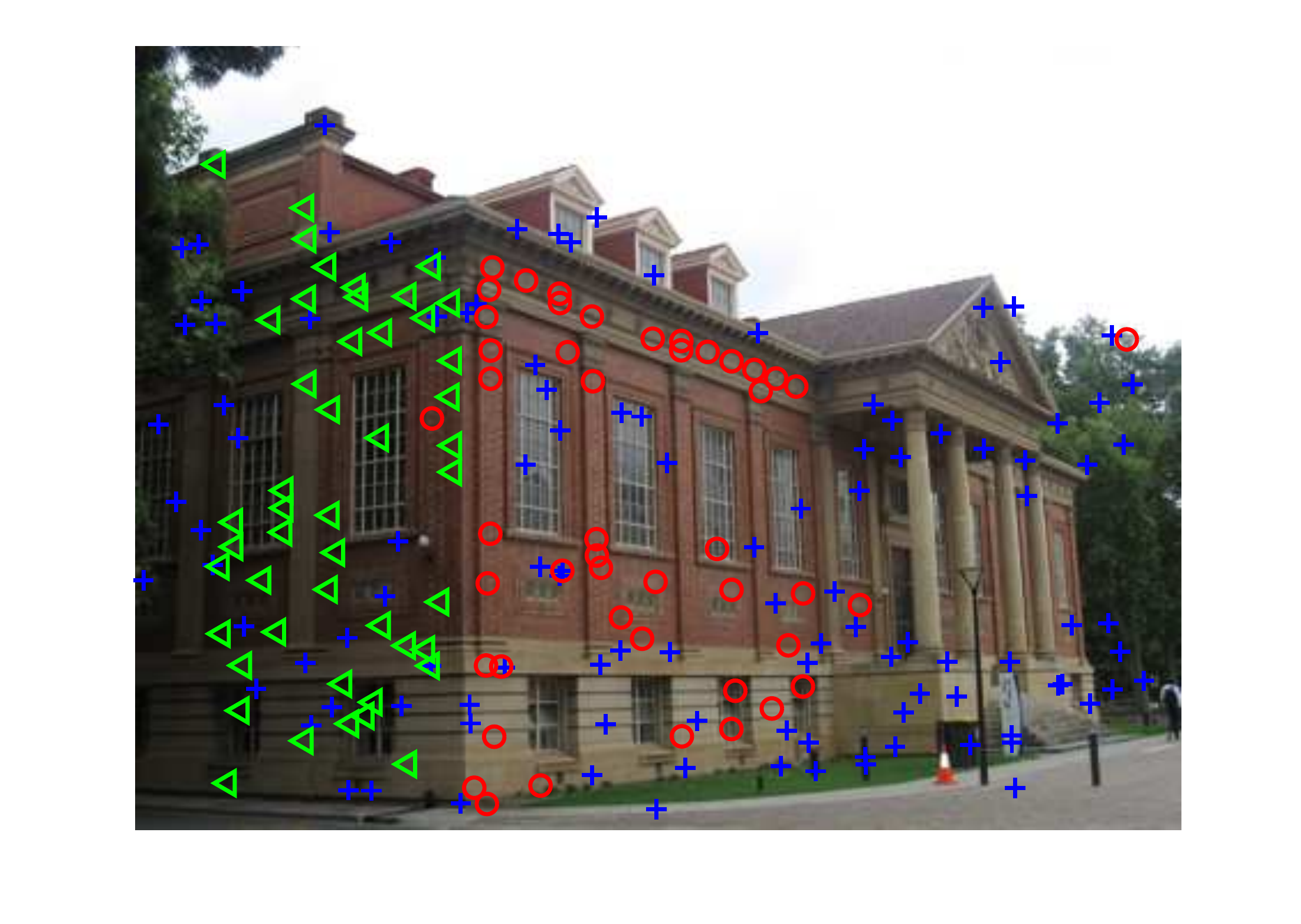}}
 \centerline{\footnotesize (d) Library }
  \centerline{\includegraphics[width=1.16\textwidth]{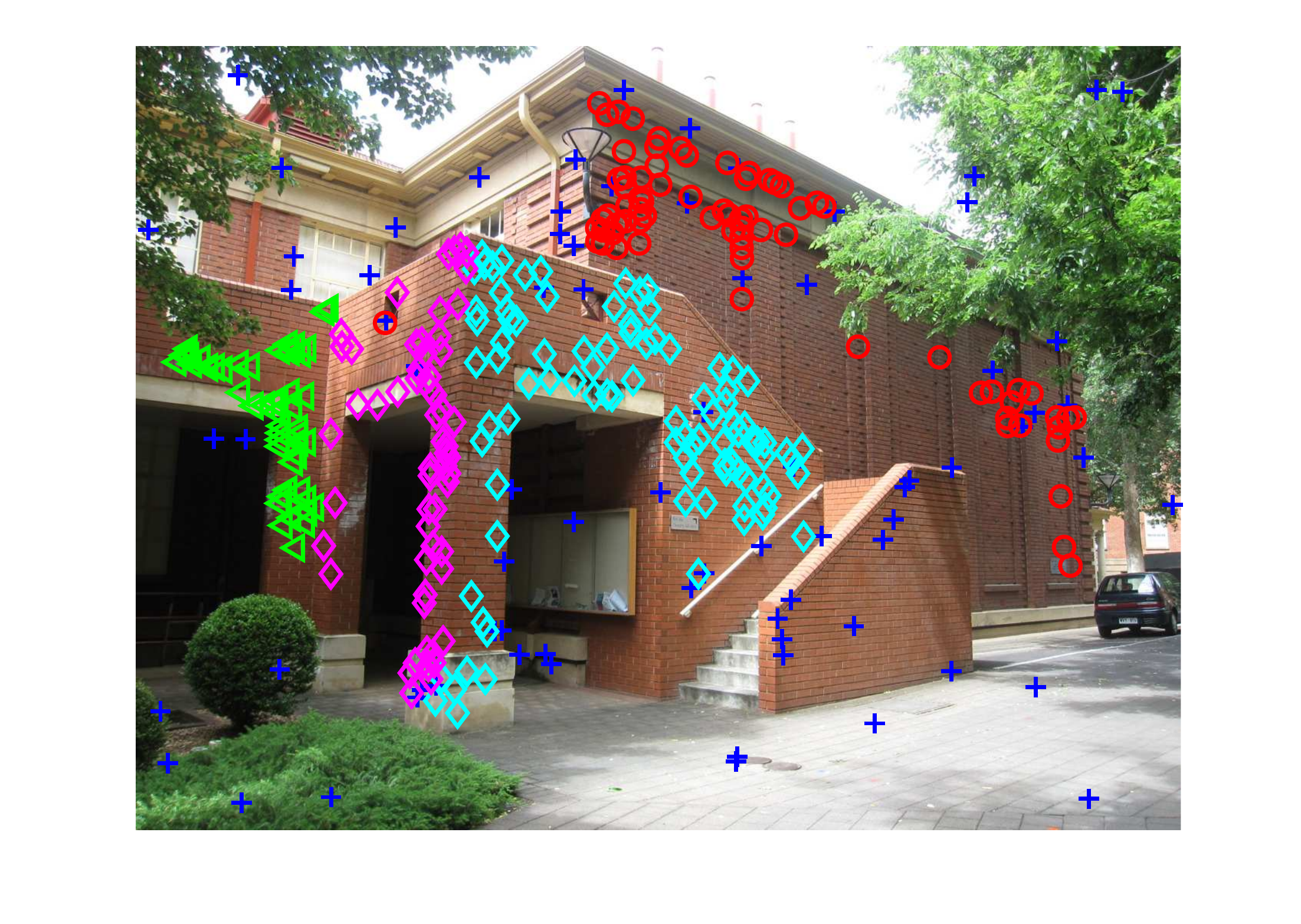}}
\centerline{\footnotesize (i) Johnsona}
\end{minipage}
\begin{minipage}[t]{.19\textwidth}
  \centering
  \centerline{\includegraphics[width=1.16\textwidth]{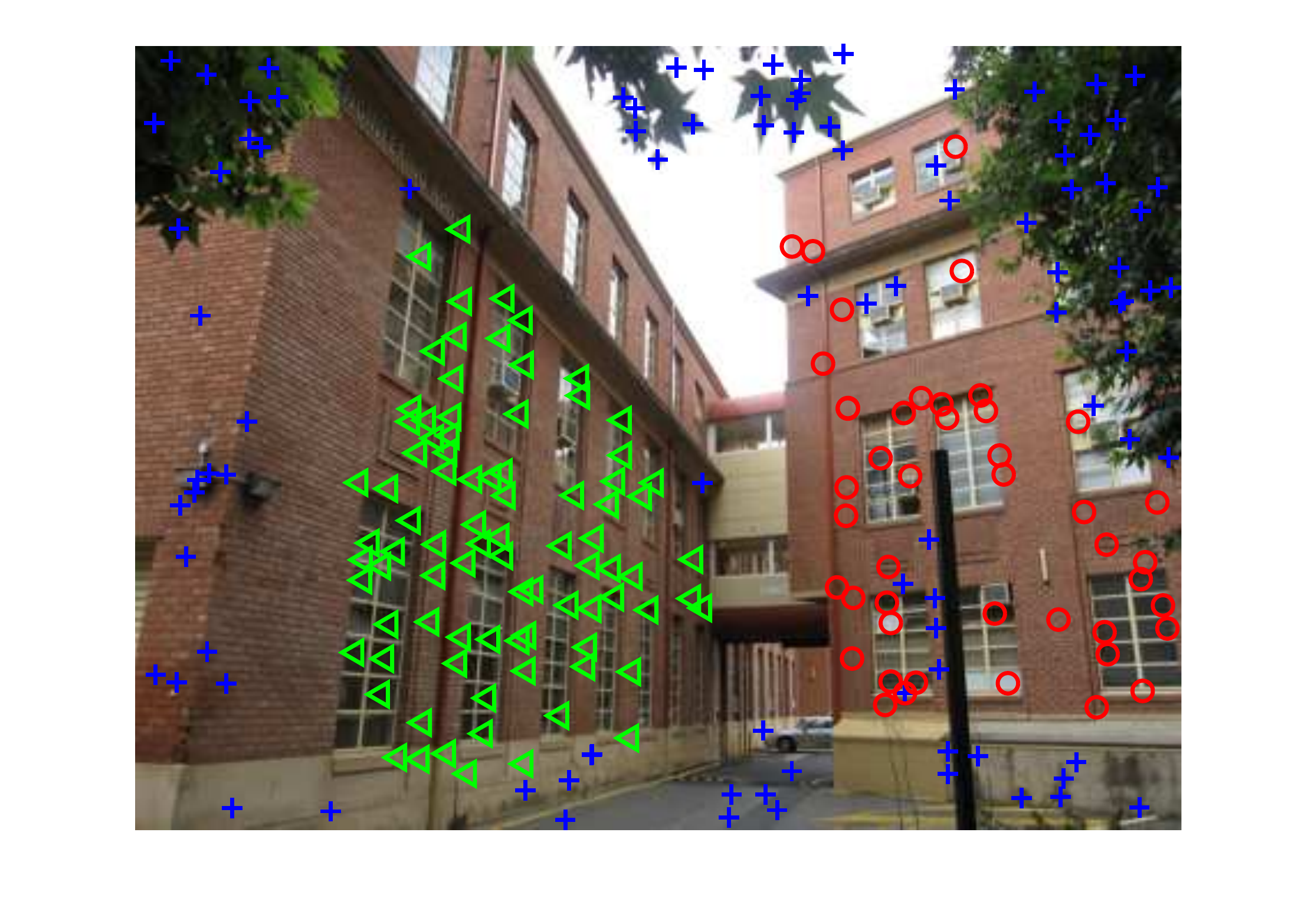}}
 \centerline{\footnotesize (e) Sene}
  \centerline{\includegraphics[width=1.16\textwidth]{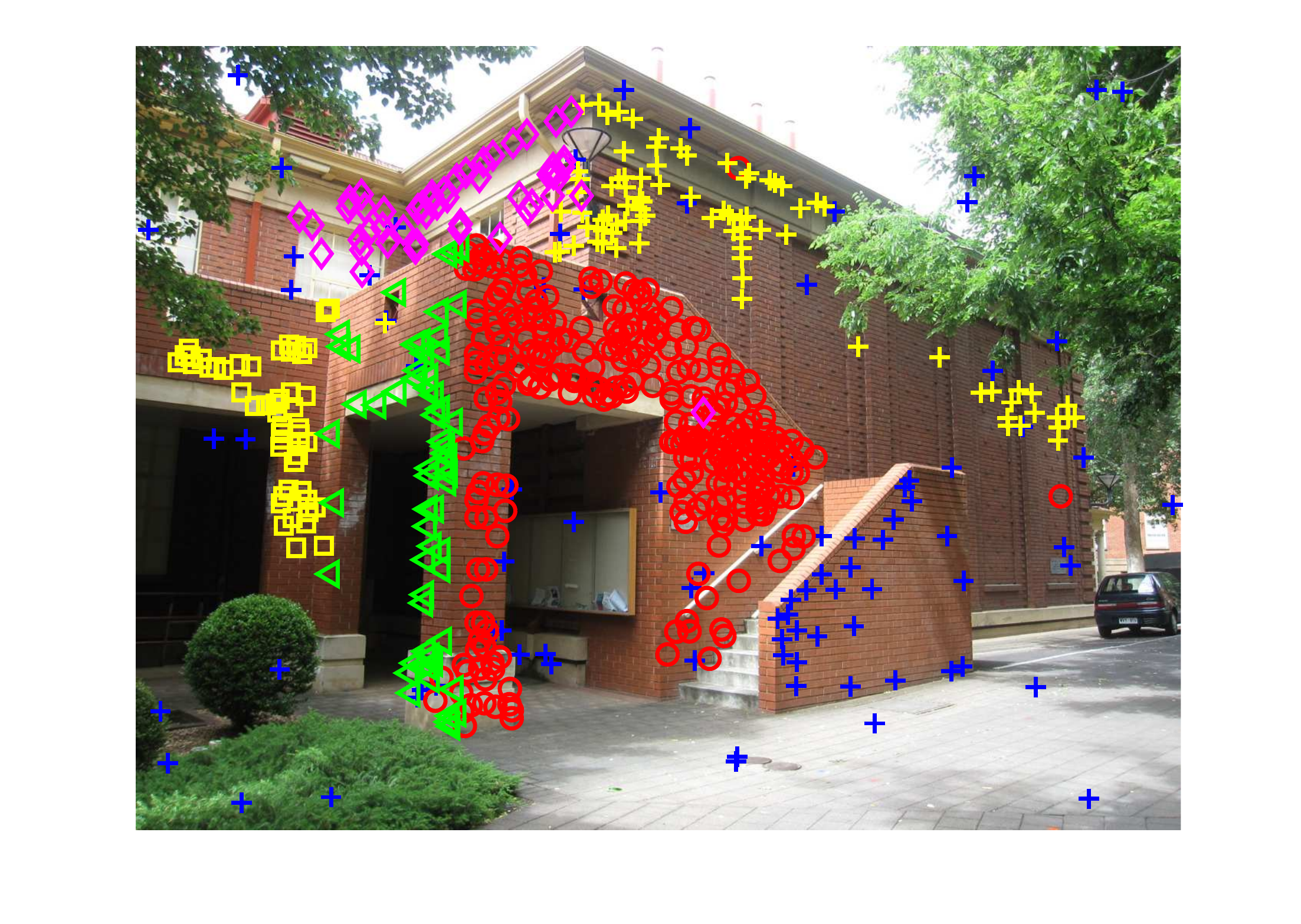}}
\centerline{\footnotesize (j) Johnsonb}
\end{minipage}
\caption{{Some fitting results obtained by MSHF2 for homography based segmentation on the AdelaideRMF dataset.}}
\label{fig:homography}
\end{figure*}
\begin{figure*}[t]
\centering
\begin{minipage}[t]{.19\textwidth}
  \centering
  \centerline{\includegraphics[width=1.16\textwidth]{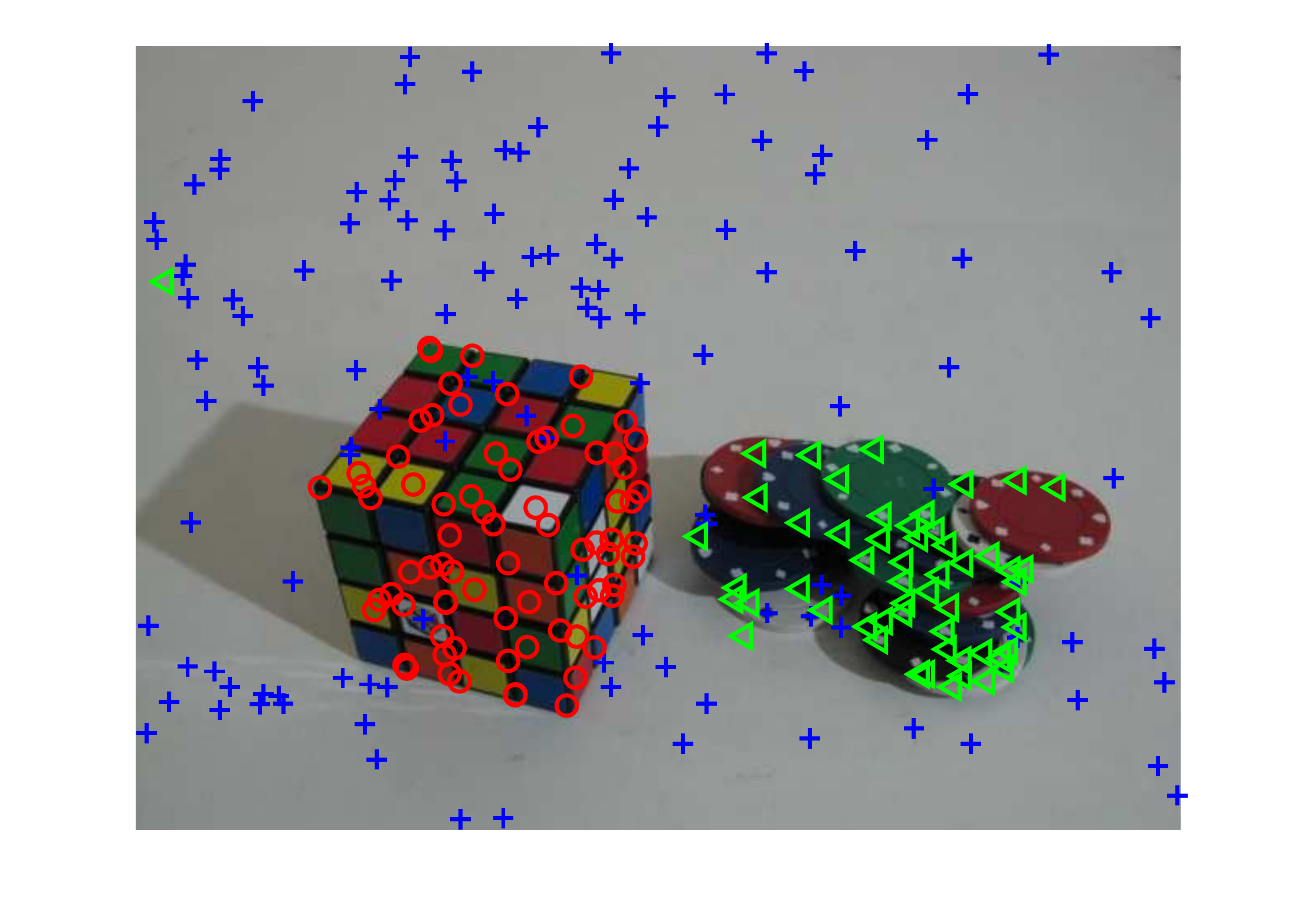}}
\centerline{\footnotesize (a) Cubechips }
  \centerline{\includegraphics[width=1.16\textwidth]{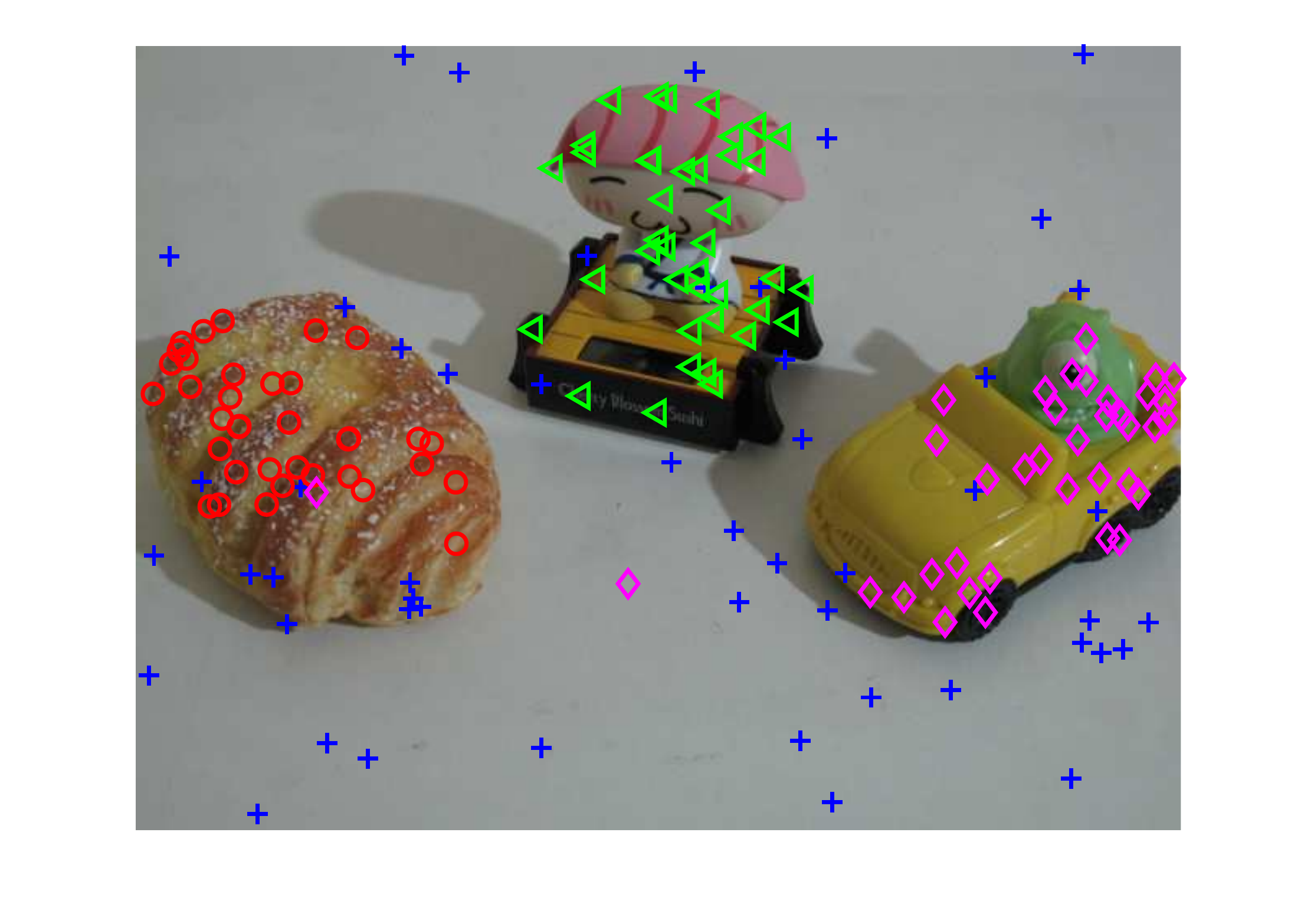}}
  \centerline{\footnotesize (f) Breadtoycar}
\end{minipage}
\begin{minipage}[t]{.19\textwidth}
  \centering
  \centerline{\includegraphics[width=1.16\textwidth]{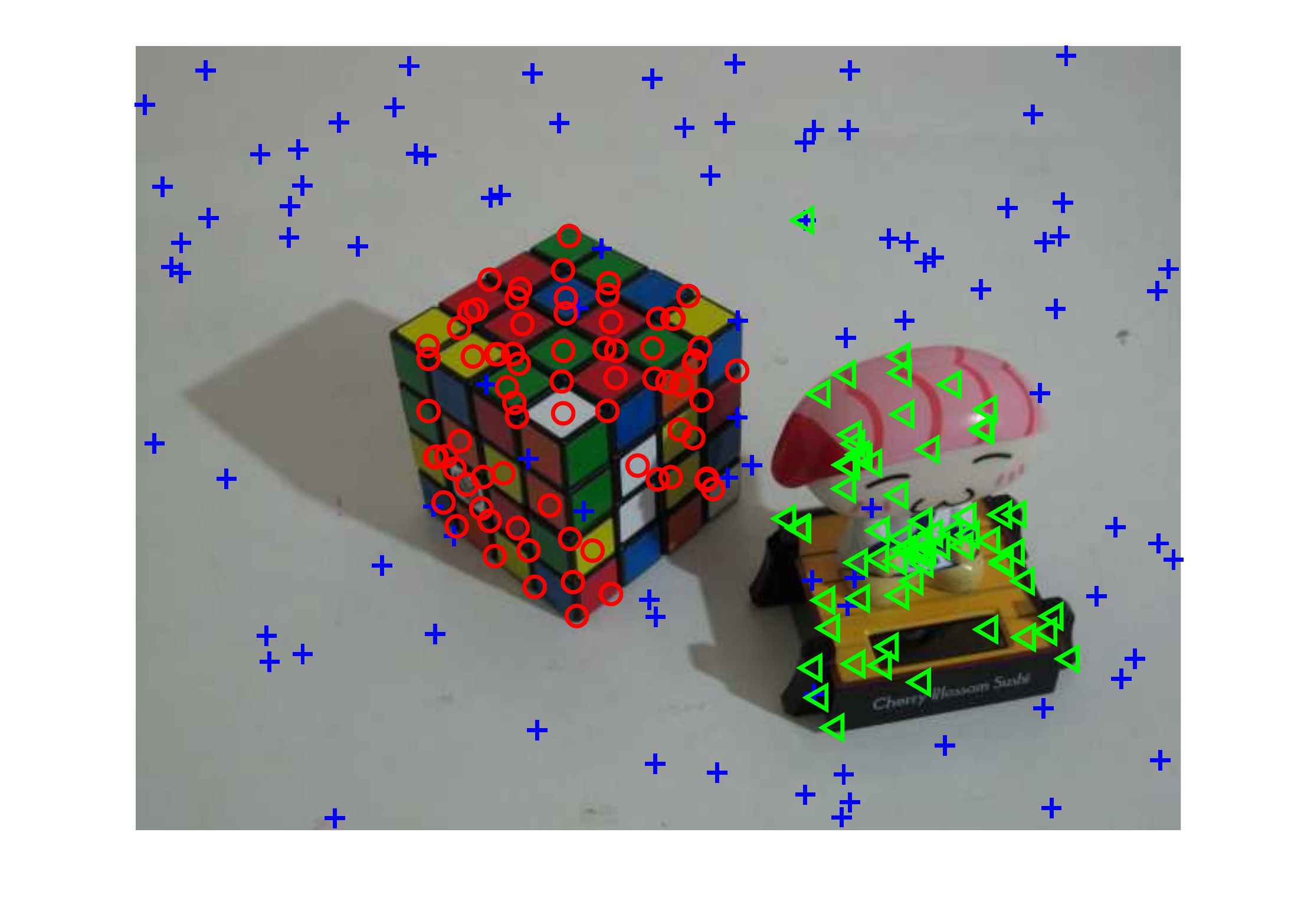}}
 \centerline{\footnotesize (b) Cubetoy }
  \centerline{\includegraphics[width=1.16\textwidth]{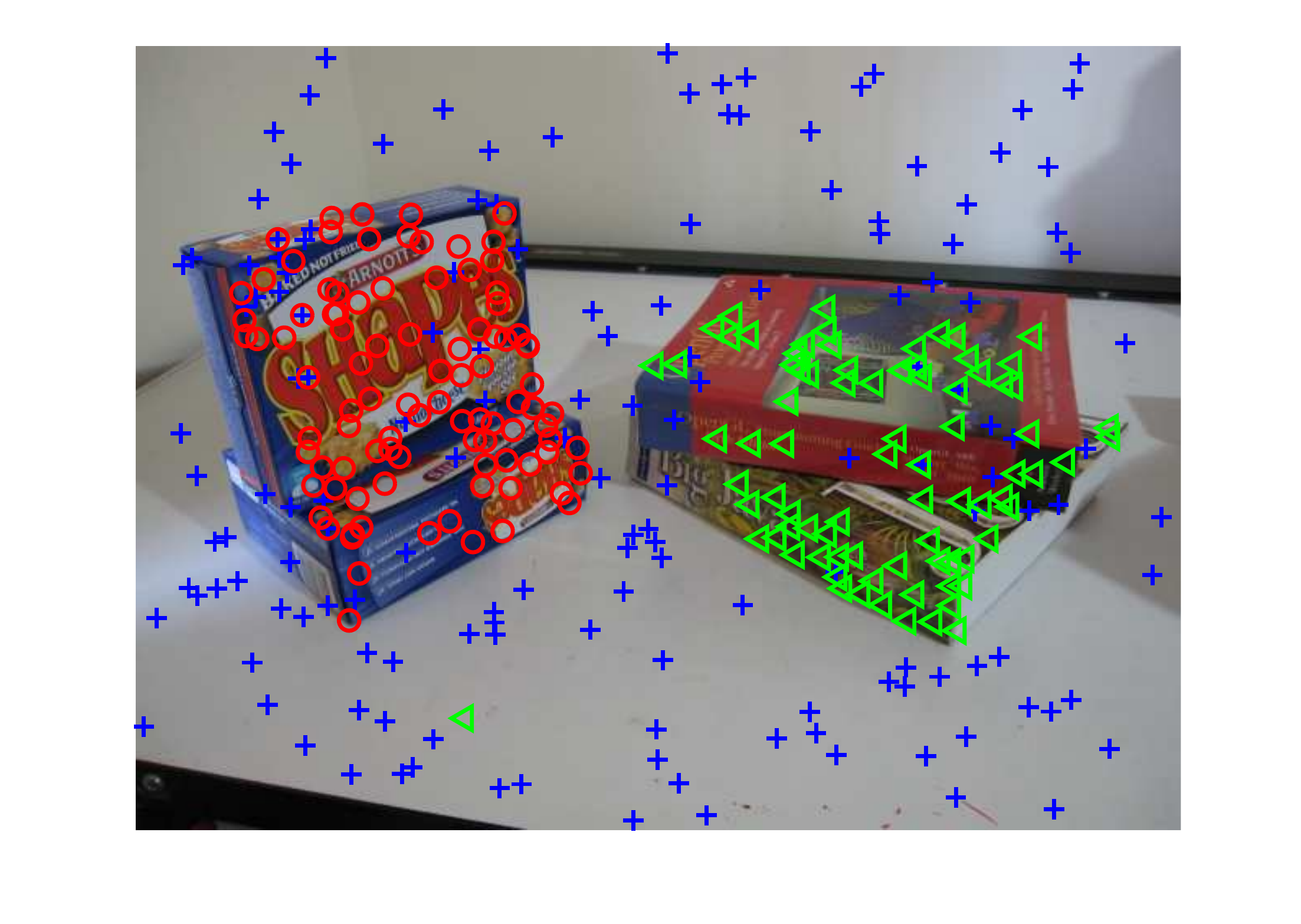}}
\centerline{\footnotesize (g) Biscuitbook}
\end{minipage}
\begin{minipage}[t]{.19\textwidth}
  \centering
  \centerline{\includegraphics[width=1.16\textwidth]{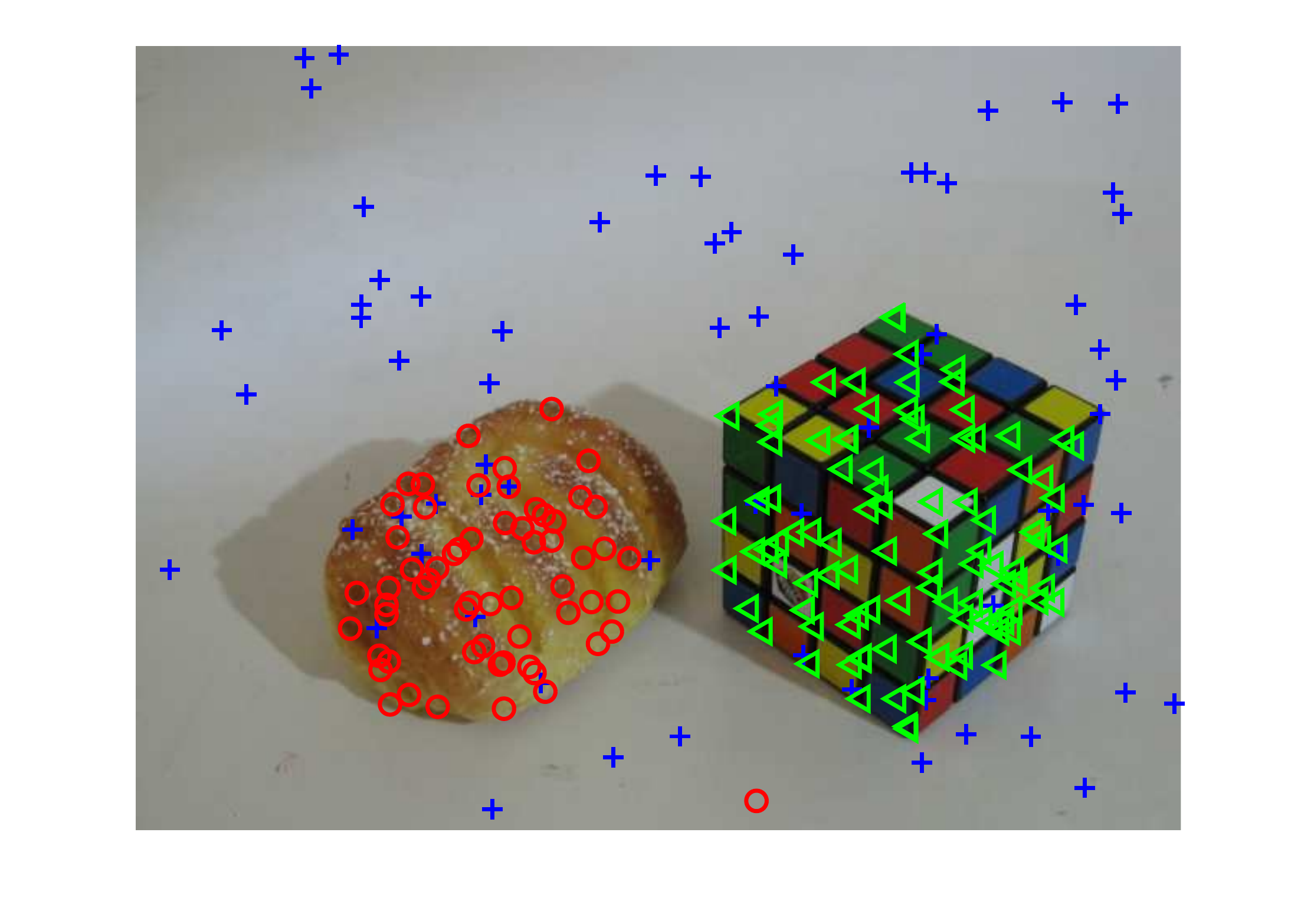}}
\centerline{\footnotesize (c) Breadcube }
  \centerline{\includegraphics[width=1.16\textwidth]{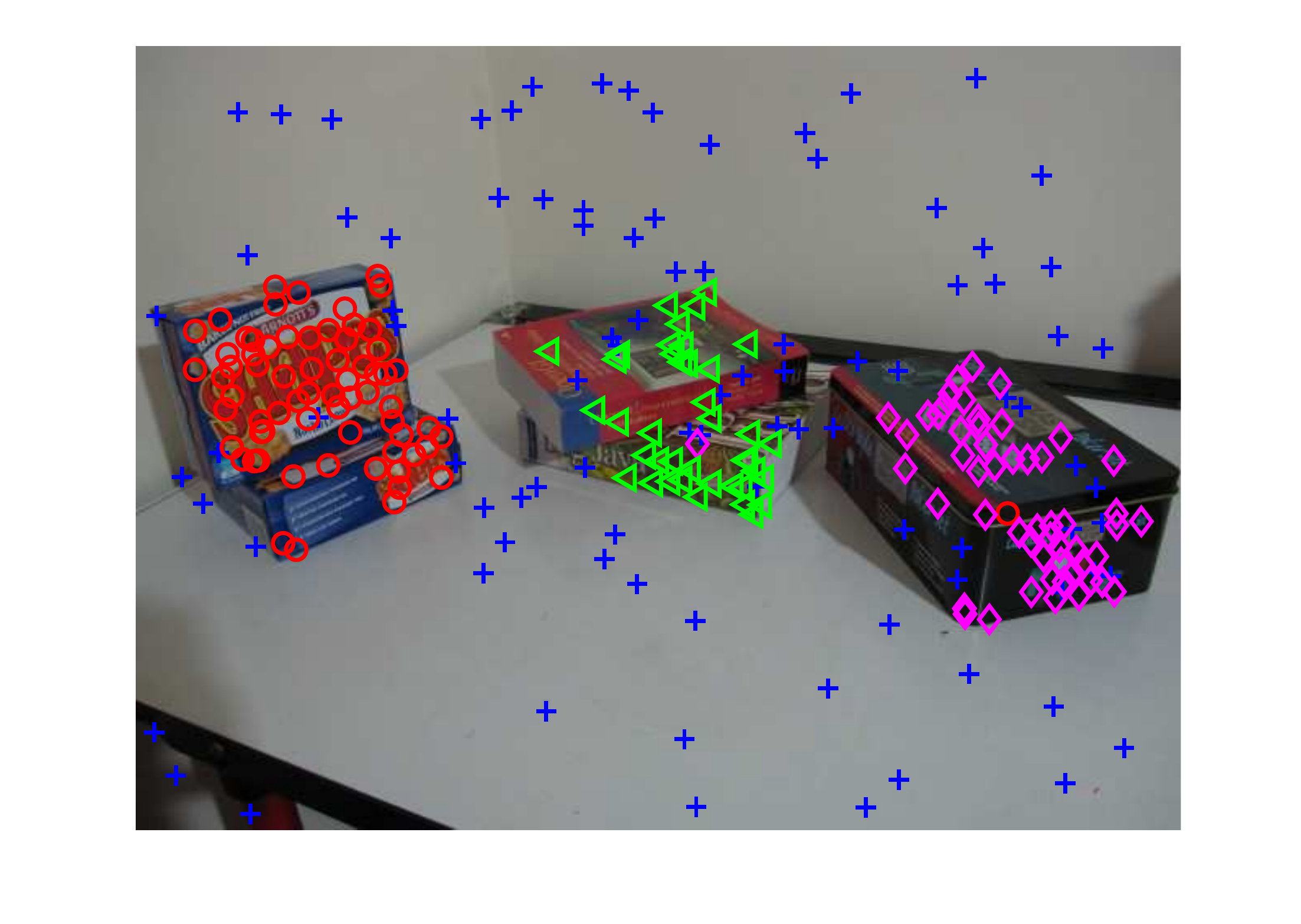}}
\centerline{\footnotesize (h) Biscuitbookbox}
\end{minipage}
\begin{minipage}[t]{.19\textwidth}
  \centering
  \centerline{\includegraphics[width=1.16\textwidth]{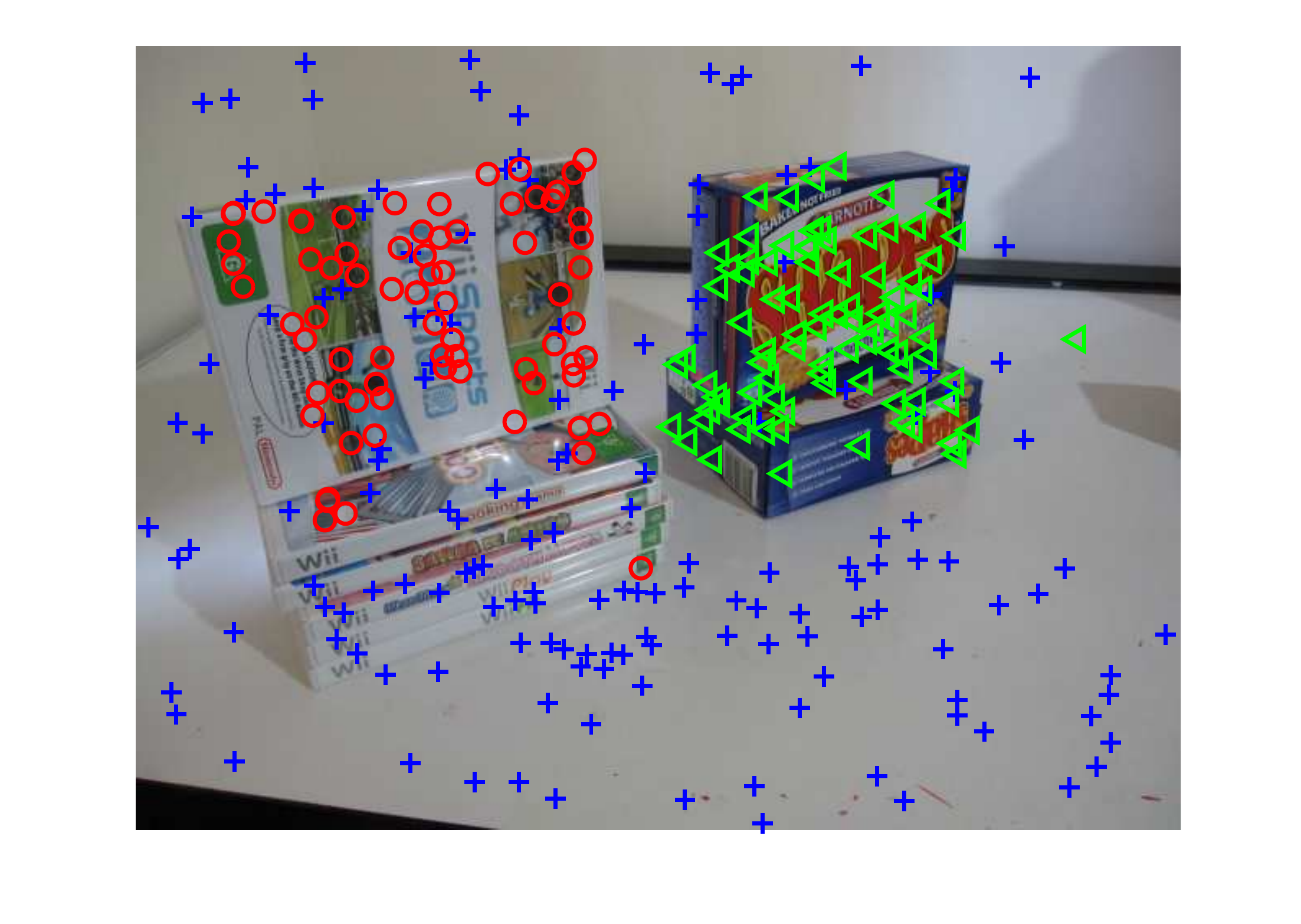}}
 \centerline{\footnotesize (d) Gamebiscuit}
  \centerline{\includegraphics[width=1.16\textwidth]{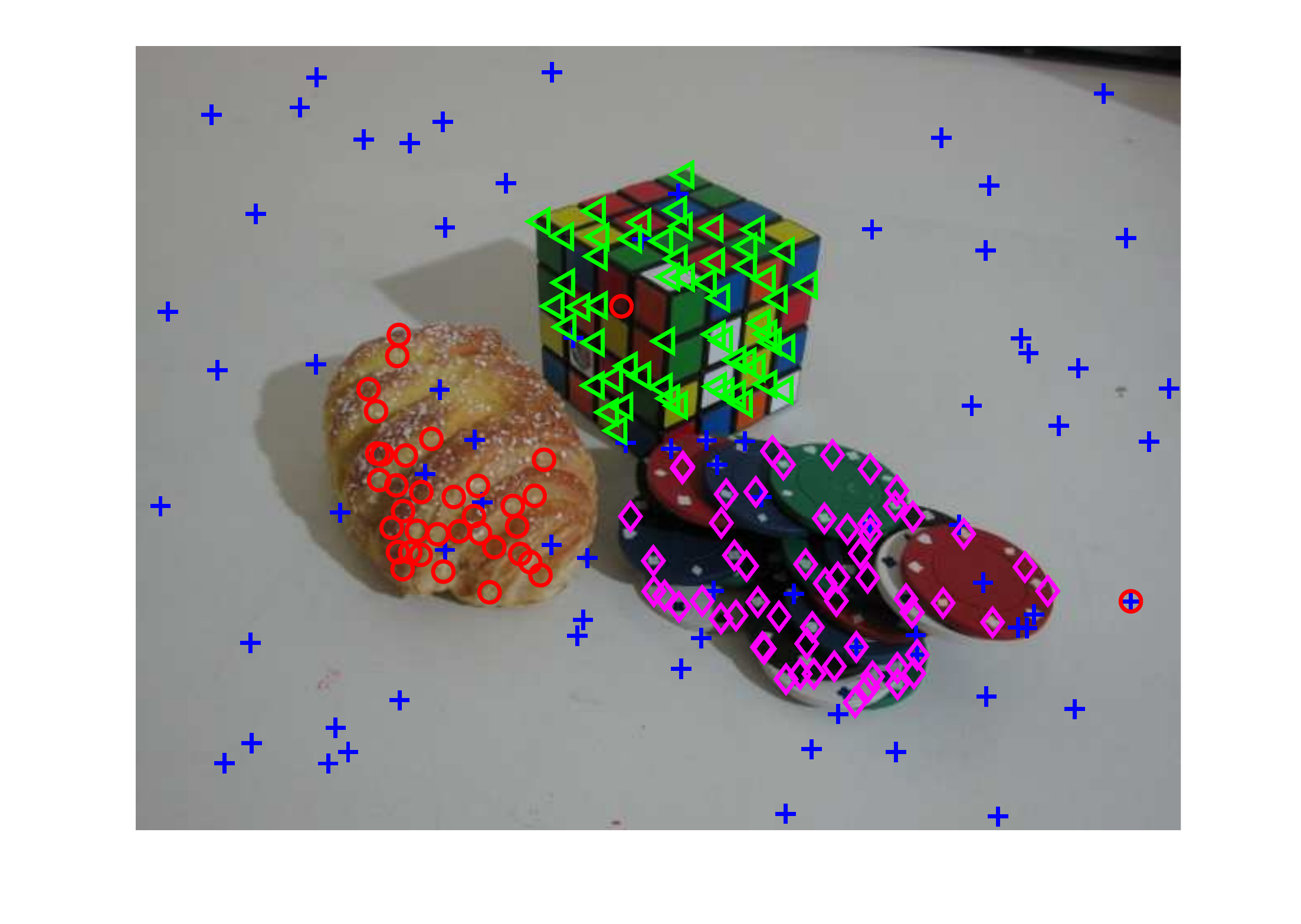}}
\centerline{\footnotesize (i) Breadcubechips}
\end{minipage}
\begin{minipage}[t]{.19\textwidth}
  \centering
  \centerline{\includegraphics[width=1.16\textwidth]{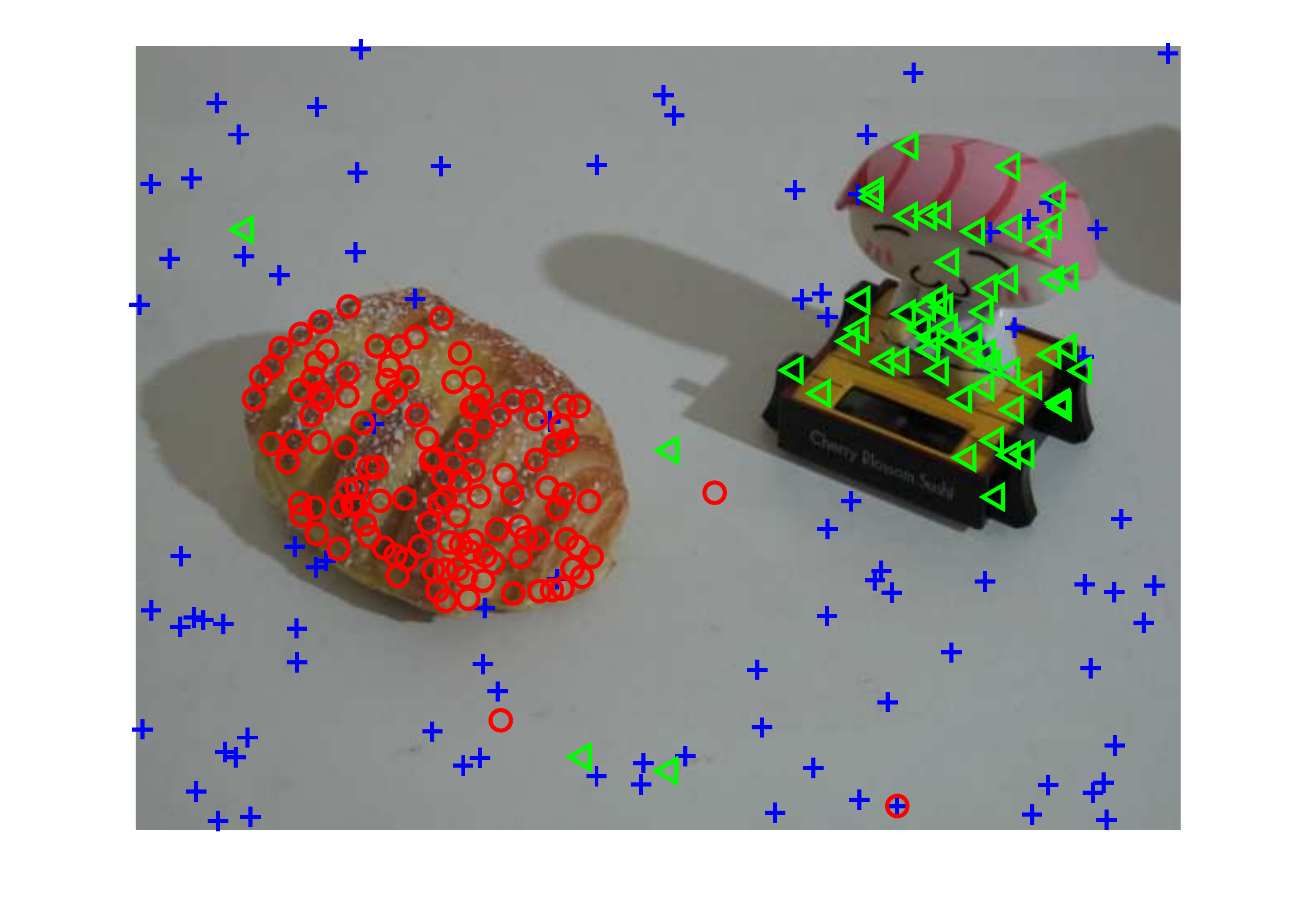}}
 \centerline{\footnotesize (e) Breadtoy}
  \centerline{\includegraphics[width=1.16\textwidth]{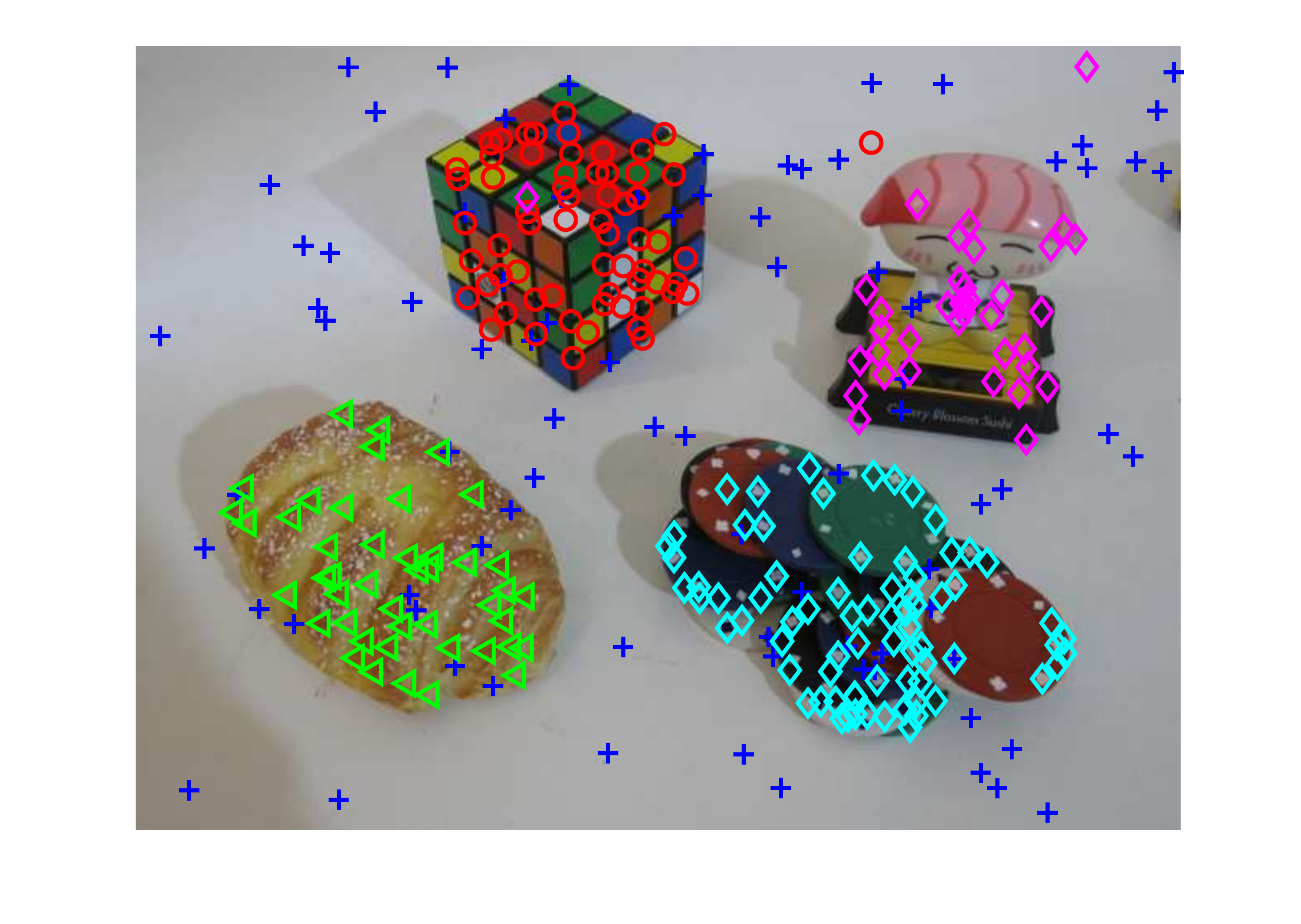}}
\centerline{\footnotesize (j) Cubebreadtoychips}
\end{minipage}
\caption{{Some fitting results obtained by MSHF2 for two-view based motion segmentation on the AdelaideRMF dataset.}}
\label{fig:motionsegmentation}
\end{figure*}
\subsubsection{Circle Fitting}
\label{sec:circlefitting}
{Next we} evaluate the performance of the {seven} fitting methods using real images for circle fitting (see Fig.~\ref{fig:circlefiting}). For the ``Coins" image, which includes five circles with {similar numbers} of inliers, there are {totally} $4,595$ edge points detected by the Canny operator. As shown in Fig.~\ref{fig:circlefiting} {and Table~\ref{table:realimagetable}}, AKSWH, T-linkage, {MSH and MSHF1/MSHF2} correctly fit all the five circles, {and MSH and MSHF1/MSHF2 are the top-three fastest among all the seven fitting methods}. In contrast, two model hypotheses estimated by KF correspond to one circle, and RCG correctly fits only four out of the five circles.

For the ``Bowls" image, which includes four circles with {significantly} unbalanced numbers of inliers, {$1,565$} edge points are detected by the Canny operator. We can see that two circles estimated by both KF and RCG overlap in the image. AKSWH correctly fits three circles but it misses one circle, because most of the model hypotheses generated for the circle with a small number of {inliers} are removed during the process that AKSWH selects significant model hypotheses. In contrast, T-linkage, {MSH and MSHF1/MSHF2} succeed in fitting all the four circles in this challenging case. {However, MSH and MSHF1/MSHF2 are much faster than T-linkage (see Table~\ref{table:realimagetable}).}

\begin{table}[th]
\caption{{Quantitative comparison results of homography based segmentation on $19$ image pairs. `\#' denotes the actual number of model instances in data.}}
\centering
\scalebox{0.90}{\tabcolsep0.04in
\begin{tabular}{|l|c|c|c|c|c|>{\columncolor{mygray}}c|>{\columncolor{mygray}}c|>{\columncolor{mygray}}c|}
\hline
Data (\#)        &                  & M1&M2&M3&M4 & M5&M6&M7 \\
\hline\hline

 &Std.      &{5.29}&{5.56}& {0.71}  &{16.54}&{\bf0.01}&{\bf0.01}&{\bf0.01}\\
   Bonython (1)                      &Avg.      &{31.86}&{5.66}& {4.54}  &{28.28}&{\bf0.00}&{\bf0.00}&{\bf0.00}\\
                      &Time      &1.01&{\bf0.49}&1.44&11.65&3.40&3.63& 0.96\\
            \hline
 &Std.      &{0.29}&{\bf0.01}& {1.91}  &{14.87}&{\bf0.01}&{\bf0.01}&{\bf0.01}\\
   Physics (1)       &Avg.      &{10.47}&{\bf0.00}& {22.54}  &{39.43}&{\bf0.00}&{\bf0.00}&{\bf0.00}\\
                       &Time      &3.02&{\bf0.24}&3.45&12.68 &1.43&2.33& 1.75\\
            \hline
                 &Std.      &{2.78}&{\bf0.01}& {1.64}  &{26.66}&{\bf0.01}&{\bf0.01}&{\bf0.01}\\
   Unionhouse (1)  &Avg.      &{27.16}&{\bf 0.30}& {2.74}  &{24.81}&{\bf0.30}&{\bf0.30}&{\bf0.30}\\
                  &Time      &1.42&{1.08}&1.63&38.05 &2.44&2.11&{\bf 1.02}\\
            \hline

 &Std.      &{0.72}&{0.37}& {\bf0.15}  &{0.45}&{\bf0.15}&{\bf0.15}&{\bf0.15}\\
   Elderhalla (2)               &Avg.      &{12.15}&{10.37}& {0.98}  &{1.17}&{\bf0.93}&{\bf0.93}&{\bf0.93}\\
                       &Time      &3.34&{\bf1.66}&2.79&15.28 &3.38& 4.57& 2.16\\
            \hline
             &Std.      &{\bf0.00}&{2.42}& {1.37}  &{0.58}&{0.96}&{1.10}&{1.10}\\
     Elderhallb (3)                &Avg.     &{34.51}&{10.12}&{13.06}&{12.63}&{3.37}&{\bf2.94}&{\bf2.94} \\
                       &Time      &3.29&{\bf1.14}&2.34&30.47&2.63&2.87&2.18\\
            \hline
            &Std.      &{7.07}&{2.91}& {0.43}  &{0.32}&{2.98}&{\bf0.31}&{\bf0.31}\\
     Hartley (2)                &Avg.      &{15.31}&{4.88}&{4.06}&{2.50}&{2.81}&{\bf1.90}&{\bf1.90} \\
                       &Time      &2.92&{\bf1.21}& 2.12&62.16&2.01&2.14&1.63\\
            \hline
             &Std.      &{6.24}&{\bf0.01}& {5.11}  &{3.38}&{5.89}&{0.82}&{0.82}\\
    Library (2)            &Avg.     &{13.19}&{9.77}&{5.79}&{4.65}&2.79&{\bf2.37}&{\bf2.37}\\
                    &Time      &3.34&{\bf 1.71}&2.13&16.04&2.31&3.65&1.80\\
            \hline
             &Std.      &{6.12}&{5.44}& {5.78}  &{0.37}&{\bf0.13}&{\bf0.13}&{\bf0.13}\\
   Sene (2)             &Avg.      &{12.08}&{10.00}&{2.00}&{0.44}&{\bf0.24}&{\bf0.24}&{\bf0.24}\\
                    &Time      &5.24&{\bf0.91}&2.73&22.78&2.59&2.11&1.80\\
            \hline
                     &Std.      &{9.57}&{\bf0.01}& {0.55}  &{0.51}&{0.27}&{0.27}&{0.27}\\
 Nese (2)   &Avg.      &{28.03}&{36.61}&{3.54}&{1.88}&{\bf0.20}&{\bf0.20}&{\bf0.20}\\
                    &Time      &5.40&{\bf0.67}&3.13&24.15&1.91&2.61&2.32\\
            \hline
             &Std.      &{2.67}&{\bf0.01}& {3.53}  &{2.58}&{0.81}&{0.86}&{0.86}\\
   Ladysymon (2)     &Avg.     &{16.46}&{22.36}&{5.74}&{5.06}&{2.87}&{\bf2.62}&{\bf2.62}\\
                                    &Time      &3.06&{\bf0.83}&2.87&20.86&2.76&3.44&2.39\\
            \hline
              &Std.      &{\bf0.01}&{8.34}& {0.14}  &{0.25}&{0.41}&{0.33}&{0.33}\\
    Oldclassicswing      &Avg.      &18.73&10.34&1.29&1.27&1.13&{\bf 1.08}&{\bf 1.08}\\
              (2)                      &Time      &2.80&{\bf 1.66}&2.25&74.89&2.44&3.56&1.81\\
            \hline
               &Std.      &{10.75}&{5.77}& {7.04}  &{4.96}&{5.40}&{\bf0.48}&{\bf0.48}\\
    Neem (3)                    &Avg.       &{10.25}&{11.17}&{5.56}&{3.82}&{2.90}&{\bf1.78}&{\bf1.78}\\
                         &Time      &6.32&{\bf0.83}&2.49&21.40&2.81&2.78&2.13\\
            \hline
            &Std.      &{3.18}&{1.84}& {12.45}  &{4.73}&{1.77}&{\bf1.43}&{\bf1.43}\\
     Johnsona (4)  &Avg.       &{25.74}&{23.06}&{8.55}&{4.03}&{3.73}&{\bf3.02}&{\bf3.02}\\
                         &Time      &16.53 &{\bf1.36}&2.93&57.11&3.63&2.96&2.24\\
            \hline
            &Std.      &{4.85}&{\bf1.81}& {6.45}  &{10.51}&{5.99}&{4.96}&{4.96}\\
     Johnsonb (7)              &Avg.        &{48.32}&{41.45}&{26.49}&{18.39}&{16.75}&{\bf16.61}&{\bf16.61} \\
                         &Time      &14.52&{\bf4.18} & 4.73& 261.62&5.67&6.48&5.00\\
            \hline

             &Std.      &{\bf2.49}&{5.22}& {4.26}  &{4.54}&{16.53}&{3.26}&{3.26}\\
  Napiera (2)     &Avg.     &{ 28.24}&{30.96}&{ 30.86}&{\bf23.37}&{32.51}&{27.78}&{27.78}\\
                                    &Time      &3.14&{\bf0.89}&1.94&29.88&2.76&3.44&2.18\\
            \hline
              &Std.      &{5.52}&{\bf0.01}& {0.38}  &{5.14}&{4.19}&{4.12}&{4.12}\\
    Napierb (3)      &Avg.      &30.42&33.59&36.33&19.92&14.21&{\bf 13.12}&{\bf 13.12}\\
                                    &Time      &2.45&{\bf 0.65}&3.33&21.93&2.18&3.31&1.87\\
            \hline

               &Std.      &{4.09}&{3.74}& {4.53}  &{6.65}&{15.84}&{9.50}&{9.50}\\
   Barrsmith (2)                    &Avg.       &{22.28}&{54.64}&{\bf20.08}&{29.33}&{37.80}&{24.48}&{24.48}\\
                         &Time      &6.06&{\bf0.62}&2.20&18.91&1.92&2.51&1.42\\
            \hline

            &Std.      &{8.54}&{8.60}& {\bf0.14}  &{4.98}&{5.39}&{0.38}&{0.38}\\
     Unihouse (5)  &Avg.       &{38.32}&{41.70}&{14.91}&{14.04}&{10.99}&{\bf9.29}&{\bf9.29}\\
                         &Time      &31.27 &{9.40}&8.67&2908.61&{\bf6.25}&13.50&10.50\\
            \hline

            &Std.      &{4.85}&{\bf1.81}& {5.10}  &{0.13}&{8.16}&{8.69}&{8.69}\\
     Bonhall (6)         &Avg.        &{48.32}&{41.45}&{38.77}&{\bf29.06}&{31.89}&{31.65}&{31.65} \\
                         &Time      &14.52&{4.18} & 7.58&835.38&{\bf4.15}&9.22&7.87\\
            \hline\hline
               &Mean      &{24.83}&{20.97}& {13.04}  &{13.89}&{8.70}&{\bf7.38}&{\bf7.38}\\
      Total         &Std.        &{11.94}&{16.59}&{12.43}&{12.32}&{12.30}&{\bf10.30}&{\bf10.30} \\
                         &Median      &25.74&{11.17} &5.79&12.63&2.87&{\bf2.37}&{\bf2.37}\\
\hline
\end{tabular}}
 \label{table:homographytable}
\end{table}
\begin{table}[!t]
\caption{{Quantitative comparison results of two-view based motion segmentation on 19 image pairs. `\#' denotes the actual number of model instances in data.}}
\centering
\scalebox{0.90}{\tabcolsep0.04in
\begin{tabular}{|l|c|c|c|c|c|>{\columncolor{mygray}}c|>{\columncolor{mygray}}c|>{\columncolor{mygray}}c|}
\hline
Data  (\#)      &                  & M1&M2&M3&M4 & M5&M6 &M7\\
\hline\hline
 &Std.        &{4.30}&{\bf0.35}& {0.42}  &{23.06}&{0.55}&{0.55}&{0.55}\\
 Biscuit (1)      &Avg.      &{\bf0.61}&{14.39}& {1.41}  &{25.84}&{1.30}&{1.30}&{1.30}\\
                 &Time      &6.08&{\bf2.13}&6.19&79.34&5.09&6.01&5.27\\
            \hline
                &Std.      &{0.55}&{0.73}& {1.24}  &{18.58}&{\bf0.42}&{\bf0.42}&{\bf0.42}\\
 Book (1)      &Avg.     &{5.88}&{7.54}&{3.47}&{24.54}&{\bf0.64}&{\bf0.64}&{\bf0.64} \\
               &Time      &6.02&{\bf 0.75}&5.49&24.97&4.08&6.56&4.81\\
            \hline
               &Std.      &{\bf0.22}&{0.88}& {0.86}  &{22.34}&{0.66}&{0.66}&{0.66}\\
   Cube (1)   &Avg.      &{8.70}&{22.48}&{2.21}&{23.07}&{\bf2.08}&{\bf2.08}&{\bf2.08} \\
                &Time      &7.02&{\bf1.80}&5.93&79.48&6.94&6.01&5.11\\
            \hline
              &Std.      &{\bf0.11}&{2.11}& {0.97}  &{32.65}&{1.07}&{0.74}&{0.74}\\
 Game (1)   &Avg.     &{18.81}&{19.31}&{2.61}&{38.15}& {\bf2.44}&{\bf2.44}&{\bf2.44}\\
              &Time      &7.11&{\bf 1.18}&5.87&42.70&6.99&5.50&4.95\\
            \hline
 &Std.      &{4.30}&{1.04}& {3.10}  &{\bf0.85}&{1.26}&{0.98}&{0.98}\\
 Cubechips (2)       &Avg.      &{8.42}&{13.43}& {4.72}  &{5.63}&{3.80}&{\bf3.55}&{\bf3.55}\\
                       &Time      &7.94&{\bf1.69}&5.10&64.87 &6.45&6.71&5.18\\
            \hline
             &Std.      &{10.80}&{1.38}& {3.78}  &{0.80}&{1.27}&{\bf0.79}&{\bf0.79}\\
 Cubetoy (2)      &Avg.     &{12.53}&{13.35}&{7.23}&{5.62}&{3.21}&{\bf2.16}&{\bf2.16} \\
                       &Time      &6.08&{\bf1.34}&4.97&51.65&5.74&6.30&4.89\\
            \hline
             &Std.      &{3.92}&{3.27}& {6.06}  &{1.32}&{0.95}&{\bf0.78}&{\bf0.78}\\
   Breadcube (2)                  &Avg.      &{14.83}&{12.60}&{5.45}&{4.96}&{2.69}&{\bf2.31}&{\bf2.31} \\
                       &Time      &7.07&{\bf1.53}&6.10&46.17&6.01&6.05&4.82\\
            \hline
             &Std.      &{\bf0.19}&{9.53}& {10.74}  &{1.85}&{2.71}&{0.74}&{0.74}\\
 Gamebiscuit (2)                &Avg.     &{13.78}&{9.94}&{7.01}&{7.32}&3.72&{\bf1.95}&{\bf1.95}\\
                    &Time      &7.66&{\bf 2.36}&6.44&91.49&6.93&7.55&5.81\\
            \hline
             &Std.      &{3.41}&{5.81}& {6.72}  &{\bf1.50}&{8.07}&{7.76}&{7.76}\\
Breadtoy  (2)    &Avg.      &{8.36}&{20.48}&{15.03}&{7.33}&{5.90}&{\bf4.86}&{\bf4.86}\\
                    &Time      &22.51&{\bf2.12}&15.18&68.62&9.48&13.02&5.87\\
            \hline
             &Std.      &{8.22}&{8.15}& {8.58}  &{\bf1.43}&{1.96}&{1.96}&{1.96}\\
 Breadtoycar  (3)    &Avg.     &{16.87}&{26.51}&{9.04}&{\bf4.42}&{6.63}&{5.42}&{5.42}\\
                                    &Time      &5.70&{\bf0.98}&4.56&24.15&5.48&6.18&5.06\\
            \hline
             &Std.      &{10.71}&{5.31}& {1.40}  &{\bf1.16}&{1.82}&{1.82}&{1.82}\\
Biscuitbook (2)      &Avg.      &12.90&3.82&2.89&2.55&2.60&{\bf 2.40}&{\bf 2.40}\\
                                    &Time      &7.84&{\bf2.03}&6.59&129.47&8.52&9.60&6.57\\
            \hline
             &Std.      &{4.00}&{1.98}& {3.17}  &{1.60}&{0.92}&{\bf0.90}&{\bf0.90}\\
 Biscuitbookbox (3)     &Avg.       &{16.06}&{16.87}&{8.54}&{1.93}&{\bf1.54}&{\bf1.54}&{\bf1.54}\\
             &Time      &8.50&{\bf1.71}&5.11&53.44&6.11&6.35&5.44\\
            \hline
             &Std.      &{7.26}&{6.64}& {3.41}  &{7.03}&{4.39}&{\bf1.75}&{\bf1.75}\\
 Breadcubechips (3)     &Avg.       &{33.43}&{26.39}&{7.39}&{\bf1.06}&{1.74}&{1.74}&{1.74}\\
              &Time      &16.53 &{\bf1.36}&2.93&57.11&8.35&13.28&4.35\\
            \hline
             &Std.      &{4.99}&{12.18}& { 0.95}  &{7.72}&{7.18}&{6.62}&{6.62}\\
   Cubebreadtoychips    &Avg.        &{31.07}&{37.95}&{14.95}&{\bf3.11}&{4.28}&{4.25}&{4.25} \\
              (4)                  &Time      &25.68&{\bf1.83} &5.99&91.05&9.16&13.09&4.70\\
            \hline
             &Std.      &{7.90}&{\bf1.29}& {5.26}  &{9.45}&{8.96}&{6.14}&{6.14}\\
 Breadcartoychips &Avg.      &{26.96}&{49.36}& {42.86}  &{\bf16.96}&{33.92}&{25.06}&{25.06}\\
       (4)     &Time      &6.91&{\bf1.96}&4.92&40.76&6.23&5.14&4.02\\
            \hline
             &Std.      &{9.09}&{1.59}& {1.67}  &{\bf0.44}&{5.10}&{7.56}&{7.56}\\
 Carchipscube (3)      &Avg.     &{\bf10.96}&{38.96}&{51.75}&{17.51}&{20.72}&{25.51}&{25.51} \\
            &Time      &6.52&{\bf1.58}&4.60&18.52&4.84&4.32&3.81\\
            \hline
             &Std.      &{10.23}&{\bf0.48}& {6.06}  &{1.20}&{9.74}&{6.57}&{6.57}\\
   Toycubecar (3)                  &Avg.      &{27.05}&{38.75}&{34.55}&{16.20}&{20.35}&{\bf14.00}&{\bf14.00} \\
            &Time      &7.77&{\bf0.68}&5.72&25.35&3.94&6.06&4.73\\
            \hline
             &Std.      &{9.90}&{\bf 1.50}& {5.96}  &{1.91}&{ 4.13}&{7.77}&{7.77}\\
 Boardgame (3)                   &Avg.     &{30.21}&{45.16}&{48.13}&{28.60}&21.68&{\bf21.57}&{\bf21.57}\\
                   &Time      &2.36&{\bf 1.39}&7.57&58.07&4.34&5.11&4.63\\

            \hline
                         &Std.      &{3.41}&{\bf0.92}& {10.74}  &{1.85}&{5.32}&{2.08}&{2.08}\\
 Dinobooks  (3)                &Avg.     &{30.86}&{54.27}&{24.72}&{19.52}&{\bf16.08}&{18.05}&{18.05}\\
                    &Time      &12.93&{\bf 2.33}&6.66&118.96&5.54&5.61&4.89\\
                                \hline\hline
               &Mean      &{17.27}&{24.81}& {15.47}  &{13.38}&{8.17}&{\bf7.41}&{\bf7.41}\\
        Total         &Std.        &{9.81}&{15.00}&{16.55}&{10.91}&{9.48}&{\bf8.63}&{\bf8.63} \\
                         &Median      &14.83&{20.48} &7.39&7.33&3.72&{\bf2.44}&{\bf2.44}\\

\hline
\end{tabular}}
 \label{table:motionsegmentation}
\end{table}

\subsubsection{Homography Based Segmentation}
\label{sec:homographbasedsegmentation}
We also evaluate the performance of the {seven} fitting methods using the {$19$} real image pairs from the AdelaideRMF dataset~\cite{wong2011dynamic}\footnote{\url{http://cs.adelaide.edu.au/~hwong/doku.php?id=data}} {(the dataset contains $19$ image pairs designed for homography fitting - which we use here - and $19$ image pairs for motion segmentation - which we use in Sec.~\ref{sec:motionsegmentation} devoted to that topic)} for homography based segmentation. We repeat each experiment 50 times, and show {the standard variances, the average fitting errors (in percentage)} and the average CPU time (in seconds) in Table~\ref{table:homographytable} (we exclude the time used for sampling and generating potential hypotheses, which is the same for all the fitting methods). {Some fitting results} obtained by MSHF2 are also shown in Fig.~\ref{fig:homography}.

{From Fig.~\ref{fig:homography} and Table~\ref{table:homographytable}, we can see that MSHF1/MSHF2 obtain good results, achieving the lowest average fitting errors in $16$ out of $19$ image pairs. Although MSHF1 is slightly slower than MSH, it significantly improves the fitting accuracy over MSH in $12$ out of $19$ image pairs. The reason behind this is that MSHF1 removes less vertices corresponding to model hypotheses than MSH, and thus MSHF1 takes more time to seek modes in a hypergraph. However, MSHF1 retains more good vertices corresponding to significant model hypotheses, which improves its fitting accuracy. MSHF2 achieves the same fitting errors as MSHF1, but it is faster than MSHF1 in all the 19 image pairs. In contrast, AKSWH only succeeds in fitting $10$ out of $19$ image pairs with low fitting errors. Although T-linkage can also achieve low fitting errors in most of image pairs, it is much slower than the other six competing methods. Both KF and RCG achieve bad results in most cases. We note that KF clusters many outliers together with inliers, and RCG is very sensitive to its parameters when there exist many bad model hypotheses in the generated model hypotheses. For the overall fitting errors, MSH and MSHF1/MSHF2 achieve the top-three best performance on the mean and median fitting errors among all the seven competing fitting methods. MSHF1/MSHF2 also achieve the lowest standard variances of fitting errors. For the performance of computational time, RCG achieves the lowest values in $17$ out of $19$ image pairs, but it cannot obtain low fitting errors. In short, MSH and MSHF1/MSHF2 can achieve low fitting errors within reasonable time for most image pairs.
}

\subsubsection{Two-view Based Motion Segmentation}
\label{sec:motionsegmentation}
{For the two-view based motion segmentation problem,  we use the other $19$ image pairs of the AdelaideRMF dataset to evaluate the performance of the competing fitting methods. The results are shown in Table~\ref{table:motionsegmentation} and Fig.~\ref{fig:motionsegmentation}.

From Table~\ref{table:motionsegmentation} and Fig.~\ref{fig:motionsegmentation}, we can see that both KF and RCG achieve high fitting errors and both fail in most cases. This is because when a large number of model hypotheses are generated for two-view based motion segmentation to cover all the model instances in the data, a large proportion of bad model hypotheses may lead to inaccurate similarity measure between data points, which results in the wrong estimation of the number of model instances by KF and RCG. AKSWH achieves better results than both KF and RCG on the average fitting errors. T-linkage, MSH and MSHF1/MSHF2 achieve low fitting errors, while MSHF1/MSHF2 obtain relatively better results than T-linkage and MSH (as shown in Table~\ref{table:motionsegmentation}). MSHF1/MSHF2 achieve the lowest average fitting errors in $12$ out of $19$ image pairs and the second lowest ones in $6$ out of the remaining $7$ image pairs. MSHF1/MSHF2 also achieve the best performance on the standard variances, the mean and median fitting errors for the overall results. It is worth pointing out that MSHF1/MSHF2 achieve lower average fitting errors than MSH in $11$ out of $19$ image pairs (and they achieve the same average fitting errors in $6$ out of the remaining 8 image pairs). {This benefits from the step of hypergraph reduction.}

For the computational time, MSH and MSHF1/MSHF2 do not achieve better results than RCG. However, MSH and MSHF1/MSHF2 are significantly faster than KF in most data. Compared with T-linkage, which achieves good performance on fitting accuracy, MSH and MSHF1/MSHF2 show significant superiority regarding to the computational time for all the $19$ image pairs. Note that MSHF2 is faster than MSHF1 for all the $19$ image pairs due to the use of neighboring constraint, which can effectively reduce the computational cost.
}
\begin{figure}[t]
\centering
\begin{minipage}{.24\textwidth}
\centerline{\includegraphics[width=1.00\textwidth]{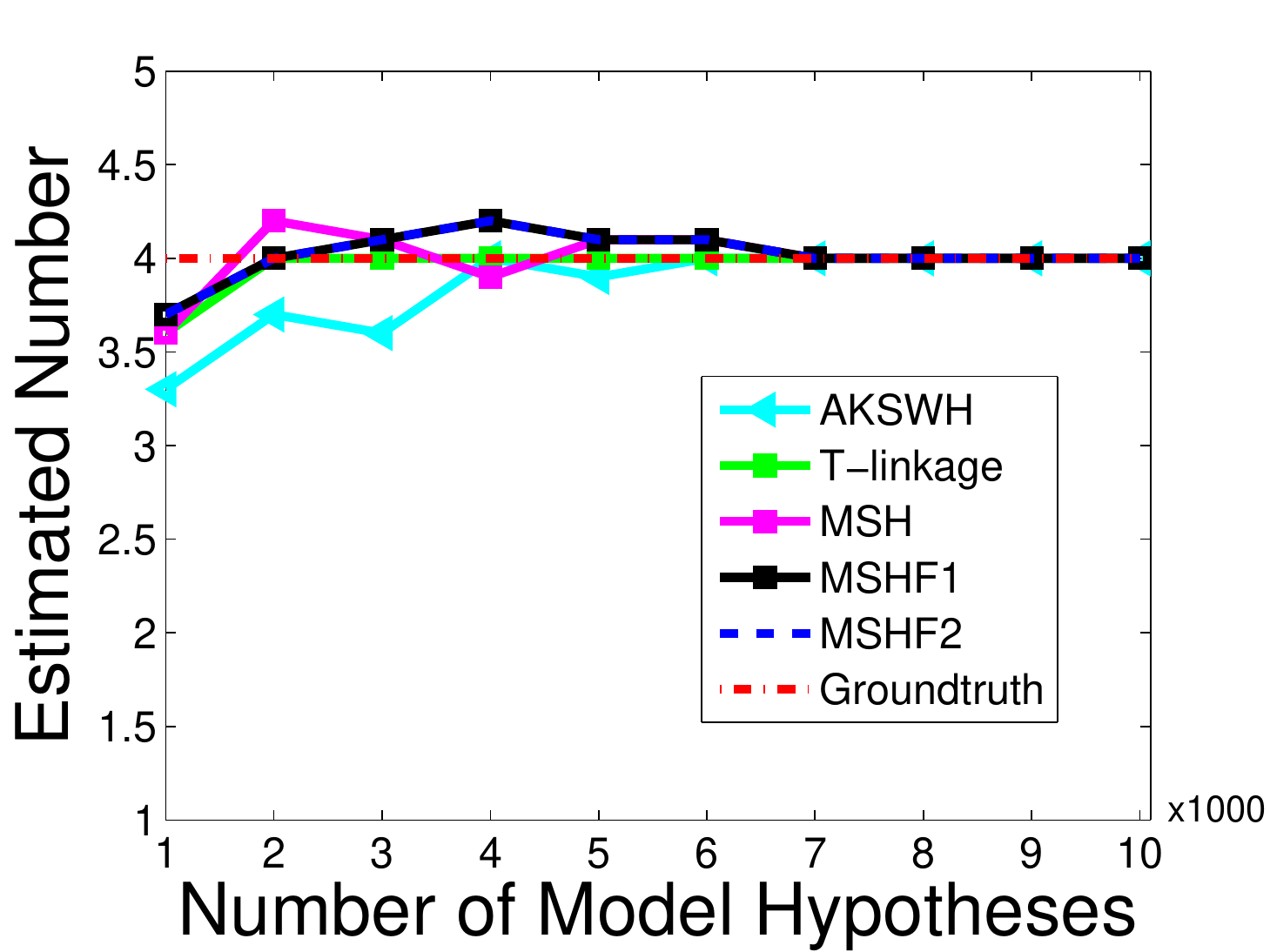}}
  \centerline{(a)}
  \centerline{\includegraphics[width=1.00\textwidth]{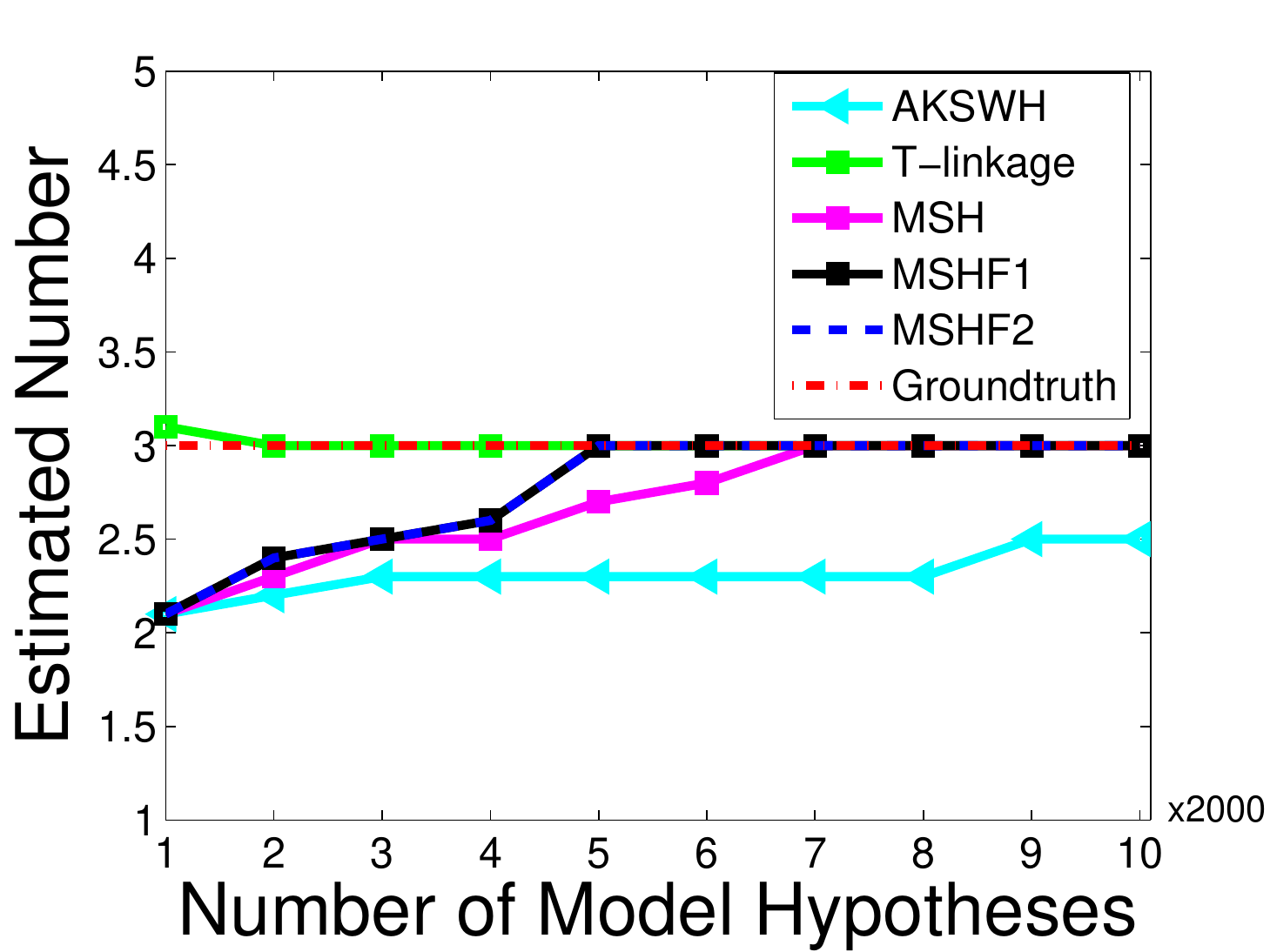}}
  \centerline{(c)}
\end{minipage}
\begin{minipage}{.24\textwidth}
\centerline{\includegraphics[width=1.00\textwidth]{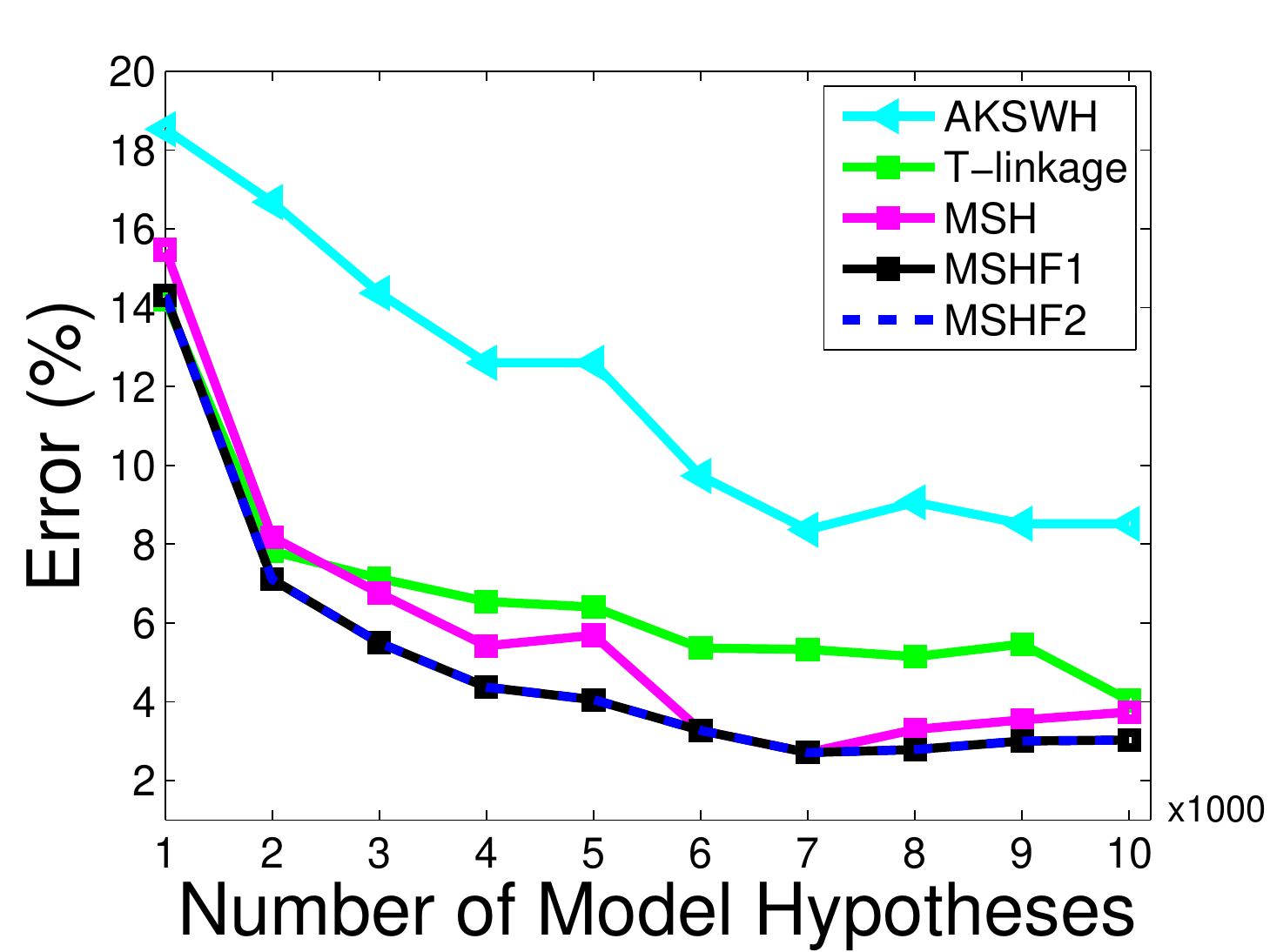}}
  \centerline{(b)}
  \centerline{\includegraphics[width=1.00\textwidth]{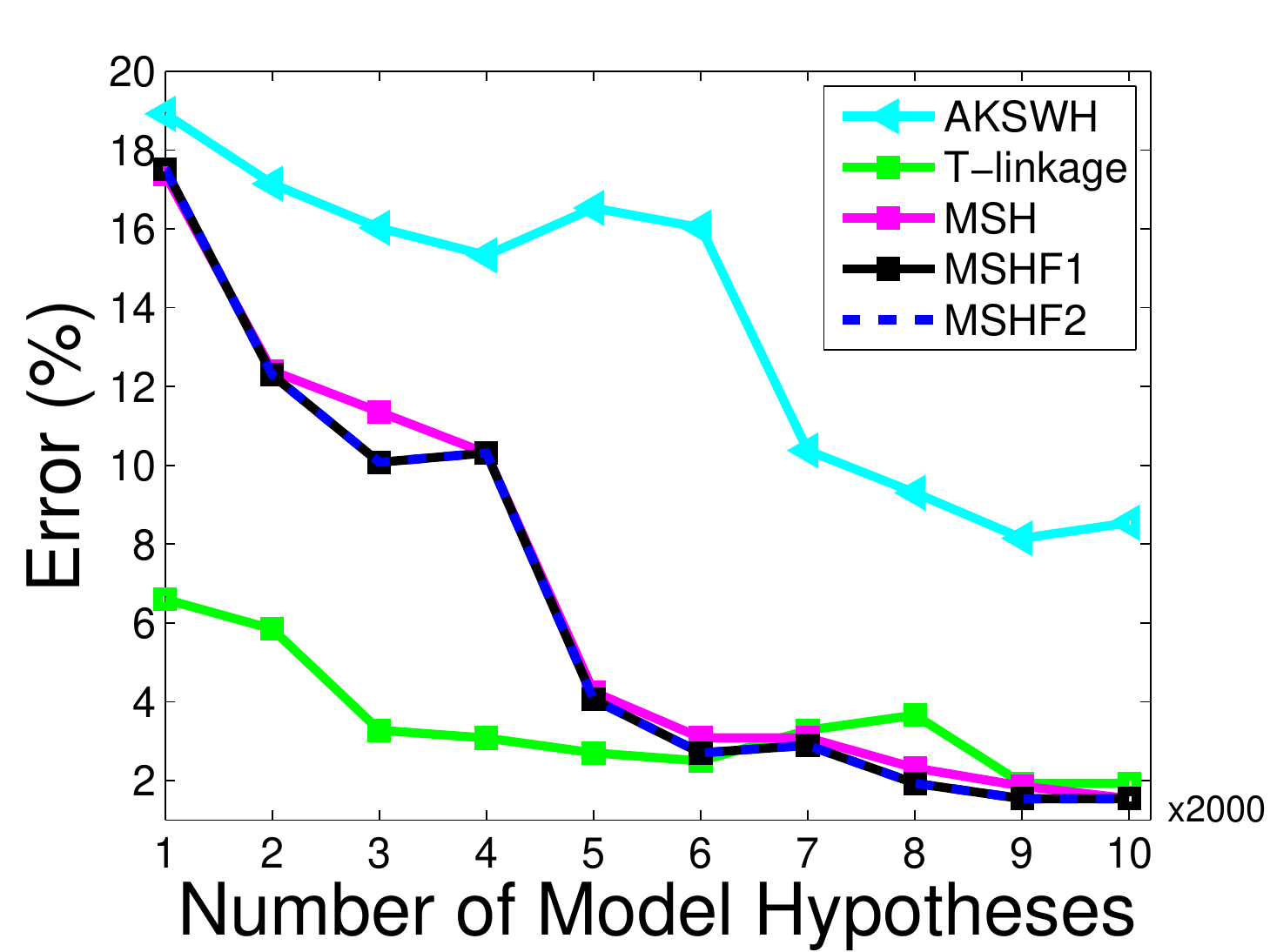}}
  \centerline{(d)}
\end{minipage}
\caption{{Quantitative comparison for the estimated number of model instances and the average fitting errors with different numbers of generated model hypotheses: (a) and (b) respectively show the performance comparison for homography based segmentation on the ``Johnsona" image pair; (c) and (d) respectively show the performance comparison for two-view based motion segmentation on the ``Biscuitbookbox" image pair.}}
\label{fig:diffnumberofhypotheses}
\end{figure}

{We also evaluate the performance of the five competing methods (i.e., AKSWH, T-linkage, MSH and MSHF1/MSHF2) on the estimated number of model instances and the average fitting errors with different numbers of generated model hypotheses for homography based segmentation and two-view based motion segmentation, as shown in Fig.~\ref{fig:diffnumberofhypotheses} (we do not show the results of KF and RCG since they cannot achieve good results). We can see that, the number of generated model hypotheses has significant influence on the results of all the five fitting methods. When the number of model hypotheses is large, all the five fitting methods achieve relatively lower fitting errors. For the estimated number of model instances, T-linkage achieves the best results for both homography based segmentation and two-view based motion segmentation. However, MSH and MSHF1/MSHF2 also correctly estimate the number of model instances when sufficient model hypotheses are generated. For the fitting errors, MSH and MSHF1/MSHF2 achieve the top-three best results for homography based segmentation, and they also achieve the top-three best results for two-view based motion segmentation when the number of model hypotheses is larger than $12,000$.
}

\section{Limitations}
\label{sec:limitations}
\begin{figure}[h!]
\centering
\begin{minipage}{.26\textwidth}
\centerline{\includegraphics[width=1.0\textwidth]{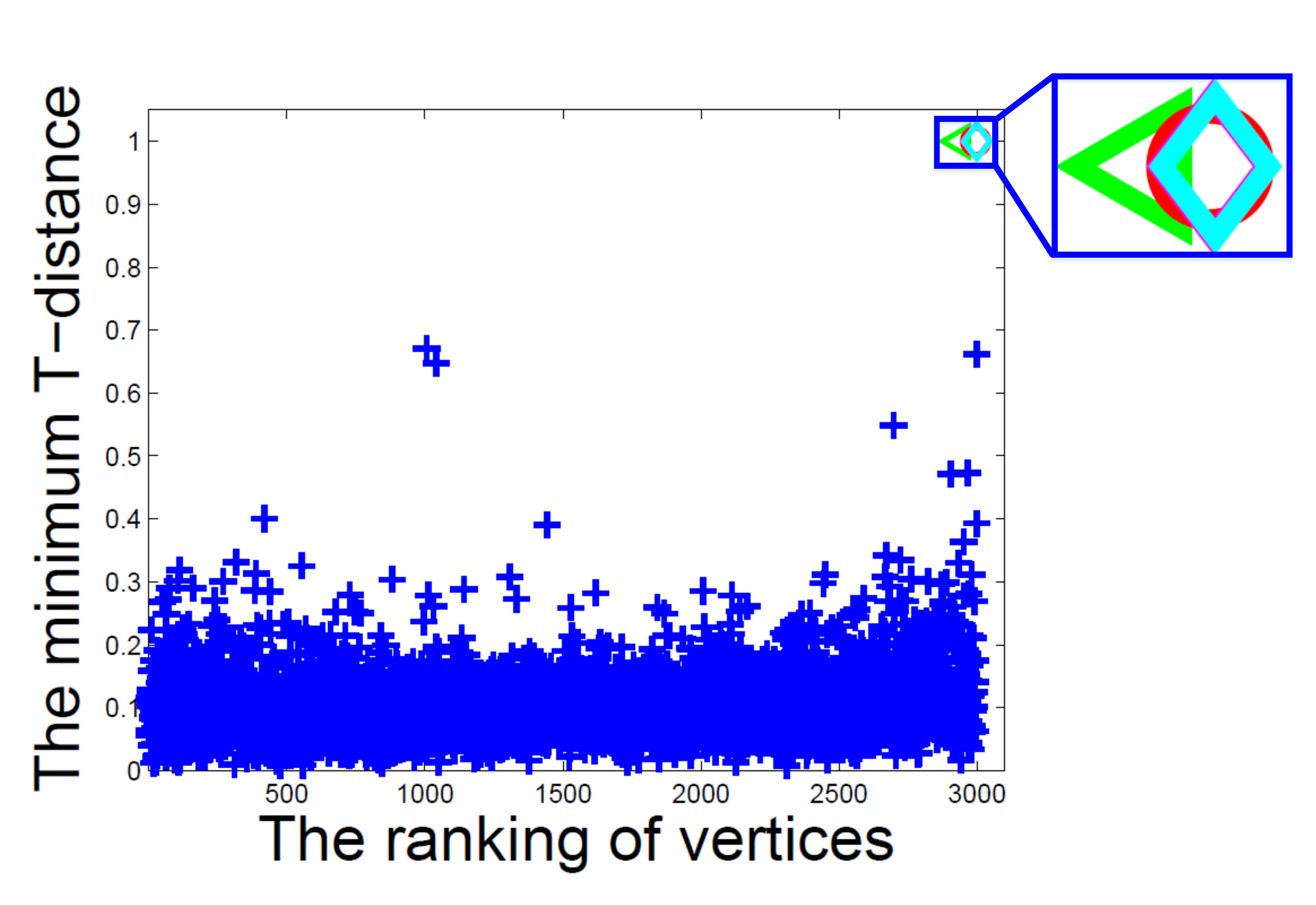}}
 \centerline{(a)}
\end{minipage}
\begin{minipage}{.22\textwidth}
\centerline{\includegraphics[width=1.1\textwidth]{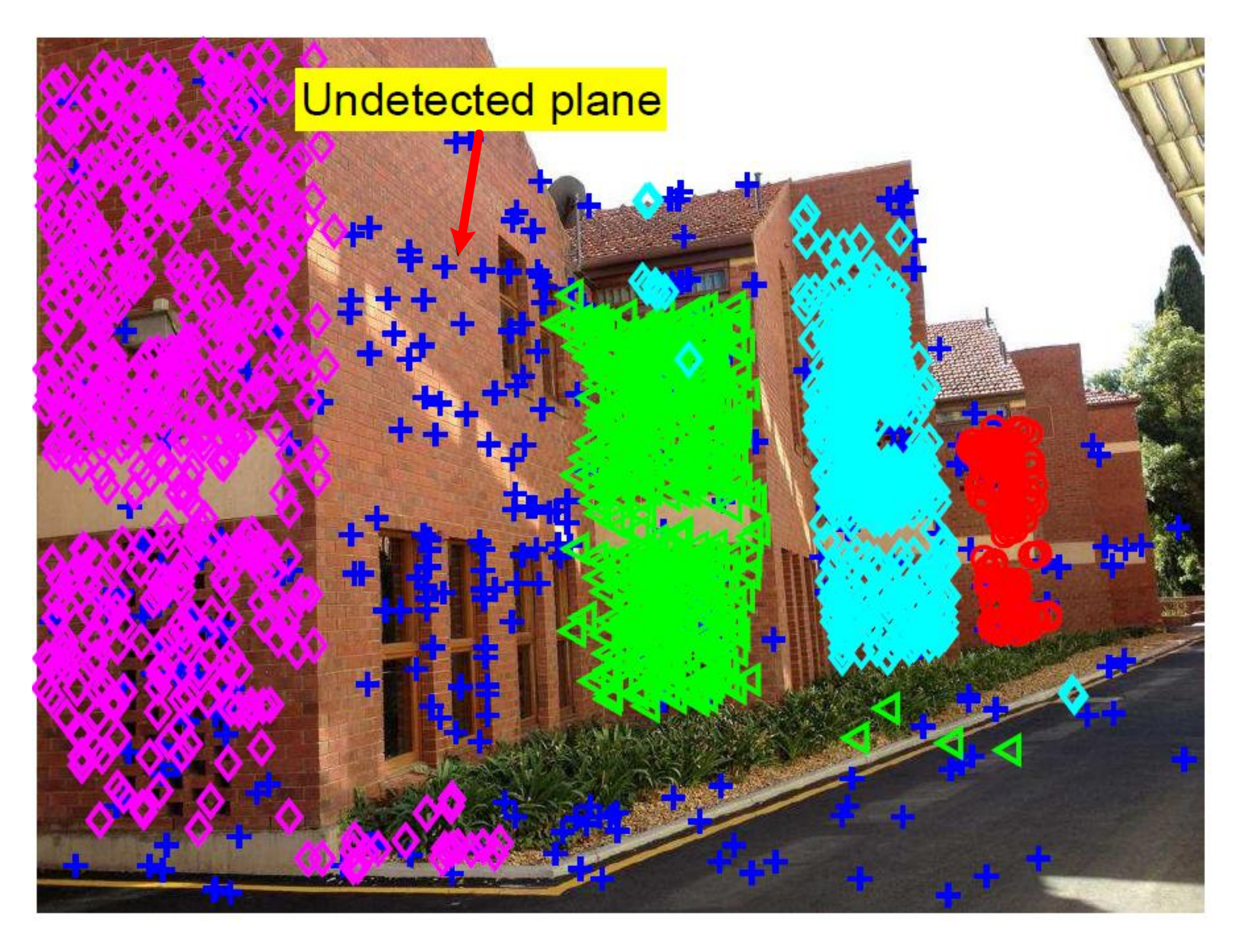}}
 \centerline{(b)}
\end{minipage}
\caption{An example shows that the proposed method fails to estimate the number of model instances in homography based segmentation on the ``Unionhouse'' data (only one of the two views is shown for each case). (a) The decision graph obtained by MSHF2. (b) The results obtained by MSHF2.}
\label{fig:undetectedplane1}
\end{figure}
The proposed method (MSHF) is a parameter space based fitting method, {which can effectively segment data points near the intersection of model instances. However, MSHF cannot effectively estimate the model instance if there are no (or there are {very low proportions of}) model hypothesis} candidates corresponding to the model instance in the initial generated model hypotheses.

For example, as shown in Fig.~\ref{fig:undetectedplane1}, the model instance corresponding to the ``undetected plane" (in Fig.~\ref{fig:undetectedplane1}(b)) {includes} only $4.17\%$ inliers, and there are only {about $0.06\%$ of generated model hypothesis} candidates corresponding to the model instance in the initial $10,000$ generated model hypotheses. MSHF cannot find the representative modes by using the decision graph (in Fig.~\ref{fig:undetectedplane1}(a)). Note that this limitation also affects the other parameter space based fitting methods, e.g., AKSWH.
\section{Conclusion}
\label{sec:conclusion}
This paper formulates geometric model fitting as a mode-seeking problem on a hypergraph, in which each vertex represents a model hypothesis and each hyperedge denotes a data point. Based on the hypergraph, we propose a novel mode-seeking algorithm, which searches for representative modes by analyzing the weighting score of vertices and the similarity between {the vertices and their neighbors. Hypergraph construction, hypergraph reduction} and mode-seeking are effectively combined by the proposed fitting method (MSHF) {to simultaneously estimate the number and the parameters of model instances in the parameter space. MSHF can also effectively alleviate sensitivity to unbalanced data.} Experimental results on both synthetic data and real images have demonstrated that the proposed method significantly outperforms several other start-of-the-art fitting methods for geometric model fitting {(i.e., line fitting, circle fitting, homography-based segmentation and motion segmentation) when the data} involve a large percentage of outliers. 

 \section*{Acknowledgment}%
{\footnotesize This work was supported by the National Natural Science Foundation of China under Grants U1605252, 61472334, 61571379 and 61702431, by the Natural Science Foundation of Fujian Province of China under Grant 2017J01127. David Suter acknowledged funding under ARC DPDP130102524.}
{\small
\bibliographystyle{IEEEtran}

}
\begin{IEEEbiography}[{\includegraphics[width=1in,height=1.2in]{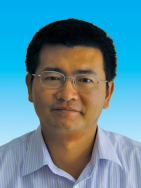}}]{Hanzi Wang}\scriptsize
is currently a Distinguished Professor of  ``Minjiang Scholars" in Fujian province and a Founding Director of the Center for Pattern Analysis and Machine Intelligence (CPAMI) at Xiamen University  in China. He received his Ph.D degree in Computer Vision from Monash University. His research interests are concentrated on computer vision and pattern recognition including visual tracking, robust statistics, object detection, video segmentation, model fitting, optical flow calculation, 3D structure from motion, image segmentation and related fields. He was an Associate Editor for IEEE Transactions on Circuits and Systems for Video Technology (T-CSVT) from 2010 to 2015 and a Guest Editor of Pattern Recognition Letters (September 2009). He was the General Chair for ICIMCS2014, Program Chair for CVRS2012, Area Chair for ACCV2016,  DICTA2010, Tutorial Chair for VALSE2017, Publicity Chair for ICIG2015 and IEEE NAS2012. He also served on the program committee (PC) of ICCV, ECCV, CVPR, ACCV, PAKDD, ICIG, ADMA, CISP, etc, and he serves on the reviewer panel for more than 40 journals and conferences.
\end{IEEEbiography}
\addtolength{\itemsep}{-35em}
\begin{IEEEbiography}[{\includegraphics[width=1in,height=1.2in]{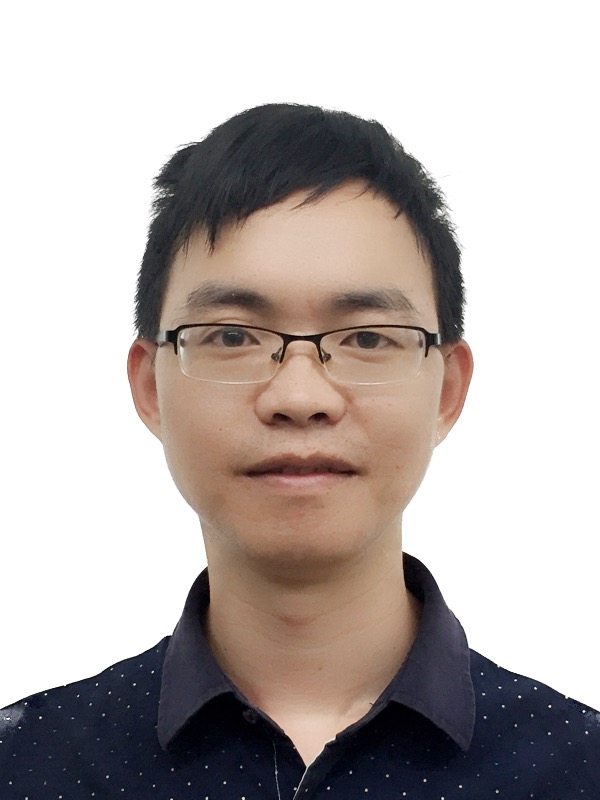}}]{Guobao Xiao}\scriptsize
is currently a research assistant in the School of Aerospace Engineering at Xiamen University, China. He received the Ph.D. degree in Computer Science and Technology from Xiamen University, China, in 2016. He has published over 10 papers in the international journals and conferences including the PR, PRL, CVIU, ICCV, ECCV, ACCV, ICARCV, etc. His research interests include computer vision and pattern recognition.
\end{IEEEbiography}
\begin{IEEEbiography}[{\includegraphics[width=1in,height=1.2in]{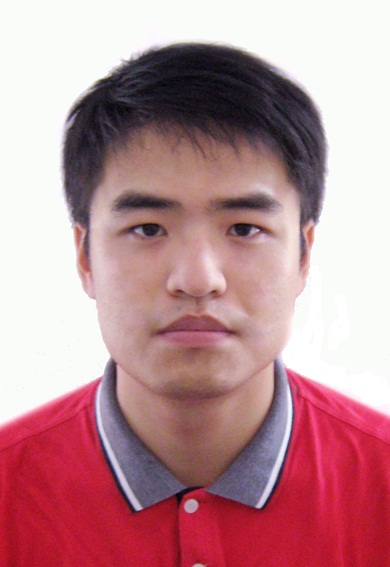}}]{Yan Yan}\scriptsize
is currently an associate professor in the School of Information Science and Engineering at Xiamen University, China. He received the Ph.D. degree in Information and Communication Engineering from Tsinghua University, China, in 2009. He worked at Nokia Japan R\&D center as a research engineer (2009-2010) and Panasonic Singapore Lab as a project leader (2011). He has published around 40 papers in the international journals and conferences including the IEEE T-IP, IEEE T-ITS, PR, KBS, ICCV, ECCV, ACM MM, ICPR, ICIP, etc. His research interests include computer vision and pattern recognition.
\end{IEEEbiography}
\begin{IEEEbiography}[{\includegraphics[width=1in,height=1.2in]{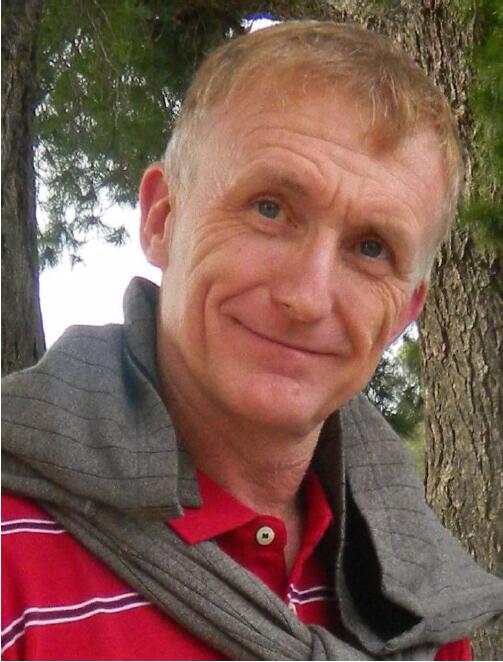}}]{David Suter}\scriptsize
received the BSc degree in applied mathematics and physics from The Flinders University of South Australia in 1977, the Graduate Diploma in Computing from the Royal Melbourne
Institute of Technology in 1984, and the PhD degree in computer science from La Trobe University in 1991. He was a lecturer at La Trobe from 1988 to 1991; and a senior lecturer in 1992, associate professor in 2001, and professor from 2006 to 2008 at Monash University, Melbourne, Australia. Since 2008, he has been a professor in the School of Computer Science, The University of Adelaide. He served on the ARC College of Experts from 2008 to 2010. He is currently the editorial board of the journal Pattern Recognition. He has previously served on the editorial boards of MVA, IJCV and JMIV. He was general co-chair of ACCV 2002 and ICIP 2013.
\end{IEEEbiography}
 \vfill
\end{document}